\documentclass[10pt,twocolumn,letterpaper]{article}
 
\usepackage[pagenumbers]{cvpr} 

\usepackage{comment}
\usepackage{multicol,multirow}
\usepackage{array}

\newcommand{\ourwork}{Prompt2NeRF-PIL}
\newcommand{\coarseF}
{$F_{\text{coarse}}$}
\newcommand{\fineF}
{$F_{\text{fine}}$}

\usepackage[dvipsnames]{xcolor}
\newcommand{\red}[1]{{\color{red}#1}}
\newcommand{\todo}[1]{{\color{red}#1}}
\newcommand{\TODO}[1]{\textbf{\color{red}[TODO: #1]}}

\DeclareMathOperator*{\argmin}{arg\,min}

\definecolor{cvprblue}{rgb}{0.21,0.49,0.74}
\usepackage[pagebackref,breaklinks,colorlinks,citecolor=cvprblue]{hyperref}

\def\paperID{7538} 
\def\confName{CVPR}
\def\confYear{2024}

\title{\ourwork:~Fast NeRF Generation via Pretrained Implicit Latent}
\renewcommand*{\thefootnote}{\fnsymbol{footnote}}
\author{Jianmeng LIU\footnotemark{}\\
HKUST\\
jliudq@connect.ust.hk \\
\and
Yuyao ZHANG\footnotemark[\value{footnote}]\\
HKUST\\
yzhangkp@connect.ust.hk\\
\and
Zeyuan MENG\footnotemark[\value{footnote}]\\
HKUST\\
zmengaf@connect.ust.hk\\
\and
Yu-Wing TAI\\
Dartmouth College\\
yu-wing.tai@dartmouth.edu\\
\and
Chi-Keung TANG\\
HKUST\\
cktang@cs.ust.hk\\
}

\begin{document}
\maketitle

\footnotetext{${}^*$Co-first authors, ranked by alphabetical order of first names}

\renewcommand*{\thefootnote}{\arabic{footnote}}

\begin{abstract}
This paper explores promptable NeRF generation (e.g., text prompt or single image prompt) for {\em direct} conditioning and {\em fast} generation of NeRF parameters for the underlying 3D scenes, thus undoing complex intermediate steps while providing full 3D generation 
with conditional control.
Unlike previous diffusion-CLIP-based pipelines that involve tedious per-prompt optimizations, \ourwork~is capable of generating a variety of 3D objects with a single forward pass, leveraging a pre-trained implicit latent space of NeRF parameters. Furthermore, in zero-shot tasks, our experiments demonstrate that the NeRFs produced by our method serve as semantically informative initializations, significantly accelerating the inference process of existing prompt-to-NeRF methods. Specifically, we will show that our approach speeds up the text-to-NeRF model DreamFusion~\cite{poole2022dreamfusion} and the 3D reconstruction speed of the image-to-NeRF method Zero-1-to-3~\cite{liu2023zero1to3} by 3 to 5 times.



\end{abstract}
\section{Introduction}
\label{sec:intro}
Recent breakthrough in text-to-visual content generation has brought unprecedented success in vision-language transfer, 
enabling communication and collaboration among individuals with diverse backgrounds and facilitating immersive virtual experiences, especially in the gaming and film industry. We have impressive works that bridge the gap between textual inputs and 2D visual representations: Stable Diffusion~\cite{rombach2022highresolution}, DALL$\cdot$E~\cite{ramesh2021zero}, and GPT-4~\cite{openai2023gpt4} are some prominent examples, which have propelled significant progresses in
Text Inversion~\cite{gal2022image}, LoRA~\cite{hu2021lora}, and ControlNet~\cite{zhang2023adding} to finetune ~\cite{rombach2022highresolution, ramesh2021zero} to achieve their specific tasks.

However, due to the scarcity of annotated 3D data and the variety of representations, text-to-3D generation remains challenging. Many works have attempted to build their pipelines upon CLIP~\cite{radford2021learning}. For example, in~\cite{jain2022zero,mohammad2022clip} the lack of data is alleviated by using CLIP to optimize the corresponding 3D models, but purely using pretrained text-image models leads to costly optimization. CLIP-Forge~\cite{sanghi2022clipforge} proposes an efficient strategy to generate 3D mesh from text without using annotated text image pairs, but the out-of-distribution performance is less satisfactory. Feature-disentangled control was incorporated in~\cite{wang2022clip, jang2021codenerf, michel2022text2mesh,wang2023nerf} using CLIP to demonstrate interpolatable latents. With the prior of pretrained text-image diffusion models such as~\cite{rombach2022highresolution}, DreamFusion~\cite{poole2022dreamfusion} and Magic-3D~\cite{lin2023magic3d} can generate 3D scenes from text prompts while using CLIP to finetune NeRF or Mesh. This approach however makes speed a severe issue, since multiple back-propagations of score distillation loss and renderings are necessary. To achieve fast 3D generation, one pioneer work~\cite{erkocc2023hyperdiffusion} attempted to implicitly generate 3D scenes by regressing  model parameters of a given NeRF network by a diffusion model in an unconditional manner. Despite its contributions in demonstrating the feasibility of implicit generation of NeRF parameters, this first  paper uses 
simple datasets with lack of user control.

\begin{figure*}[t]
    \centering
    \includegraphics[width=1\textwidth]{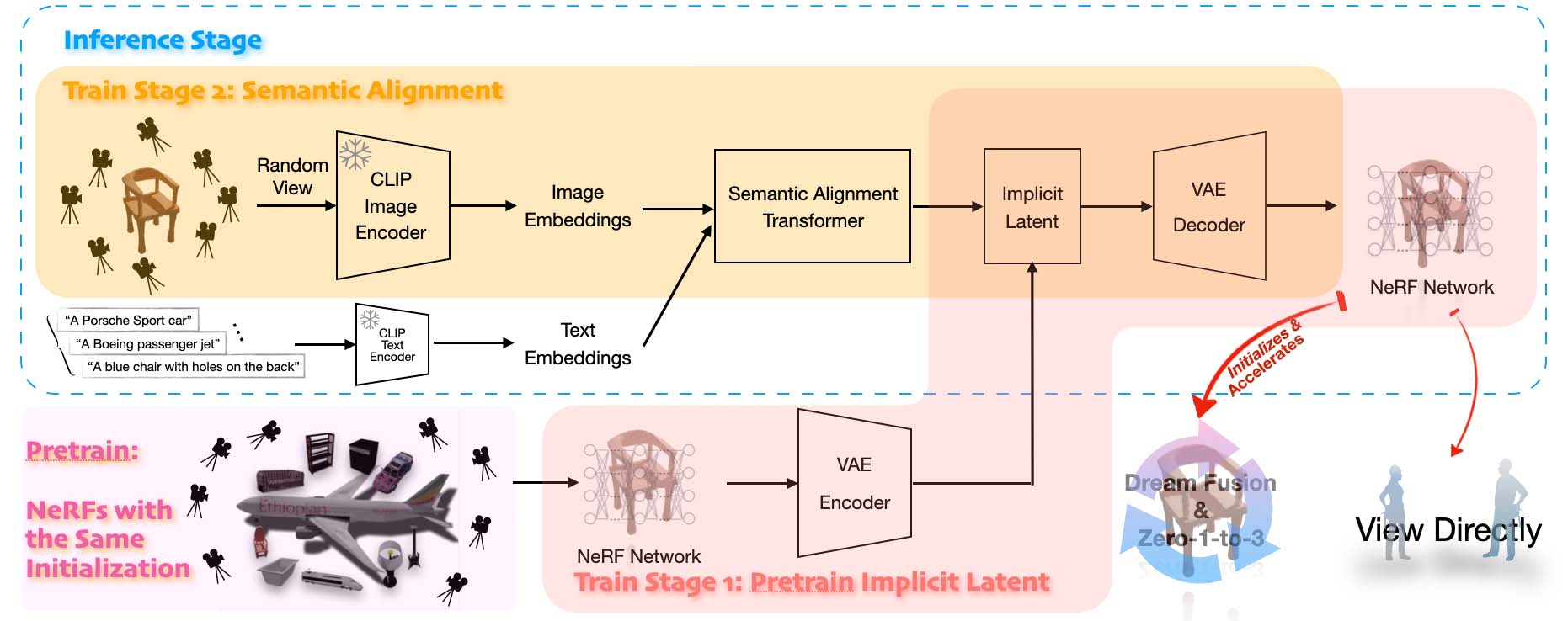}
    \caption{\ourwork's {\bf overall pipeline}, which accepts a textual or image input with a single forward pass during inference. Two main modules are shown: a semantic alignment transformer model that maps CLIP embeddings input to a pretrained implicit latent NeRF space, and then a VAE decoder transforms it into a parameter space representing a NeRF network that can either be directly rendered or serve as a good initialization used in prior text-to-NeRF works such as DreamFusion~\cite{poole2022dreamfusion} and image-to-3D works such as Zero-1-to-3~\cite{liu2023zero1to3} to drastically reduce their generation time.}
    \vspace{-0.2in}
    \label{fig:pipeline}
\end{figure*}

We propose to overcome these challenges by exploring the feasibility of {\em directly} generating NeRF parameters with only one single forward pass from text or image prompts, with an \textit{optional} optimizing process~(accelerated by our method) using diffusion-based methods such as~\cite{poole2022dreamfusion} and~\cite{liu2023zero1to3} to improve the performance on out-of-domain tasks, where the output NeRF from our model serves as a good initialization. 

Our~\ourwork~is a simple yet effective pre-training strategy, where an encoder-decoder structure is trained to encode NeRF parameters into an implicit latent space with desirable properties such as interpolability, and then using Transformer~\cite{vaswani2023attention} to align the prompt semantics with the pretrained implicit latent space, which enables direct prediction of the NeRF parameters from prompts. Inspired by CLIP-Forge~\cite{sanghi2022clipforge}, we leverage image embeddings to tackle the problem of lack of text descriptions, allowing text and image prompts to be used interchangeably.

During the testing stage, our method can generate a satisfactory in-distribution object within a few seconds by a single forward pass. While for out-of-distribution generation, our method is able to provide a good initialization of the NeRF model. We demonstrate in our comprehensive experiments 
that such initialization can boost the convergence speed of DreamFusion~\cite{poole2022dreamfusion} and Zero-1-to-3~\cite{liu2023zero1to3} by 3 to 5 times, along with better and more consistent results in more generalized scene categories. Extensive experimental results and definitions of test scenarios (in-distribution and out-of-distribution) will be presented.

We summarize our contributions as follows:
\begin{enumerate}
    \item As far as we know, \ourwork~is the first work to enable implicit NeRF parameters generation of diverse objects from text and image prompts in a single forward pass, without any per-prompt training required.
    \item Our simple yet effective method can serve as a good initialization and guidance to accelerate other main-stream text-to-3D generation methods such as~\cite{poole2022dreamfusion} and image-to-3D methods such as 3D reconstruction using~\cite{liu2023zero1to3}, allowing for faster NeRF generation in higher-quality.
    \item Our pretrained implicit latent demonstrates the possibility of endowing the topological space of the parameters with semantics. And we proposed a new dataset for NeRF training.
\end{enumerate}

\section{Related Works}
\textbf{Neural Radiance Field.}\quad NeRF, or Neural Radiance Field  introduced in~\cite{mildenhall2021nerf} is a generative model that establishes a mapping from accurate camera poses to a 3D scene given a set of images. Unlike traditional representations such as voxels and meshes which highly depend on explicit 3D models, the foundation of NeRF is a radiance field, which is a function of how a batch of light rays travels through a 3D volume.
NeRF can be formulated as:
\begin{align*}
    F:(\textbf{x}, \textbf{d}) \rightarrow (\textbf{c}, \boldsymbol{\sigma})
\end{align*}
where ${\bf x}=(x,y,z)$, ${\bf d}=(\Theta,\phi)$ are respectively the location coordinate and direction in a predefined space, ${\bf c}=(r,g,b)$ is the color at the position, and $\boldsymbol{\sigma}$ can be thought of the probability that a light ray is occluded in 3D space.
Following this novel representation, 
Mip-NeRF~\cite{barron2021mip} alternatively uses light cones to reduce artifacts, and Bungee-NeRF~\cite{xiangli2022bungeenerf} uses multi-resolution images to train Mip-NeRF progressively, which enables city-scale scenes rendering. Existing works also use explicit representations to obtain a hybrid NeRF. For example, Instant-NGP~\cite{muller2022instant} uses a multi-resolution hash encoding structure; TensoRF~\cite{chen2022tensorf} adopts factorizations to achieve a compact encoding space; Plenoxels~\cite{yu_and_fridovichkeil2021plenoxels} leverages a sparse grid for NeRF synthesis. In this experimental work, we adopt the MLP similarly done in the first NeRF work~\cite{mildenhall2021nerf} as the foundation model in coarse-to-fine refinement.

\noindent\textbf{Variational AutoEncoders.}\quad The Variational Autoencoder (VAE)~\cite{kingma2013auto} 
learns a low-dimensional latent space to encode the underlying distribution of the input data, thus allowing sampling from the latent to enable the generative feature. 
VAE demonstrates robustness in handling noisy data, thus, in this work, we leverage its inherent properties that facilitate smooth interpolation between latent variables, so that meaningful semantic relationships can be preserved. 

\noindent \textbf{CLIP and CLIP-Based 3D Generation} \quad Contrastive Language-Image Pre-Training (CLIP)~\cite{radford2021learning} offers a zero-shot capability that bridges the gap between the latent spaces of text and image. To date, it has been utilized for various zero-shot downstream applications~\cite{luo2021clip4clip, fang2021clip2video, lei2021more, sanghi2022clipforge}. As CLIP bridges text and vision, many works~\cite{sanghi2022clipforge, wang2022clip, jang2021codenerf, michel2022text2mesh, wang2023nerf} attempt to use it to optimize 3D models. However, despite their innovations, most of these works require a considerable amount of time to achieve satisfactory results. Recently, an image-to-3D work Zero-1-to-3~\cite{liu2023zero1to3} tweaks Stable Diffusion~\cite{rombach2022highresolution} to adopt a vector that combines clip image embedding and pose information as input, and employs the images generated by their model as input to train NeRF model. Other methods such as~\cite{poole2022dreamfusion, lin2023magic3d} have attempted to leverage the vast information within pretrained generative models via score distillation sampling under CLIP's guidance, by incorporating Stable Diffusion. Though having better performances on zero-shot generations, these approaches still require a long time for self-optimization and suffer from poor view consistency due to the use of 2D supervision without any initialization or view information. Therefore, aligning CLIP's embedding space with the 3D structure space is necessary.

\noindent \textbf{Concurrent Works}\quad At the time of writing, we noticed one concurrent work also investigates the ability to implicitly generate NeRF structures. ATT3D~\cite{lorraine2023att3d}, which adopts the InstantNGP~\cite{muller2022instant} NeRF backbone, optimizes a set of prompts at the same time to save training time and share the feature between the prompts to enables unseen generation. However, the amortization direction is quite limited as the result figures shown in the paper, and that the implicit part only lies in the feature grid of InstantNGP. Other text-to-3D models such as~\cite{shi2023mvdream,wang2023prolificdreamer} aim to solve the 3D consistency problem whose setting is thus different from ours.

\section{Method}
\label{sec:method}

\Cref{fig:pipeline} shows \ourwork's overall architecture, which generates a NeRF from a textual or image input with merely a single forward pass during inference. Our framework consists of two main modules: an alignment model $H$ that maps input ${\bf c}$ to an implicit NeRF latent space ${\bf z} = H({\bf c})$, and then a decoder $D$ to transform ${\bf z}$ back into a parameter space representing a NeRF $F = D\big({\bf z}\big)$. The NeRF obtained can either be directly rendered for in-distribution generation tasks, or provide a good initialization for out-of-distribution tasks to significantly accelerate the generation of prior text-to-NeRF works such as DreamFusion~\cite{poole2022dreamfusion}, and image-to-3D works such as Zero-1-to-3~\cite{liu2023zero1to3}.

\subsection{NeRF Parameters Dataset}
\label{sec:method-nerf-dataset}
In the absence of prior implicit generation tasks, as noted by~\cite{erkocc2023hyperdiffusion}, it is necessary to develop a new dataset of NeRF parameters. To construct the dataset, we first 
use Blender to render 100 different camera views
for each 3D scene $S_i, i\in \{1,2,\cdots, N\}$ selected from: chair objects from Objaverse~\cite{deitke2023objaversexl} and 7 other categories from ShapeNetCore~\cite{chang2015shapenet}, resulting in a total of $N=4071$ three-dimensional scenes across 8 categories. We then train a NeRF model $F_{\Theta_i}(\cdot)$ for each instance $S_i$ on the obtained camera views. 

To reduce the training time for each NeRF, we adopt a similar approach in~\cite{erkocc2023hyperdiffusion}, where a simplified 7-layer MLP is used instead of the Vanilla NeRF in~\cite{mildenhall2021nerf}, which reduces the training time for each NeRF to around 10 minutes on a single RTX 4090 board. Then we partition and save the parameters of the MLPs ${\Theta_i}, i\in \{1,2,\cdots,N\}$ into 14 components, as $\Theta_i$ consists of 7 weights ${\bf W}_j$ and 7 biases ${\bf b}_j$ for each layer $j=1,2,\cdots,7$. 

We randomly split the datasets into 90\% training data and 10\% testing data. The optimization process is halted at 15,000 iterations, as our empirical evidence demonstrates satisfactory PSNR levels (33.82 on average) can usually be achieved. Moreover, to avoid generating a prohibitive solution space of NeRF parameters, we use the same NeRF initialization for all data as inspired by~\cite{erkocc2023hyperdiffusion} where, in differential geometry 
all 3D genus-zero objects can be deformed into the same topological sphere. We will demonstrate the effectiveness of this initialization as it yields favorable topological properties in our latent space in~\cref{sec:ablation-nerf}.

\subsection{~\ourwork\  Pipeline}
\label{sec:method-pipeline}

\textbf{Stage 1: Pretrain Implicit Latent} \quad
In order to achieve a smoothly interpolatable latent representation, we first encode NeRF parameters into a latent space. Inspired by~\cite{sanghi2022clipforge} that adds Gaussian noise to Autoencoder to improve robustness, we instead utilize a Variational Autoencoder (VAE~\cite{kingma2013auto}) that learns the latent space as a distribution to furthur enhance smoothness and robustness. Specifically, VAE first encodes the input NeRF parameter $\Theta_i$ in 14 partitions into two latent vectors ${\boldsymbol \mu}_i \in \mathbb{R}^{14\times d}, {\boldsymbol \sigma}_i \in \mathbb{R}^{14\times d}$ representing respectively the mean and variance of the distribution,
where $d$ is the dimension of the latent space. 
Then we randomly sample from this distribution and decode back to the reconstructed NeRF network $\Theta_i$. Denote the Encoder and Decoder respectively as $E(\cdot; \alpha), D(\cdot; \beta)$, where $\alpha, \beta$ are the corresponding parameters. In this stage, we use Mean Square Error loss (MSE) as supervision. The whole process can be formulated as the following optimization:
\begin{align}
    \hat{\alpha}, \hat{\beta} = \argmin_{\alpha, \beta} \frac{1}{N} \sum_{i=1}^N \left\| D\big(E(\Theta_i)\big)-\Theta_i \right\|^2
\end{align}


\noindent\textbf{Stage 2: Semantic Alignment}\quad
Our semantic alignment Transformer model $H$ leverages the structure introduced in~\cite{vaswani2023attention} that maps a semantic embedding ${\bf c}_i \in \mathbb{R}^{d}$ that represents a 3D scene to ${\bf z}_i \in \mathbb{R}^{14\times d}$, its corresponding NeRF parameters in the latent space obtained from VAE. 
We use the CLIP image encoder to obtain the embeddings from images. Since one image cannot thoroughly represent information for one 3D scene, for each scene $S_i$, we randomly select one image $S_i(\boldsymbol{x},\boldsymbol{d})$ from different viewpoints $(\boldsymbol{x},\boldsymbol{d})$ each epoch during training to increase alignment robustness. In such a way, a rotation-invariant mapping is learned from a general representation $R_i$ invariant to the viewpoint and rotation $(\boldsymbol{x},\boldsymbol{d})$ chosen of each 3D scene $S_i$
in the CLIP embeddings space,
to the implicit NeRF latent space\footnote{We include more results on this in supplementary materials.}. 

During inference, an image from an arbitrary view can be used as input for one specific scene. For text prompts, we notice that even though CLIP effectively maps images and texts into the same space, huge discrepancies still exist between their embeddings. We will show in supplementary material that for semantically highly-matched text and image pairs, the cosine similarity of their clip embeddings is still low, typically around 30\% to 40\%, indicating it is infeasible to directly use text CLIP embedding for inference. To get around this issue, when prompted with text, we first find the semantically nearest scene with the given text prompt in our training set and use the CLIP embedding of images of that scene instead.


\section{Experiments}
\label{sec:experiments}
We will first briefly describe the datasets in~\cref{sec:dataset}, and summarize the implementation details in~\cref{sec:implementation}. Then we present the baselines and metrics we used in~\cref{sec:exp-baselines} and~\cref{sec:exp-metrics}, respectively, followed by our experimental results in two aspects in~\cref{sec:results-indomain} and~\cref{sec:results-outdomain}. Specifically, we evaluate \ourwork~in two settings: 1)~in-distribution generation, and 2)~out-of-distribution generation and acceleration for current prompt-to-NeRF works. To clarify, we define ``in-distribution" as prompting \ourwork~using a) images of unseen objects in the test split from our dataset, or b) texts that describe objects whose category is present in our datasets, since our training set does not contain any text description of the objects. For ``out-of-distribution" generation, we use a)~images from other sources whose categories are not included in the dataset, and b)~texts that describe objects whose categories are not included in our dataset.

\subsection{Datasets}
\label{sec:dataset}
For all our in-distribution experiments, we use the dataset built by ourselves mentioned in \Cref{sec:method-nerf-dataset} which contains 8 categories and the ratio of the training set to the testing set is 9:1. We further modified our dataset to construct an image-NeRF parameter pair dataset, where the image is rendered from the corresponding NeRF scene.

\subsection{Implementation Details}
\label{sec:implementation}

In this section, we highlight some implementation details of our method. More detailed information will be provided in supplementary materials.

For VAE training, we used a learning rate of 
$3\times 10^{-4}$ and trained for $10000$ epochs, while for semantic alignment, we used a learning rate of $10^{-6}$ and trained for $100000$ epochs until convergence. Adam optimizer~\cite{kingma2017adam} was used in both cases. In our experiments, we utilized the CLIP ViT-B/32 model to generate image and text embedding, thus the dimension of VAE's latent space is naturally designed to $512$.

All experiments were run on NVIDIA RTX 4090 graphic cards. Unless specified, all times reported are recorded when running the experiment on one RTX 4090 card.

\begin{figure}[ht]
    \centering
    \includegraphics[width=0.95\linewidth]{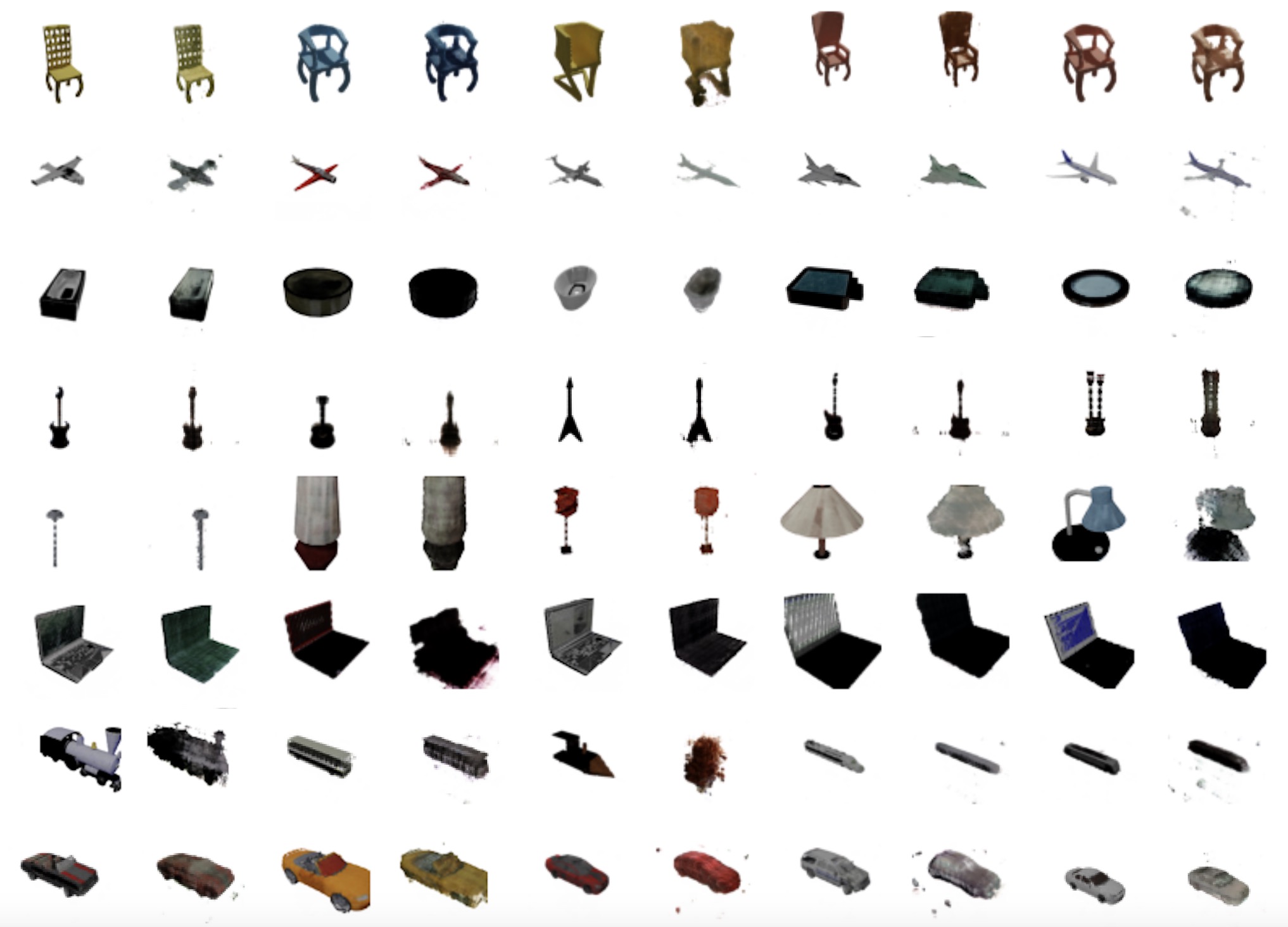}
    \caption{Results on {\bf in-distribution generation from image prompts} randomly selected in test split. Each row presents one category (8 categories in total), where each pair 
    represents the ground-truth image and the model-generated result, respectively. 
    }
    \label{fig:testset11by5_5}
    \vspace{-0.2in}
\end{figure}

\begin{figure*}[ht]
    \centering  \includegraphics[width=0.95\textwidth]{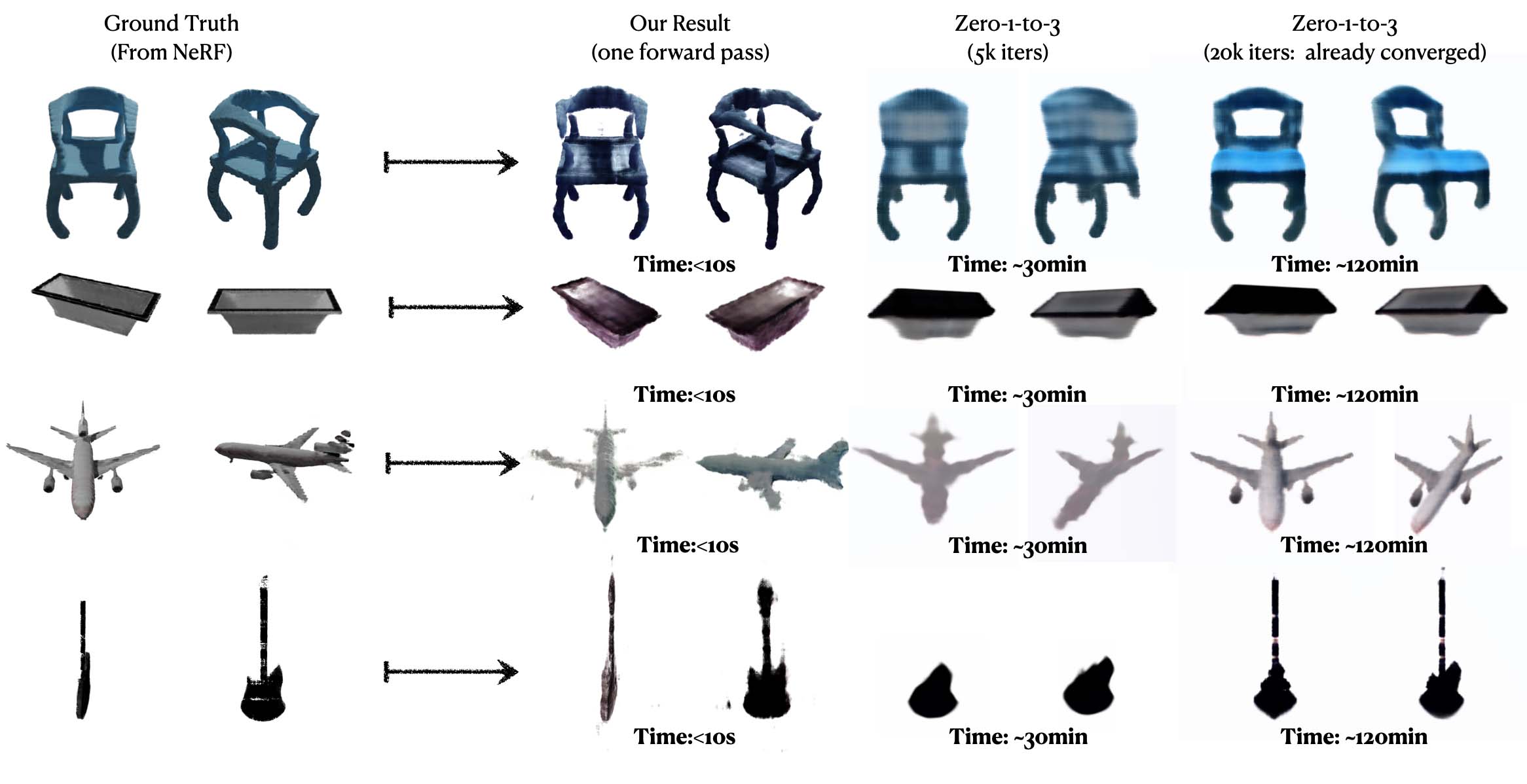}
    \vspace{-0.2in}
    \caption{{\bf In-distribution generation results from image prompts}. The first column contains the images directly rendered by our ground truth NeRFs; the second column contains the single forward inference results of our models; the third and fourth columns are the results generated by Zero-1-to-3, trained for 5k iterations and until convergence, respectively. For the blue chair and the guitar, our results are clearly of higher quality. As for the plane and the bathtub, although the quality seems less satisfying at first glance, our result is a consistent 3D solid model, while the other lookalikes are revealed as planar cardboard with little 3D consistency when viewed sideways.} 
    \label{fig:indomain-image-compare}
    \vspace{-0.2in}
\end{figure*}

\subsection{Baselines}
\label{sec:exp-baselines}
We compare our results with prior works DreamFusion~\cite{poole2022dreamfusion} for text-to-NeRF tasks and Zero-1-to-3~\cite{liu2023zero1to3} for image-to-NeRF tasks. For their implementations, we leveraged the open-source project Stable-Dreamfusion~\cite{stable-dreamfusion}, a popular unofficial implementation of DreamFusion, as DreamFusion has not yet released their implementation and Stable-Dreamfusion also supports Zero-1-to-3 as guidance, while at the same time, a NeRF can be output instead of the official implementation of Zero-1-to-3 that can only synthesis novel views\footnote{Moreover, in the official implementation of Zero-1-to-3, the authors directly link the Stable-Dreamfusion's repository for 3D Reconstruction task.}.
To ensure fairness, we replaced their NeRF network structure with ours, hence the validity of time comparison will not be undermined by different rendering times due to inconsistent NeRF structures.
Moreover, we observed occasional instability in the optimization process of DreamFusion and Zero-1-to-3 using the implementation in~\cite{stable-dreamfusion}. To address this, we ran it with five different seeds for all experiments, then selected and presented the best result obtained. Despite this approach, it is important to note that we still encountered some cases where the optimization did not converge or failed to generate a scene relevant to the given prompt.

\subsection{Metrics}
\label{sec:exp-metrics}
For quantitative results, we use two different metrics: Fréchet Inception Distance (FID)~\cite{heusel2018gans} and cosine similarity score to validate that the NeRFs generated by our model are close to ground-truth NeRFs.
For FID, we compare the generated NeRFs with the ground truth NeRFs by computing the distance between the feature distributions of real and generated images rendered by corresponding NeRF using the Fréchet distance. Notice that due to inconsistent networks being used, the results may not be comparable to those in other works. Specifically, we used~\cite{pytorch-fid} with dim 192. 
Due to the lack of high-quality image captions in our dataset, we use the cosine similarity between the prompt and the embeddings of the rendered views of the generated NeRF as an analogy of the CLIP retrieval score~\cite{Park2021BenchmarkFC} to evaluate the quality of our output. 

\subsection{Results on In-Distribution Generation}
\label{sec:results-indomain}

\begin{table}[ht]
   \small
   \centering
   \resizebox{\linewidth}{!}{%
   \begin{tabular}{cccc}
   \toprule
   \multirow{2}{*}{\textbf{Metrics}} & \textbf{\textit{Ours}} & \textbf{Ours} & \multirow{2}{*}{\textbf{Zero-1-to-3}} \\ 
    & \textbf{\textit{(train split)}} & \textbf{(test split)} & \\
   \midrule
   \multirow{1}{*}{Similarity $\uparrow$} 
   & {\it (83.1\%)} & 81.5\% (81.3\%) & 80.7\% \\
   \multirow{1}{*}{FID $\downarrow$} 
   & {\it (1.334)} & 4.189 (3.071) & 13.525 \\
   \bottomrule
   \end{tabular}%
   }
   \caption{{\bf Quantitative results on in-distribution image inference task.} We randomly sample 40 images and compare the CLIP embedding similarity and FID score of our method with Zero-1-to-3. The Scores in parentheses represent the evaluation of the entire train/test split, and the scores evaluated on the train split serve as an upper bound reference (in {\it italics}). 
   Our method outperforms Zero-1-to-3 in higher similarity and lower FID score. Though Zero-1-to-3 also has high similarity scores, the low FID scores may indicate that it suffers from view inconsistency problems. Moreover, the performance of our method demonstrates comparable performance to the train split, indicating an effective alignment learned by our model.}
   \label{tab:exp1_image_quant}
   \vspace{-0.15in}
\end{table}

\begin{table}[ht]
\footnotesize
    \newcommand{\dummyImg}     
    {\fbox{\rule{0pt}{0.9in} \rule{0.65\linewidth}{0pt}} }
    \centering
\resizebox{0.99\linewidth}{!}{    
    \begin{tabular}{>{\centering\arraybackslash}m{0.20\linewidth}|>{\centering\arraybackslash}m{0.24\linewidth}|>{\centering\arraybackslash}m{0.22\linewidth}|>{\centering\arraybackslash}m{0.24\linewidth}}
    \toprule
        Text Prompt & Ours & DreamFusion (5k iters.) & DreamFusion (converge)\\
    \midrule
        \vspace{0.25in}  
        {\it ``A green chair with armrests and a hollow back.''}  
        & \includegraphics[width=0.95\linewidth]{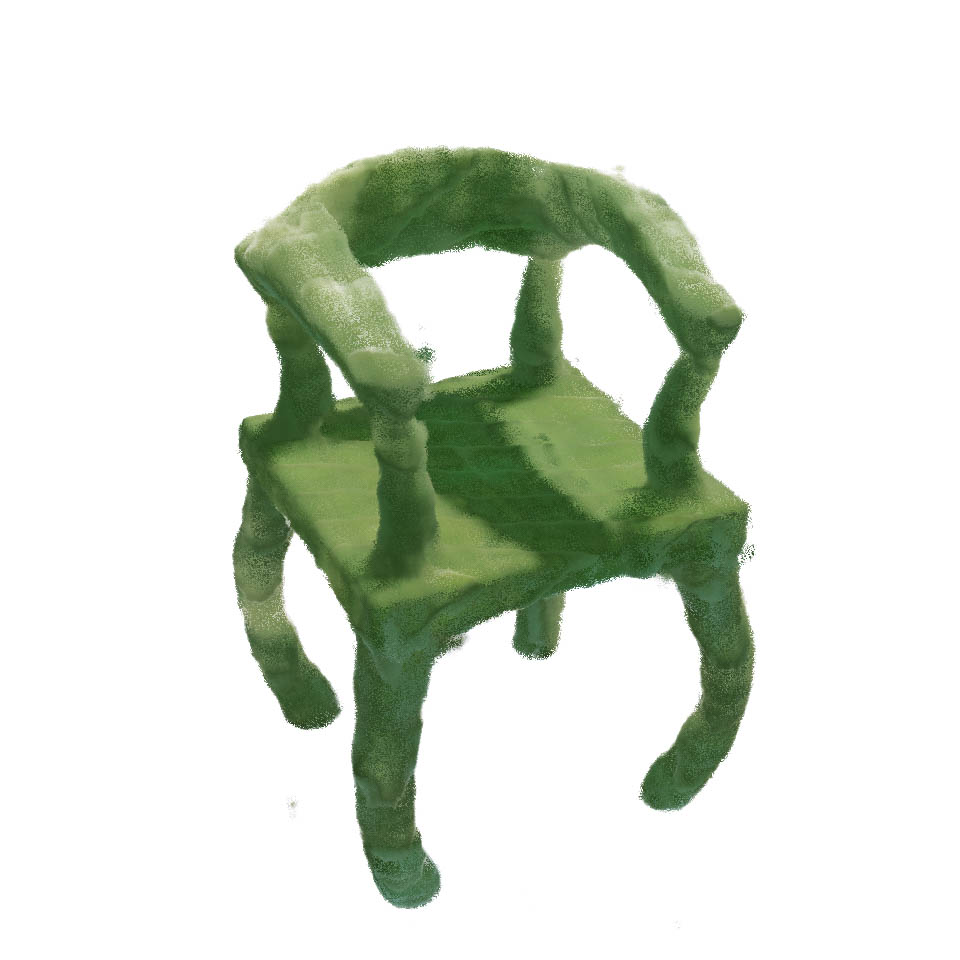}
        & \includegraphics[width=0.95\linewidth]{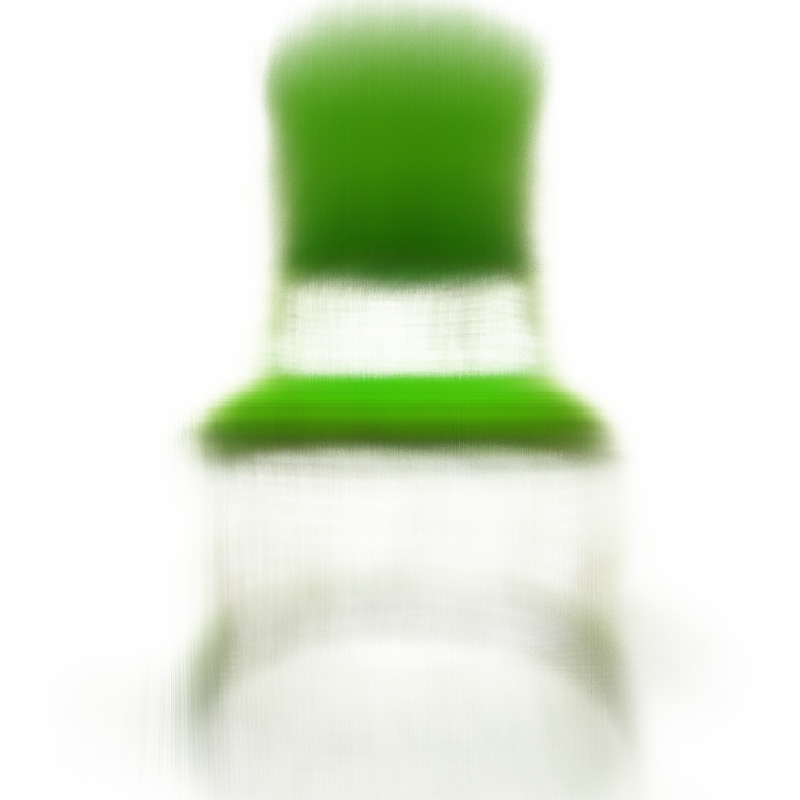}
        & \includegraphics[width=0.95\linewidth]{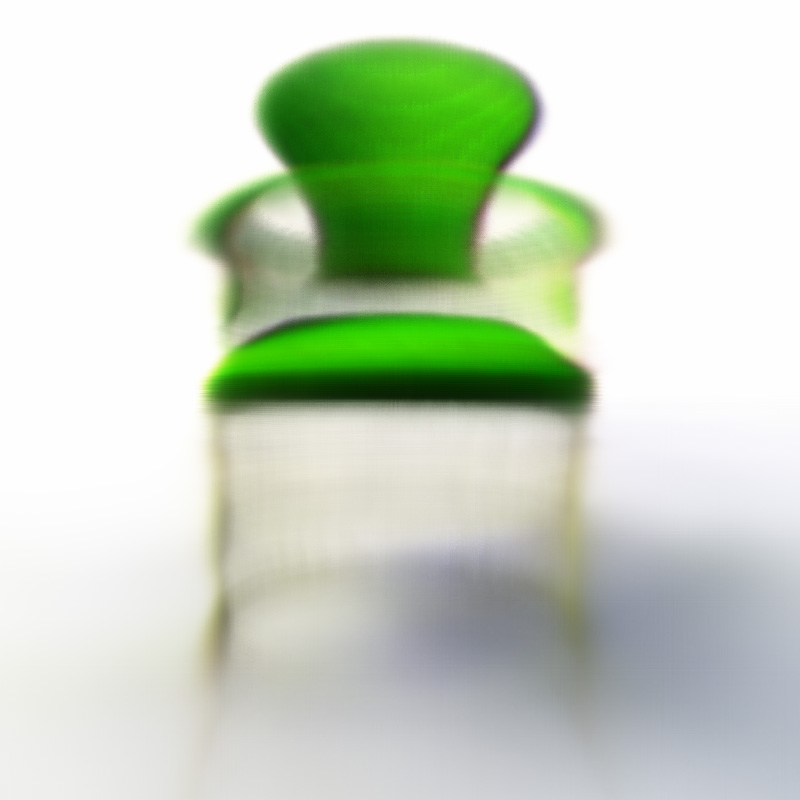}\\
         & Sim: 35.6\% & Sim: 32.7\% & Sim: 29.5\%\\
         & Time: $< 10$s & Time: $\sim 30$min & Time: $\sim 180$min\\\hline
        \vspace{0.25in}  
        {\it ``An F16 fighter jet.''}  
        & \includegraphics[width=0.75\linewidth]{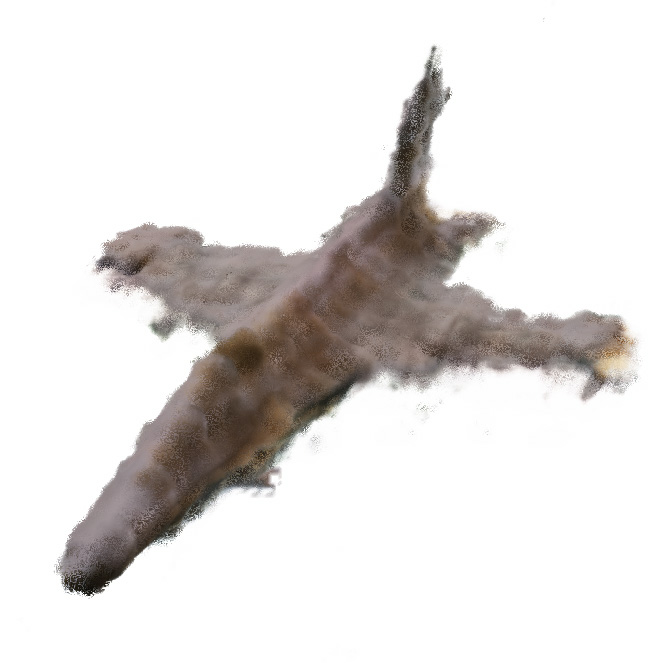}
        & \includegraphics[width=0.58\linewidth]{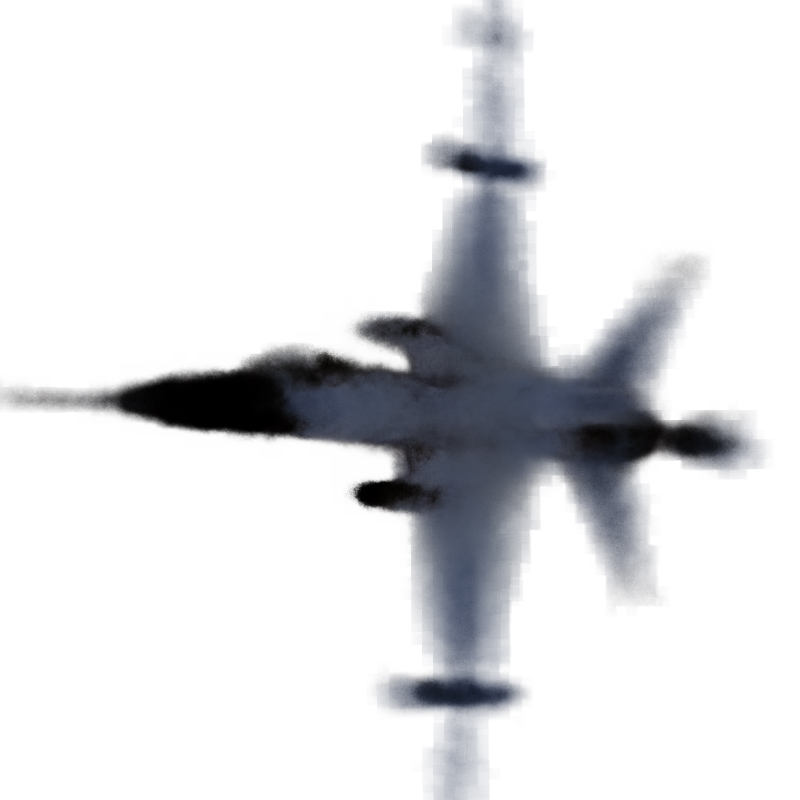}
        & \includegraphics[width=0.55\linewidth]{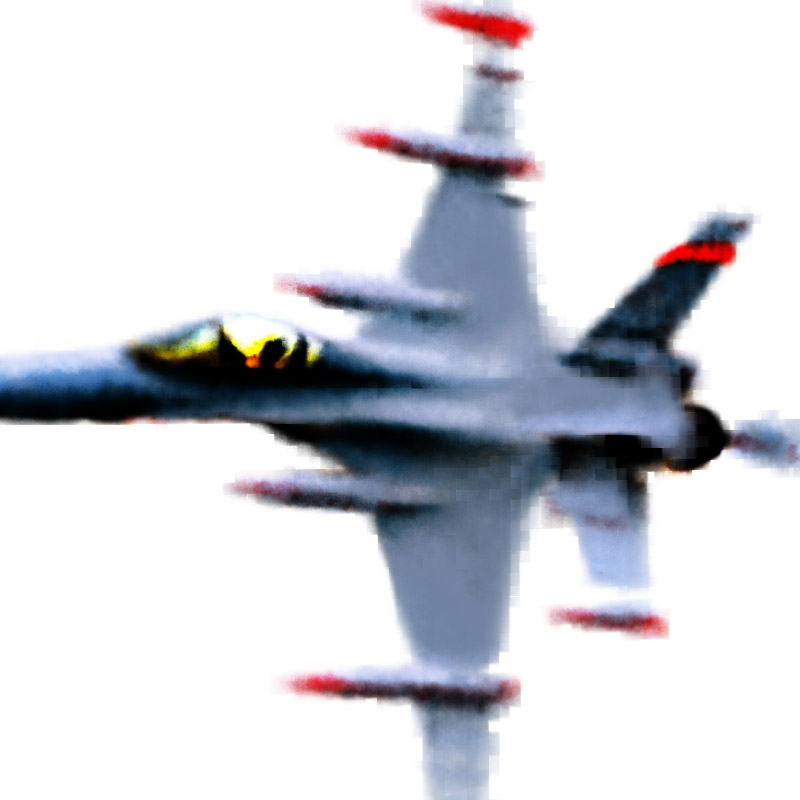}\\
         & Sim: 28.2\% & Sim: 29.4\% & Sim: 28.2\%\\
         & Time: $< 10$s & Time: $\sim 30$min & Time:  $\sim 190$min\\\hline
        \vspace{0.25in}  
        {\it ``A purple and blue sport car.''}  
        & \includegraphics[width=0.65\linewidth]{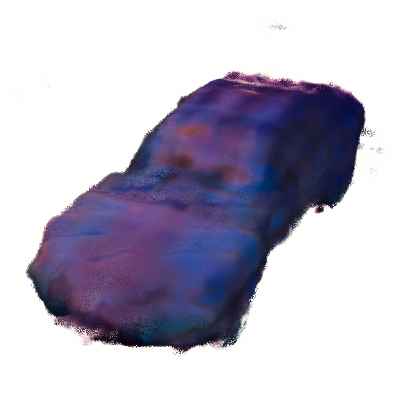}
        & \includegraphics[width=0.7\linewidth]{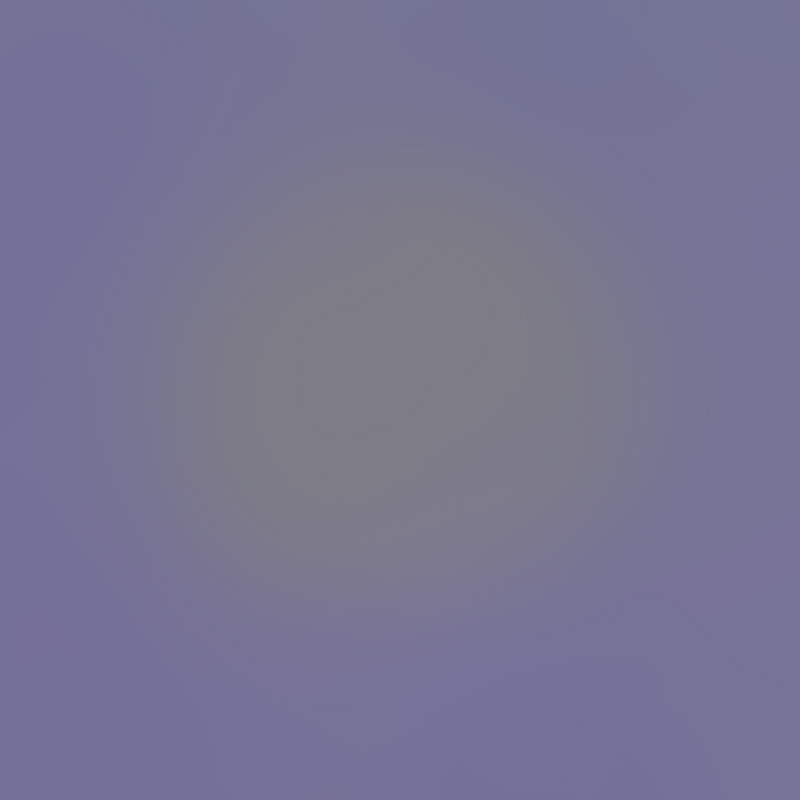}
        & \includegraphics[width=0.66\linewidth]{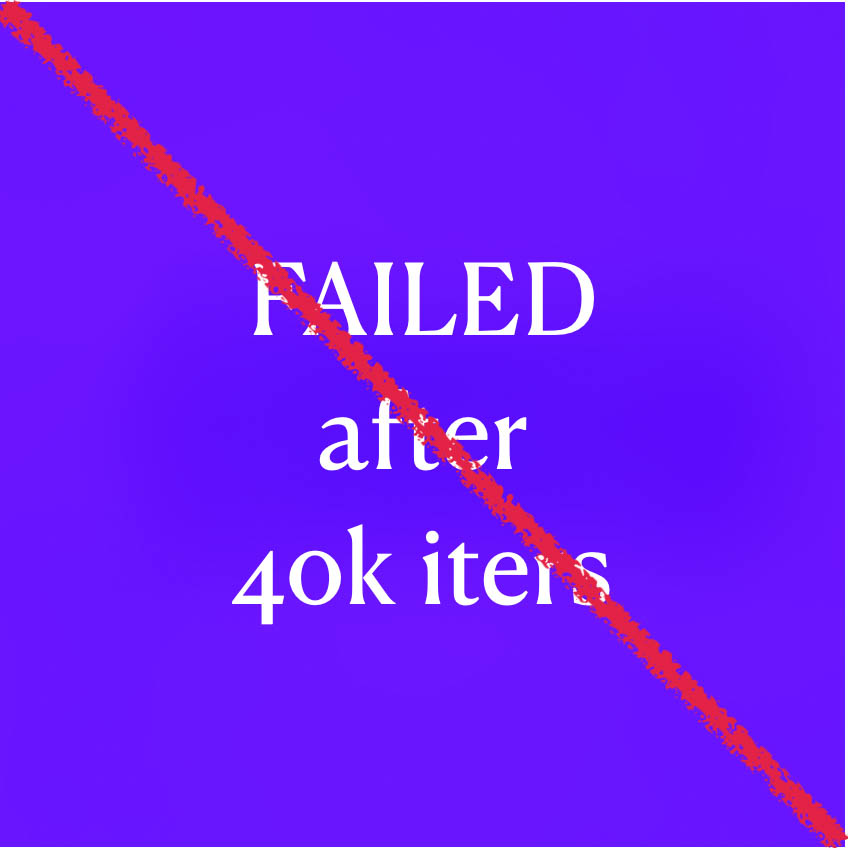}\\
         & Sim: 27.0\% & Sim: 22.2\% & Sim: 26.6\%\\
         & Time: $<10$s & Time: $\sim30$min & Time: $>250$min\\
    \bottomrule
    \end{tabular}
    }
    \caption{Comparison of {\bf in-distribution inference result with text prompts}. 
    For DreamFusion results, we optimized each prompt for both 5000 iterations and until convergence. Our method almost consistently achieves higher CLIP embedding cosine similarities compared to DreamFusion, and it does so in significantly shorter time frames, demonstrating the effectiveness of our method on the text inference task. For the last prompt, DreamFusion failed to converge to a satisfying result after several trials, as explained in~\cref{sec:exp-baselines}, even though the similarity is high, indicating that sometimes CLIP guidance may mislead DreamFusion in unfavorable directions.}
    \label{tab:in-dist-text-prompt-compare}
    \vspace{-0.2in}
\end{table}

\begin{table*}[ht]
    \newcommand{\dummyImg}{\fbox{\rule{0pt}{0.7in} \rule{0.7\linewidth}{0pt}} }
    \centering
    \begin{tabular}{>{\centering\arraybackslash}m{0.13\linewidth}|>{\centering\arraybackslash}m{0.15\linewidth}|>{\centering\arraybackslash}m{0.15\linewidth}|>{\centering\arraybackslash}m{0.15\linewidth}|>
    {\centering\arraybackslash}m{0.15\linewidth}|>
    {\centering\arraybackslash}m{0.15\linewidth}}
    \toprule
        \multirow{2}{*}{Prompt} & Ours & Zero-1-to-3 (S) & Zero-1-to-3 (P) & Zero-1-to-3 (S) & Zero-1-to-3 (P)\\
         & (converge) & (same iterations) & (same iterations) & (converge) & (converge) \\
    \midrule
        \vspace{0.2in}  
        \includegraphics[width=0.82\linewidth]{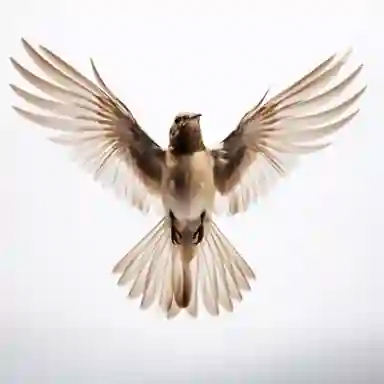}
        & \includegraphics[width=0.82\linewidth]{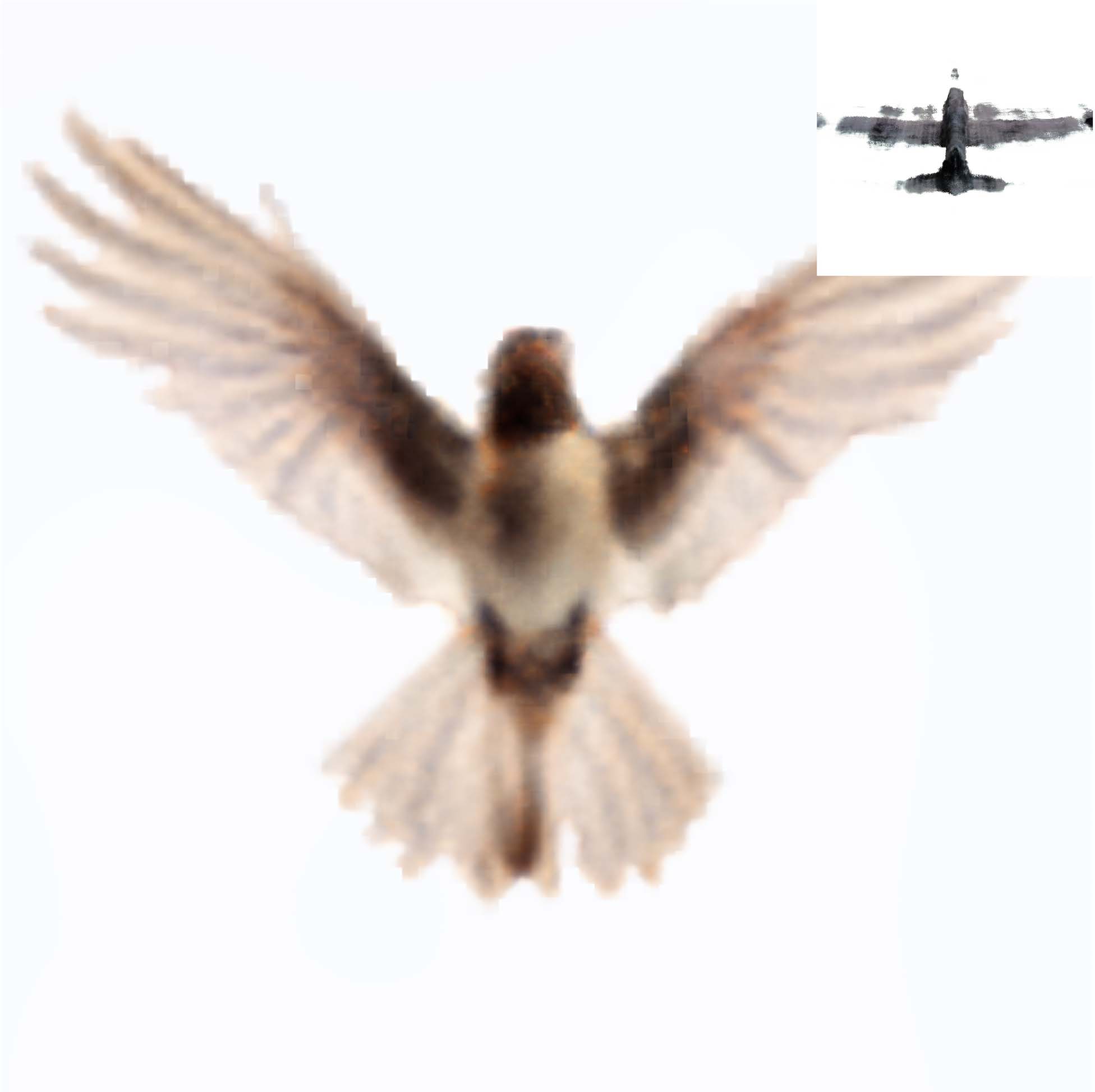}
        & \includegraphics[width=0.82\linewidth]{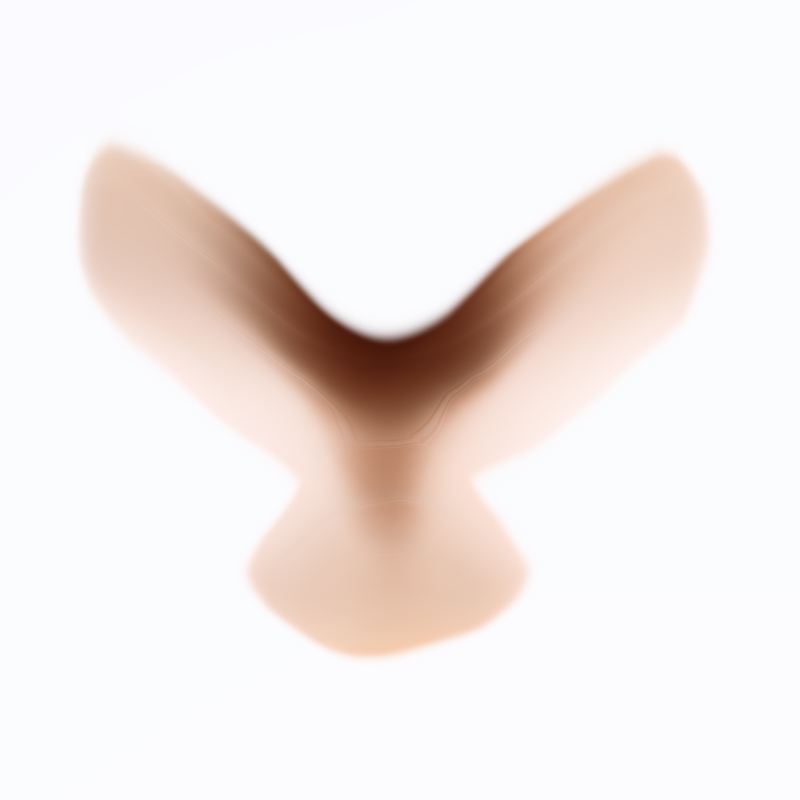}
        & \includegraphics[width=0.82\linewidth]{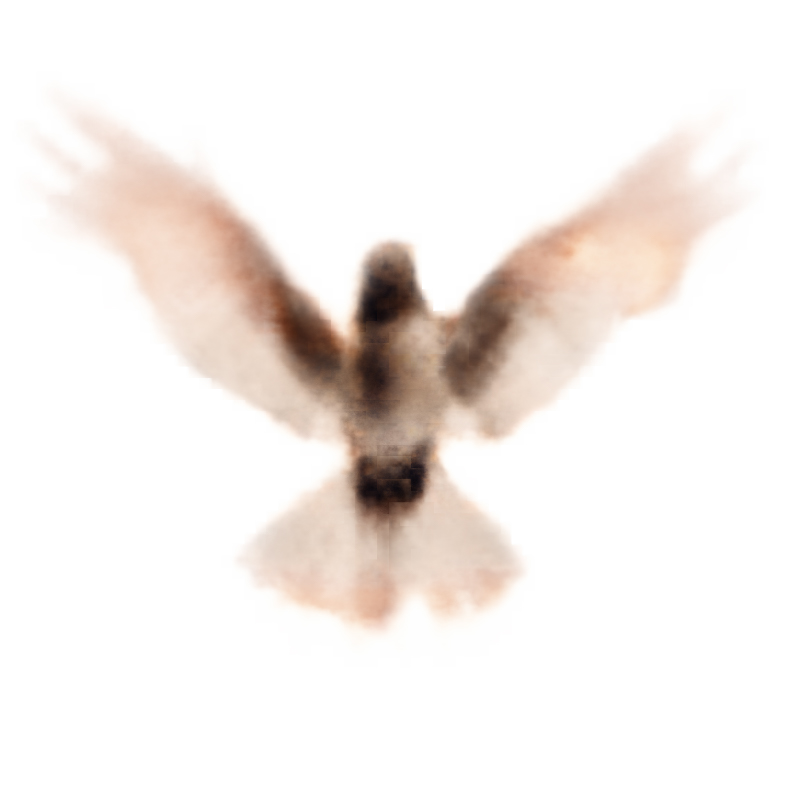}
        & \includegraphics[width=0.82\linewidth]{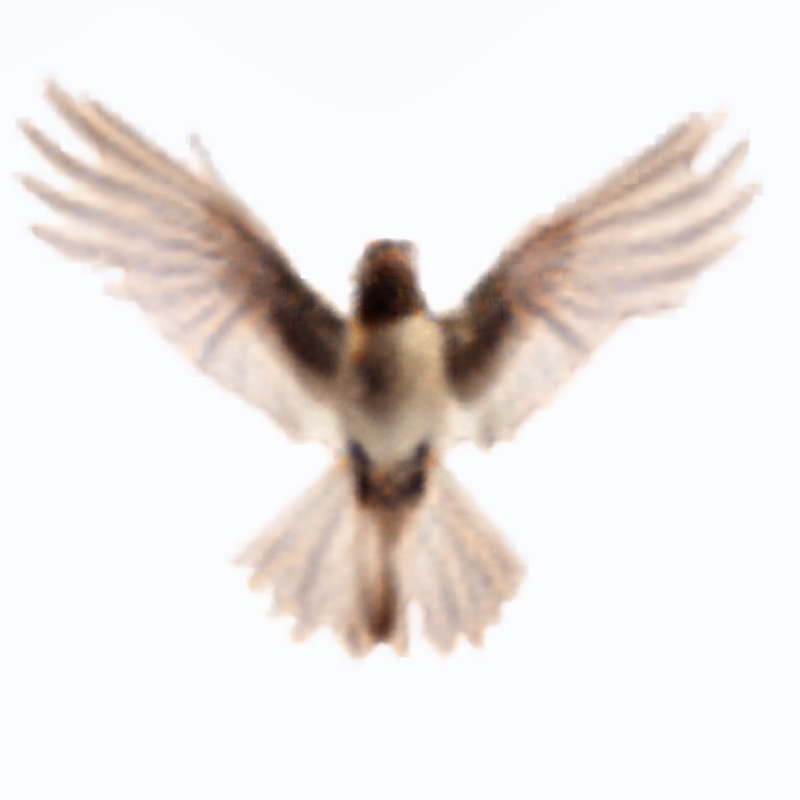}
        & \includegraphics[width=0.82\linewidth]{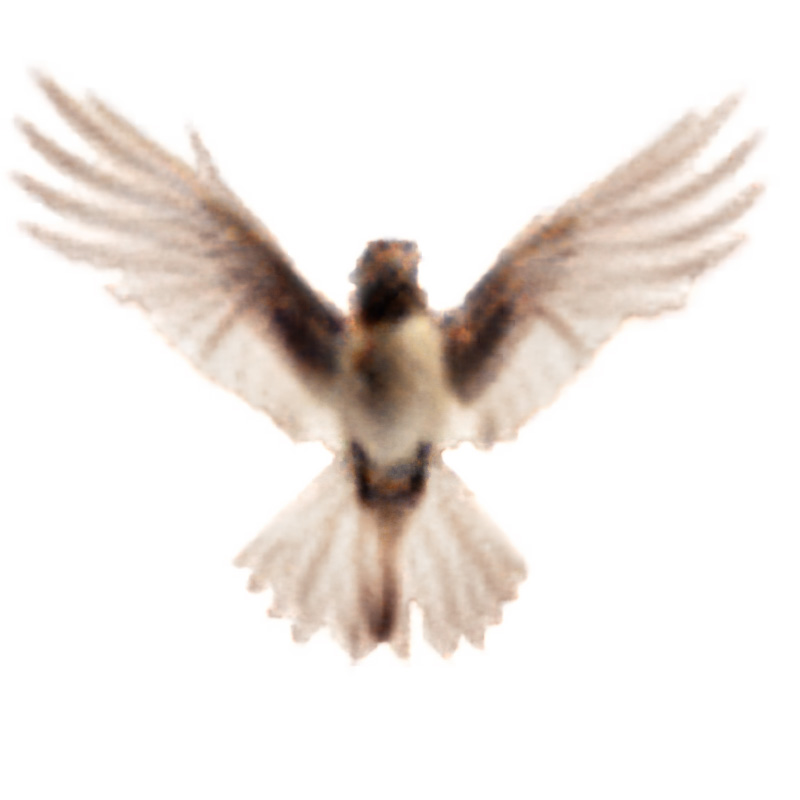}\\
         & Similarity: 89.1\% & Similarity: 71.0\% & Similarity: 83.9\% & Similarity: 84.9\% & Similarity: 88.4\%\\
         & Iterations: 6.1k & Iterations: 6.1k & Iterations: 6.1k & Iterations: 20.7k & Iterations: 18.9k\\\hline

        \vspace{0.2in}  
        \includegraphics[width=0.82\linewidth]{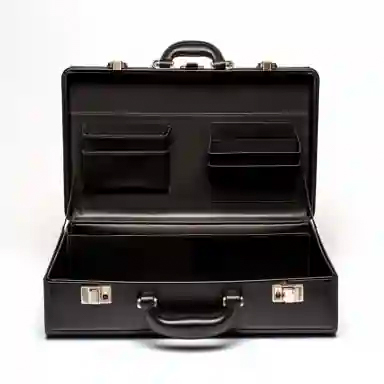}
        & \includegraphics[width=0.82\linewidth]{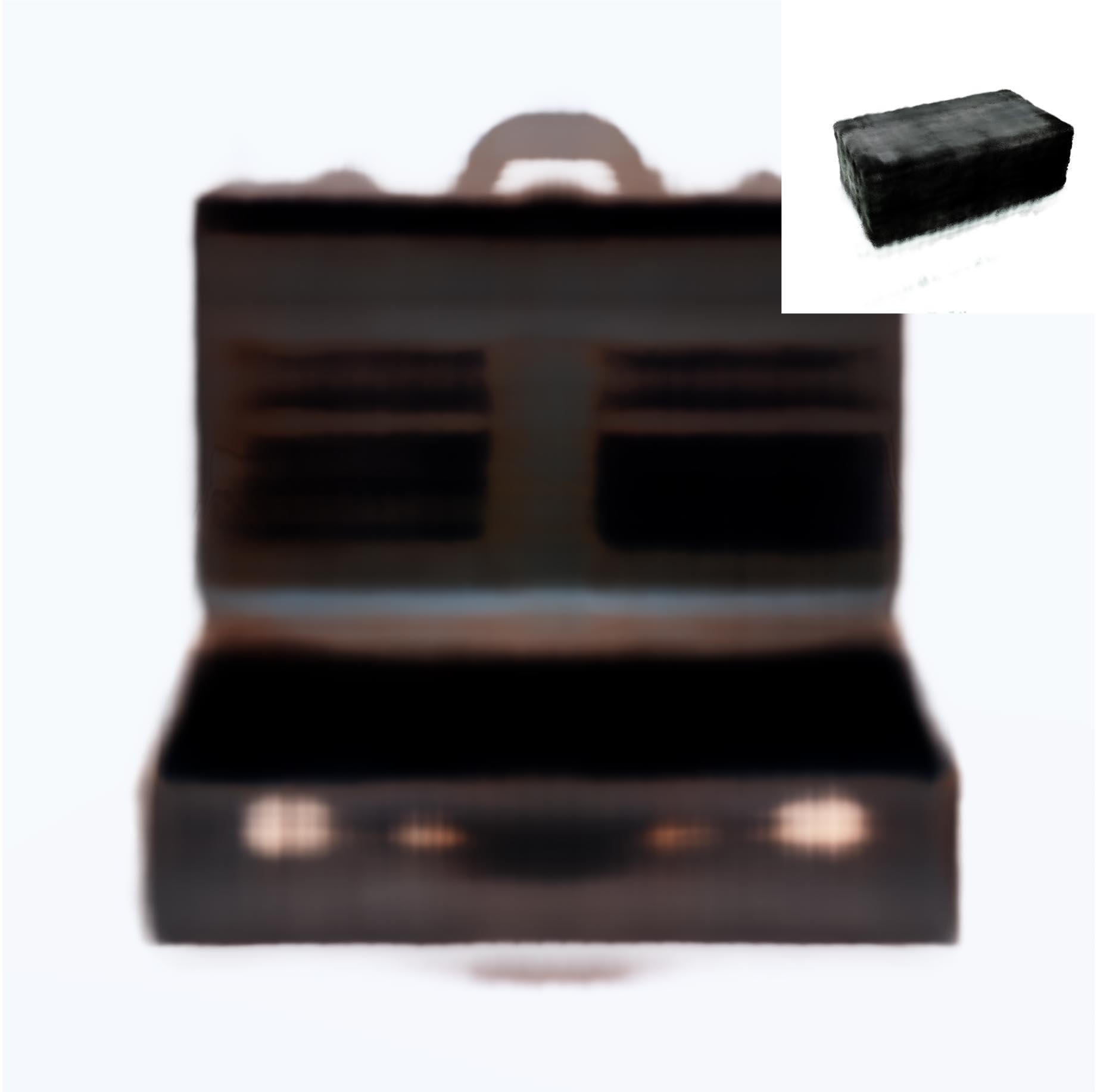}
        & \includegraphics[width=0.82\linewidth]{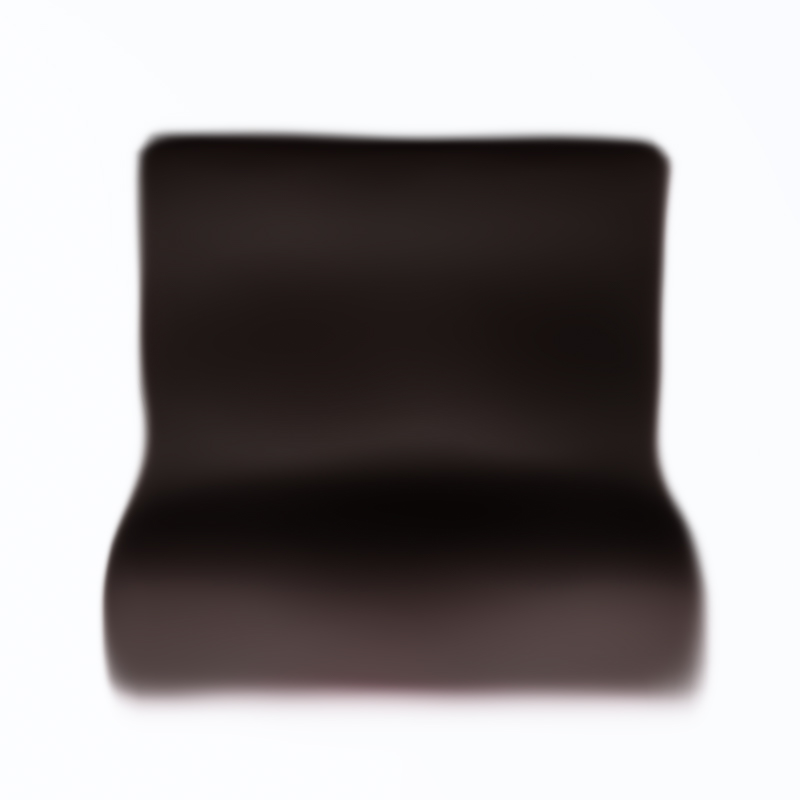}
        & \includegraphics[width=0.82\linewidth]{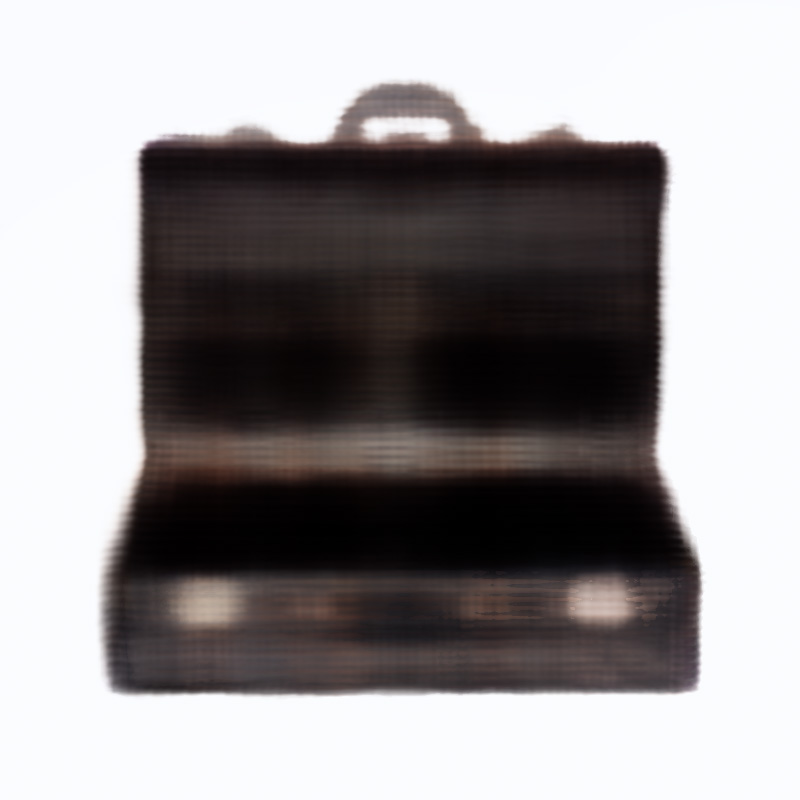}
        & \includegraphics[width=0.82\linewidth]{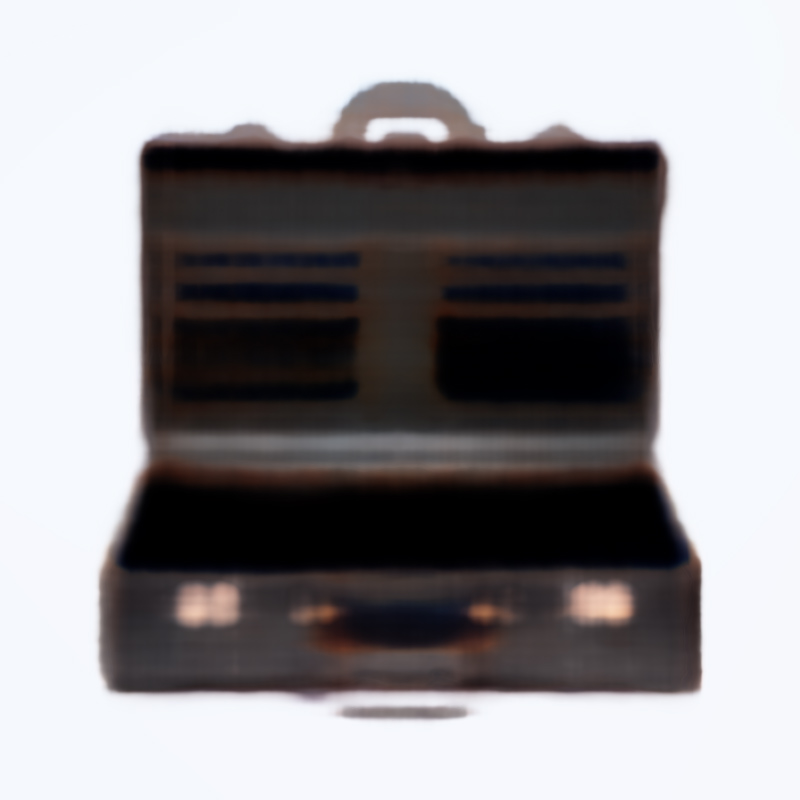}
        & \includegraphics[width=0.82\linewidth]{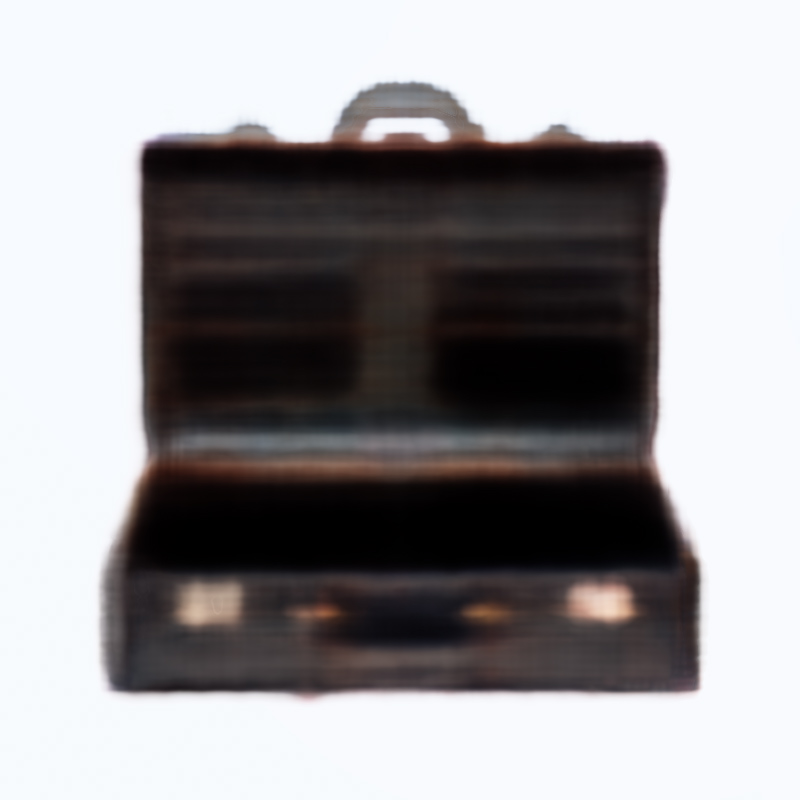}\\
         & Similarity: 80.8\% & Similarity: 73.7\% & Similarity: 79.8\% & Similarity: 78.3\% & Similarity: 78.6\%\\
         & Iterations: 7.5k & Iterations: 7.5k & Iterations: 7.5k & Iterations: 24.6k & Iterations: 22.3k\\
    \bottomrule
    \toprule
        \multirow{2}{*}{Prompt} & Ours & DreamFusion (S) & DreamFusion (P) & DreamFusion (S) & DreamFusion (P)\\
         & (converge) & (same iterations) & (same iterations) & (converge) & (converge) \\
    \midrule

        \vspace{0.25in}  
        {\it ``A brown table.''}  
        & \includegraphics[width=0.75\linewidth]{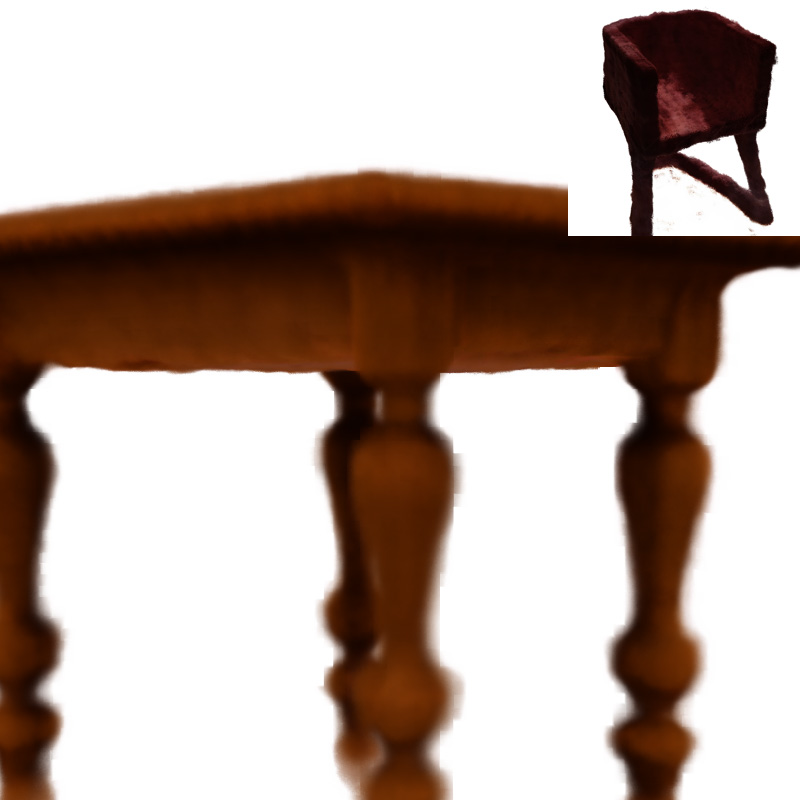}
        & \includegraphics[width=0.75\linewidth]{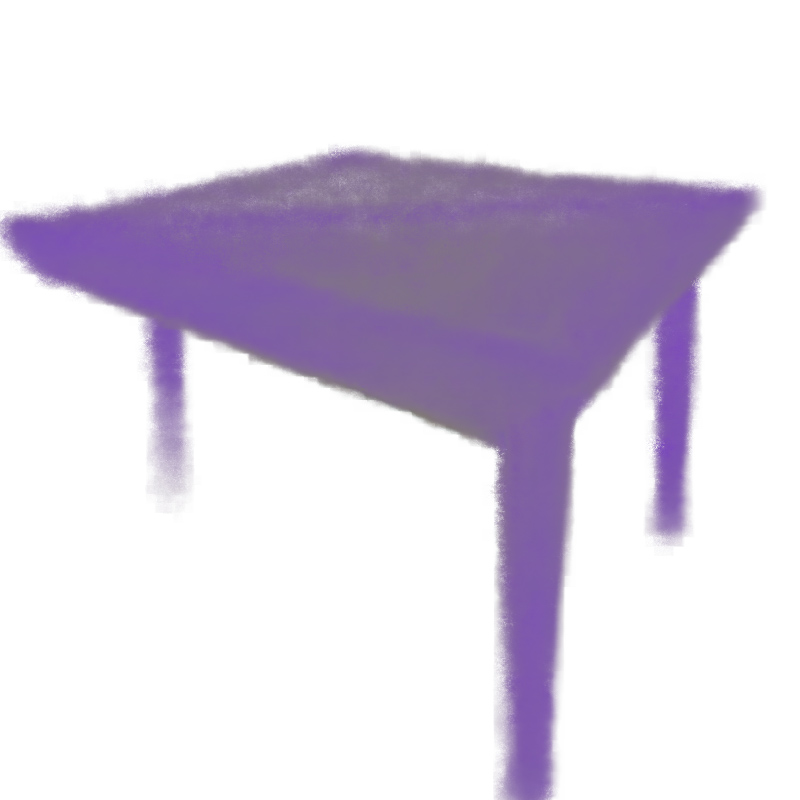}
        & \includegraphics[width=0.75\linewidth]{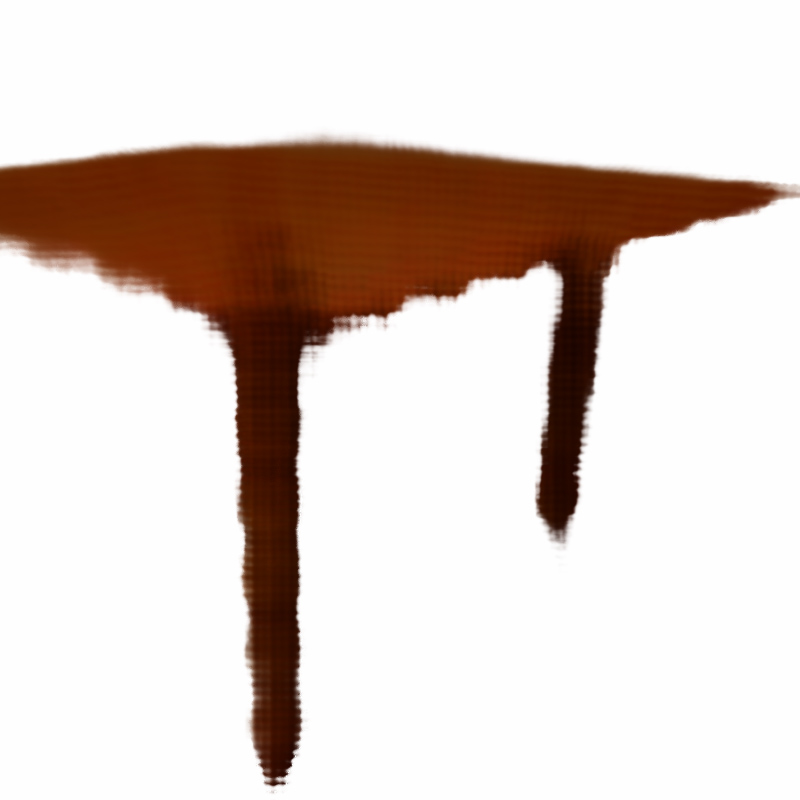}
        & \includegraphics[width=0.75\linewidth]{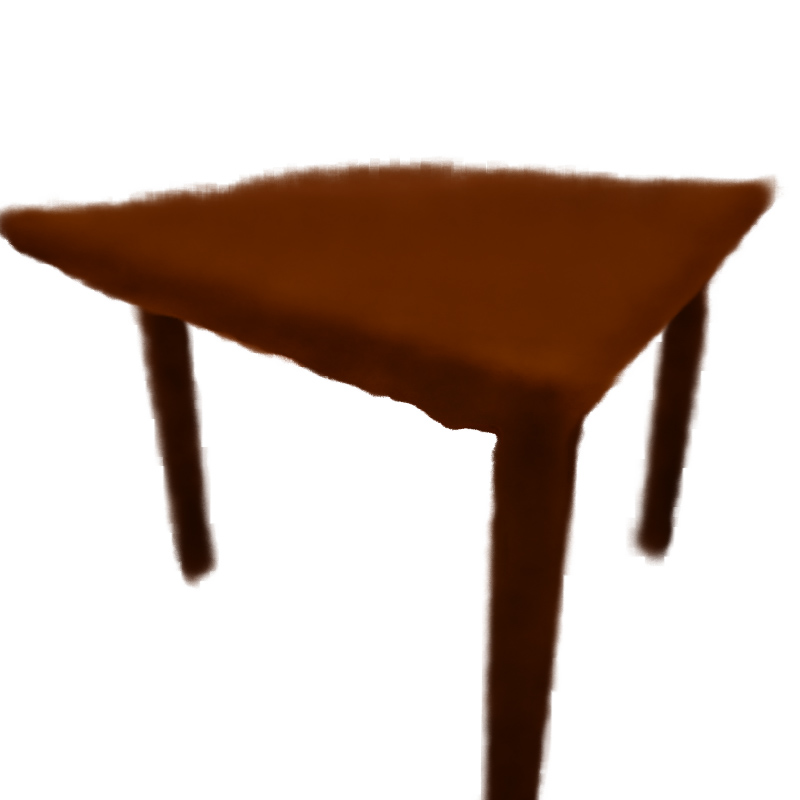}
        & \includegraphics[width=0.75\linewidth]{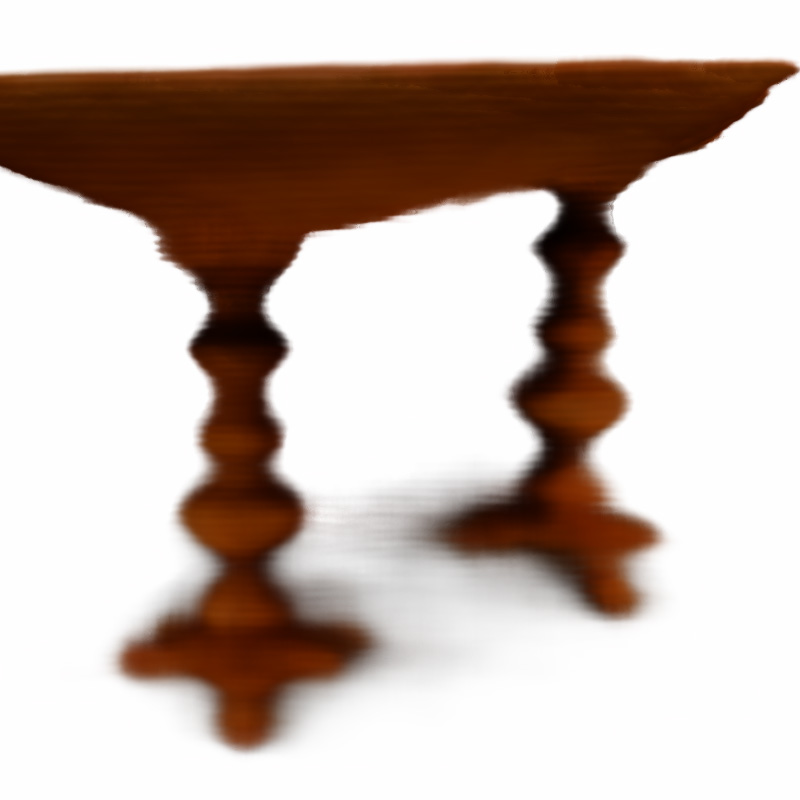}\\
         & Similarity: 32.3\% & Similarity: 28.1\% & Similarity: 32.3\% & Similarity: 30.9\% & Similarity: 33.3\%\\
         & Iterations: 5.5k & Iterations: 5.5k & Iterations: 5.5k & Iterations: 24k & Iterations: 20.5k\\\hline

        \vspace{0.25in}  
        {\it ``A round fountain.''}  
        & \vspace{0.05in} \includegraphics[width=0.72\linewidth]{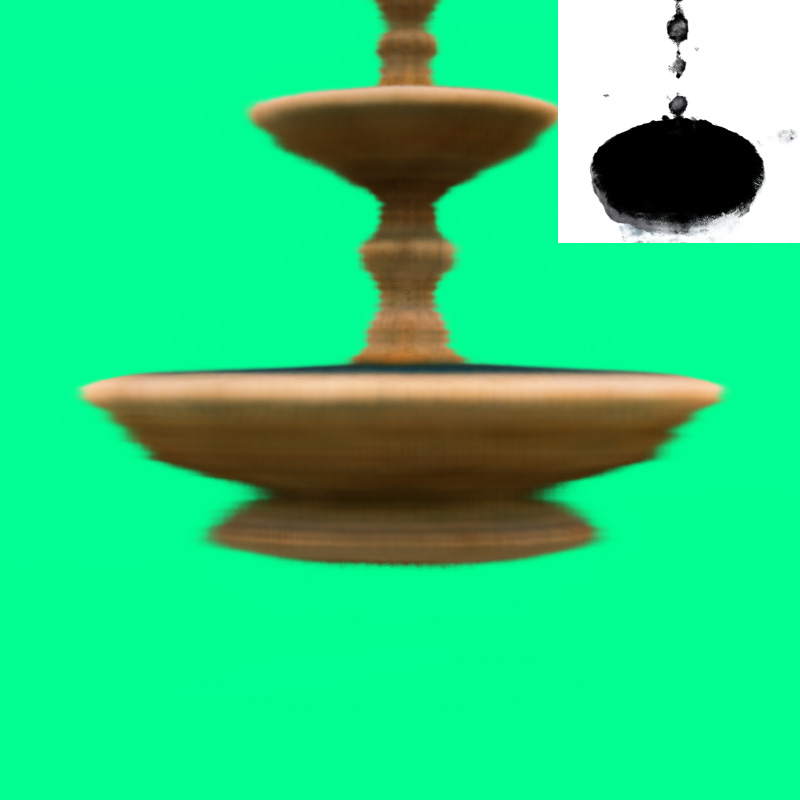}
        & \includegraphics[width=0.72\linewidth]{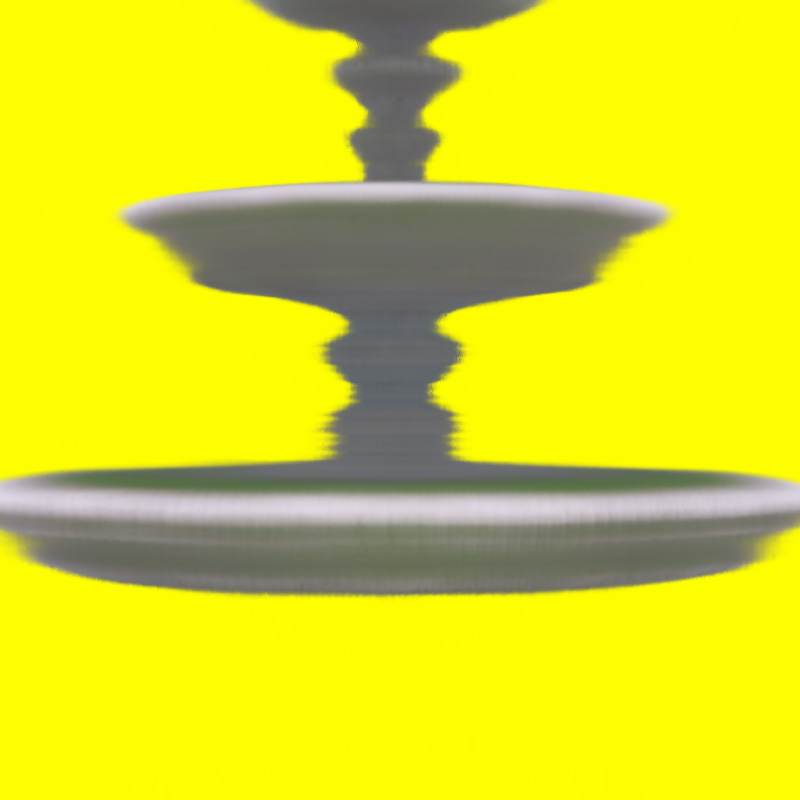}
        & \includegraphics[width=0.72\linewidth]{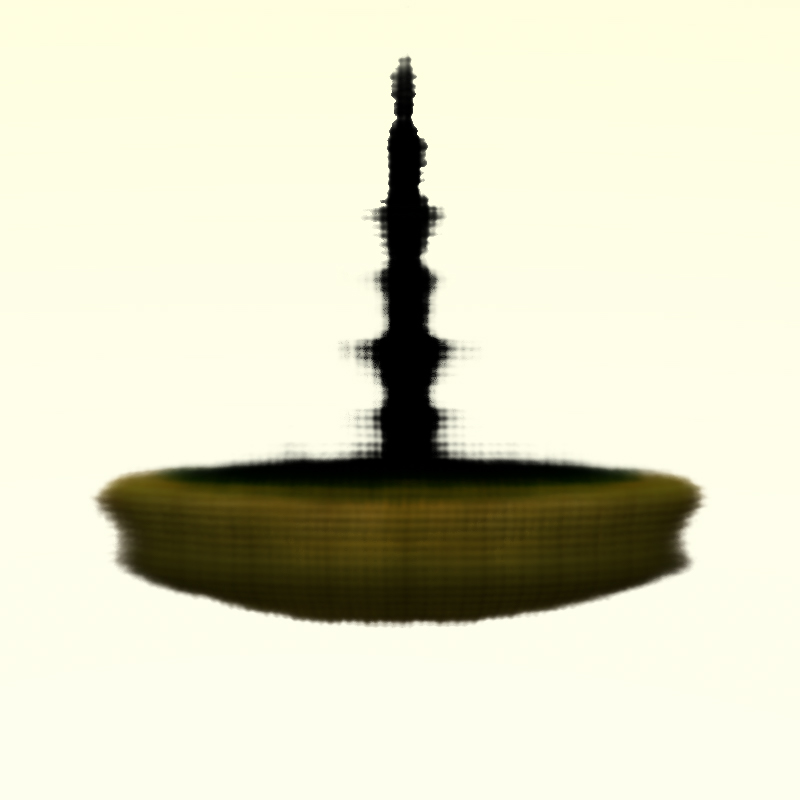}
        & \includegraphics[width=0.72\linewidth]{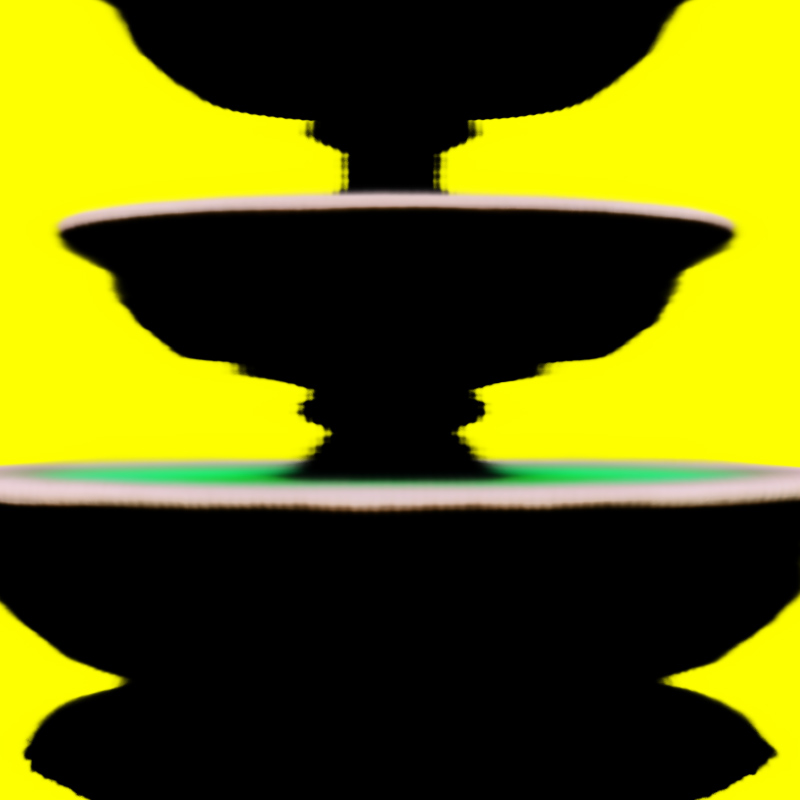}
        & \includegraphics[width=0.72\linewidth]{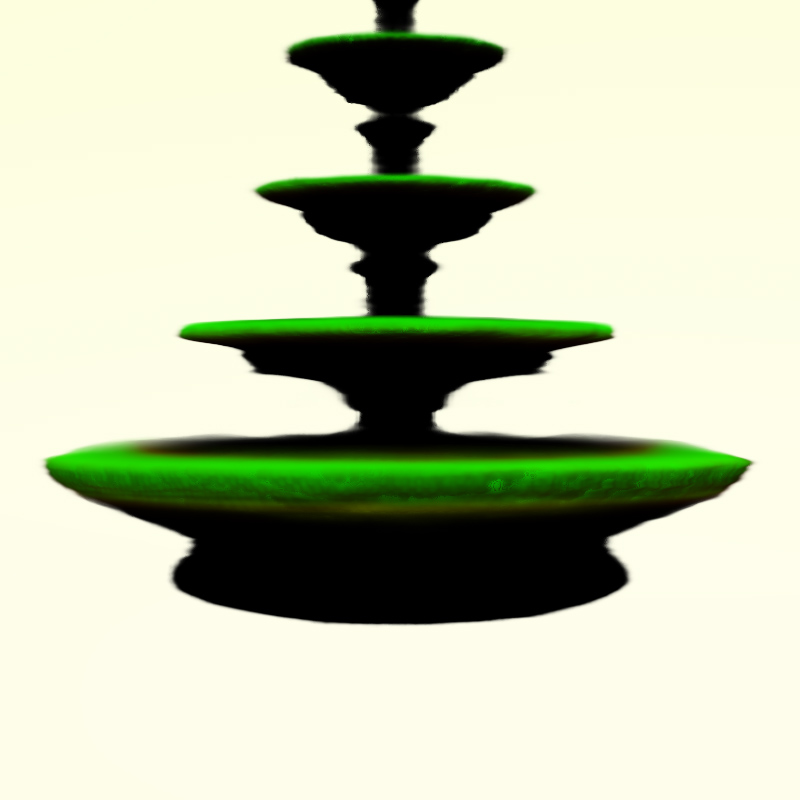}\\
         & Similarity: 31.6\% & Similarity: 29.6\% & Similarity: 26.9\% & Similarity: 30.3\% & Similarity: 30.6\%\\
         & Iterations: 7k & Iterations: 7k & Iterations: 7k & Iterations: 32k & Iterations: 30k\\
    \bottomrule
    \end{tabular}
    \caption{Comparisons of {\bf out-of-distribution inference result, using either image or text prompts}. For each prompt, we record the convergence iteration count using our initialization zero-shot, from one single forward pass (top-right inset). Scratch (S) or pre-trained (P) baselines (Zero-1-to-3 for image-to-NeRF tasks, and DreamFusion for text-to-NeRF tasks) are trained for the same number of iterations and until convergence, and their results are displayed.
    Our method significantly accelerates the baseline optimization, achieving a 3 to 5 times speed boost while almost consistently yielding better results. Moreover, our method provides a semantically meaningful initialization, as pretrained (P) baselines still take an extremely long time to converge though initialized with a similar scene.
    Due to space limit, more results can be found in supplementary materials.}
    \label{tab:out-dist-text-prompt-compare}
    \vspace{-0.2in}
\end{table*}



\noindent {\bf Generation from Image Prompts}\quad First, we use images on the test split as prompts to verify the in-distribution generation ability of our model. \Cref{fig:testset11by5_5} provides an overview of inference results of our model on test split, with 5 scenes randomly selected from each category. We then examine both qualitative and quantitative results in details as follows.



For qualitative results, we randomly sampled one scene from randomly chosen 4 categories (all cases are tested): chair, bathtub, plane, and guitar, and compared the results of our model with the performance of Zero-1-to-3~\cite{liu2023zero1to3}. To ensure fairness, we allow Zero-1-to-3 to train until convergence, eliminating the disadvantage of lack of prior knowledge of the distribution of our dataset. From the comparison in \cref{fig:indomain-image-compare}, our model demonstrates a strong ability to generate in-distribution NeRF scenes from image prompts, producing much clearer and more realistic results compared to Zero-1-to-3 even with one single forward pass. Besides, it can be seen that Zero-1-to-3 sometimes produces view-inconsistent results. For example, the plane it generated lacks height when viewed from one side, and their generated guitar exhibits inconsistent shapes when viewed from the frontal and side angles. 





To further evaluate performance quantitatively, we randomly selected 40 images\footnote{10\% from our entire test split; we cannot afford too many since Zero-1-to-3 takes a long time to optimize until convergence and sometimes even fails, as mentioned in~\cref{sec:exp-baselines}.} from our test split as prompts for both our method and Zero-1-to-3, then compute the cosine similarity and FID score by comparing with ground-truth images from different views.
To provide more references, we also compute these two scores on the entire test and train split using our method, where the train split result can serve as an upper-bound performance. \Cref{tab:exp1_image_quant} shows the comparison, illustrating that the performance of our model on the test split does not drastically decrease compared to its counterpart on the train split. Also, our model has higher similarity and lower FID scores compared with Zero-1-to-3, demonstrating our model's strong capability in generating view-consistent NeRFs.


\noindent {\bf Generation from Text Prompts}\quad Next, we present the generation results using text prompts. To restrict the evaluation to an in-distribution scope, all texts used in this section are designed to describe objects from the 8 categories in our dataset. \Cref{tab:in-dist-text-prompt-compare} compares the cosine similarity of the text prompt and generated images, as well as the inference time of our results with DreamFusion's, where we let DreamFusion train for 5k iterations and until convergence.  We can see that our method is significantly faster than DreamFusion since only one single forward pass is needed, while at the same time being able to generate scenes with almost always better quality (higher similarity) compared to DreamFusion.


\subsection{Results on Out-of-Distribution Generation}
\label{sec:results-outdomain}

As real-world prompts feature unseen geometries and attributes during the training phase, we anticipate that generating high-quality out-of-distribution 3D scenes without optimization would be impractical in our pre-trained settings. Instead, we discovered that the NeRFs generated by our model serve as good initializations for prior works that require per-prompt optimization to speed up their optimization time. In this section, we will show that with our initialization, the inference process of DreamFusion (for text-to-NeRF tasks) and the 3D generation process of Zero-1-to-3 (for image-to-NeRF tasks) will converge faster while almost always yielding better quality. Here we also provide the demo results of inferencing from text and image prompts similar to \cref{sec:results-indomain}, but they are now out of distribution. Thus, in both cases, we find the semantically nearest scene with the given prompt in our training set. Other initialization methods will be explored in~\cref{sec:ablation-how-to-init}.

In this out-of-distribution experiment, we selected image and text prompts that describe scenes whose categories were not present in our dataset. Qualitative results are displayed through generated demo images, while quantitatively, we calculate the cosine similarity of the generated image views with the given prompt. Moreover, we report the optimization time of baselines with/without our initialization to highlight the remarkable inference acceleration provided by our initialization for both text and image prompts.

\Cref{tab:out-dist-text-prompt-compare} shows the comparison of our method with baselines, using two image prompts and two text prompts\footnote{Due to space limit, more results can be found in supplementary materials.}. We first let our model directly generate a NeRF from the given prompt in a zero-shot manner, by inferencing from the semantically nearest scene in our train split, then use the generated NeRF as an initialization to further optimize DreamFusion or Zero-1-to-3 models with the prompt until converge. The top-right inset images in the first column depict the zero-shot generation result of our model. 

For comparison, we evaluate two kinds of baselines in our study: 1) Baseline (S), which is the baseline model trained from {\bf s}cratch with the given prompt, and 2) Baseline (P), which is the baseline model first {\bf p}re-trained to a semantically close scene with our initialization, and then optimized with the given prompt.

Our results show that ourwork~converges 3 to 5 times faster than baselines on text and image prompts. Moreover, when optimized for the same iterations, as demonstrated in columns 2 and 3, we observe that both baselines only learn geometry features, while our method already learns fine textural details.
By comparing the results of our method with baseline (P), we conclude that even though they are pre-trained into semantically close scenes, our method still converges much faster than the baseline (P), thus eliminating the possibility that the acceleration by our method is merely due to the reason that a NeRF initialization will yield faster convergence compared to a random initialization, while at the same time, demonstrating that our method has learned a semantically meaningful alignment, thereby providing a closer initialization compared to baseline (P)\footnote{We include further evidence showing the semantics our model learns in supplementary materials by performing some interpolation experiments.}.

\begin{figure}[h]
\vspace{-0.15in}
    \centering
\includegraphics[width=0.45\textwidth]{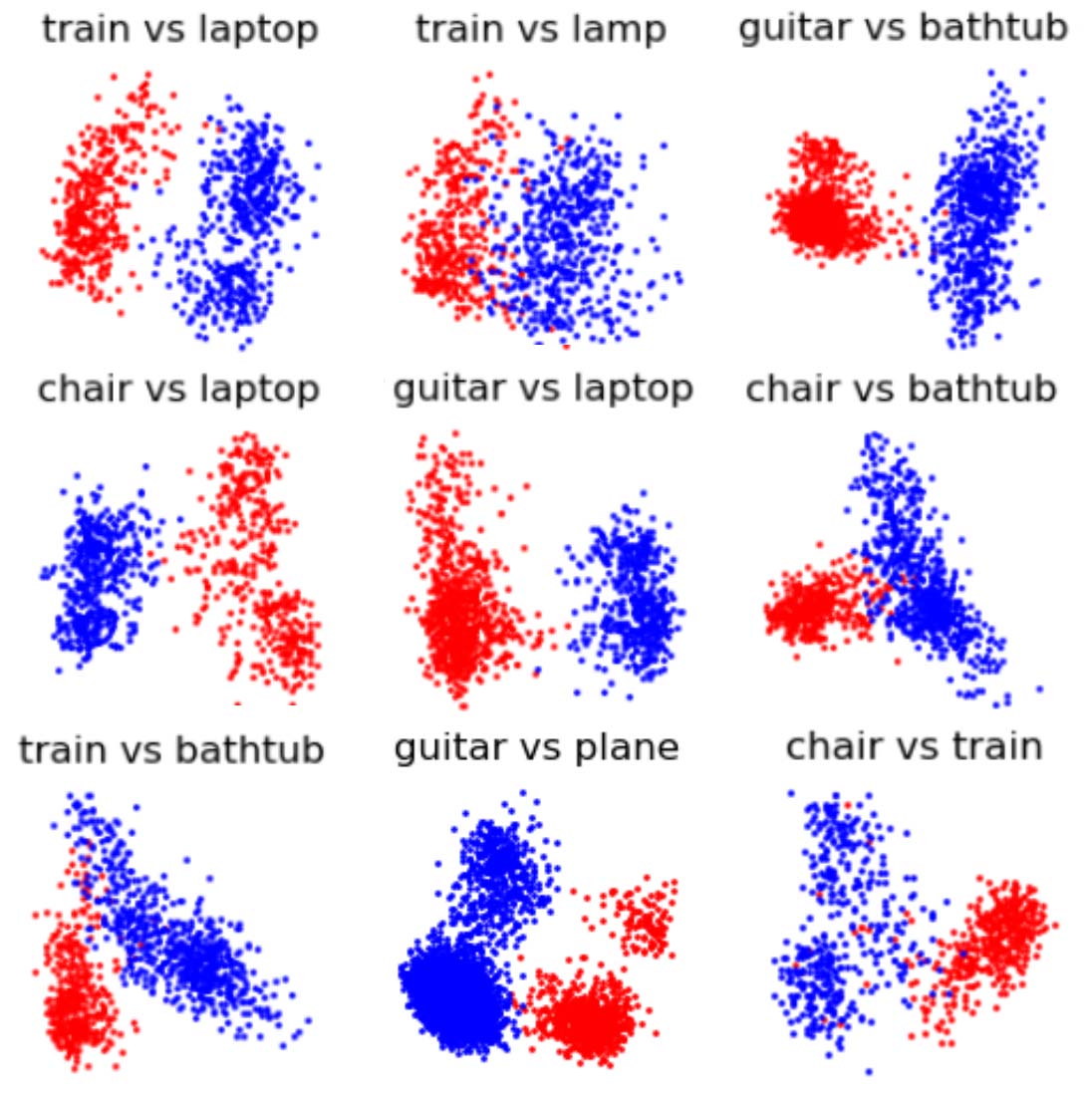}
    \vspace{-0.1in}
    \caption{Sample pairwise PCA of our trained NeRF parameters. The NeRF parameters are clearly separated among different classes, and clustered within the same class.}
    \label{pca_weights}
    \vspace{-0.1in}
\end{figure}

\section{Ablation Study}
\label{sec:ablation}

\subsection{Good Topology of NeRF Parameter Space}
\label{sec:ablation-nerf}
This section verifies that using the same (our) initialization while training NeRFs leads to better topological structures.
\Cref{pca_weights} visualizes the pairwise Principal Component Analysis (PCA) result of the trained NeRF parameters, where the parameters within each class are clustered with different centers. The intra-class affinity and inter-class separability, manifested as the blue and red clusters, are clearly visible. 

\begin{figure}[ht]
    \centering
    \includegraphics[width=0.49\linewidth]{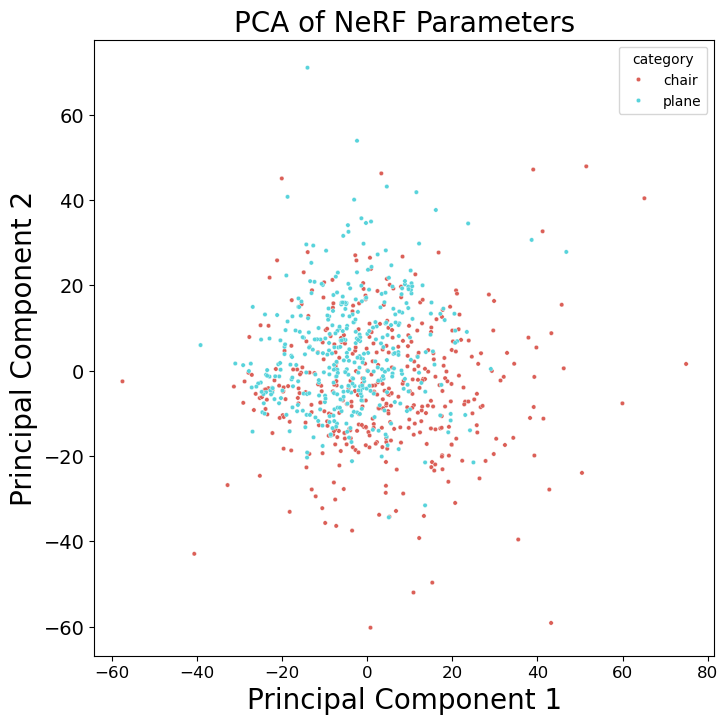} \includegraphics[width=0.49\linewidth]{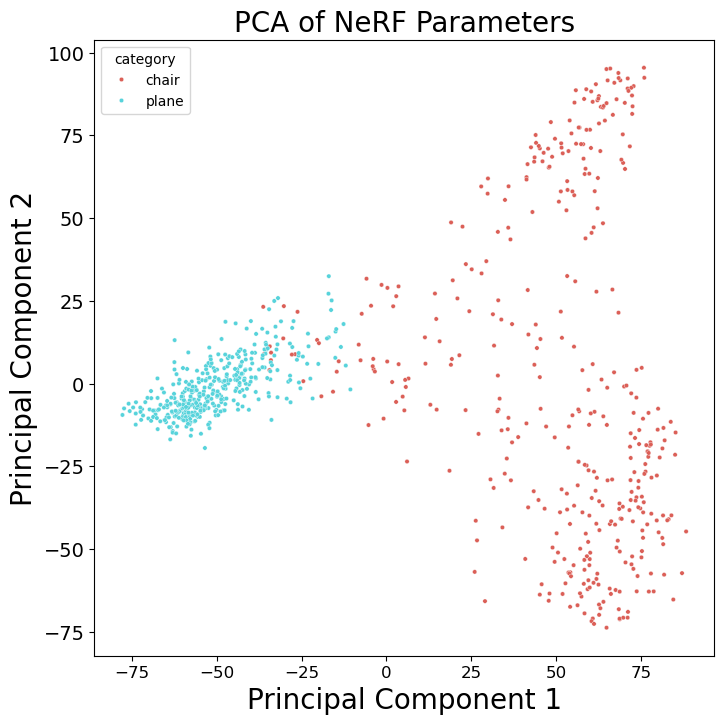}
    \caption{The PCA plots of plane and chair NeRF parameters with (right) and without (left) using the same initialization, where the left plot fails to separate the NeRF parameters for two categories.}
    \label{fig:chair-plane-pac}
    \vspace{-0.2in}
\end{figure}

Instead, if NeRFs are trained with random initializations, such property will not occur. Take the chair and plane's PCA plot as an example, Figure~\ref{fig:chair-plane-pac} illustrates that in the left plot (with random initialization), two classes are mixed together, while in the right plot (with the same initialization), the NeRF weights are separated apart for the two categories.
These trained parameters can thus result in a good manifold for our pretrained implicit latent space for better interpolation within a given class as well as latent alignment in later steps.

\subsection{How to initialize Dreamfusion}
\label{sec:ablation-how-to-init}
To determine the most effective initialization method for our out-of-distribution generation experiment that accelerates Dreamfusion~\cite{poole2022dreamfusion} or Zero-1-to-3~\cite{liu2023zero1to3}, we explore two options: 1) Ours (NRS): use the NeRF inferenced from the semantically {\bf n}ea{\bf r}e{\bf s}t image on our dataset to the prompts, which is extensively used in~\cref{sec:experiments} and 2) Ours (DI): use the NeRF {\bf di}rectly generated from the prompts. We evaluate the performance of both methods on text-to-NeRF tasks in terms of visual quality, cosine similarity between generated images and text prompts, as well as convergence speed. Using the per-prompt optimization of DreamFusion from scratch as a baseline, this comparison will allow us to determine which initialization method yields optimal results.

\begin{table}[ht]
    \newcommand{\dummyImg}{\fbox{\rule{0pt}{0.9in} \rule{0.65\linewidth}{0pt}} }
    \centering
    \begin{tabular}{>{\centering\arraybackslash}m{0.13\linewidth}|>{\centering\arraybackslash}m{0.22\linewidth}|>{\centering\arraybackslash}m{0.22\linewidth}|>{\centering\arraybackslash}m{0.22\linewidth}}
    \toprule
        Text Prompt & Ours (NRS) & Ours (DI) & DreamFusion (Scratch)\\
    \midrule
        \vspace{0.25in}  
        {\it ``Iron man on a blue chair''}  
        & \includegraphics[width=\linewidth]{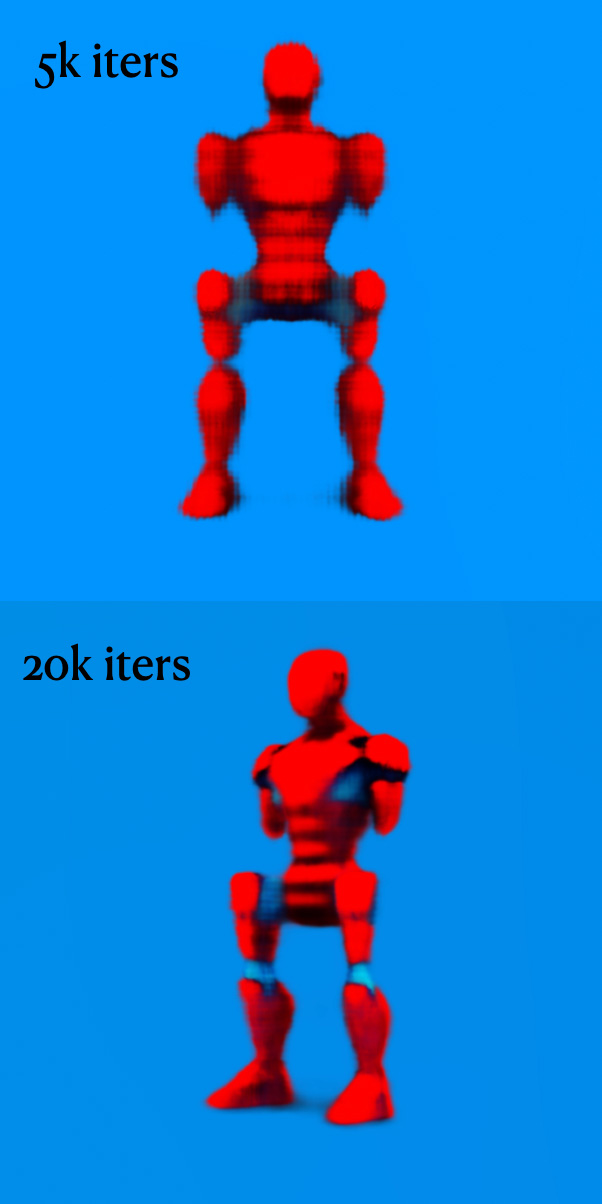}
        & \includegraphics[width=\linewidth]{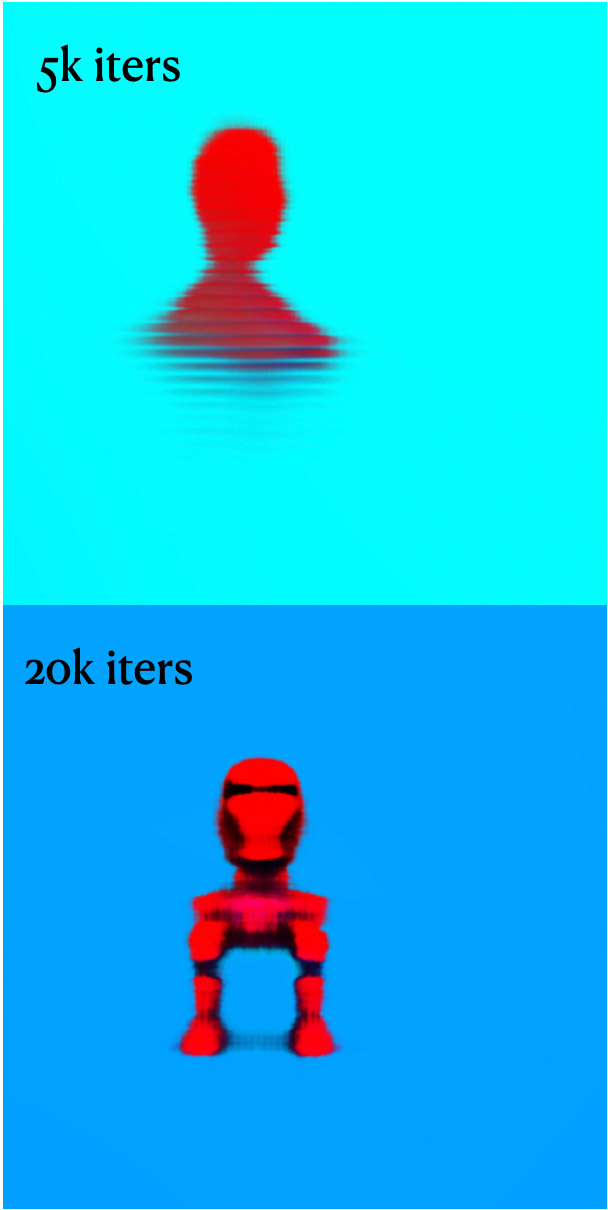}
        & \includegraphics[width=\linewidth]{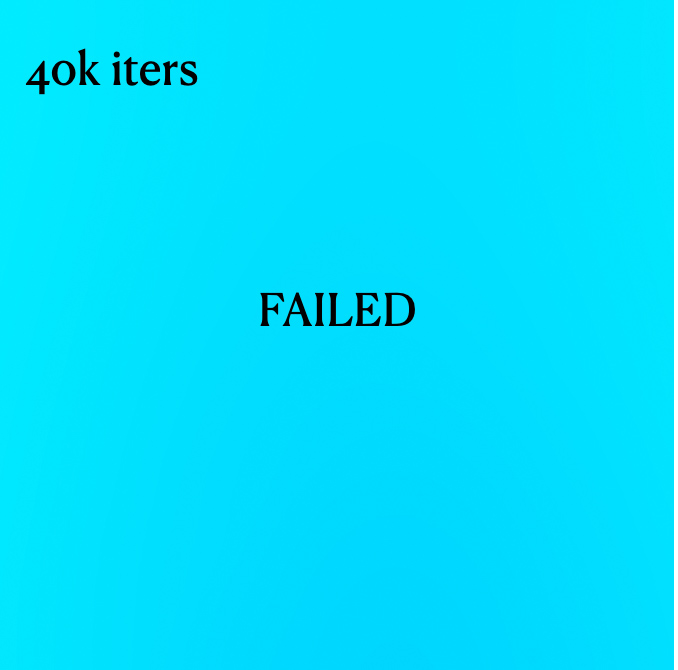}\\
          & Similarity: 31.74\% & Similarity: 29.44\% & Similarity: Failed\\
         \hline
        \vspace{0.25in}  
        {\it ``A burning torch''}  
        & \includegraphics[width=0.9\linewidth]{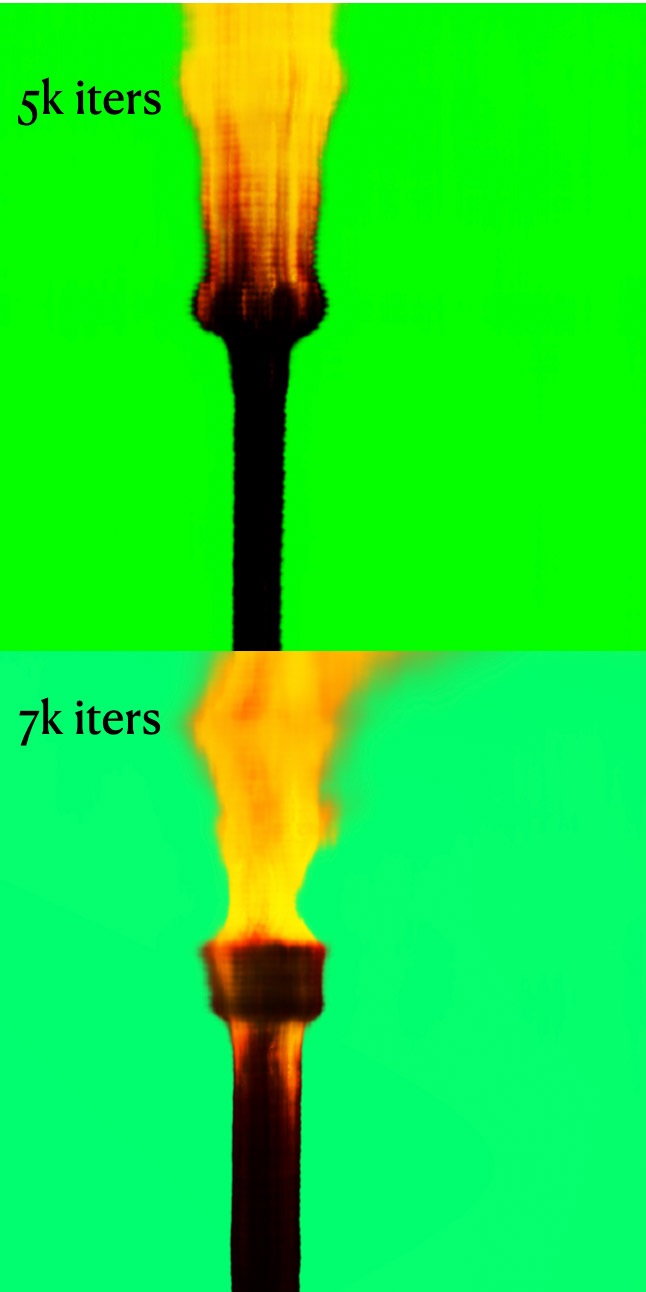}
        & \includegraphics[width=0.9\linewidth]{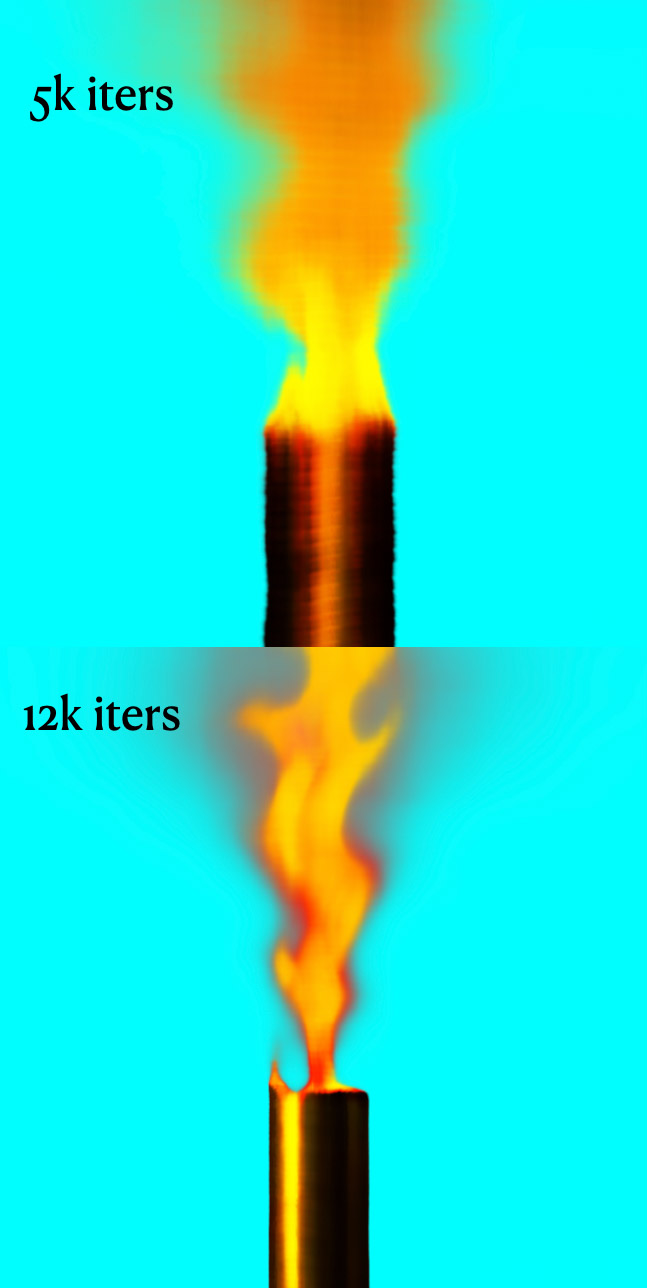}
        & \includegraphics[width=0.9\linewidth]{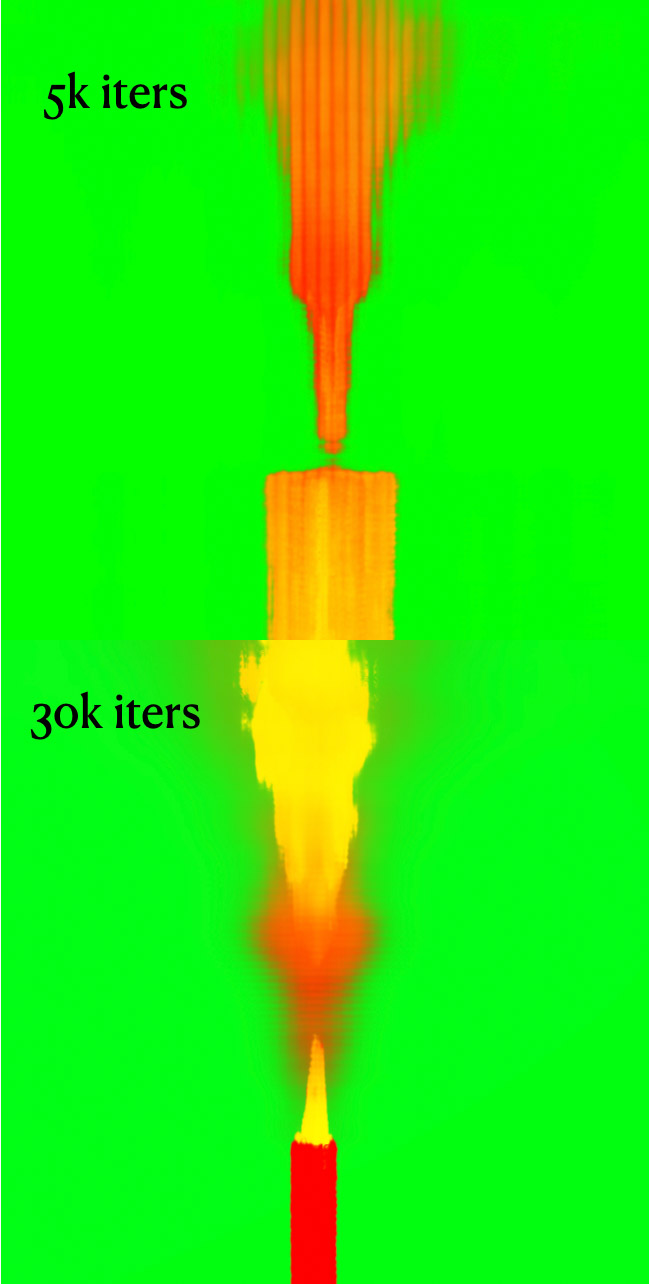}\\
          & Similarity: 31.47\% & Similarity: 31.01\% & Similarity: 32.20\% \\
         
    \bottomrule
    \end{tabular}
    \caption{This experiment qualitatively demonstrates two results of comparison on  DreamFusion trained from scratch (baseline) against training with our two {\bf initialization} methods. The top images demonstrate the results after 5k iterations; the bottom are trained either to converge or capped at maximum iterations (20k for our methods, 40k for baseline). We can see that our method (NRS)  
    produces good results in only 5k iterations, and converges a lot earlier than other methods.}
    \label{tab:ablation-study-for-init}
    \vspace{-0.2in}
\end{table}

We designed 10 prompt texts that include both some extensions of our dataset categories, such as ``Ironman on a blue chair'', and some entirely out-of-distribution categories, such as ``Thor's hammer'' and ``A burning torch''. For example, in~\cref{tab:ablation-study-for-init}, using the Ironman prompt, we can see that DreamFusion initialized with NeRFs generated using the semantically closest image results in earlier convergence (similar quality at 5k iterations to direct initialization at 20k iterations) and higher semantic similarity, than with the text prompt than the result initialized from NeRFs generated directly from texts. However, both methods are much faster and produce more satisfying results than optimizing DreamFusion from scratch (last column of the figure), which fails to converge even trained to 40k iterations with multiple tries. More results will be presented in the supplementary materials.

\section{Conclusion}

Concurrent to contemporary works investigating implicit NeRF generation indicating its potential high impact, 
this experimental paper explores alternatives to conventional NeRF
generation where typical input consists of multiple 2D images, as a springboard
transitioning into promptable NeRF generation (e.g., text prompt or single image prompt) for {\em direct} conditioning and {\em fast} generation of NeRF parameters for the underlying 3D scenes, tapping into promptable generative NeRF model. 
From our experiments, we hope~\ourwork~will ignite future works in faster 3D scene generation and easy-to-use 3D content creation. 

{
    \small
    \bibliographystyle{ieeenat_fullname}
    \bibliography{main}
}

\setcounter{section}{0}
\renewcommand{\thesection}{\Roman{section}}
\clearpage
\setcounter{page}{1}
\maketitlesupplementary

\section{Rotation-invariant Mapping}
\label{sec:supp-rotation-invariant}

As mentioned in \cref{sec:method-pipeline}, 
here, we demonstrate that by randomly selecting images from different viewpoints during the semantic alignment training, our model can learn a rotation-invariant mapping, i.e., from a general semantic representation invariant to viewpoint and rotation to the implicit NeRF ~\cite{mildenhall2021nerf} parameter space.

\begin{figure*}[ht]
    \centering
    \includegraphics[width=0.12\textwidth]{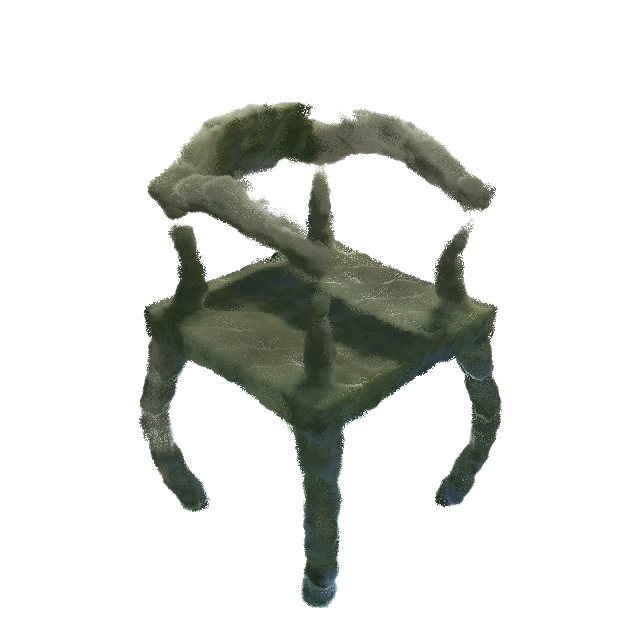}
    \includegraphics[width=0.12\textwidth]{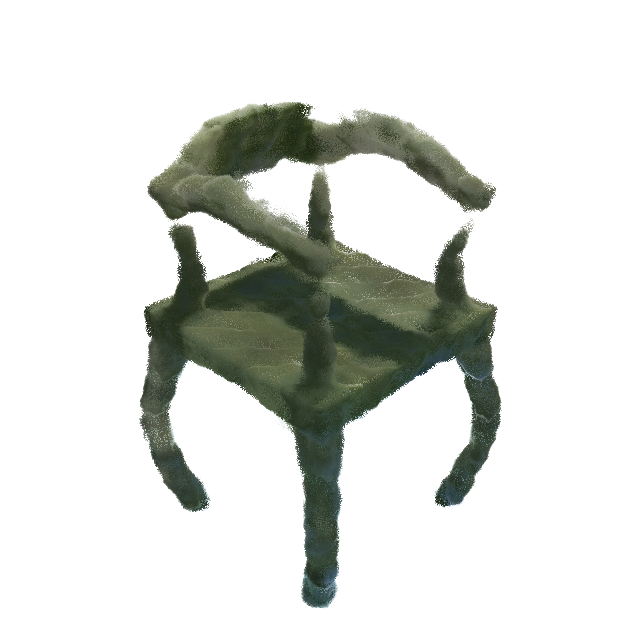}
    \includegraphics[width=0.12\textwidth]{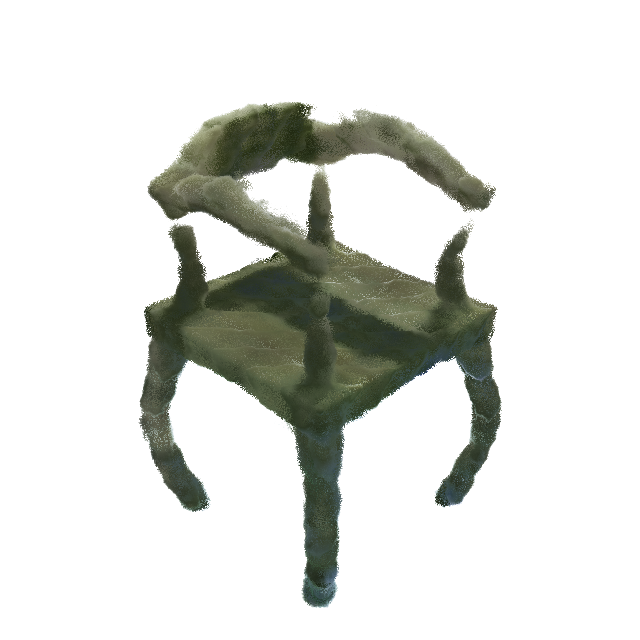}
    \includegraphics[width=0.12\textwidth]{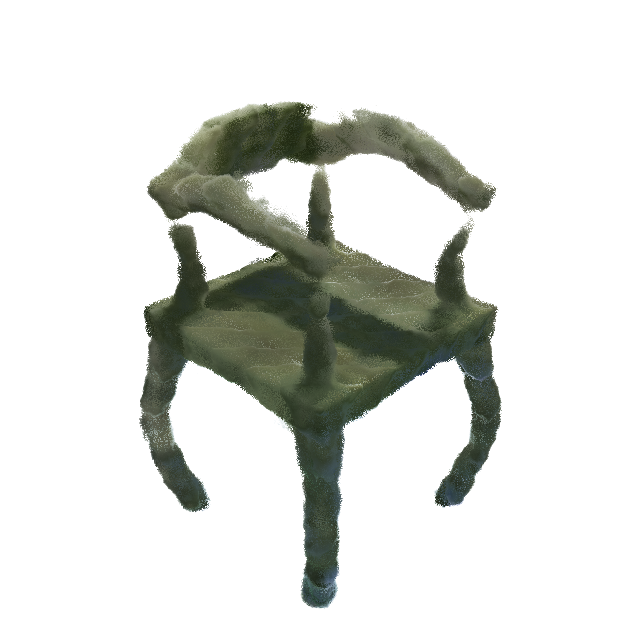}
    \includegraphics[width=0.12\textwidth]{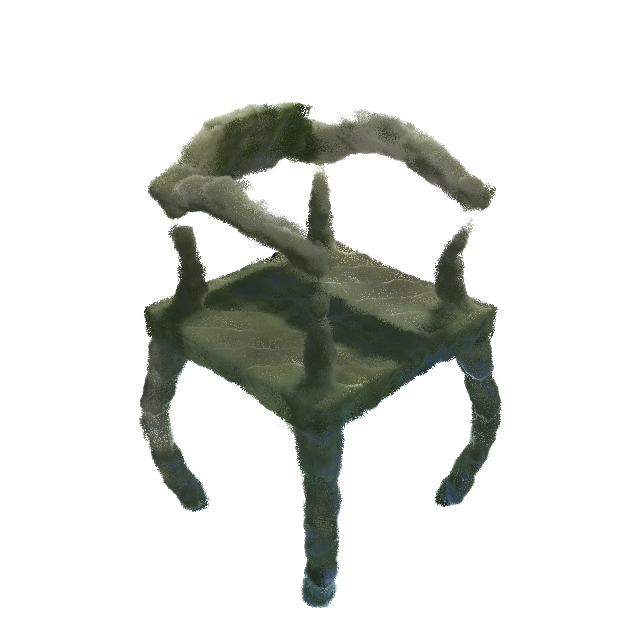}
    \includegraphics[width=0.12\textwidth]{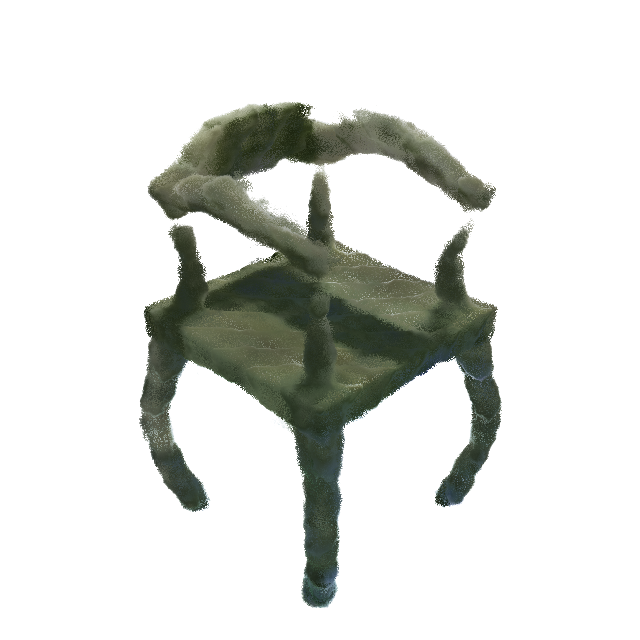}
    \includegraphics[width=0.12\textwidth]{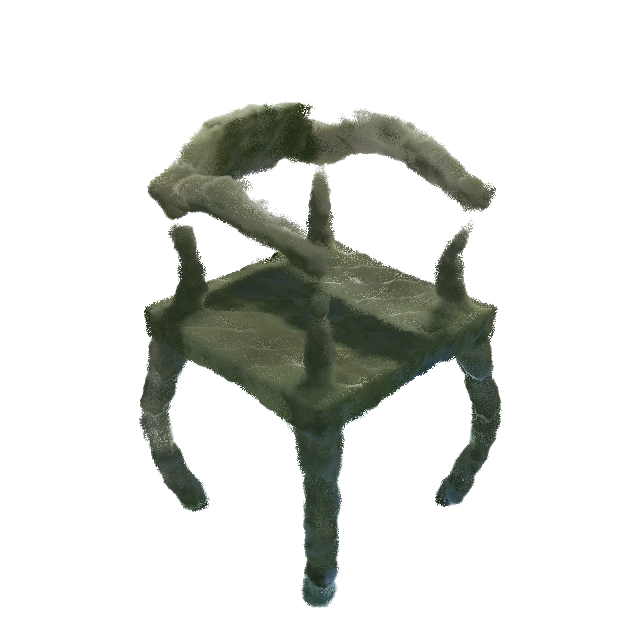}
    \includegraphics[width=0.12\textwidth]{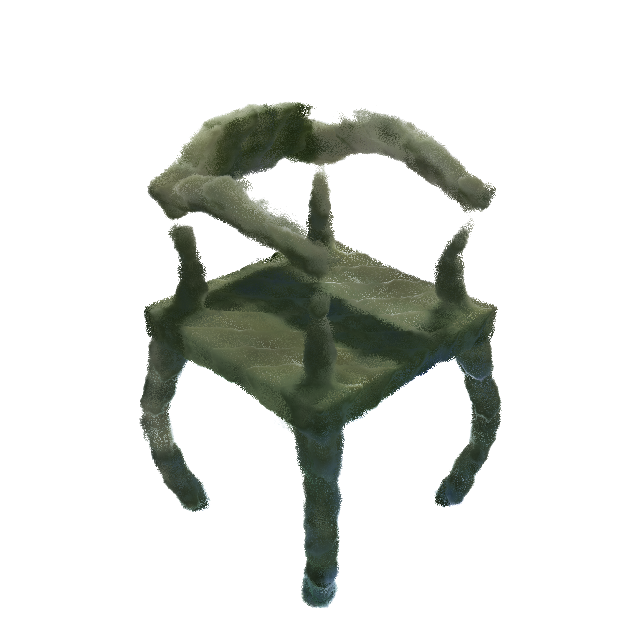}\\
    \includegraphics[width=0.12\textwidth]{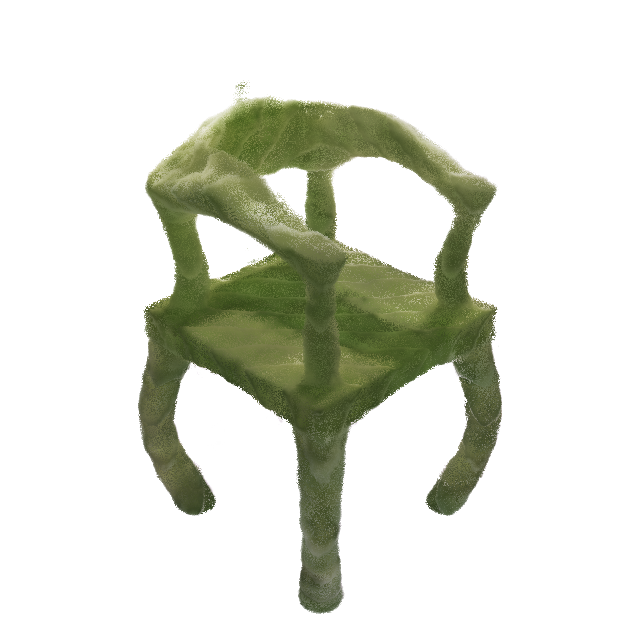}
    \includegraphics[width=0.12\textwidth]{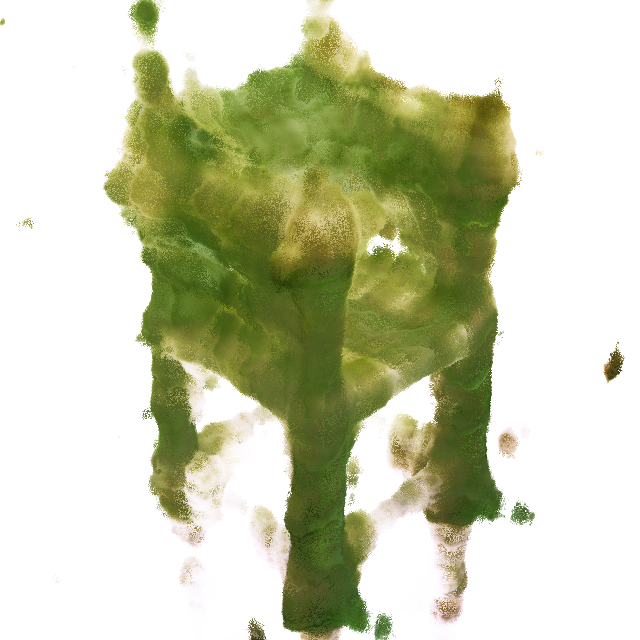}
    \includegraphics[width=0.12\textwidth]{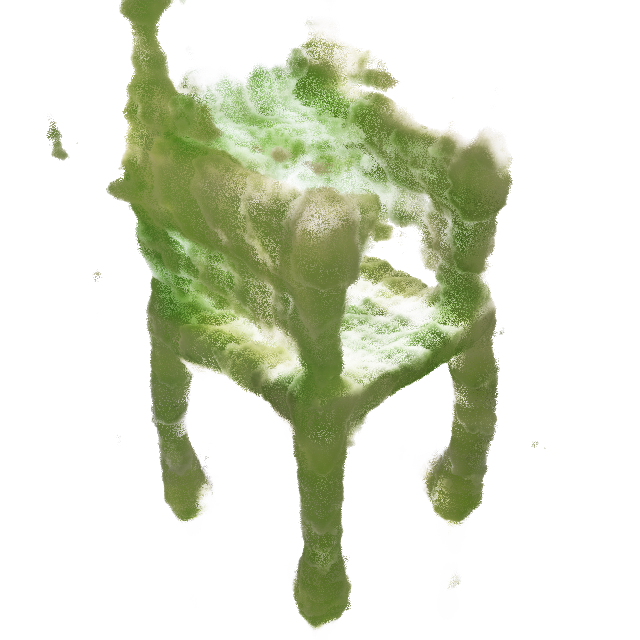}
    \includegraphics[width=0.12\textwidth]{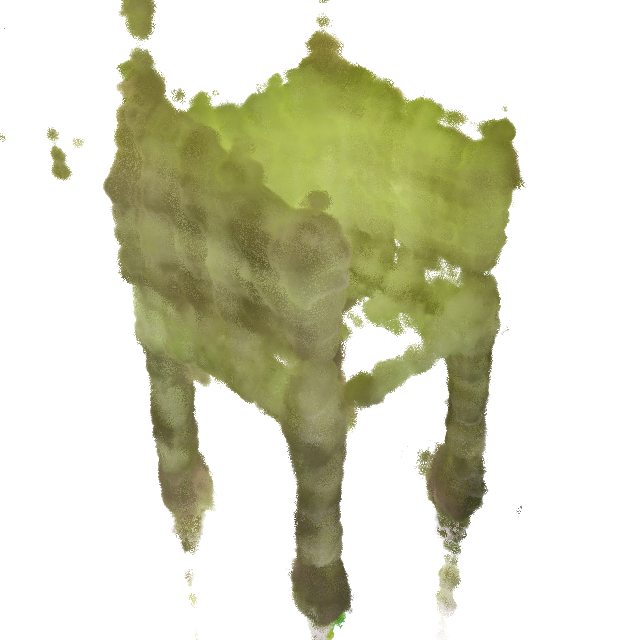}
    \includegraphics[width=0.12\textwidth]{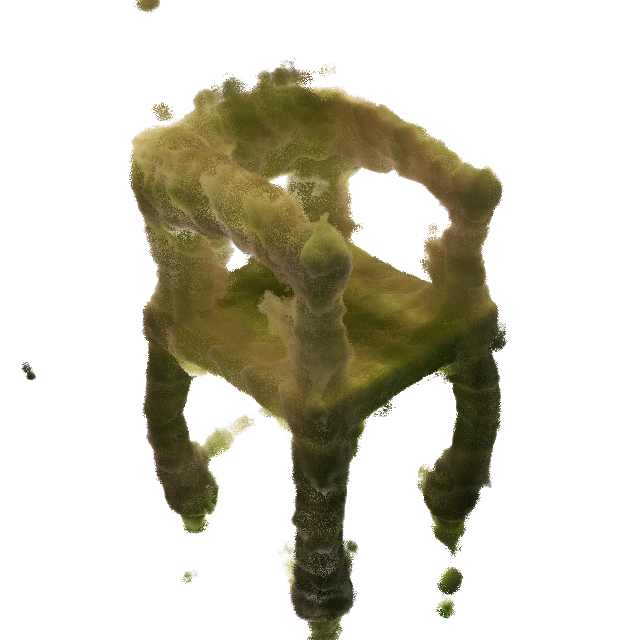}
    \includegraphics[width=0.12\textwidth]{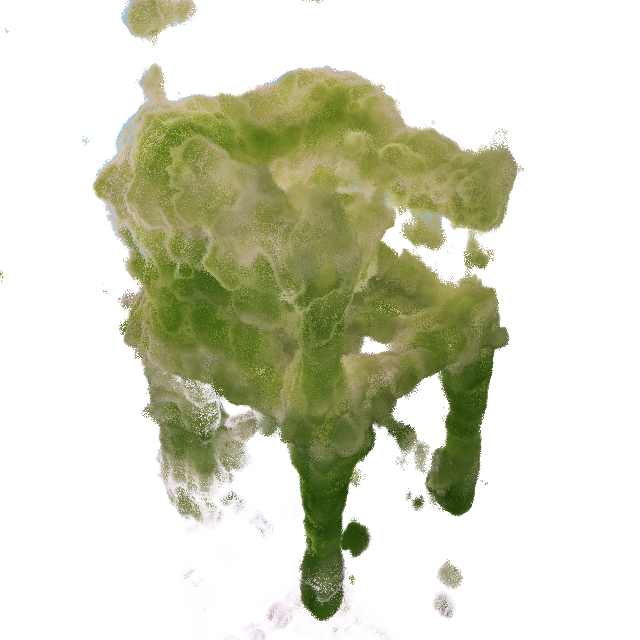}
    \includegraphics[width=0.12\textwidth]{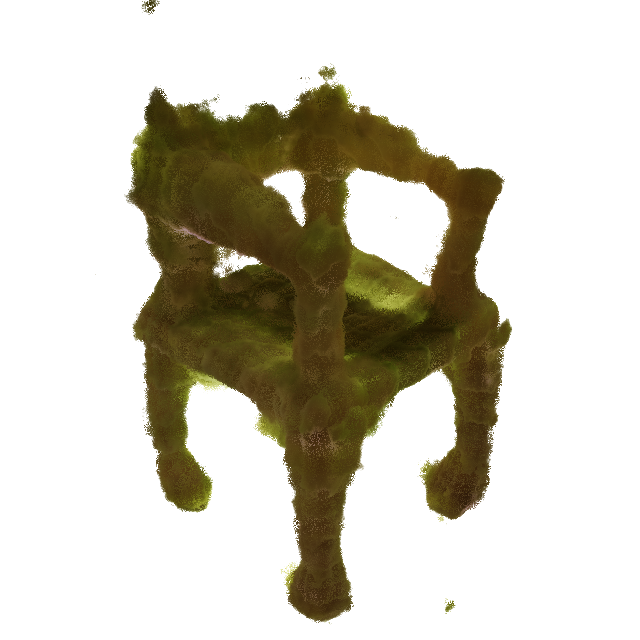}
    \includegraphics[width=0.12\textwidth]{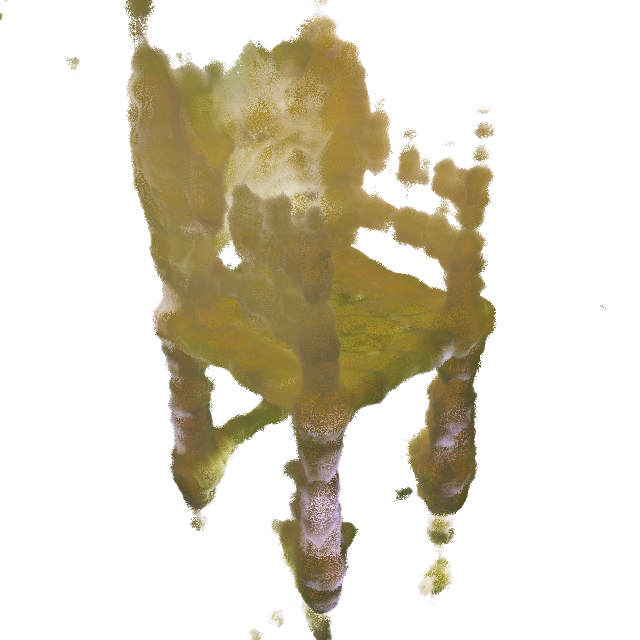}
    \caption{Comparison of inference results when using randomly-selected views (top) and fixed view (bottom) during semantic alignment. The results in column $i$, $i=1,2,\cdots 8$ are obtained by using the clip embedding of the $i$-th view during inference. Notice that in the fixed-view method, the 1st view is selected during training, hence we can observe the result in column 1 is much better than the other columns for fixed-view method. On the contrary, no matter which view is used in inference, the method trained using randomly-selected views can always produce satisfying and consistent results.}
    \label{fig:supp-rot-inv}
\end{figure*}

The presented findings can be observed in~\cref{fig:supp-rot-inv}. The upper section of the figure exhibits eight images, each representing inference results obtained from randomly selected viewpoints. These images demonstrate a high degree of similarity, indicating consistent outcomes across different viewpoints. Conversely, in the lower section of the figure, only the first image, which derived from the training viewpoint, depicts satisfactory results. However, when inferring from alternative viewpoints, the generated outcomes lack consistency and fail to align with the desired expectations.
To ensure fairness, the two checkpoints we used for comparison were trained to similar accuracies, with the fixed-view checkpoint having a slightly higher accuracy since it is more stable during training.

\section{Low CLIP Similarity on Highly-matched Image-Text Pairs}
\label{sec:supp-clip-sim}
\begin{figure}[ht]
    \centering
    \includegraphics[width=0.95\linewidth]{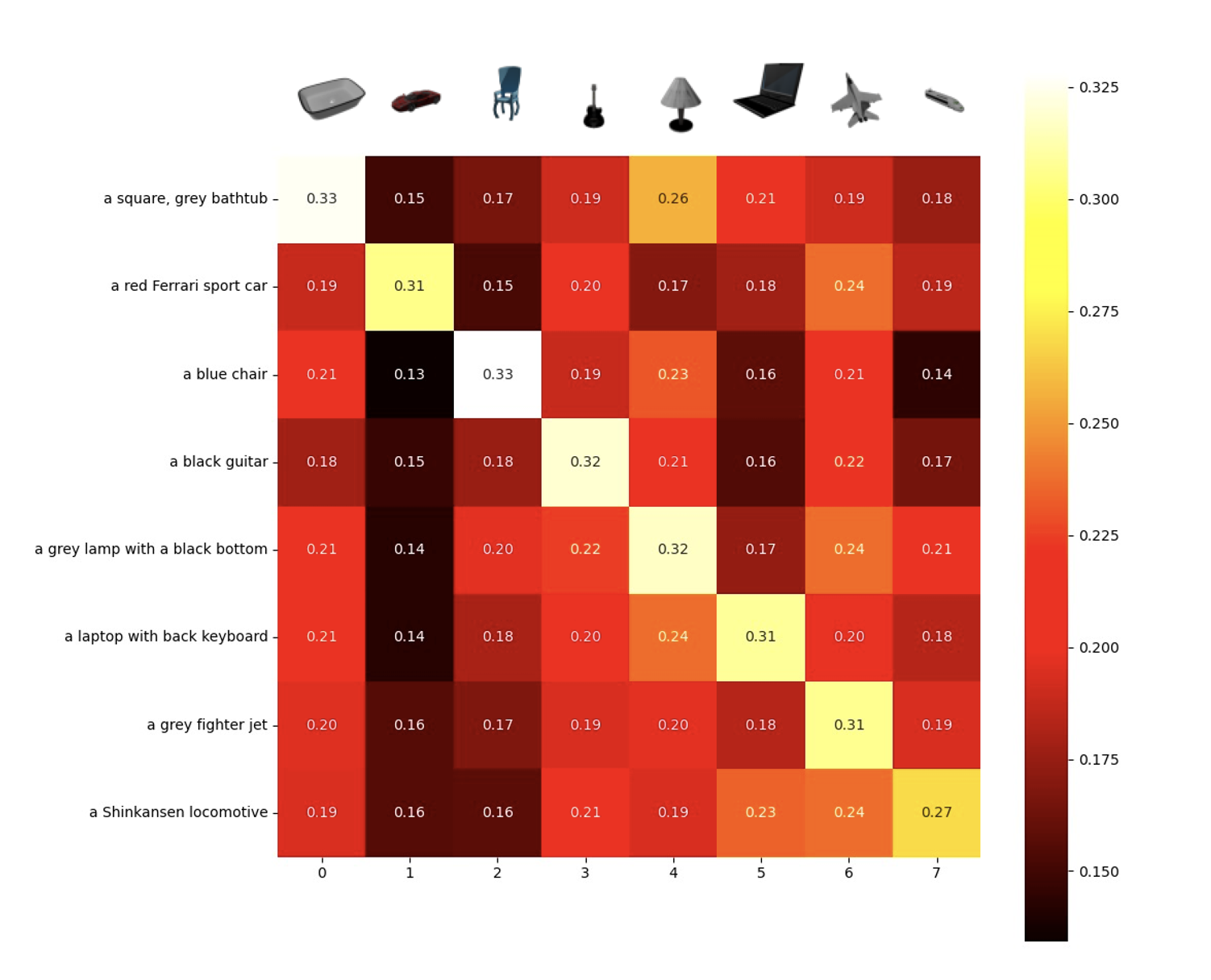}
    \caption{Result of some clip embedding cosine similarity of text description and image in our train set. Semantically highly matched pairs are placed on the diagonal. 
    }
    \label{fig:low_clip_similarity_on_match_scenes}
\end{figure}
As we have mentioned in \cref{sec:method-pipeline}, 
although CLIP~\cite{radford2021learning} effectively maps texts and images onto the same space, semantically highly matched image and text pairs still receive low (typically around 30\% to 40\%) cosine similarity, hence making it unlikely for our model trained on pure image embeddings to receive text embeddings directly during inference stage.

We noticed that some prior works such as \cite{Gonsalves_2023} have already demonstrated and discussed such a phenomenon, and in this section, we prove that this holds true on our dataset as well. \Cref{fig:low_clip_similarity_on_match_scenes} shows some examples of image and text pairs, where the cosine similarities of each pair are calculated and displayed in a heat map. We may observe that the pairs on the diagonal are semantically highly matched, yet only receive similarities of around 30\%.

\section{Implementation Details}
\label{sec:supp-impl-detail}

This section provides more implementation details of our model, as an extension of ~\cref{sec:implementation}

\noindent \textbf{VAE implementation detail} 
As detailed in \cref{sec:method-pipeline}, our NeRF network employs 14 distinct tokens for representation. To facilitate this, we employ 14 independent Variational Autoencoders (VAEs) ~\cite{kingma2013auto} to learn a separate implicit latent space for each token. Each VAE is comprised of an encoder and a decoder, each of which is composed of a 3-layer Multilayer Perceptron (MLP). Two linear layers are added between each encoder and decoder to learn a mean $\boldsymbol{\mu}$ and variance $\boldsymbol{\Sigma}$ for each token, thereby emulating a Gaussian distribution $X \sim \mathcal{N}(\boldsymbol{\mu}, \boldsymbol{\Sigma})$ representing each implicit latent space, subsequently samples an input to the decoder.

\noindent \textbf{Semantic Alignment implementation detail}
We utilized the Transformer~\cite{vaswani2023attention} model to attain semantic alignment, where we regard the semantic clip embedding ${\bf c}\in \mathbb{R}^d$ as the source sequence and its corresponding NeRF parameter in our pretrained latent space ${\bf z}\in \mathbb{R}^{14\times d}$ as the target sequence. In the training phase, a zero-valued beginning-of-sequence ([BOS]) token is appended before the target sequence, and the prediction is performed sequentially. During the inference stage, akin to the training phase, only a zero-valued [BOS] token is provided at the beginning as the target sequence. The Transformer model predicts the next target token based on the provided target sequence (only [BOS] at this stage) and the source sequence. The predicted token is then concatenated after the previous target sequence, and the Transformer model continues to predict the next token, and this process is repeated until all 14 target sequence tokens have been predicted.



\section{More Results for Out-of-distribution Inference}
\label{sec:supp-more-eg-ood}

In this section, we provide more out-of-distribution inference results, from both image and text prompts, as an extension to ~\cref{tab:out-dist-text-prompt-compare} in main paper.

\Cref{tab:more-eg-ood-img} shows four more out-of-distribution inference results from image prompts that fall out of our eight categories, and~\cref{tab:more-eg-ood-text} shows four corresponding results from text prompts. We can observe the 3 to 5 times significant acceleration that our initialization brings to the baseline models, either Zero-1-to-3 ~\cite{liu2023zero1to3} for the image-to-NeRF task or DreamFusion ~\cite{poole2022dreamfusion} for the text-to-NeRF task.

\begin{table*}[ht]
    \newcommand{\dummyImg}{\fbox{\rule{0pt}{0.7in} \rule{0.7\linewidth}{0pt}} }
    \centering
    \begin{tabular}{>{\centering\arraybackslash}m{0.13\linewidth}|>{\centering\arraybackslash}m{0.15\linewidth}|>{\centering\arraybackslash}m{0.15\linewidth}|>{\centering\arraybackslash}m{0.15\linewidth}|>
    {\centering\arraybackslash}m{0.15\linewidth}|>
    {\centering\arraybackslash}m{0.15\linewidth}}
    \toprule
        \multirow{2}{*}{Prompt} & Ours & Zero-1-to-3 (S) & Zero-1-to-3 (P) & Zero-1-to-3 (S) & Zero-1-to-3 (P)\\
         & (converge) & (same iterations) & (same iterations) & (converge) & (converge) \\
    \midrule
        \vspace{0.2in}  
        \includegraphics[width=0.82\linewidth]{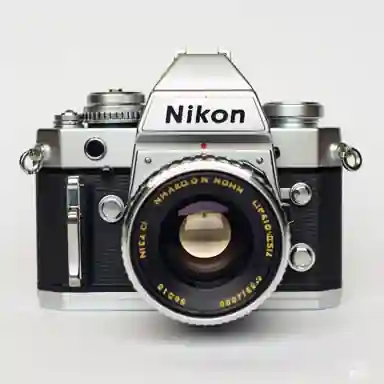}
        & \includegraphics[width=0.82\linewidth]{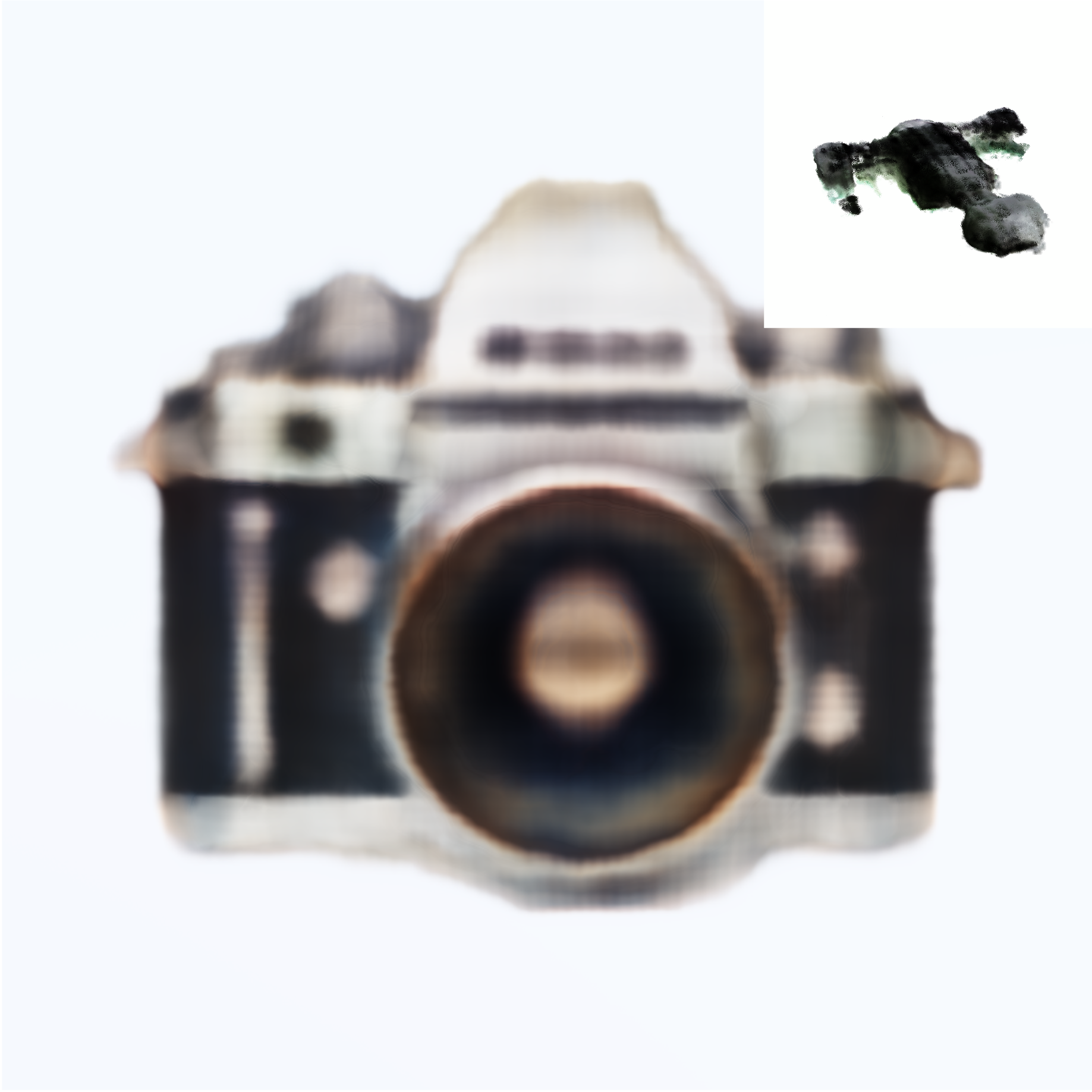}
        & \includegraphics[width=0.82\linewidth]{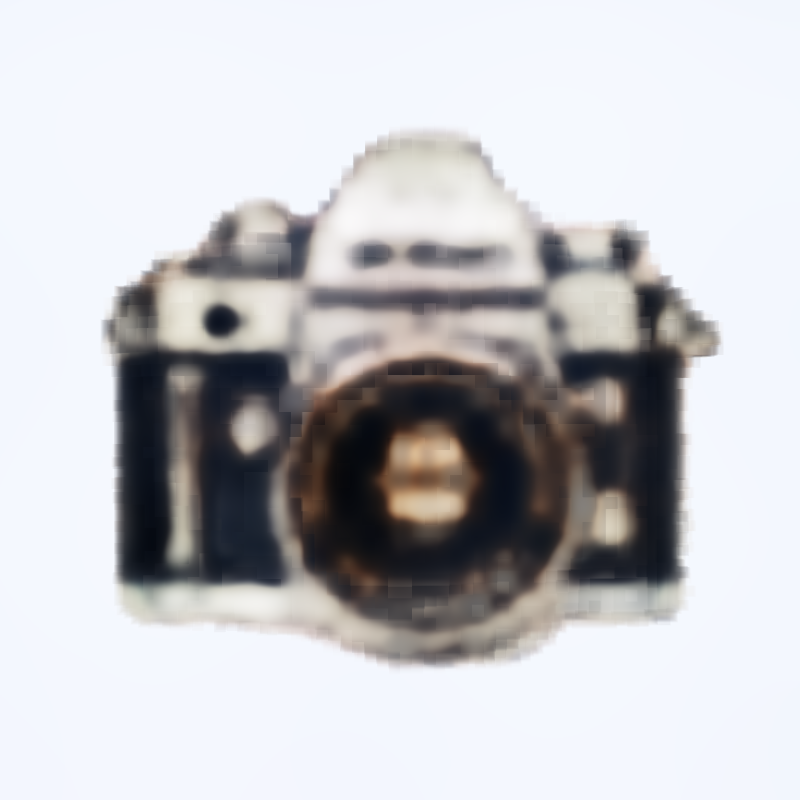}
        & \includegraphics[width=0.82\linewidth]{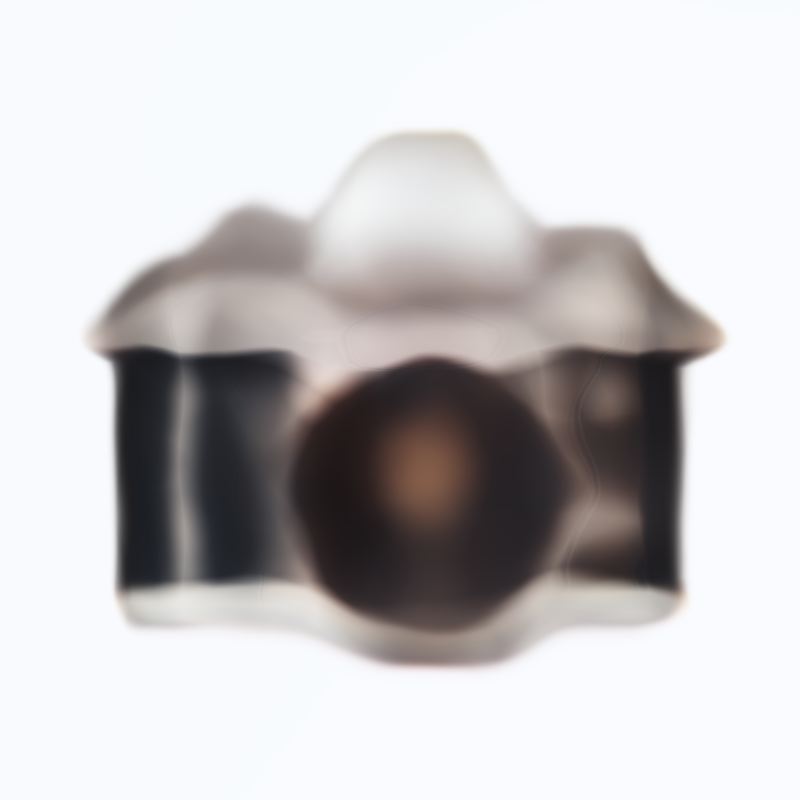}
        & \includegraphics[width=0.82\linewidth]{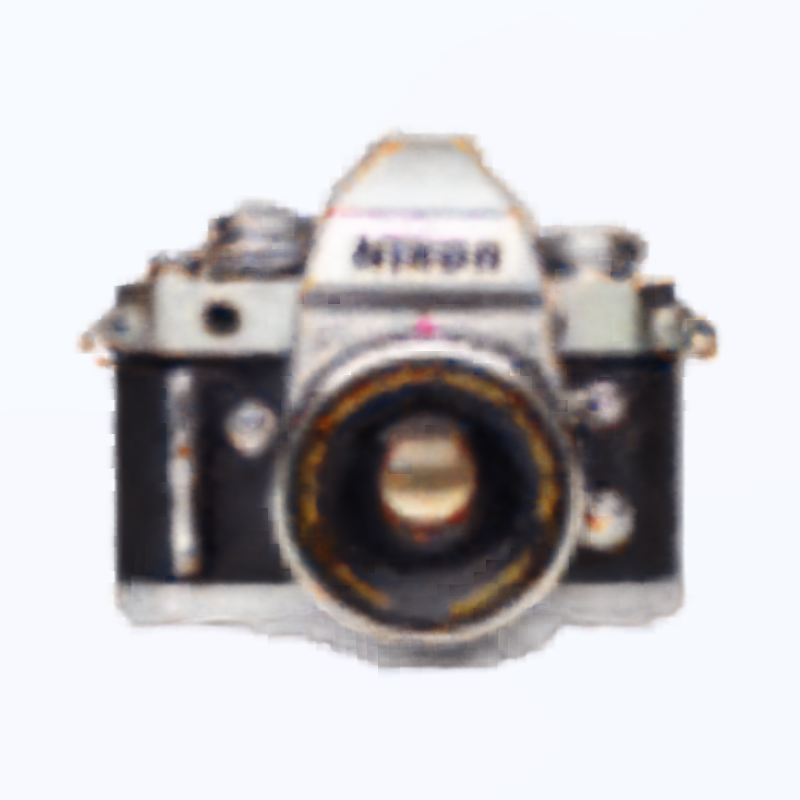}
        & \includegraphics[width=0.82\linewidth]{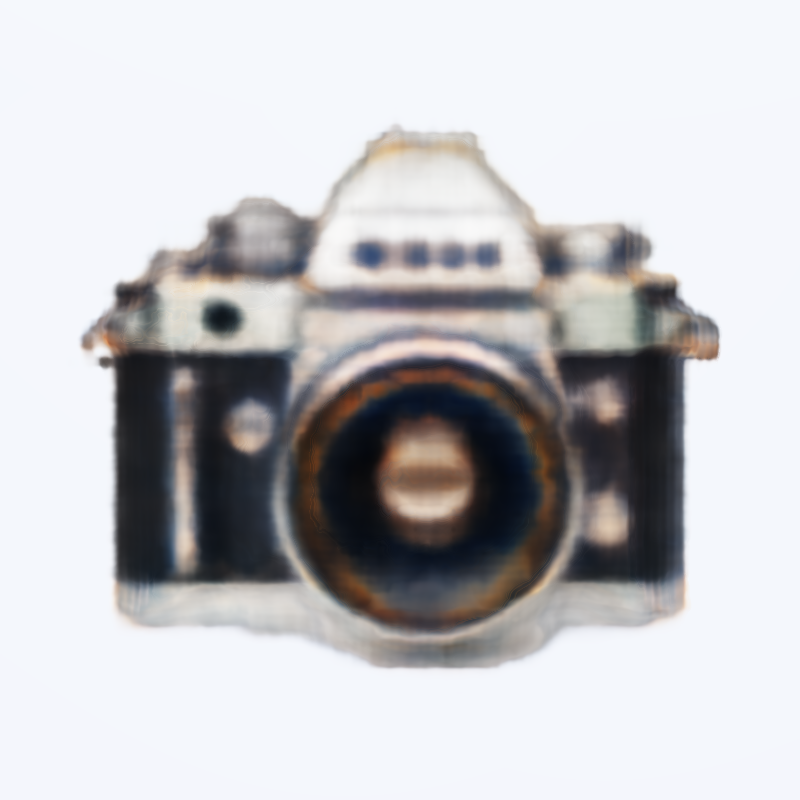}\\
         & Similarity: 69.6\% & Similarity: 68.1\% & Similarity: 71.3\% & Similarity: 68.4\% & Similarity: 68.3\%\\
         & Iterations: 12.6k & Iterations: 12.6k & Iterations: 12.6k & Iterations: 41.2k & Iterations: 40k\\\hline

        \vspace{0.2in}  
        \includegraphics[width=0.82\linewidth]{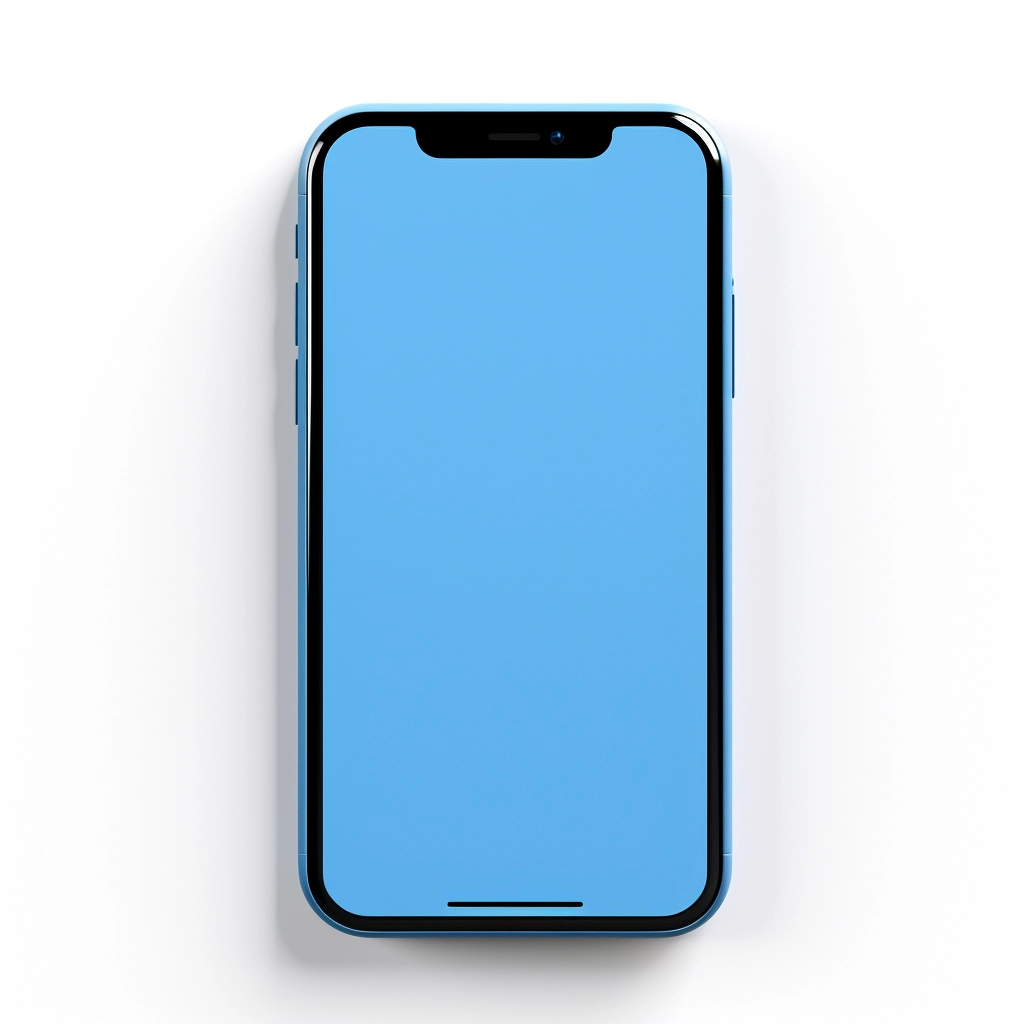}
        & \includegraphics[width=0.82\linewidth]{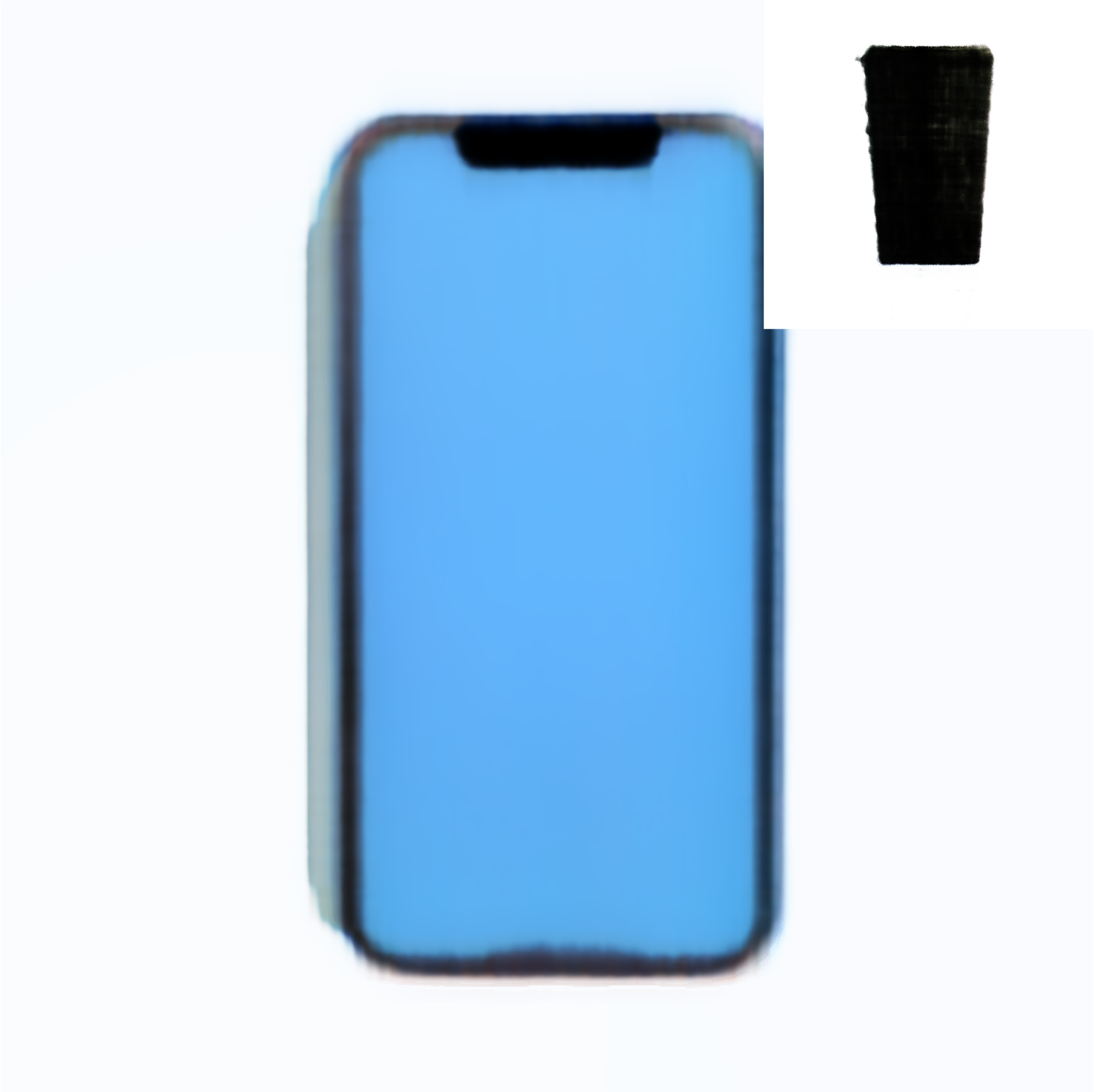}
        & \includegraphics[width=0.82\linewidth]{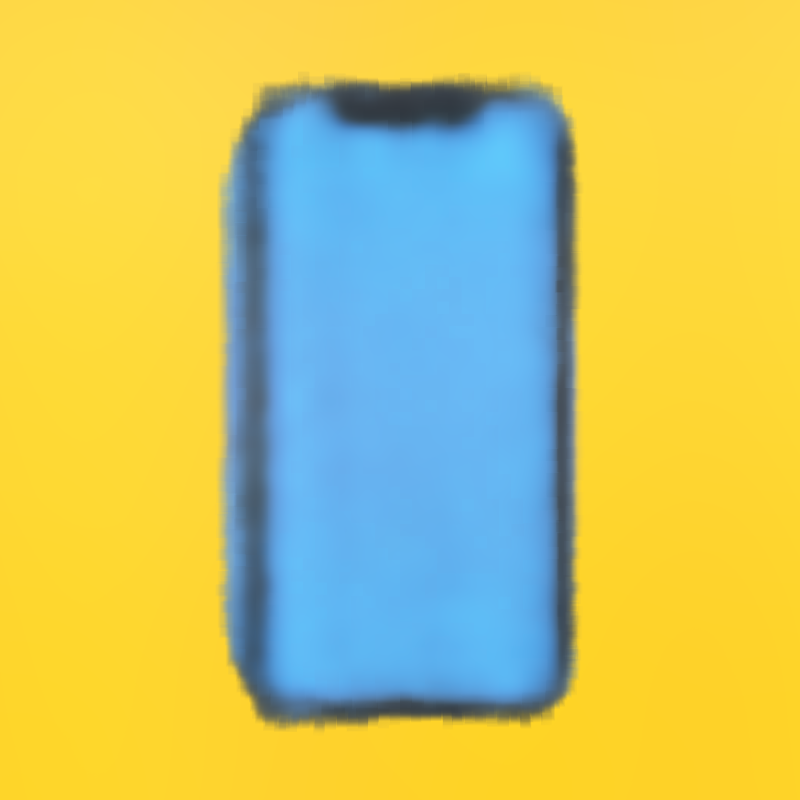}
        & \includegraphics[width=0.82\linewidth]{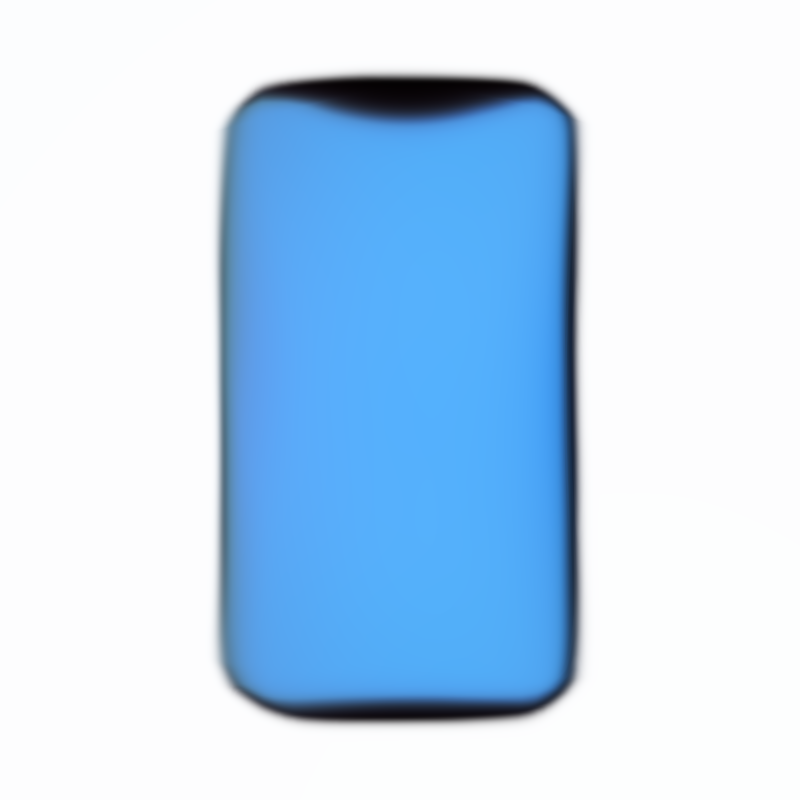}
        & \includegraphics[width=0.82\linewidth]{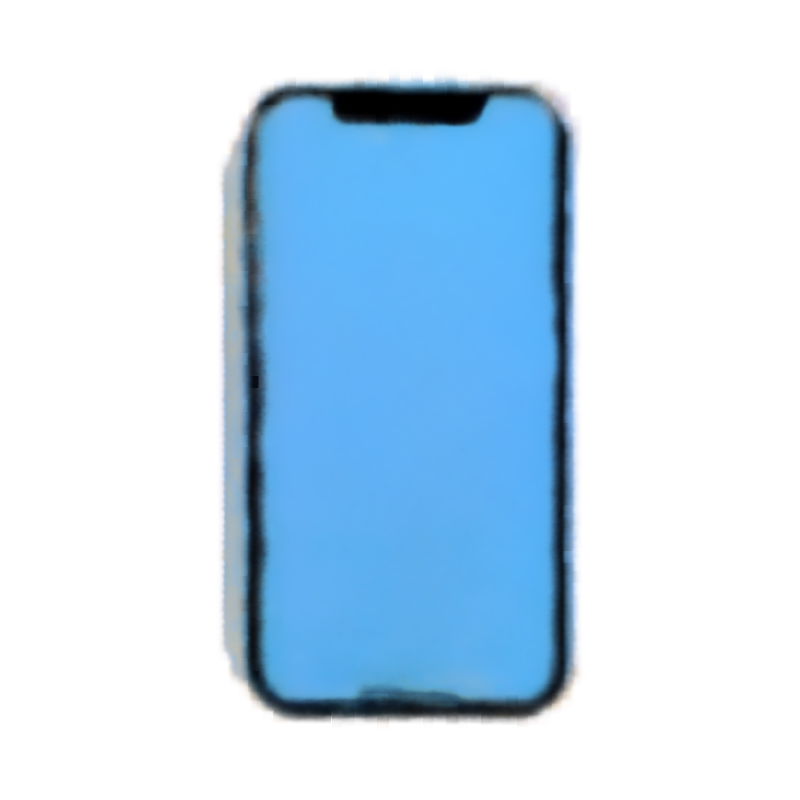}
        & \includegraphics[width=0.82\linewidth]{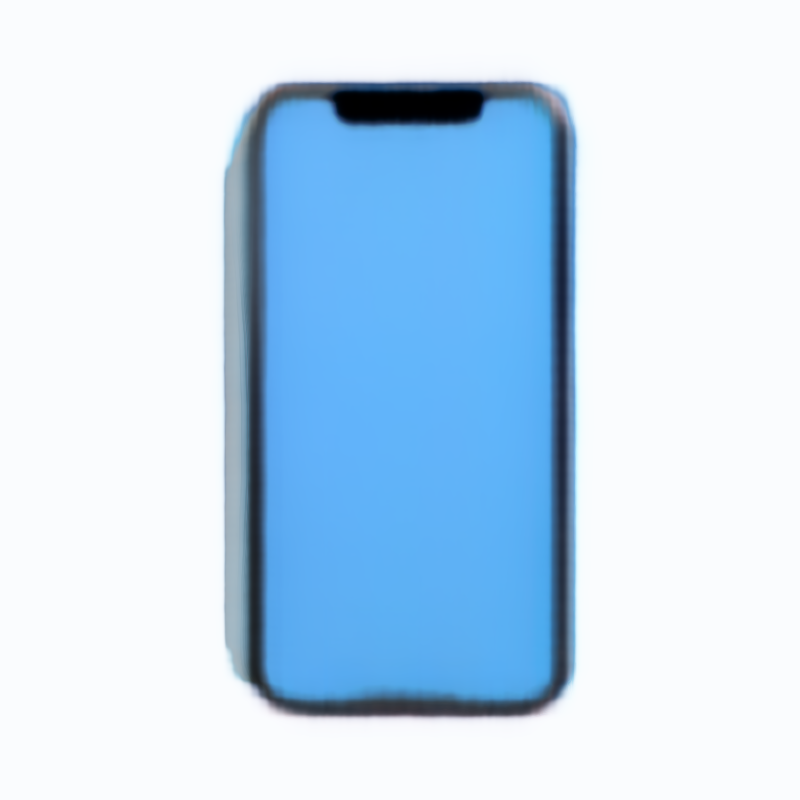}\\
         & Similarity: 93.5\% & Similarity: 84.0\% & Similarity: 90.1\% & Similarity: 92.0\% & Similarity: 92.9\%\\
         & Iterations: 4k & Iterations: 4k & Iterations: 4k & Iterations: 20k & Iterations: 17.9k\\\hline

        \vspace{0.2in}  
        \includegraphics[width=0.82\linewidth]{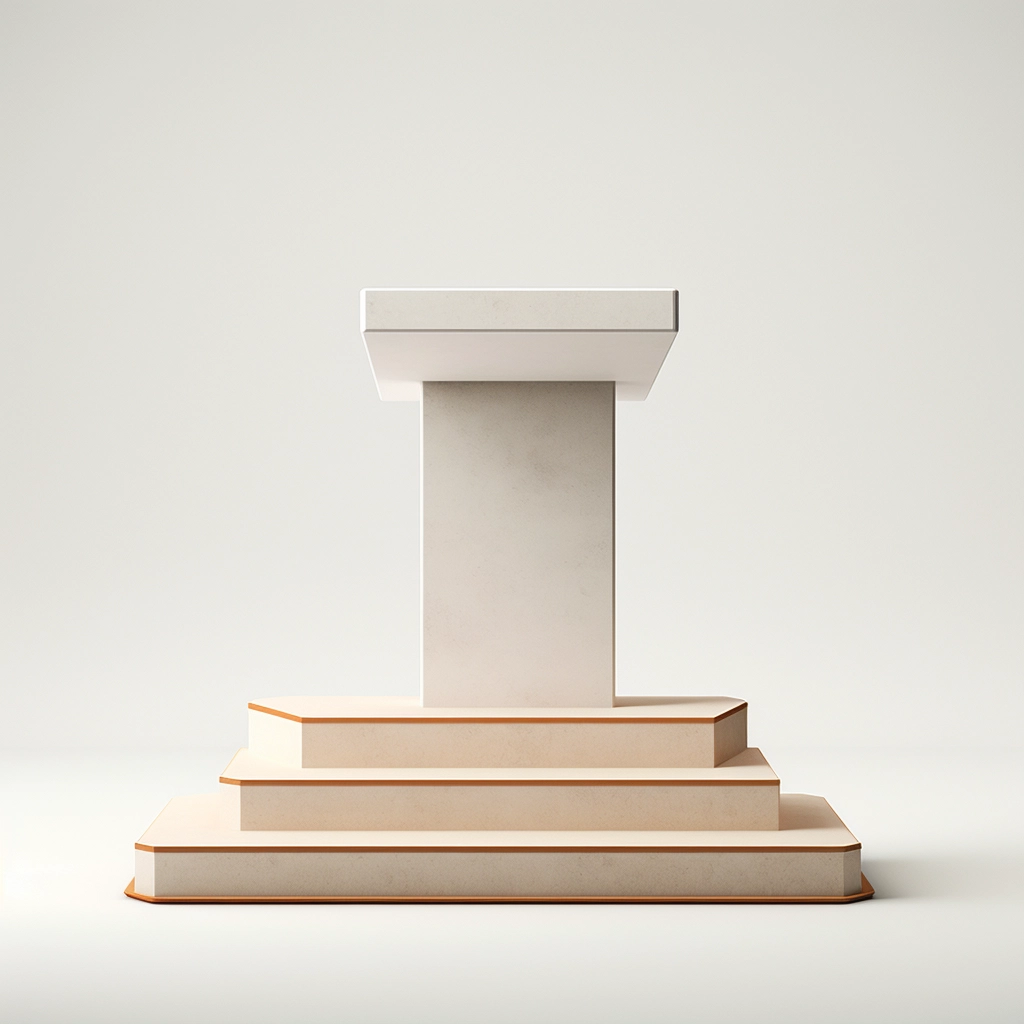}
        & \includegraphics[width=0.82\linewidth]{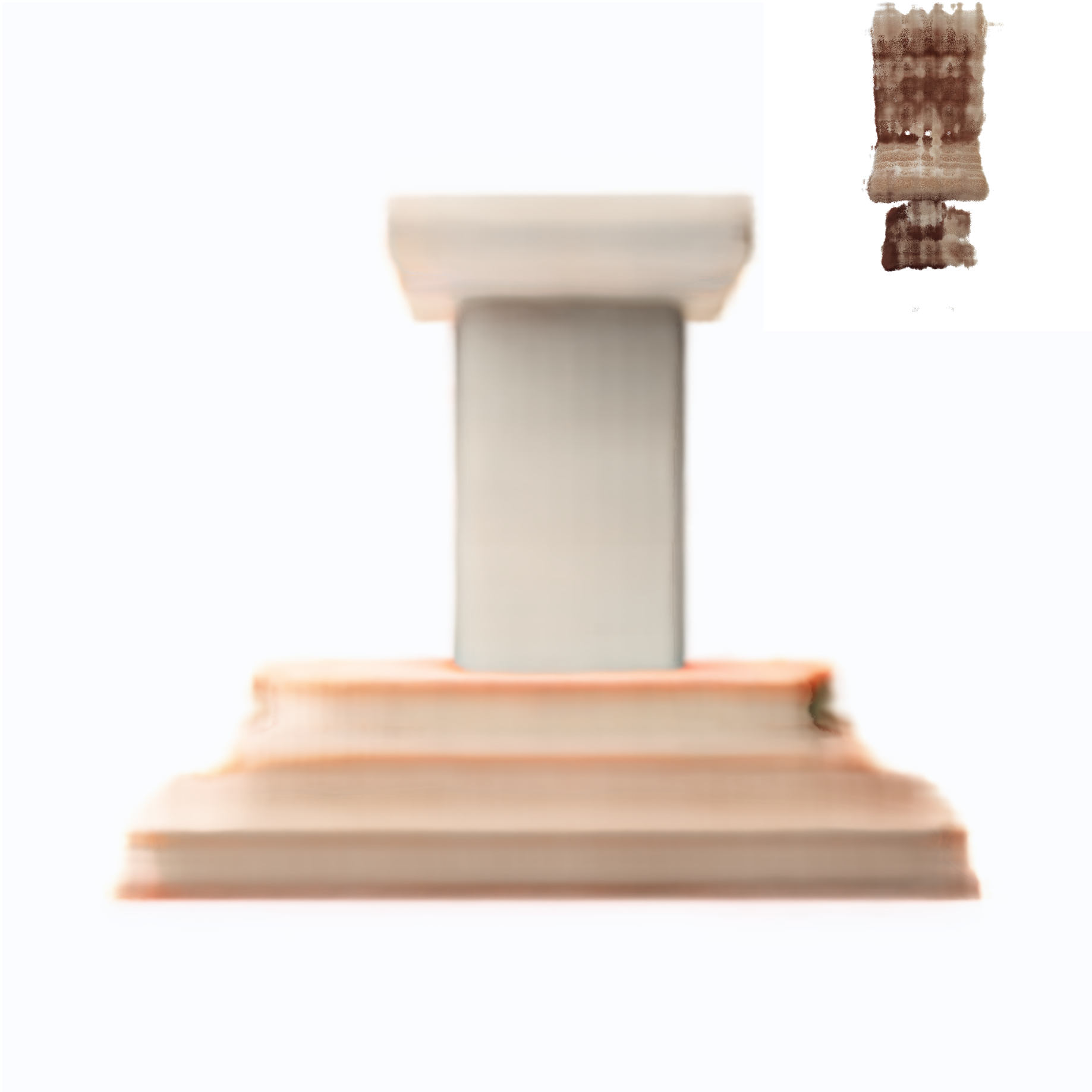}
        & \includegraphics[width=0.82\linewidth]{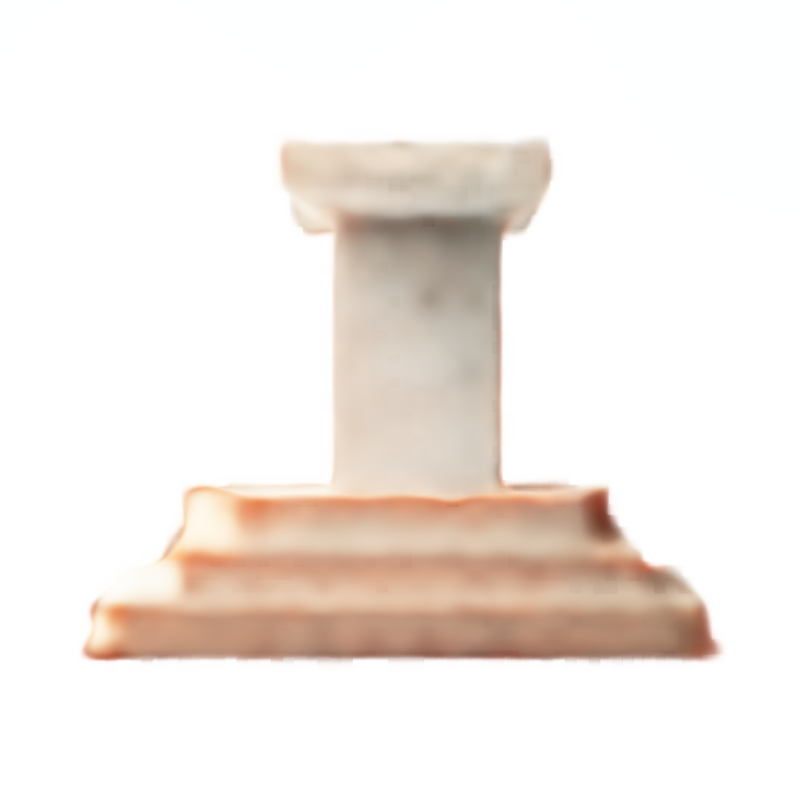}
        & \includegraphics[width=0.82\linewidth]{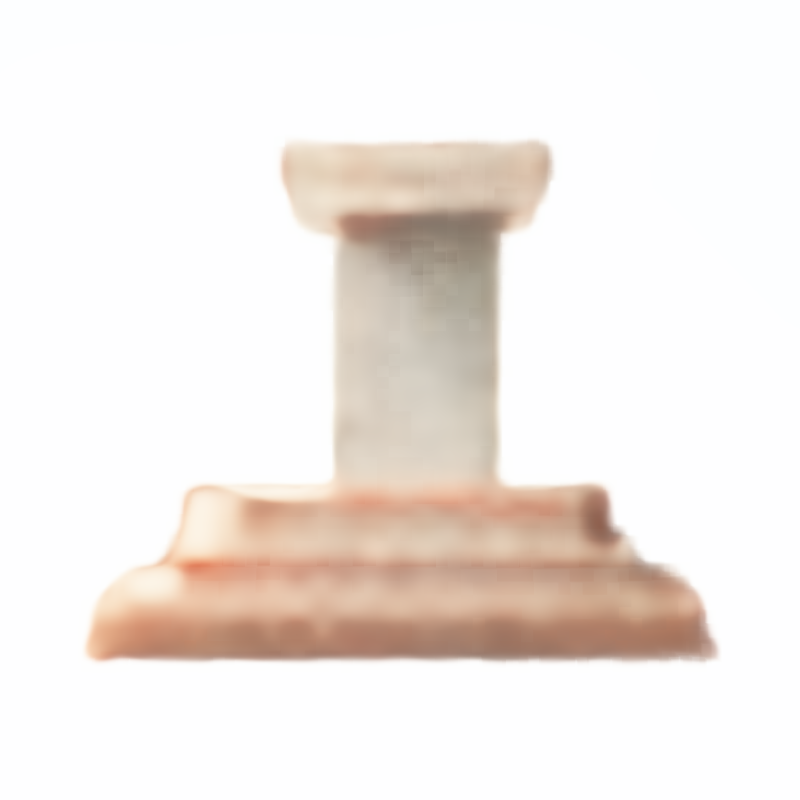}
        & \includegraphics[width=0.82\linewidth]{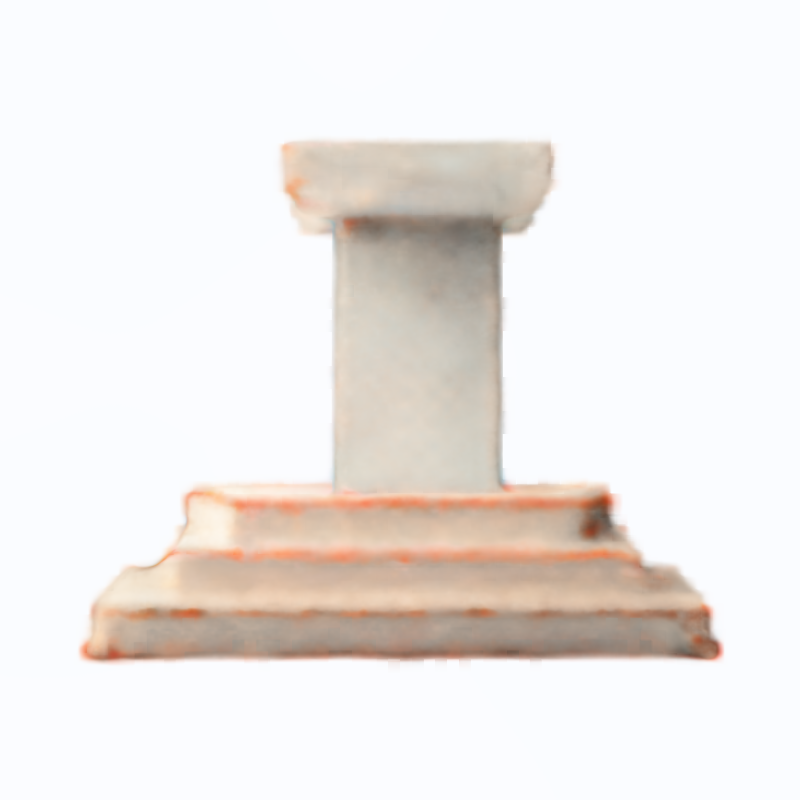}
        & \includegraphics[width=0.82\linewidth]{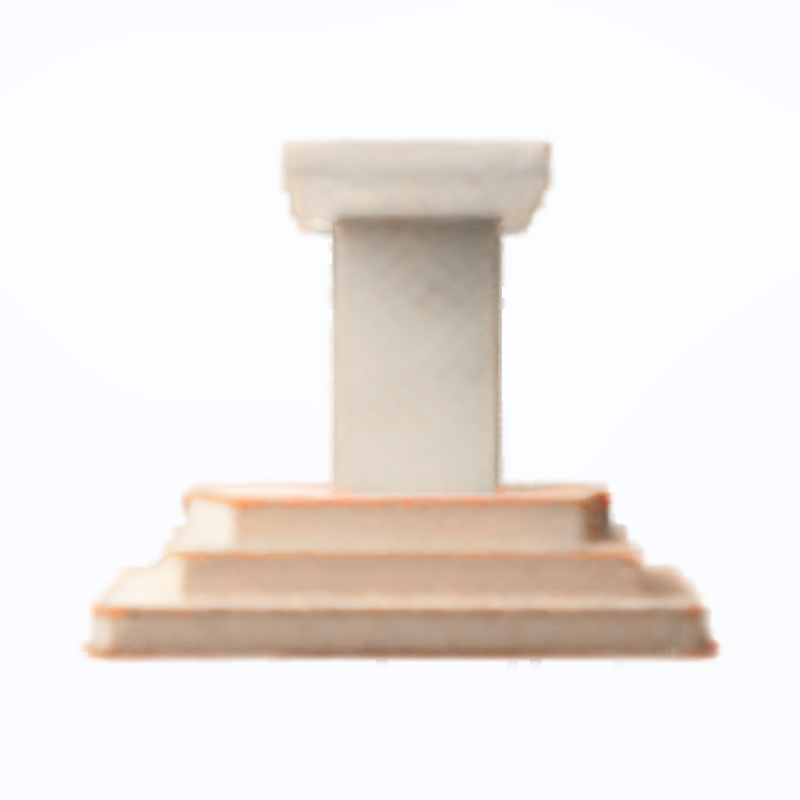}\\
         & Similarity: 85.8\% & Similarity: 78.6\% & Similarity: 74.1\% & Similarity: 81.7\% & Similarity: 86.3\%\\
         & Iterations: 6.9k & Iterations: 6.9k & Iterations: 6.9k & Iterations: 20.1k & Iterations: 21.2k\\\hline

        \vspace{0.2in}  
        \includegraphics[width=0.82\linewidth]{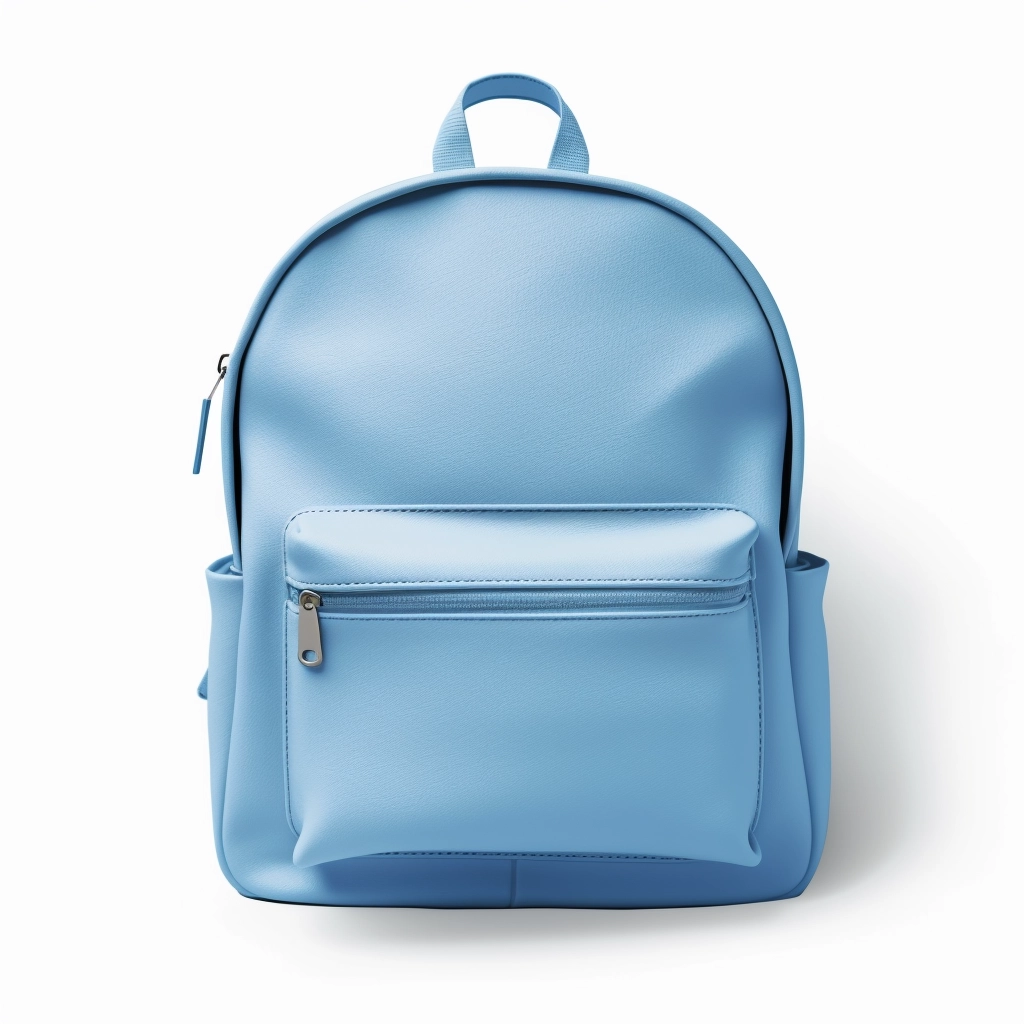}
        & \includegraphics[width=0.82\linewidth]{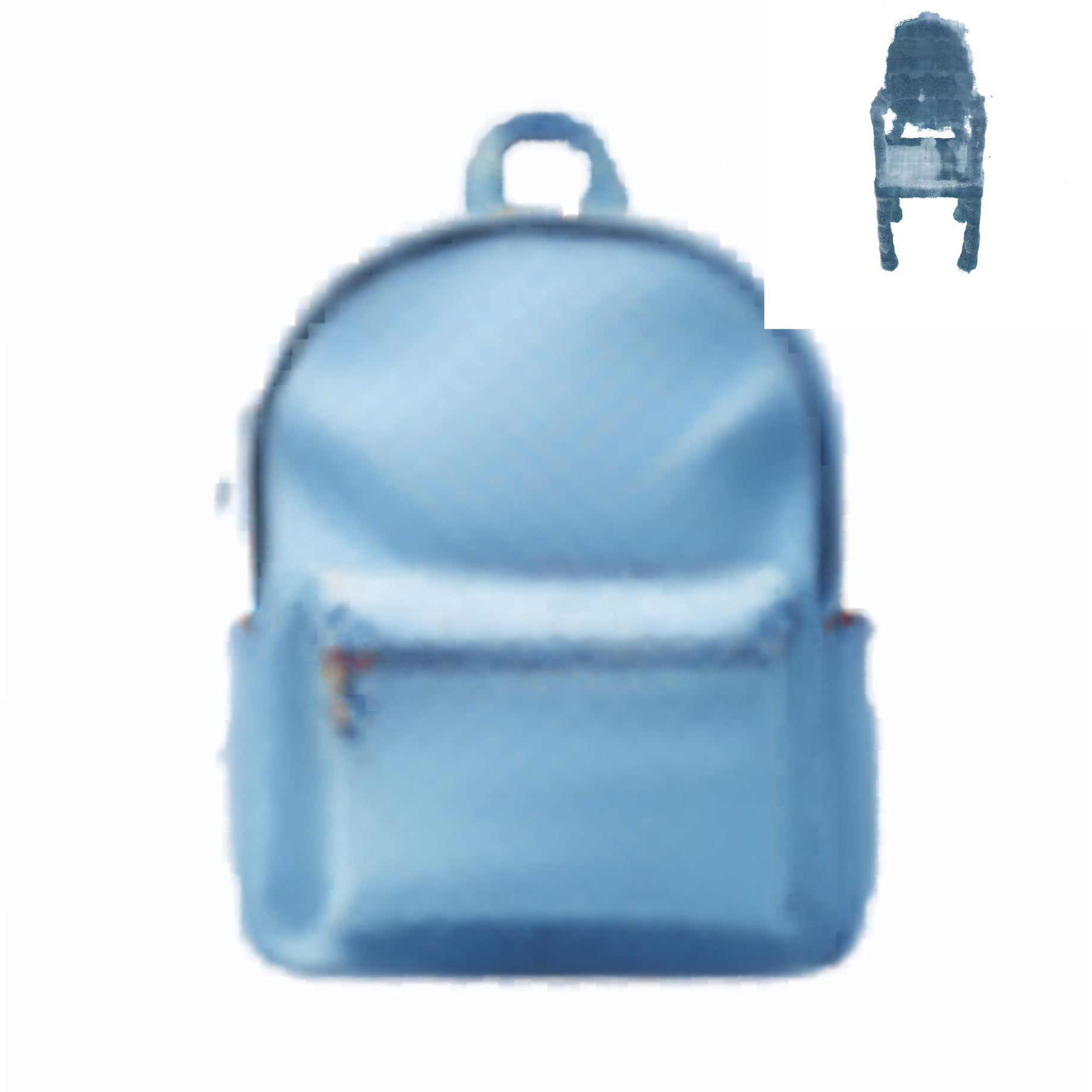}
        & \includegraphics[width=0.82\linewidth]{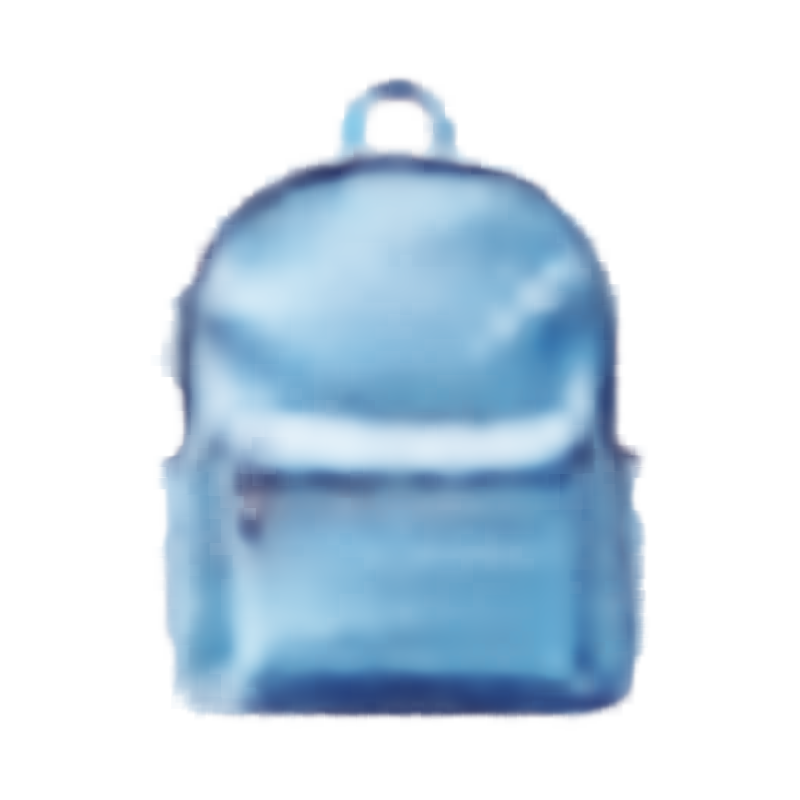}
        & \includegraphics[width=0.82\linewidth]{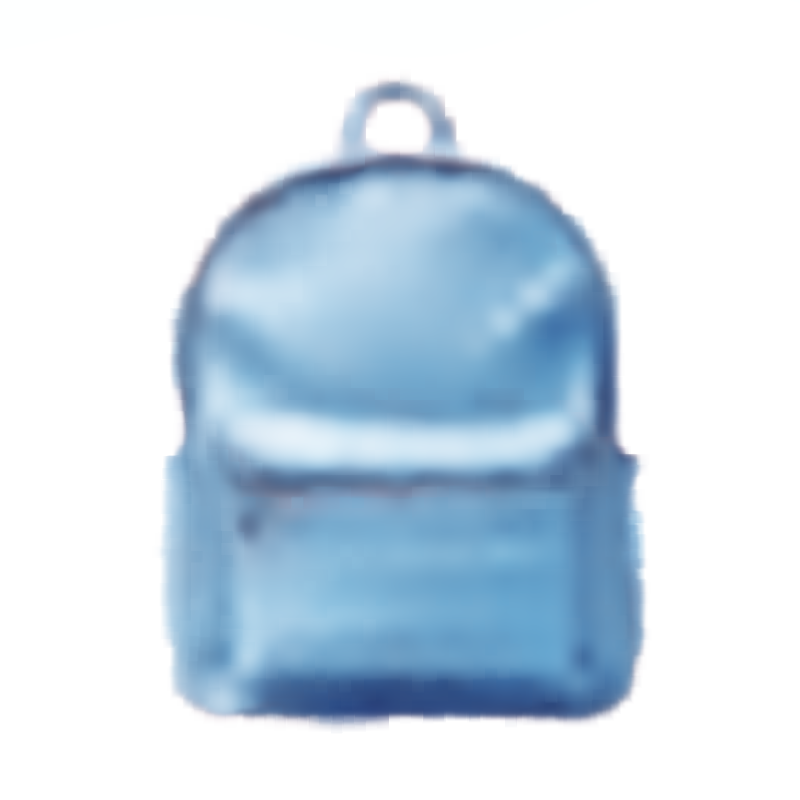}
        & \includegraphics[width=0.82\linewidth]{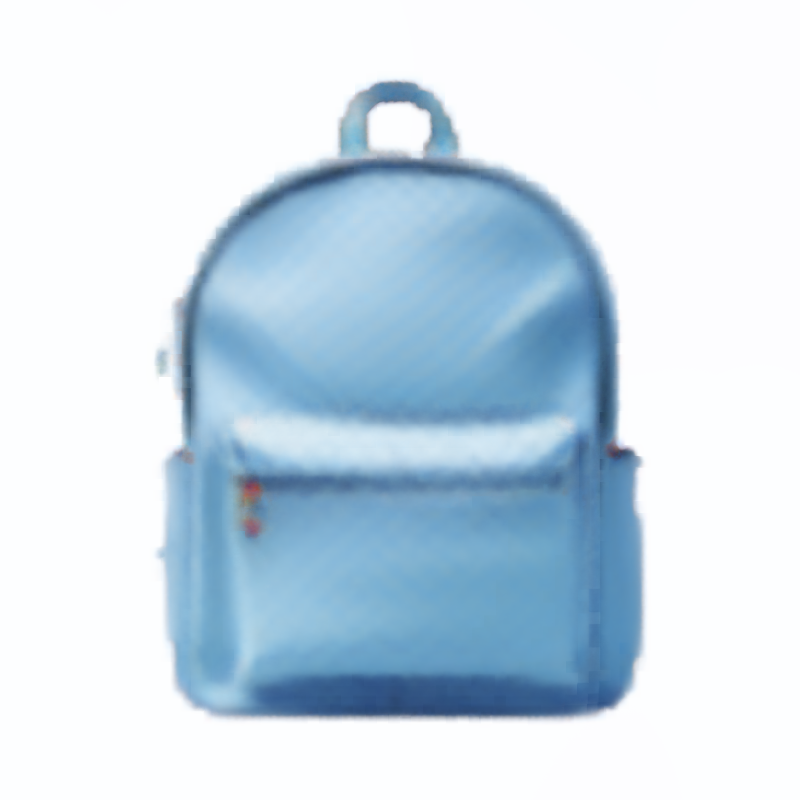}
        & \includegraphics[width=0.82\linewidth]{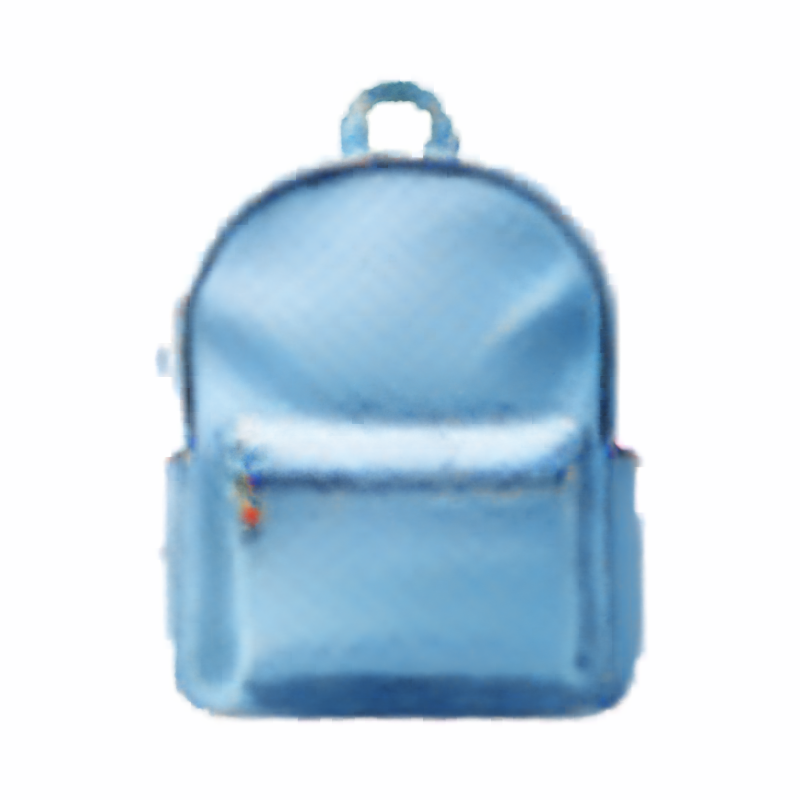}\\
         & Similarity: 94.8\% & Similarity: 89.7\% & Similarity: 90.3\% & Similarity: 94.1\% & Similarity: 93.8\%\\
         & Iterations: 10.7k & Iterations: 10.7k & Iterations: 10.7k & Iterations: 34.7k & Iterations: 33.4k\\
    \bottomrule
    \end{tabular}
    \caption{More examples of {\bf out-of-distribution inference result, using  image prompts}. For each prompt, we record the convergence iteration count using our initialization zero-shot, from one single forward pass (top-right inset). Scratch (S) or pre-trained (P) Zero-1-to-3 are respectively trained for the same number of iterations and until convergence with their respective results displayed above.
    Our method significantly accelerates the Zero-1-to-3's optimization, achieving 3 to 5 times speed boost while almost consistently yielding better results. Moreover, our method provides a semantically meaningful initialization, as pretrained (P) Zero-1-to-3 still takes an extremely long time to converge though initialized with a similar scene.}
    \label{tab:more-eg-ood-img}
    \vspace{-0.2in}
\end{table*}

\begin{table*}[ht]
    \newcommand{\dummyImg}{\fbox{\rule{0pt}{0.7in} \rule{0.7\linewidth}{0pt}} }
    \centering
    \begin{tabular}{>{\centering\arraybackslash}m{0.13\linewidth}|>{\centering\arraybackslash}m{0.15\linewidth}|>{\centering\arraybackslash}m{0.15\linewidth}|>{\centering\arraybackslash}m{0.15\linewidth}|>
    {\centering\arraybackslash}m{0.15\linewidth}|>
    {\centering\arraybackslash}m{0.15\linewidth}}
    \toprule
        \multirow{2}{*}{Prompt} & Ours & DreamFusion (S) & DreamFusion (P) & DreamFusion (S) & DreamFusion (P)\\
         & (converge) & (same iterations) & (same iterations) & (converge) & (converge) \\
    \midrule

        \vspace{0.25in}  
        {\it ``A colorful Boeing passenger plane.''}  
        & \includegraphics[width=0.75\linewidth]{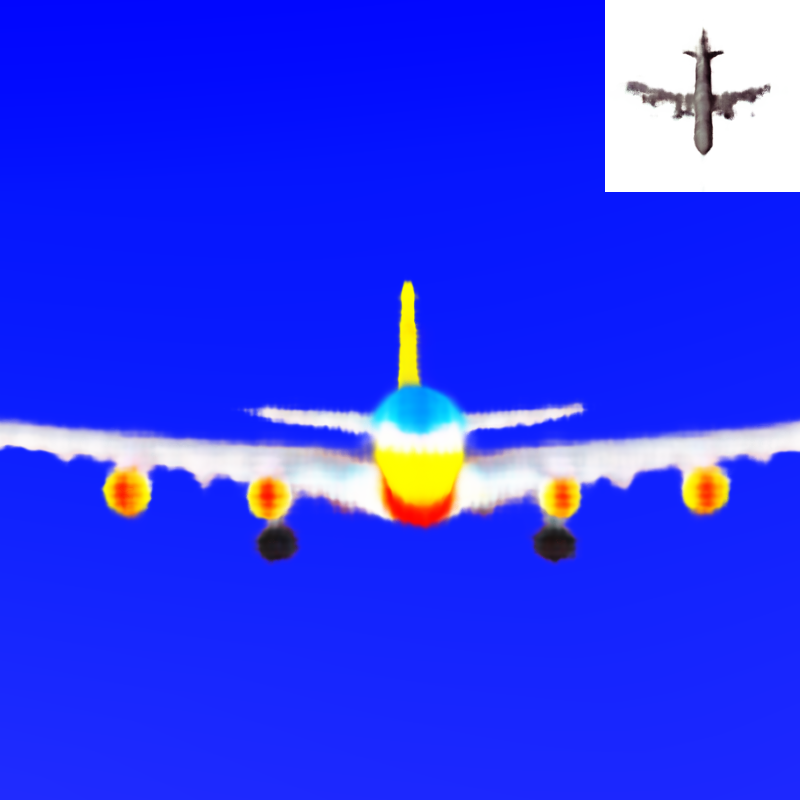}
        & \includegraphics[width=0.75\linewidth]{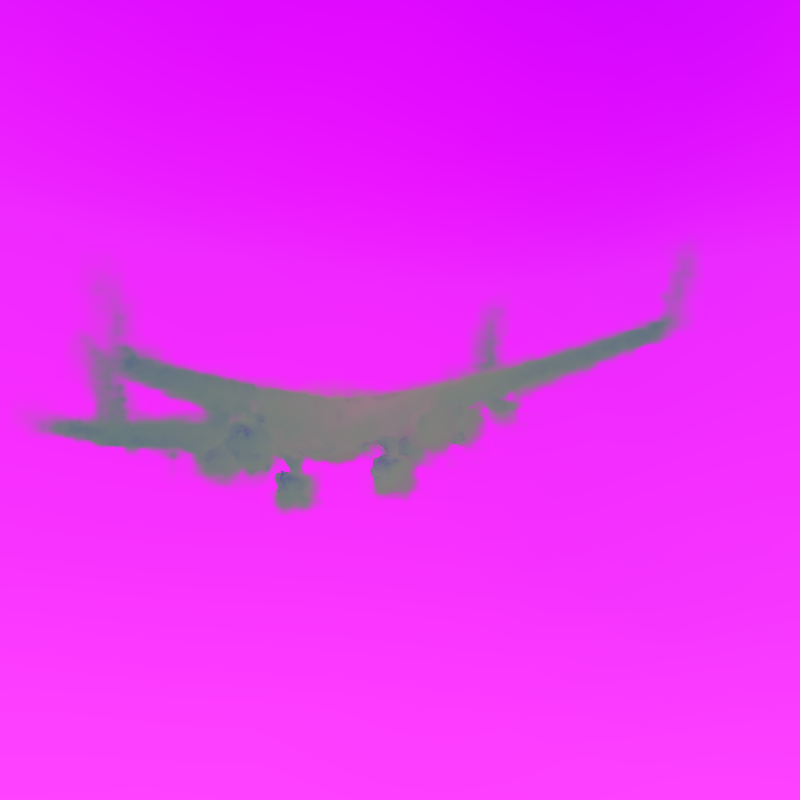}
        & \includegraphics[width=0.75\linewidth]{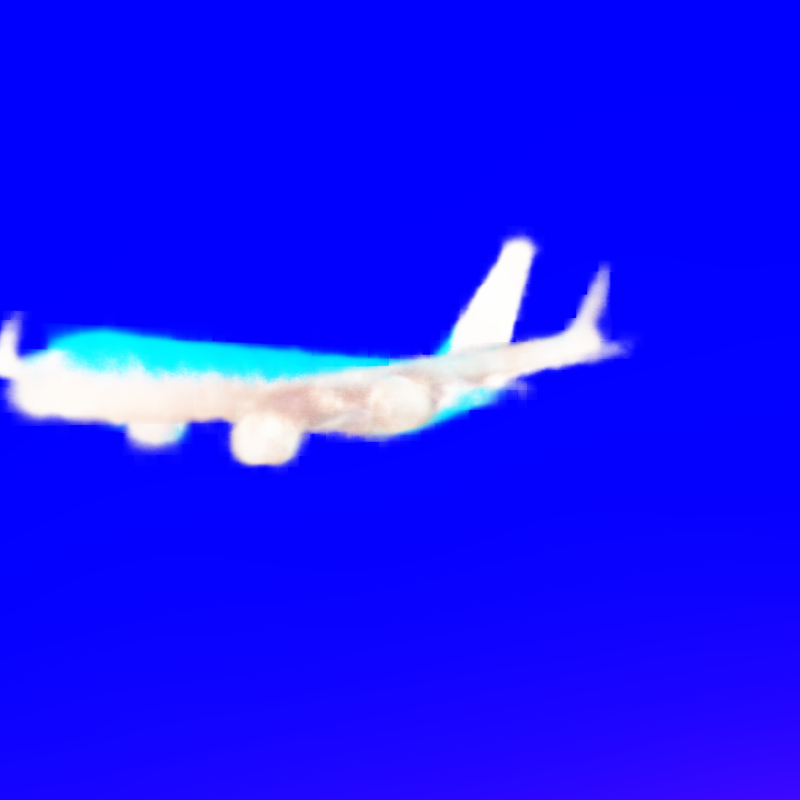}
        & \includegraphics[width=0.75\linewidth]{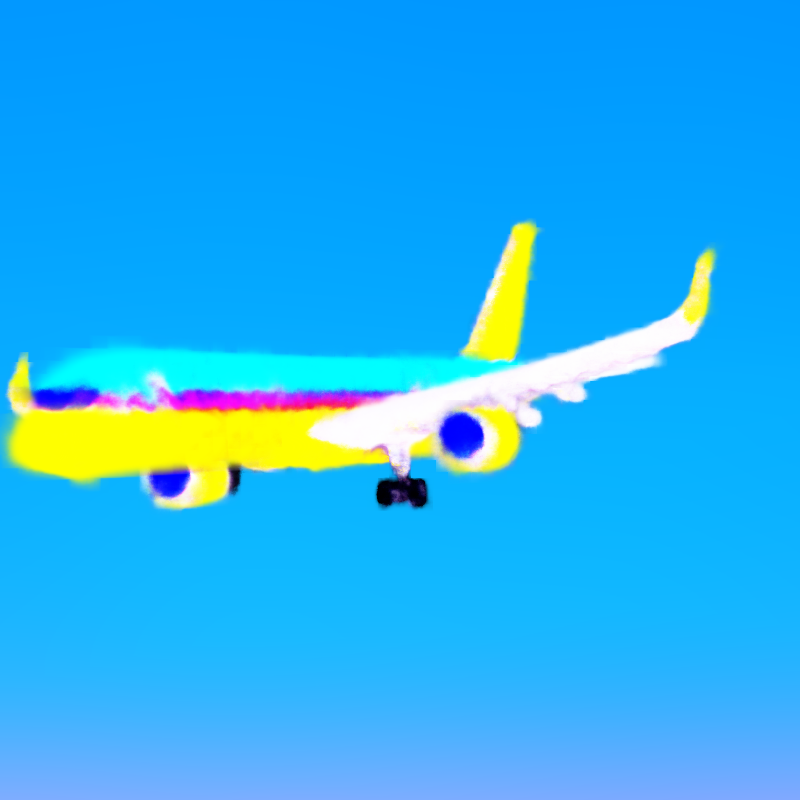}
        & \includegraphics[width=0.75\linewidth]{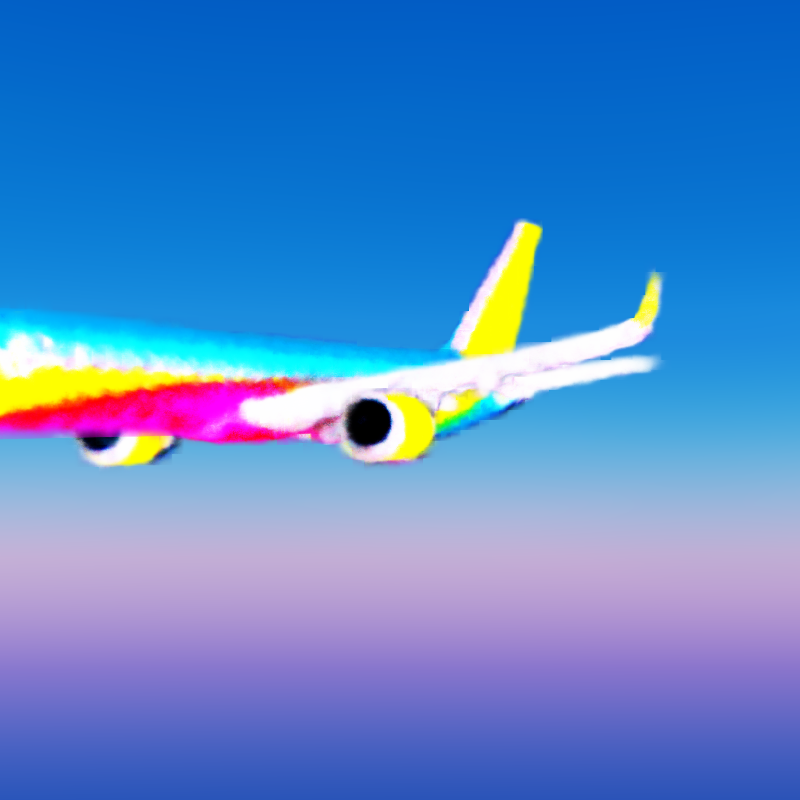}\\
         & Similarity: 31.5\% & Similarity: 29.6\% & Similarity: 28.6\% & Similarity: 33.6\% & Similarity: 33.6\%\\
         & Iterations: 7.5k & Iterations: 7.5k & Iterations: 7.5k & Iterations: 30k & Iterations: 31.5k\\\hline

        \vspace{0.25in}  
        {\it ``Iron man on a blue chair.''}  
        & \vspace{0.05in} \includegraphics[width=0.72\linewidth]{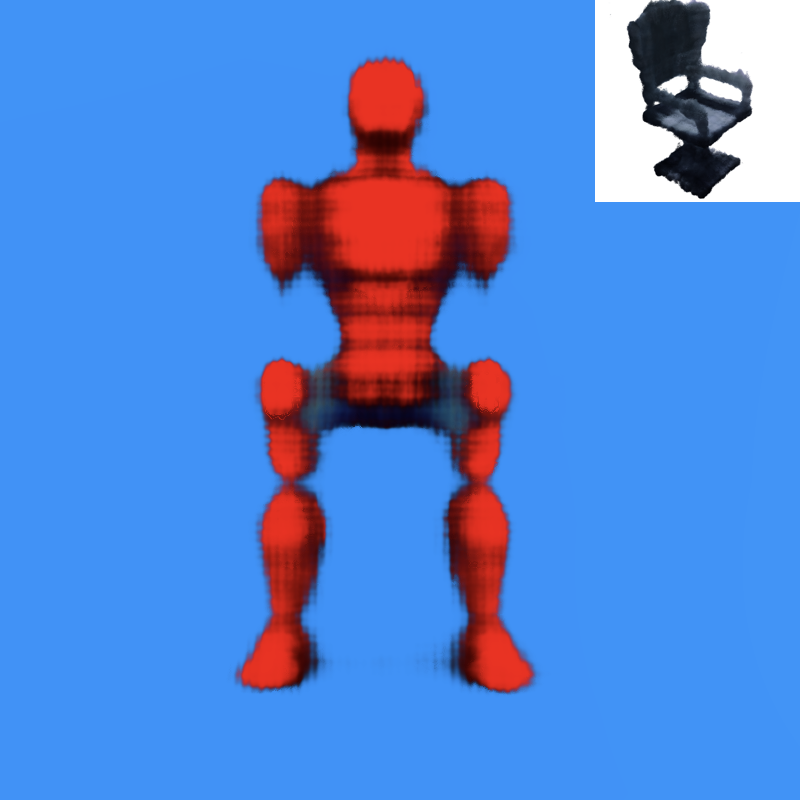}
        & \includegraphics[width=0.72\linewidth]{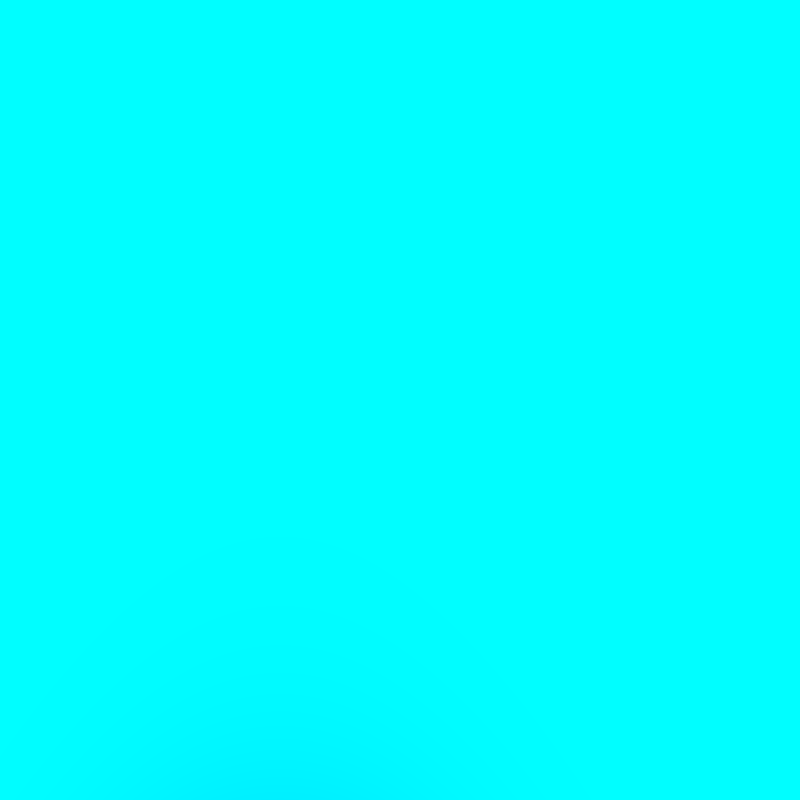}
        & \includegraphics[width=0.72\linewidth]{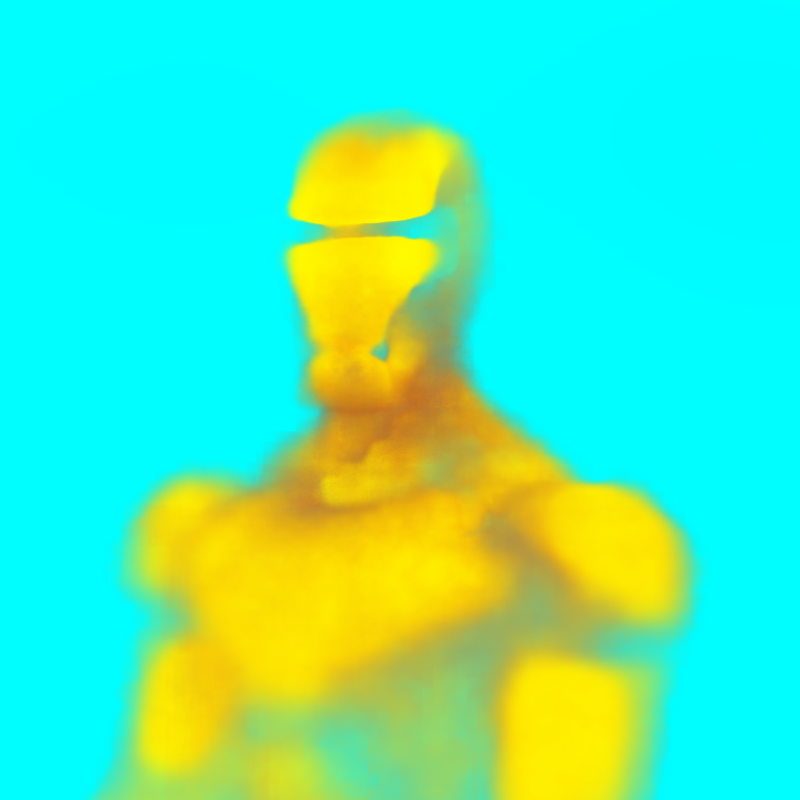}
        & \includegraphics[width=0.72\linewidth]{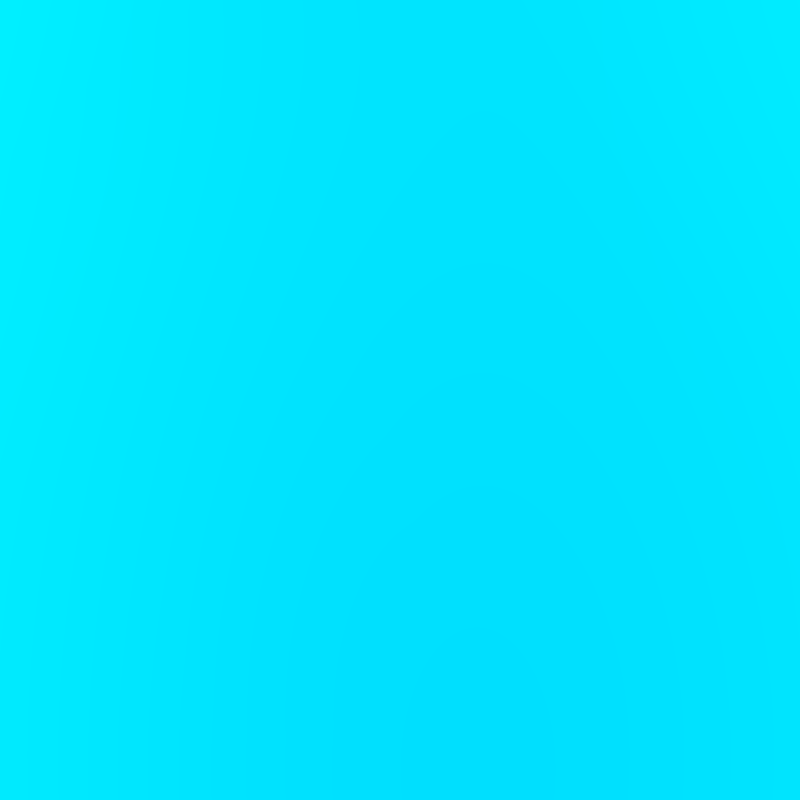}
        & \includegraphics[width=0.72\linewidth]{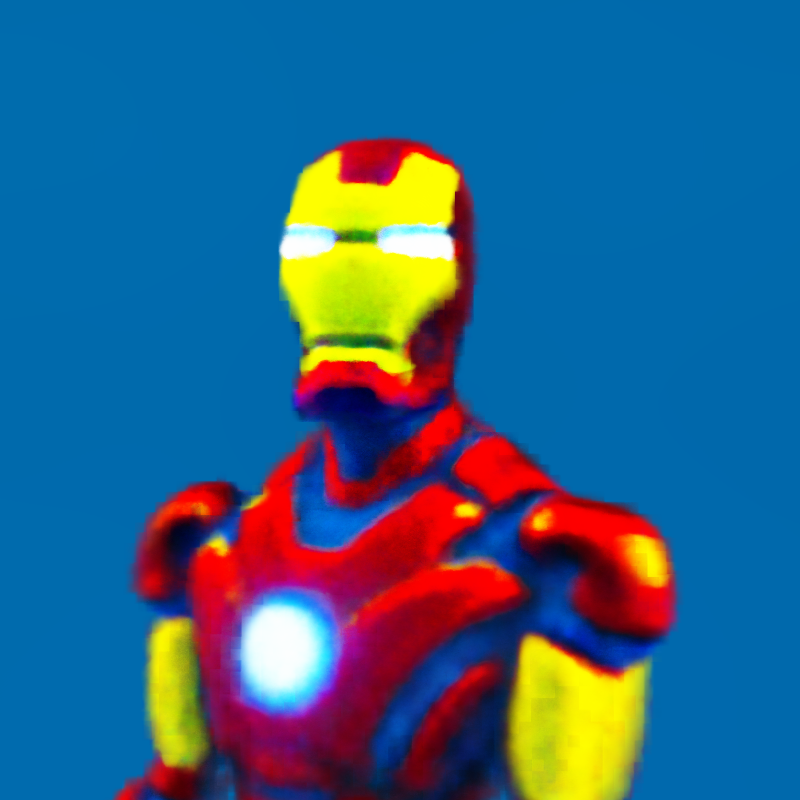}\\
         & Similarity: 28.9\% & Similarity: 23.4\% & Similarity: 26.1\% & Similarity: 23.2\% & Similarity: 31.0\%\\
         & Iterations: 5k & Iterations: 5k & Iterations: 5k & Iterations: 40k & Iterations: 15.5k\\\hline

        \vspace{0.25in}  
        {\it ``A burning torch.''}  
        & \vspace{0.05in} \includegraphics[width=0.72\linewidth]{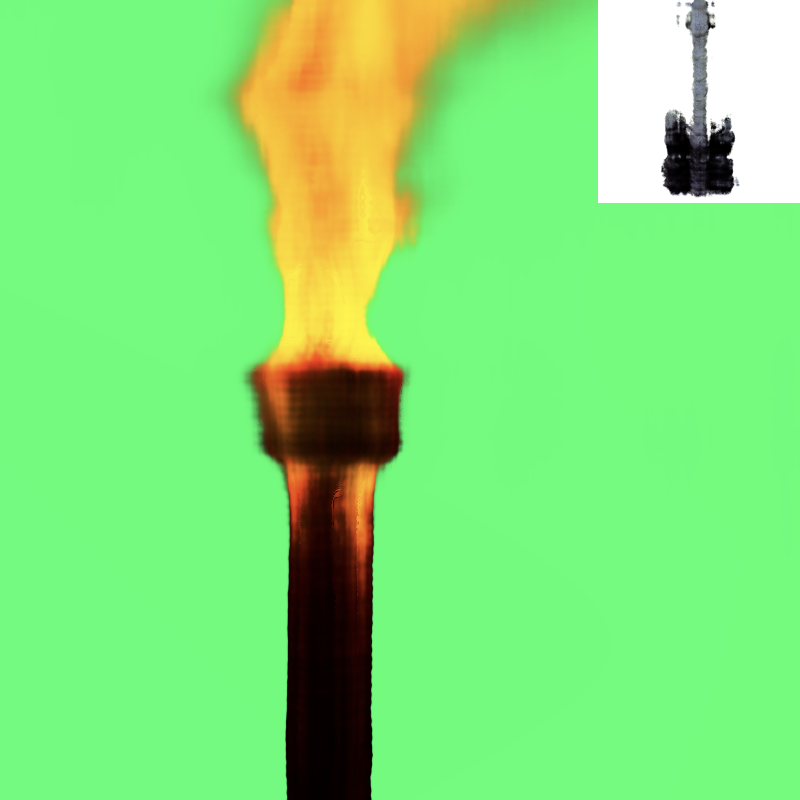}
        & \includegraphics[width=0.72\linewidth]{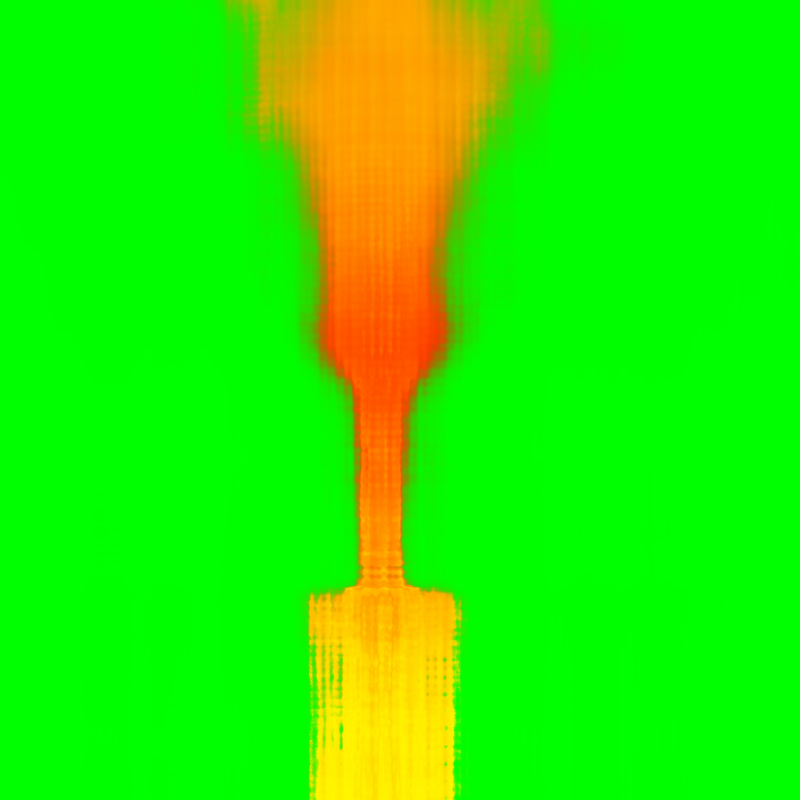}
        & \includegraphics[width=0.72\linewidth]{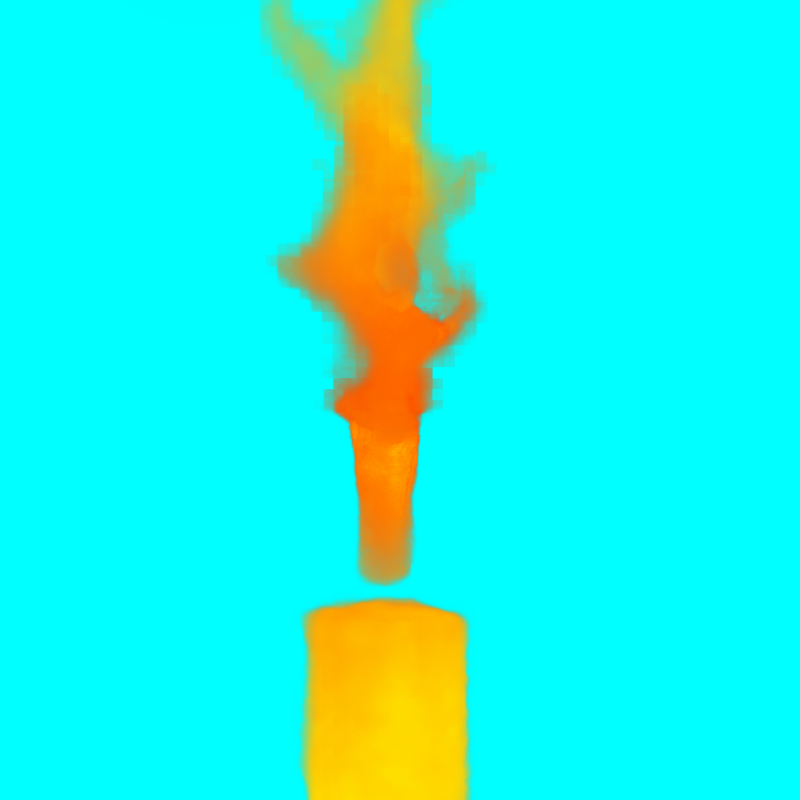}
        & \includegraphics[width=0.72\linewidth]{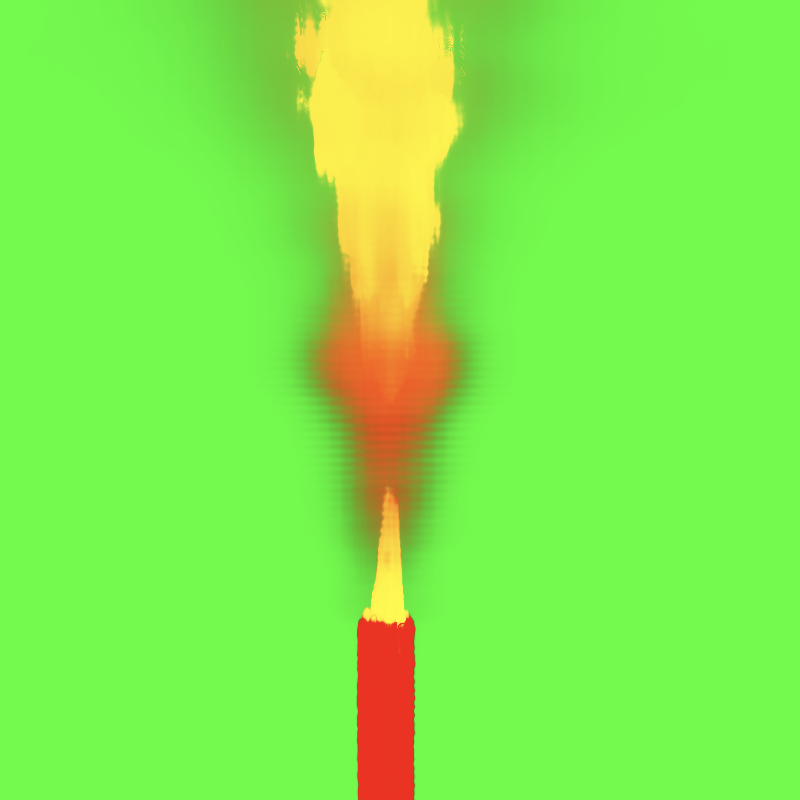}
        & \includegraphics[width=0.72\linewidth]{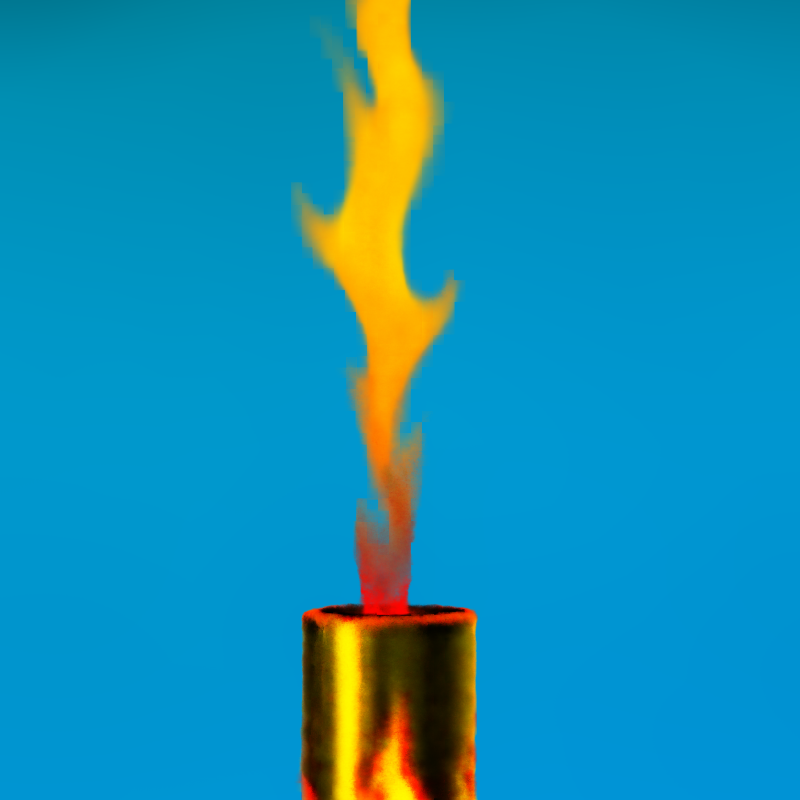}\\
         & Similarity: 32.2\% & Similarity: 30.1\% & Similarity: 29.7\% & Similarity: 32.3\% & Similarity: 31.4\%\\
         & Iterations: 7k & Iterations: 7k & Iterations: 7k & Iterations: 30k & Iterations: 19k\\\hline

        \vspace{0.25in}  
        {\it ``A basketball.''}  
        & \vspace{0.05in} \includegraphics[width=0.72\linewidth]{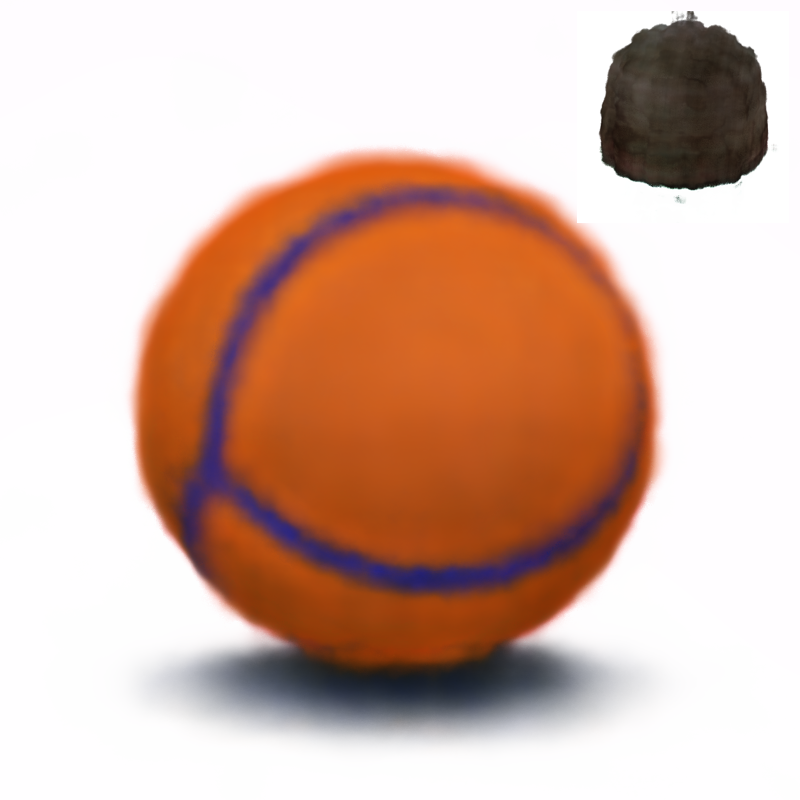}
        & \includegraphics[width=0.72\linewidth]{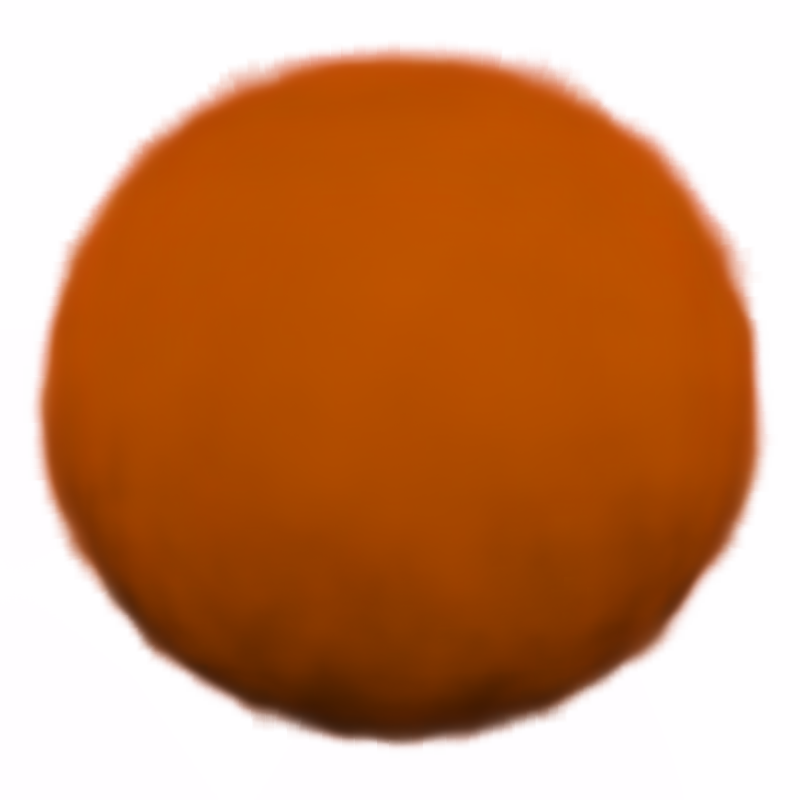}
        & \includegraphics[width=0.72\linewidth]{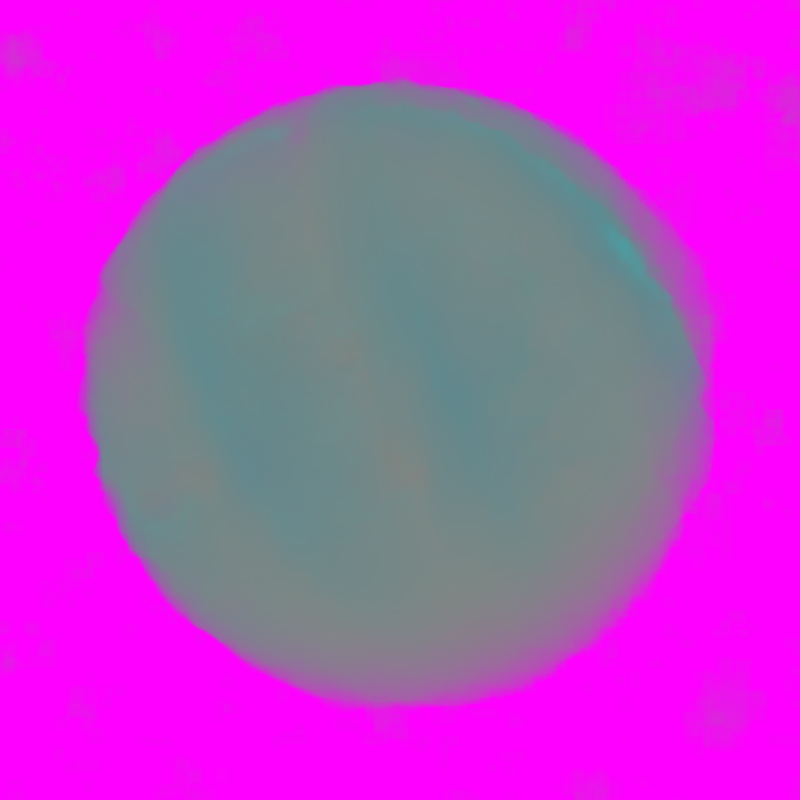}
        & \includegraphics[width=0.72\linewidth]{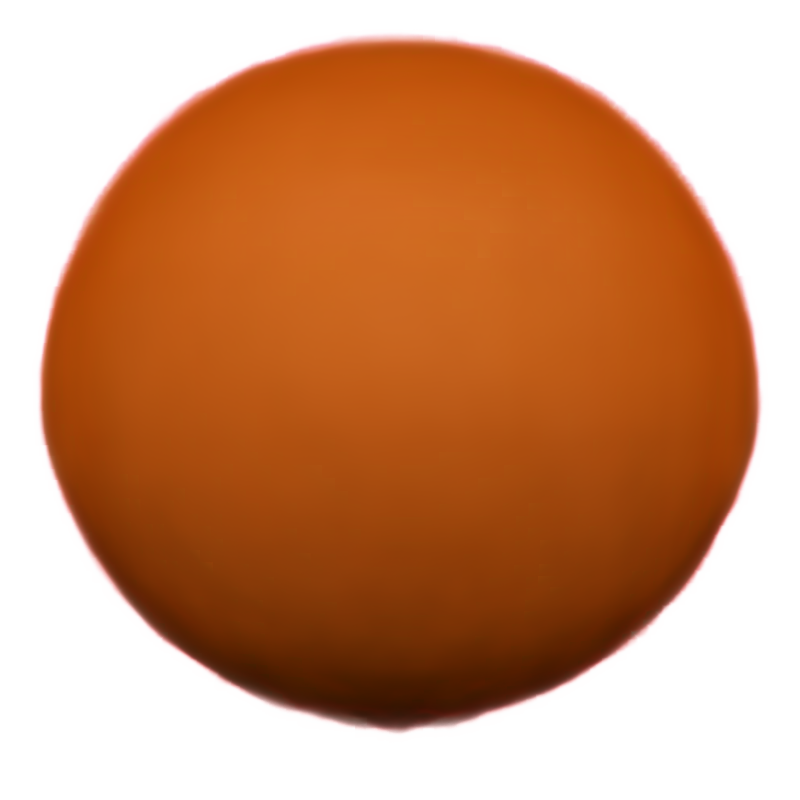}
        & \includegraphics[width=0.72\linewidth]{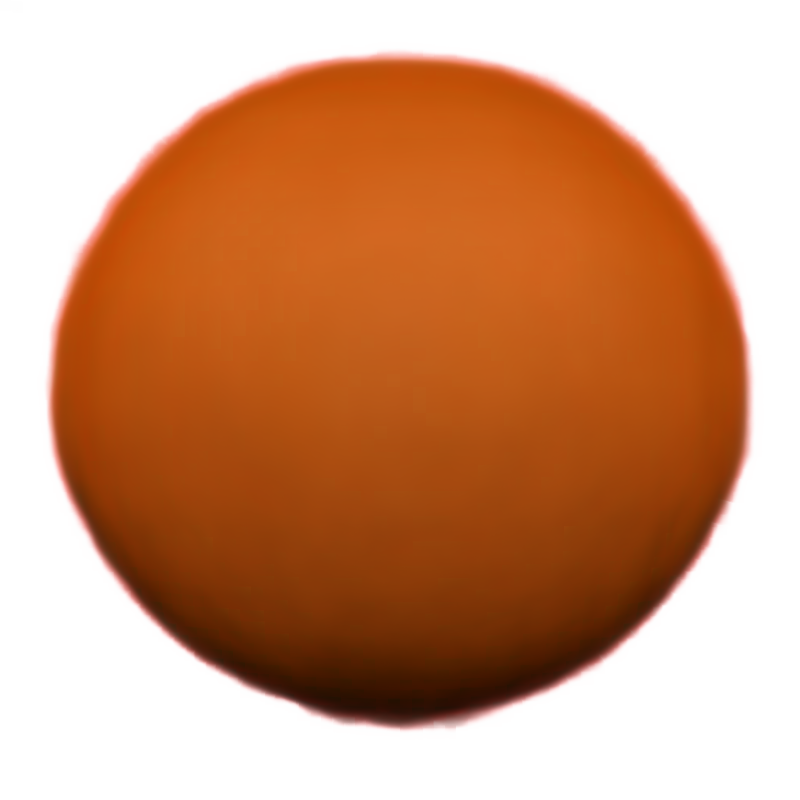}\\
         & Similarity: 30.4\% & Similarity: 26.1\% & Similarity: 24.1\% & Similarity: 27.3\% & Similarity: 27.4\%\\
         & Iterations: 7.4k & Iterations: 7.4k & Iterations: 7.4k & Iterations: 18k & Iterations: 19.5k\\
    \bottomrule
    \end{tabular}
    \caption{More examples of {\bf out-of-distribution inference result, using text prompts}. For each prompt, we record the convergence iteration count using our initialization zero-shot, from one single forward pass (top-right inset). Scratch (S) or pre-trained (P) DreamFusion are respectively trained for the same number of iterations and until convergence, with their respective results  displayed above.
    Our method significantly accelerates the DreamFusion's optimization, achieving  3 to 5 times speed boost while almost consistently yielding better results. Moreover, our method provides a semantically meaningful initialization, as pretrained (P) DreamFusion still takes an extremely long time to converge though initialized with a similar scene.}
    \label{tab:more-eg-ood-text}
    \vspace{-0.2in}
\end{table*}

\section{Interpolation Results}
\label{sec:supp-interpolation}

\begin{figure*}
    \centering
    \includegraphics[width=0.48\linewidth, trim={2cm 2cm 2cm 2cm},clip]{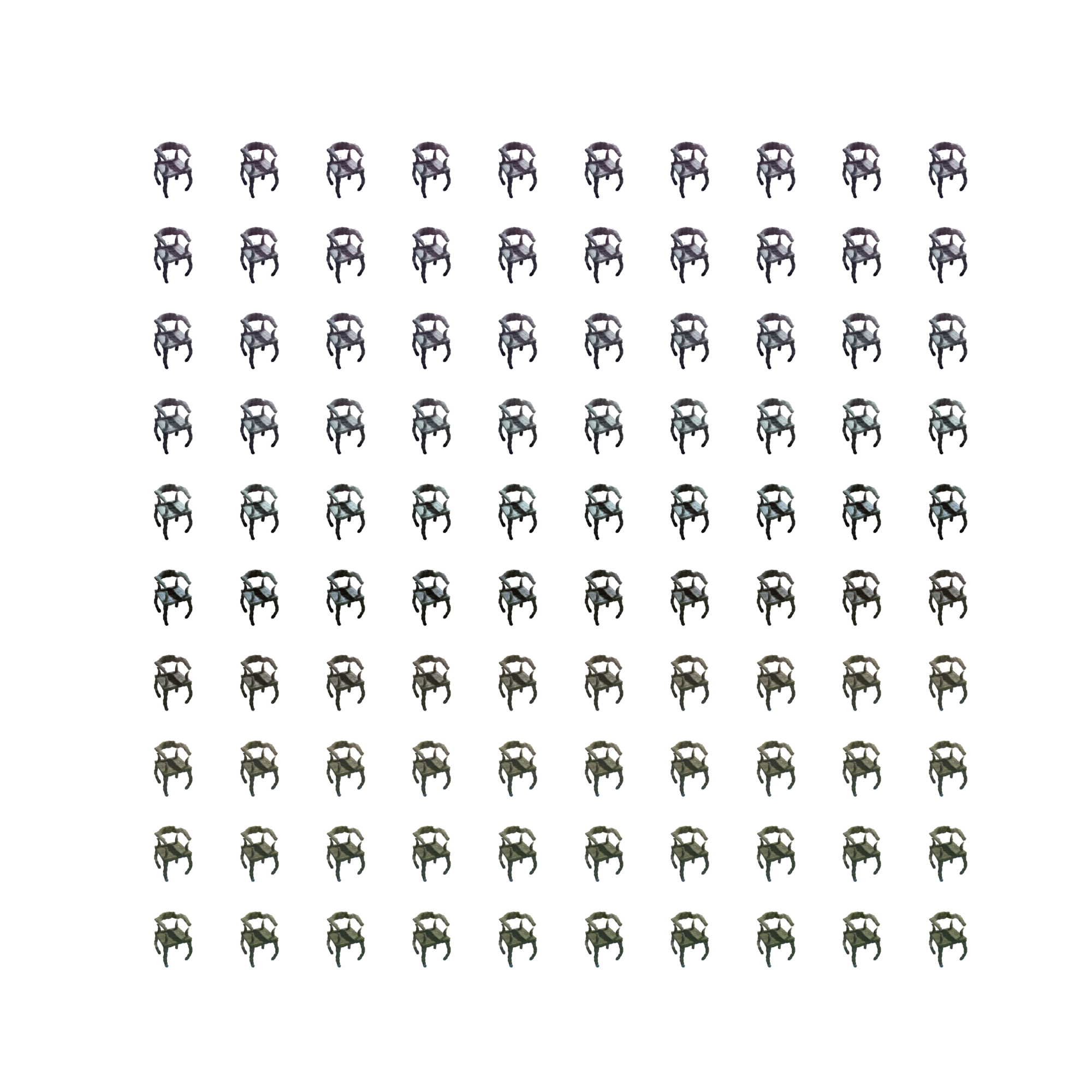}
    \includegraphics[width=0.48\linewidth, trim={2cm 2cm 2cm 2cm},clip]{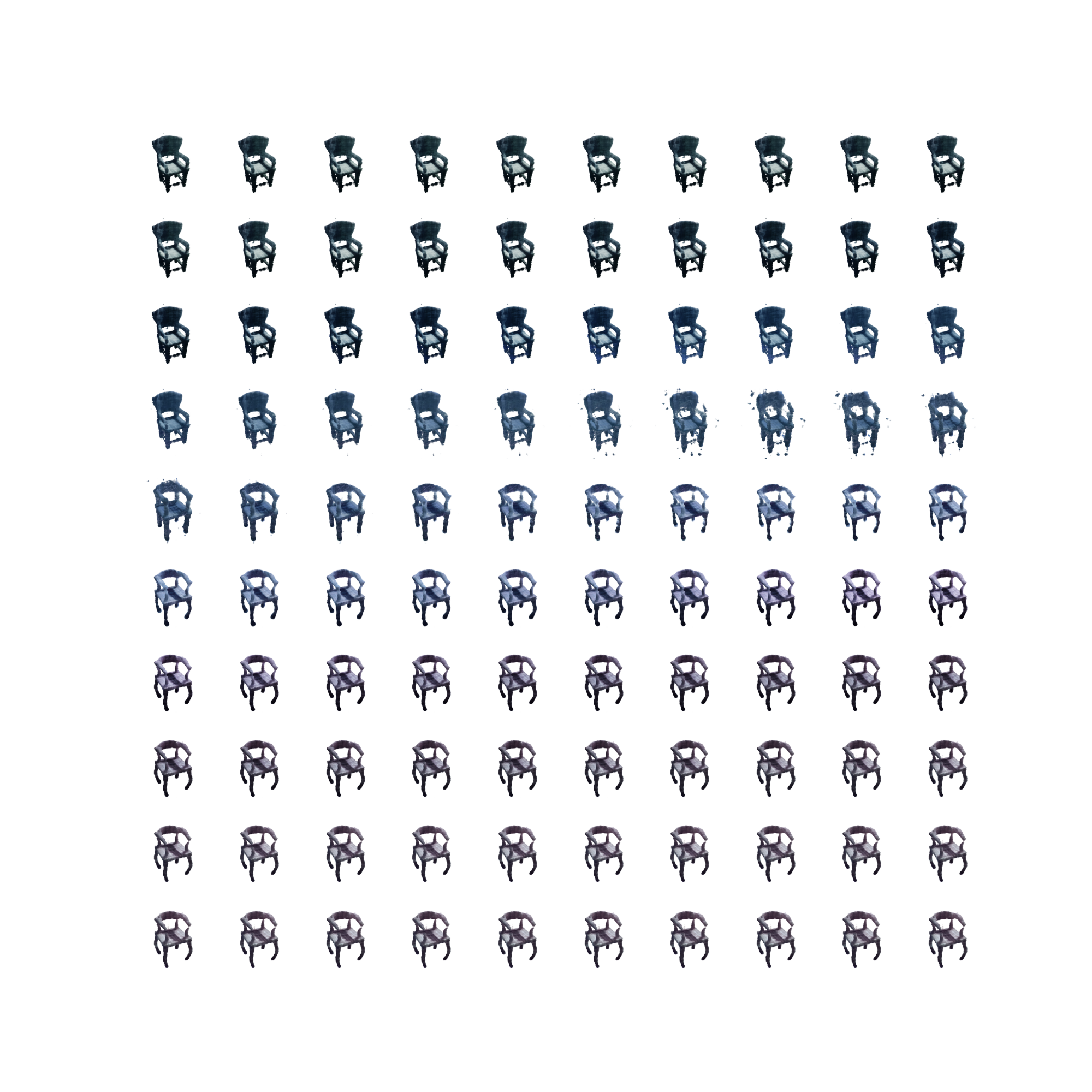}
    \caption{Interpolation results, with two objects differ only in color (left) and differ both in color and shape (right). Each time we slightly increase the weight of the first object for interpolation, decrease the weight of the second object by 0.01, and organize the results row by row, which allows us to observe a smooth interpolation result.}
    \label{fig:interpolation}
\end{figure*}

We continue the discussion in ~\cref{sec:results-outdomain}
to demonstrate that our method has learned a semantically meaningful alignment by displaying some interpolation results, thereby providing a valid explanation of why the initializations given by our method yield much faster convergence compared with the pre-trained baseline (P), which is also trained to a semantically close scene.
The results in~\cref{fig:interpolation} demonstrate that we can get not only smooth interpolation between objects with a single feature, like colors, but also a smooth morph between two objects with a difference in both geometry and color.

\section{Comparisons on Convergence Speed}
\label{sec:supp-converge-curve}

As previous results only provide the number of iterations required for convergence when demonstrating the effective acceleration ability of our initialization, here we include examples of the curves comparing the optimization process of the baseline model (Zero-1-to-3 for image-to-NeRF tasks and DreamFusion for text-to-NeRF tasks) with/without our initialization.

Specifically, after training every 100 iterations (or one epoch), we render $m$ images $I_{1}^{(i)}, I_{2}^{(i)}, \cdots, I_{m}^{(i)}$ for epoch $i$, from viewpoints evenly divided from $360^\circ$, and compute the pixel-wise mean-squared difference $D_i$ between the $m$ images rendered in the $i$-th epoch and the previous $(i-1)$-th epoch. This metric can be formulated as
\begin{align*}
    D_i &= \sum_{j=1}^{m} \left\| I_{j}^{(i)} - I_{j}^{(i-1)} \right\|_2^2,\quad {\rm for\ } i > 1\\
    D_1 &= 0
\end{align*}
where by taking the difference of two images $I$ we mean the pixel-wise difference.

We then consider the optimization process is converged if
$D_i<\epsilon$ for continuous $20$ epochs.

\begin{figure*}[h]
    \centering
    \includegraphics[width=0.48\textwidth]{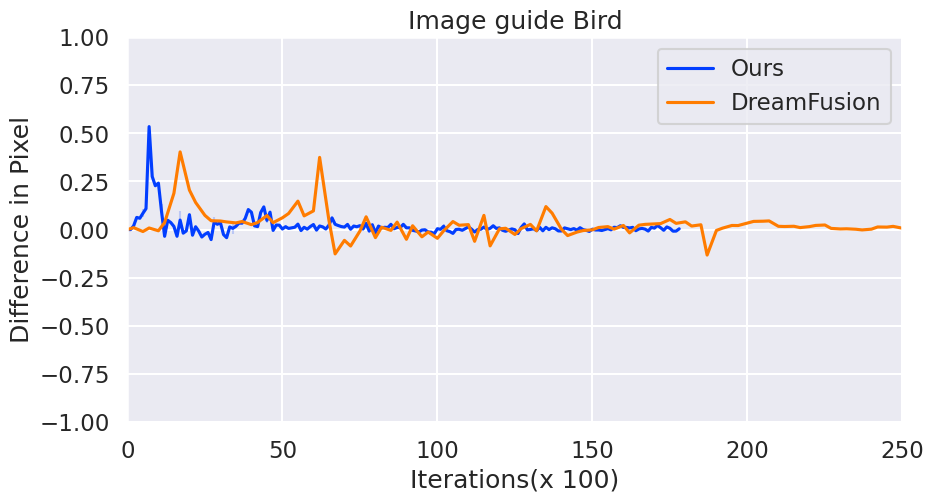}
    \includegraphics[width=0.48\textwidth]{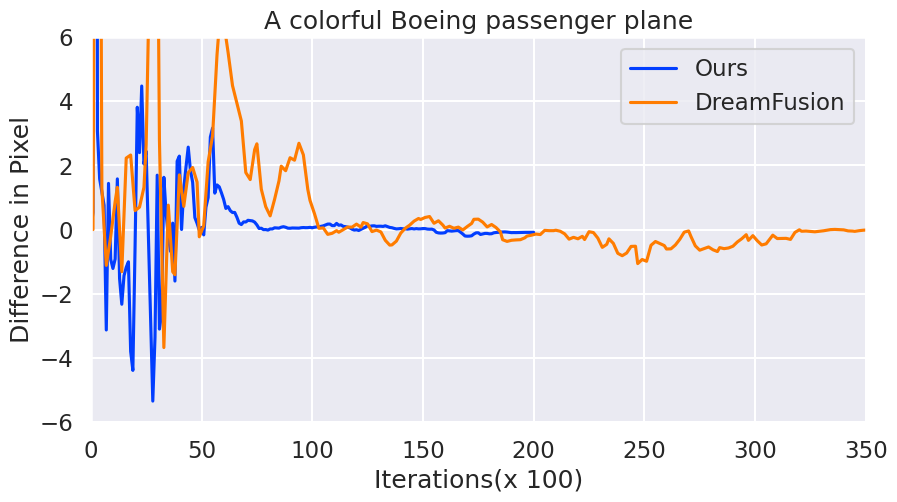}
    \caption{Comparisons of optimization convergence curves with (Ours) / without (DreamFusion) our initialization, when using image (left) and text (right) prompts for inference.}
    \label{fig:supp-converge-curve}
\end{figure*}

\Cref{fig:supp-converge-curve} shows two examples of the convergence curves with $m=8$ and $\epsilon=0.01$, using image prompt and text prompt, respectively. We may clearly observe the acceleration that our initialization brings to the baseline.

\section{Video \& Multi-view Results of Generated Scenes}
\label{sec:supp-video}

In this section, we provide some video results\footnote{will release later} of the NeRF scene we generated, as well as the 8 multi-view images used to generate the videos, demonstrating that our implicit generation method yields satisfying view-consistent NeRF scenes.

\begin{figure*}[ht]
    \centering
    \includegraphics[width=0.12\textwidth]{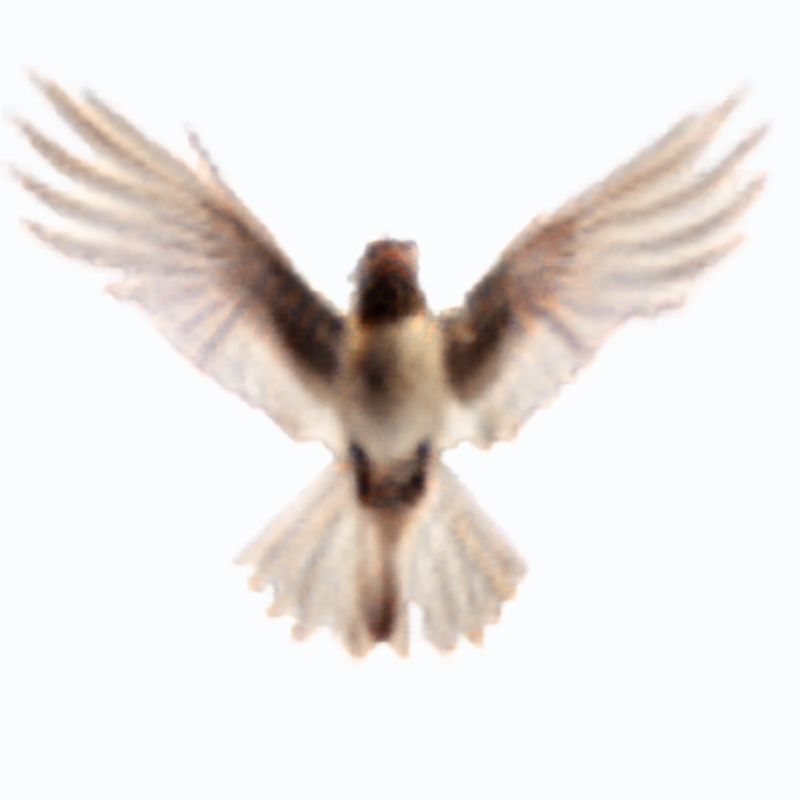}
    \includegraphics[width=0.12\textwidth]{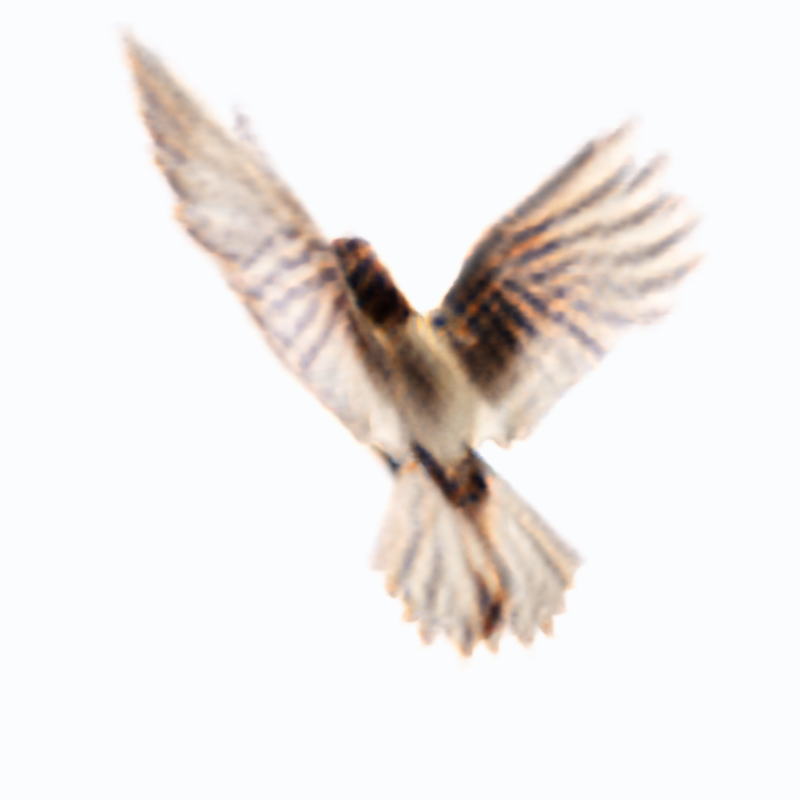}
    \includegraphics[width=0.12\textwidth]{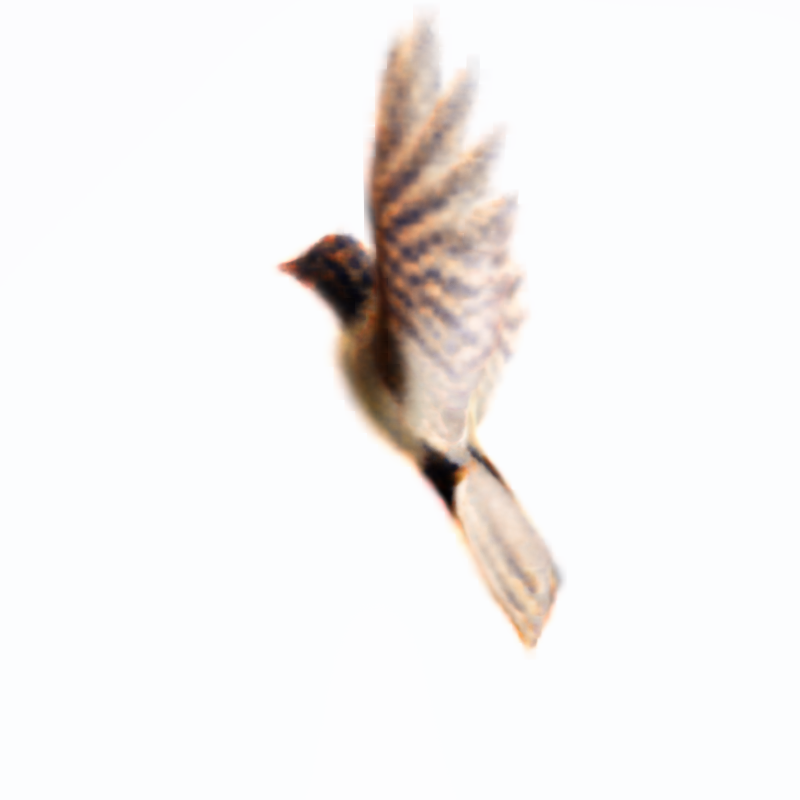}
    \includegraphics[width=0.12\textwidth]{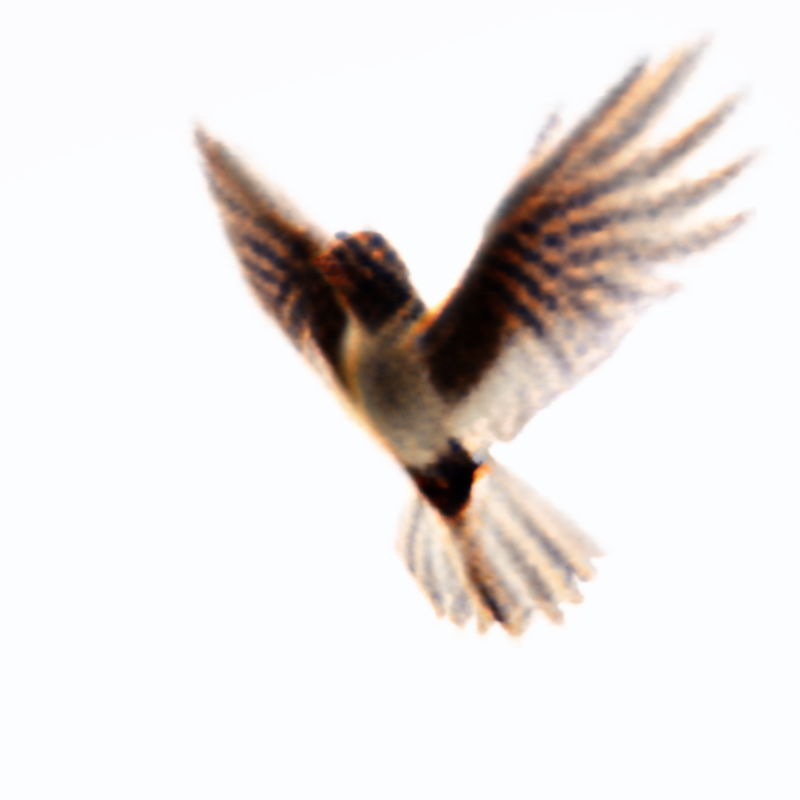}
    \includegraphics[width=0.12\textwidth]{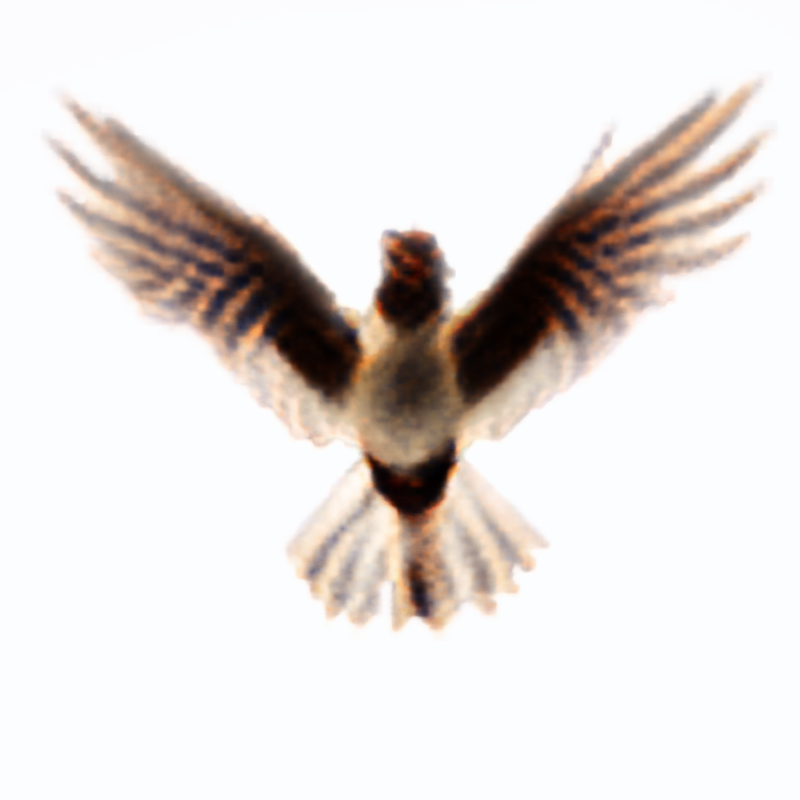}
    \includegraphics[width=0.12\textwidth]{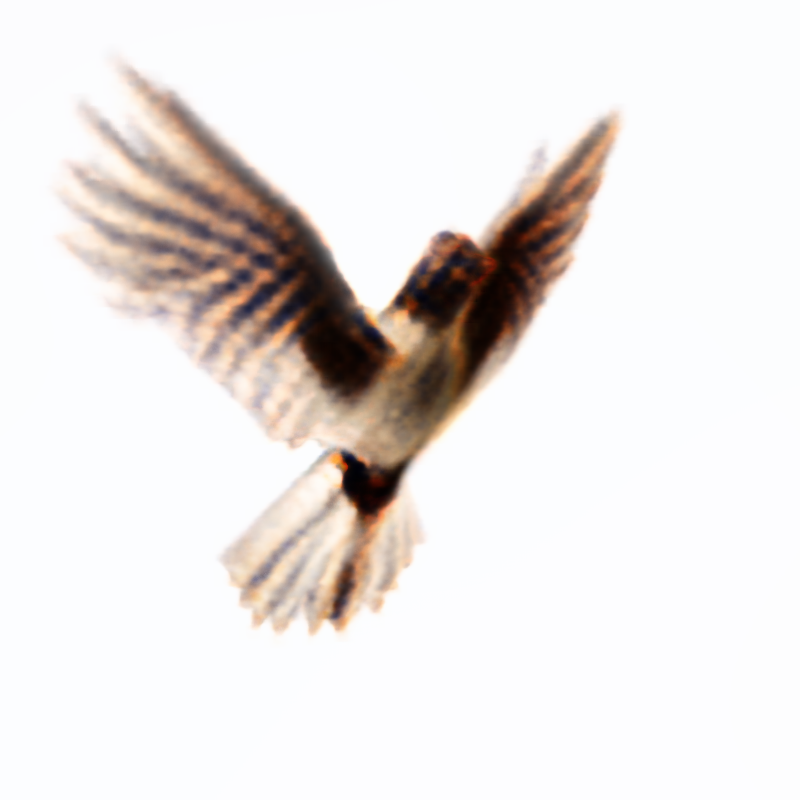}
    \includegraphics[width=0.12\textwidth]{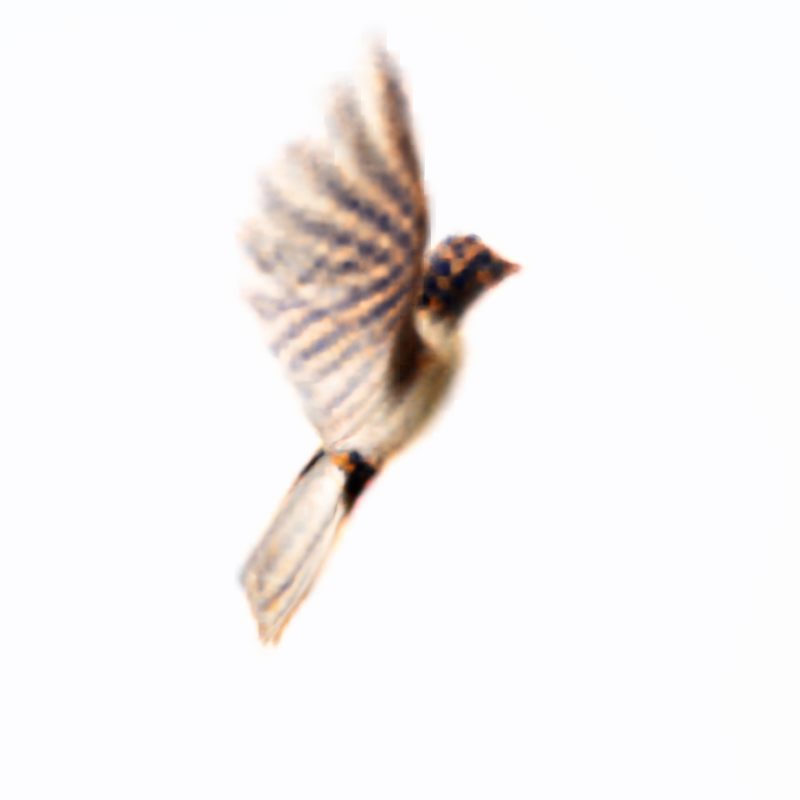}
    \includegraphics[width=0.12\textwidth]{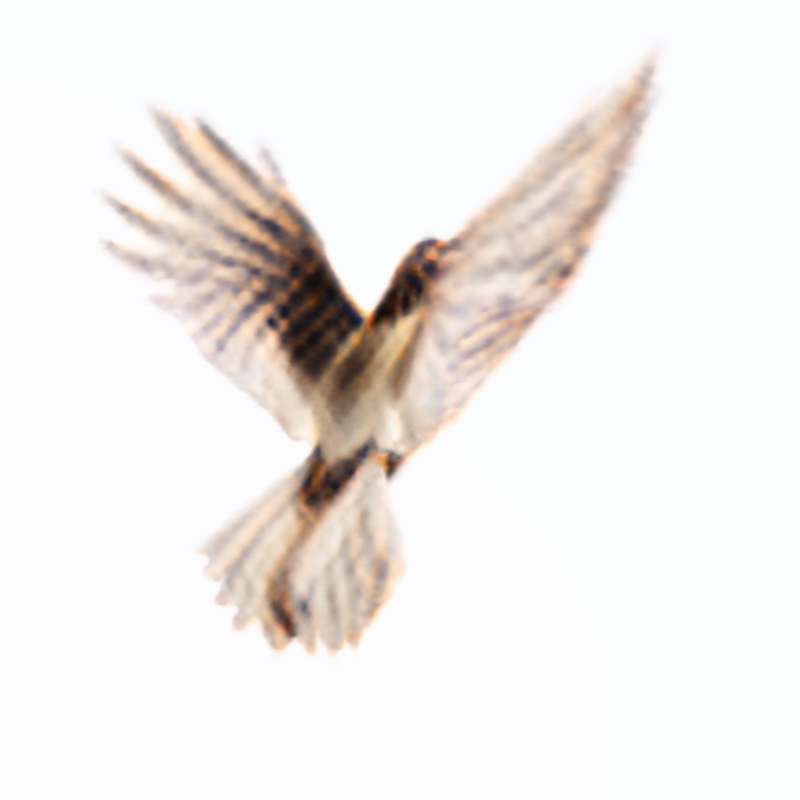}\\
    \includegraphics[width=0.12\textwidth]{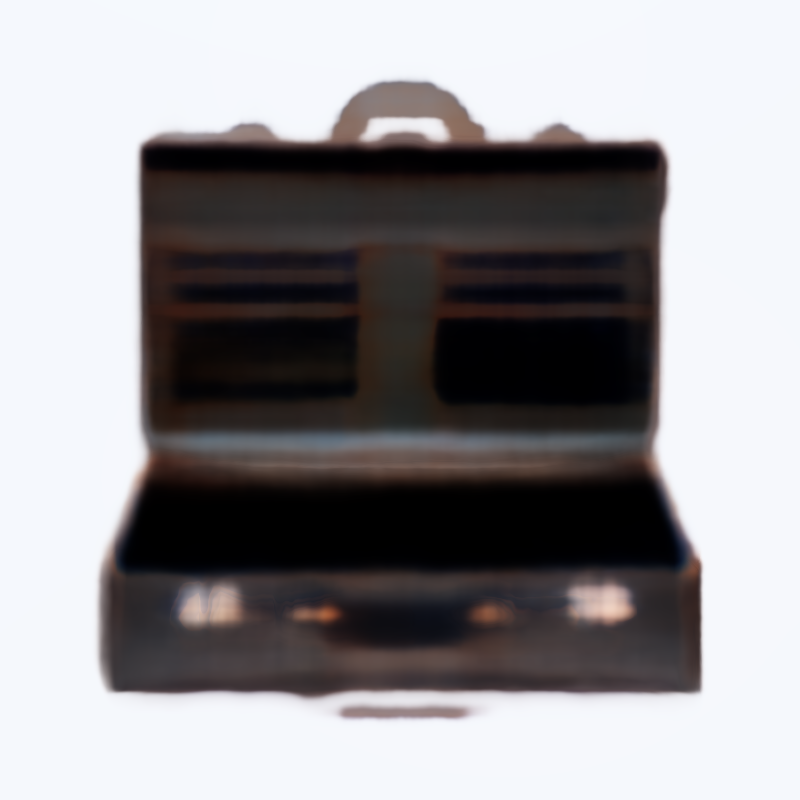}
    \includegraphics[width=0.12\textwidth]{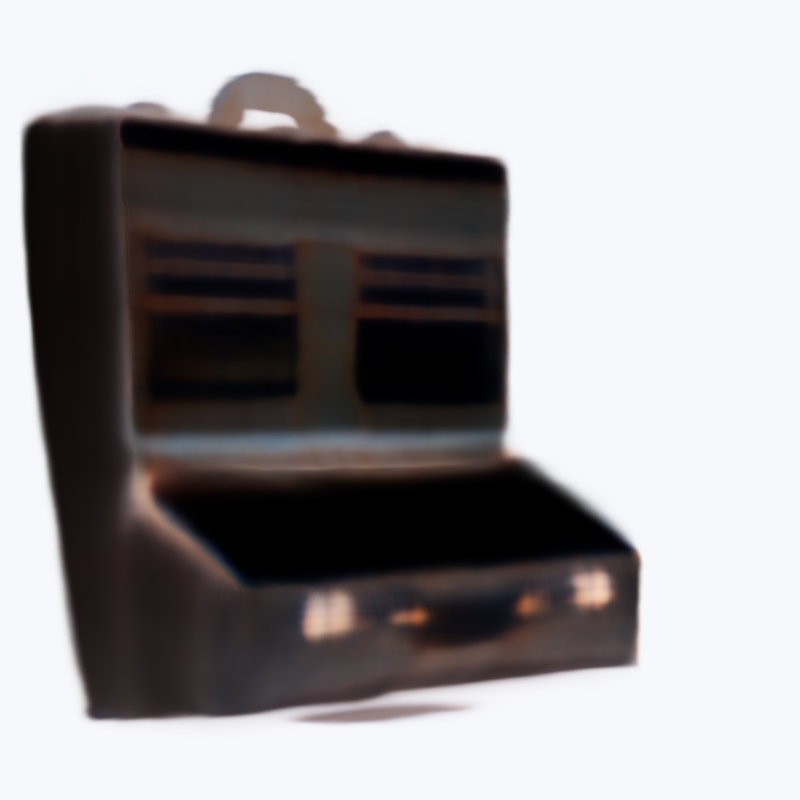}
    \includegraphics[width=0.12\textwidth]{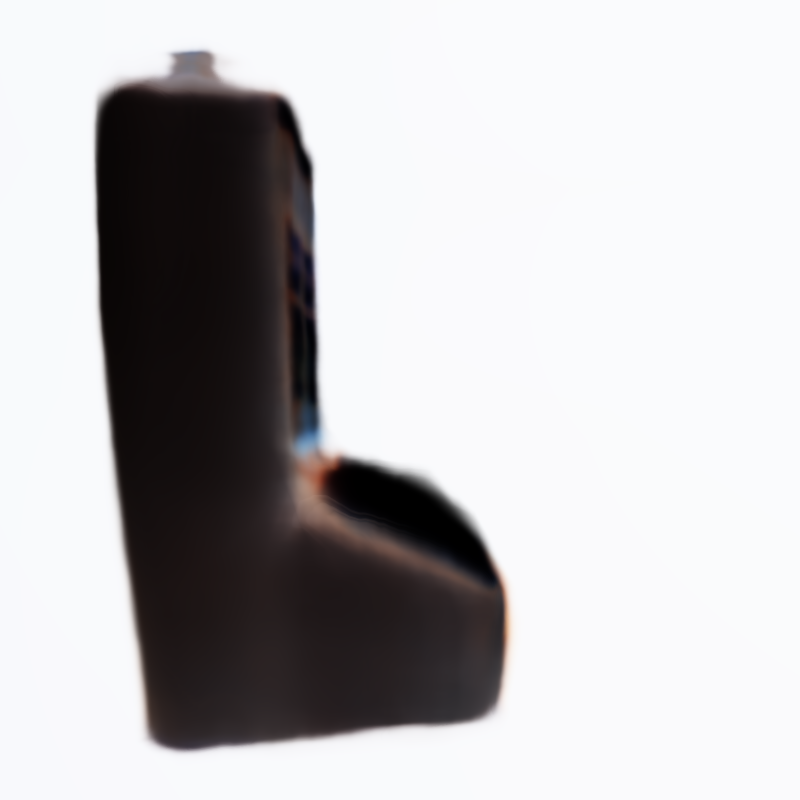}
    \includegraphics[width=0.12\textwidth]{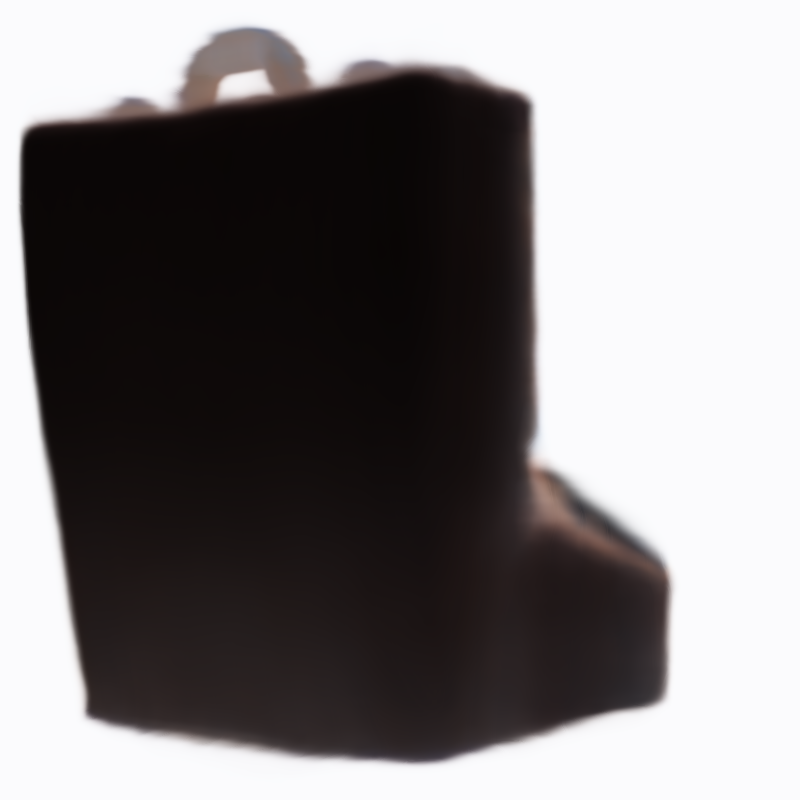}
    \includegraphics[width=0.12\textwidth]{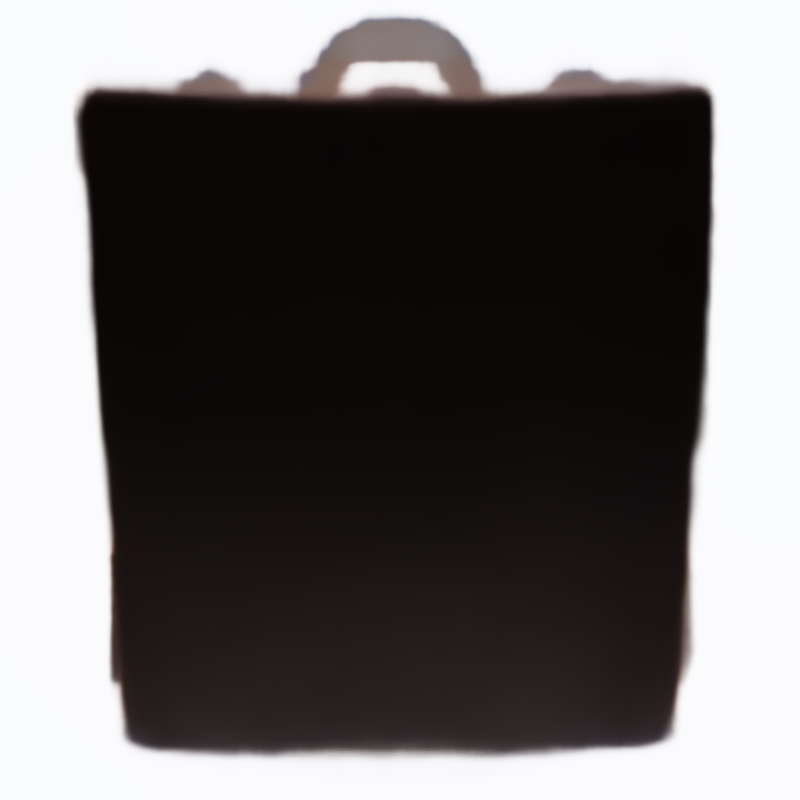}
    \includegraphics[width=0.12\textwidth]{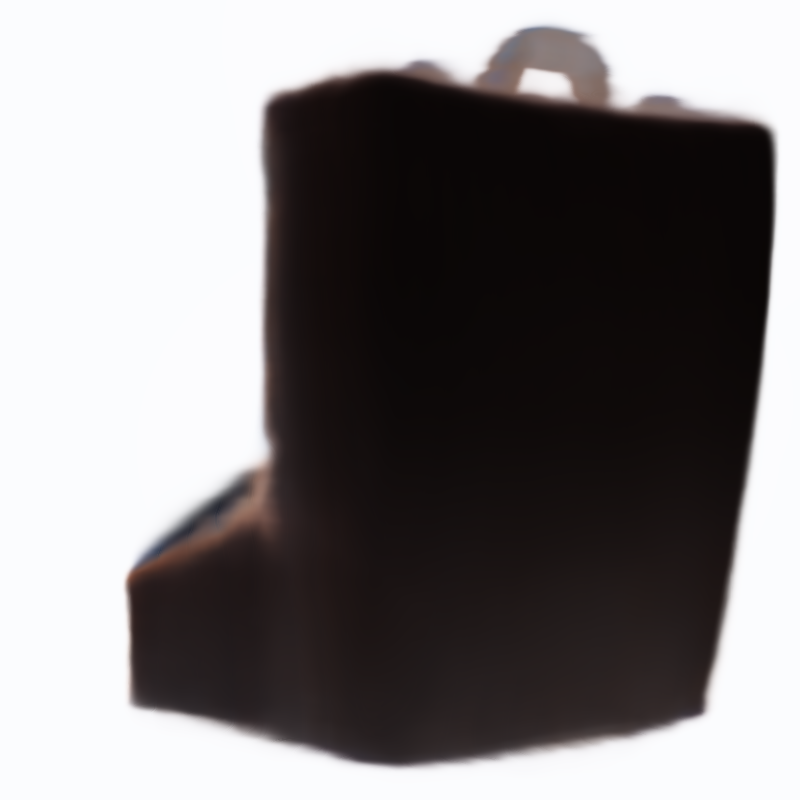}
    \includegraphics[width=0.12\textwidth]{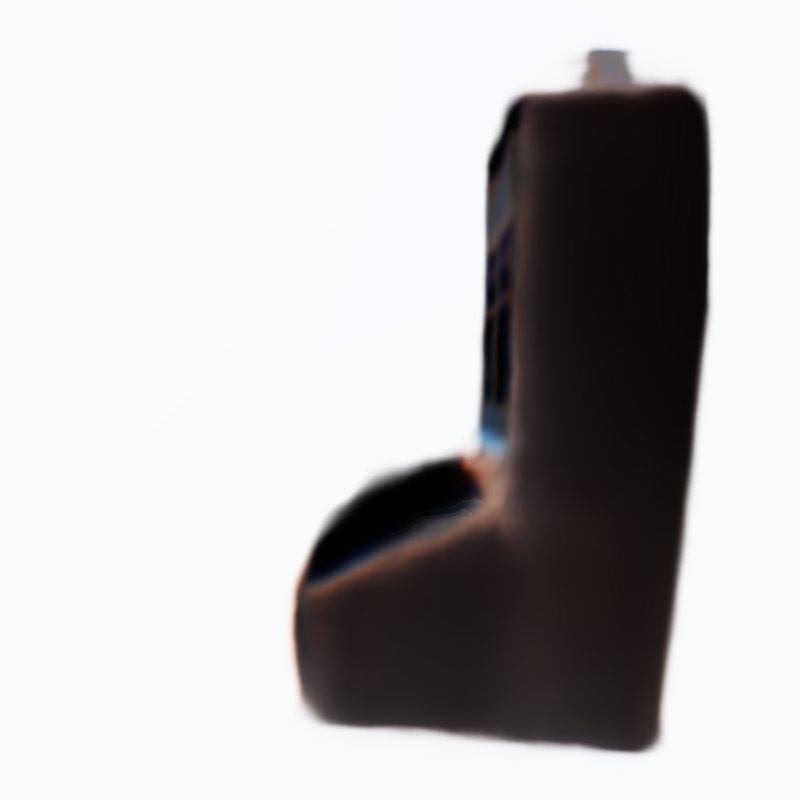}
    \includegraphics[width=0.12\textwidth]{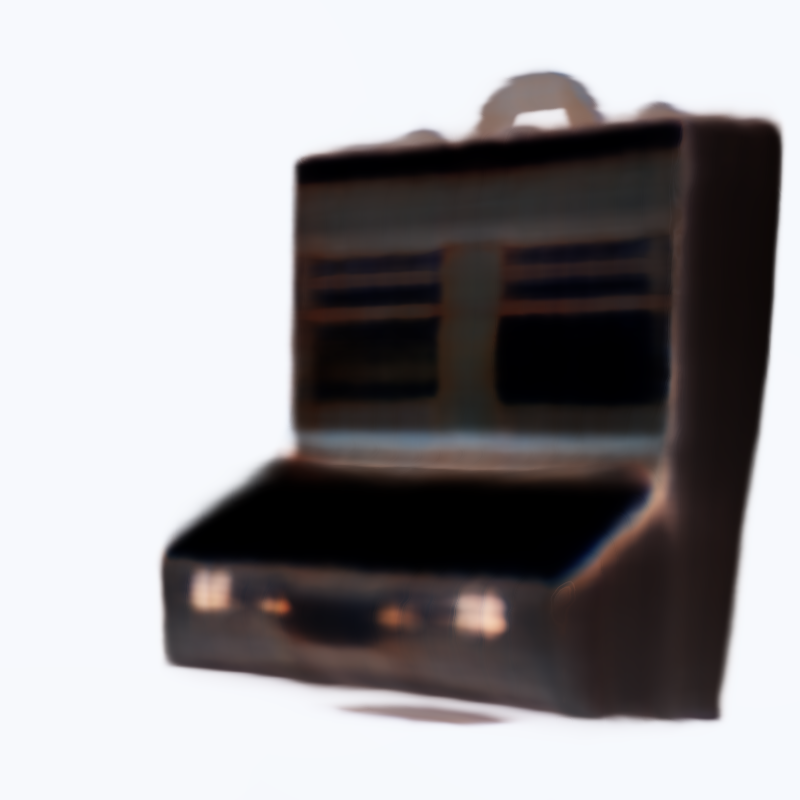}\\
    \includegraphics[width=0.12\textwidth]{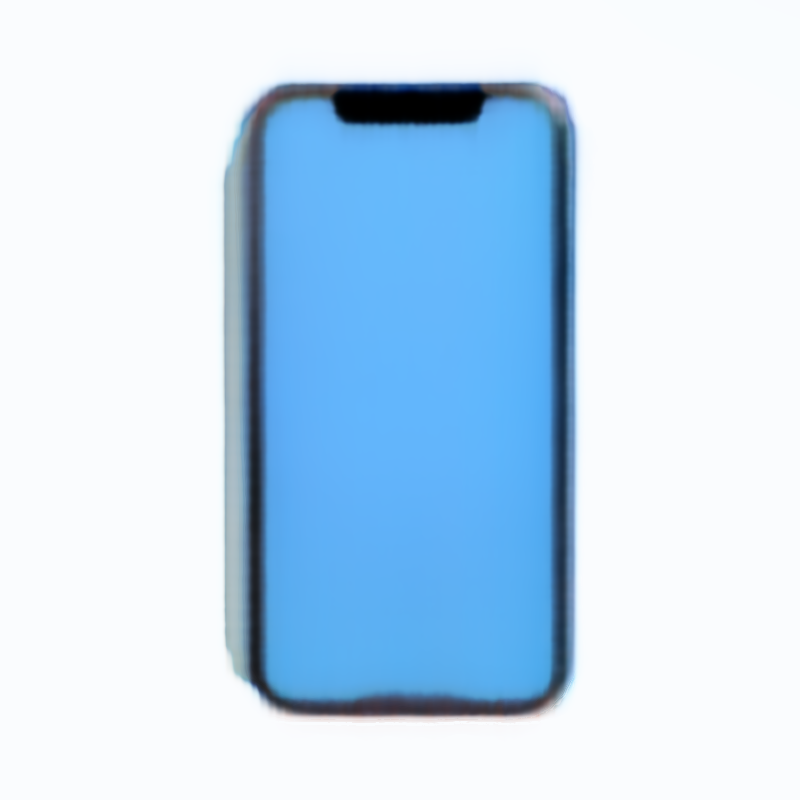}
    \includegraphics[width=0.12\textwidth]{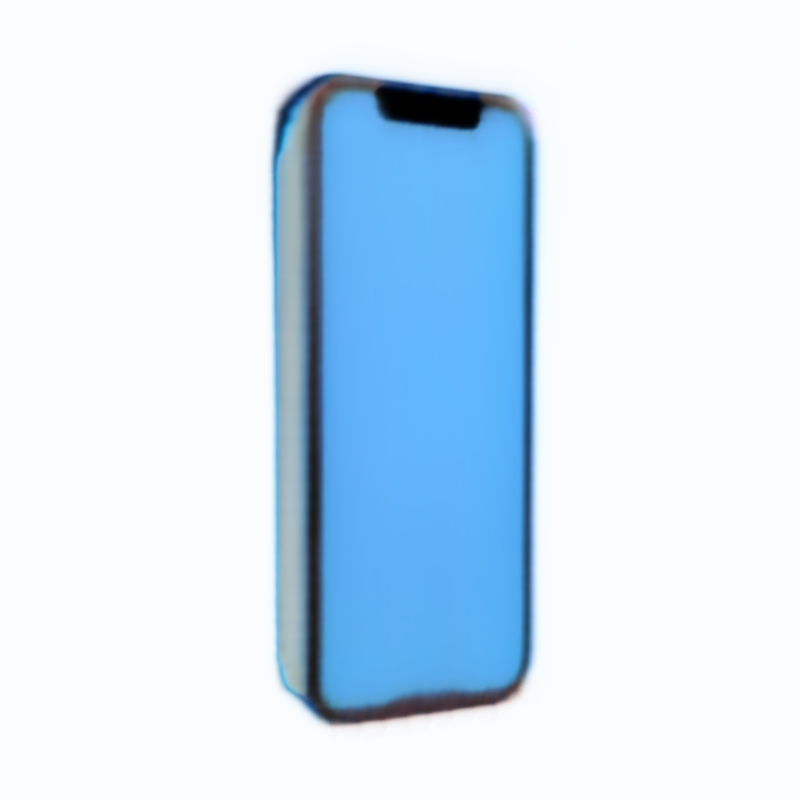}
    \includegraphics[width=0.12\textwidth]{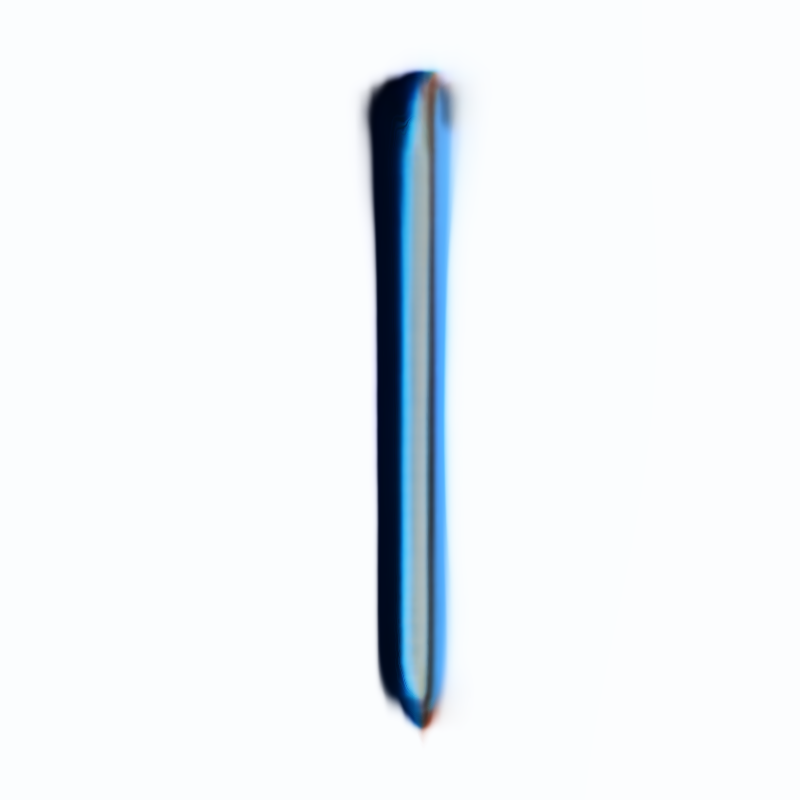}
    \includegraphics[width=0.12\textwidth]{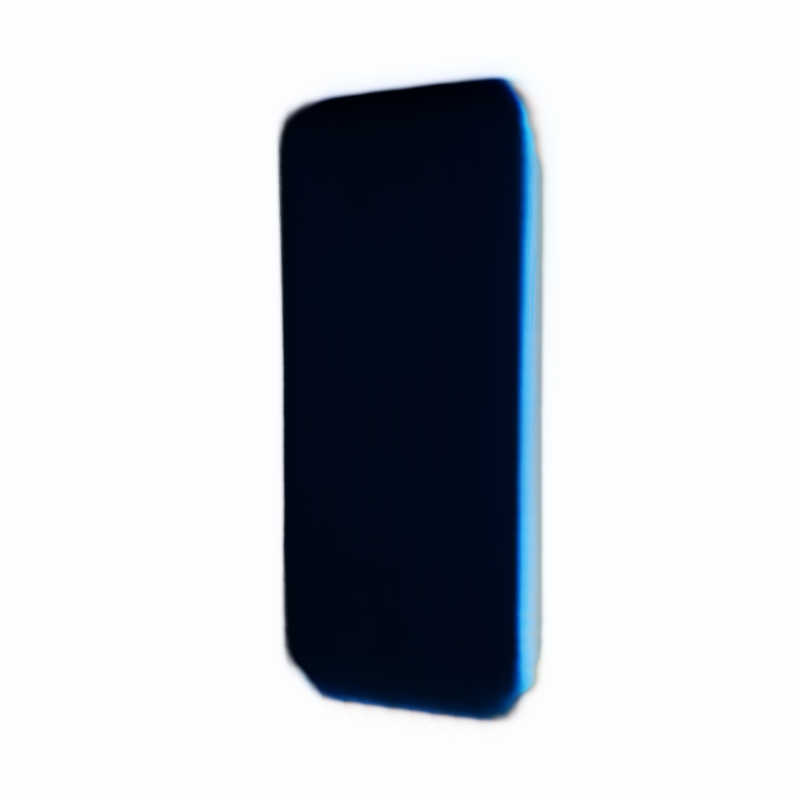}
    \includegraphics[width=0.12\textwidth]{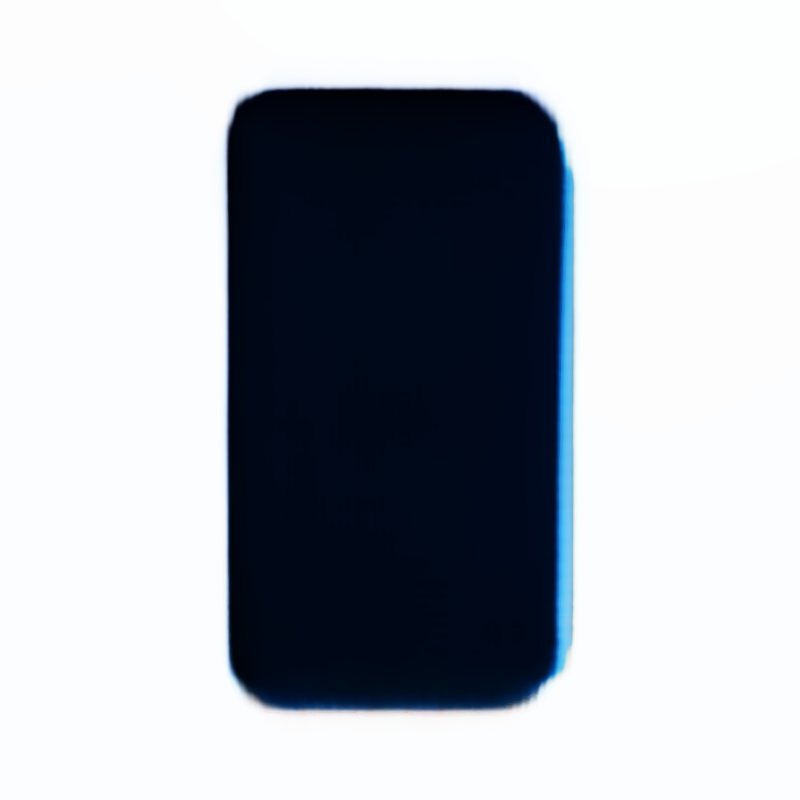}
    \includegraphics[width=0.12\textwidth]{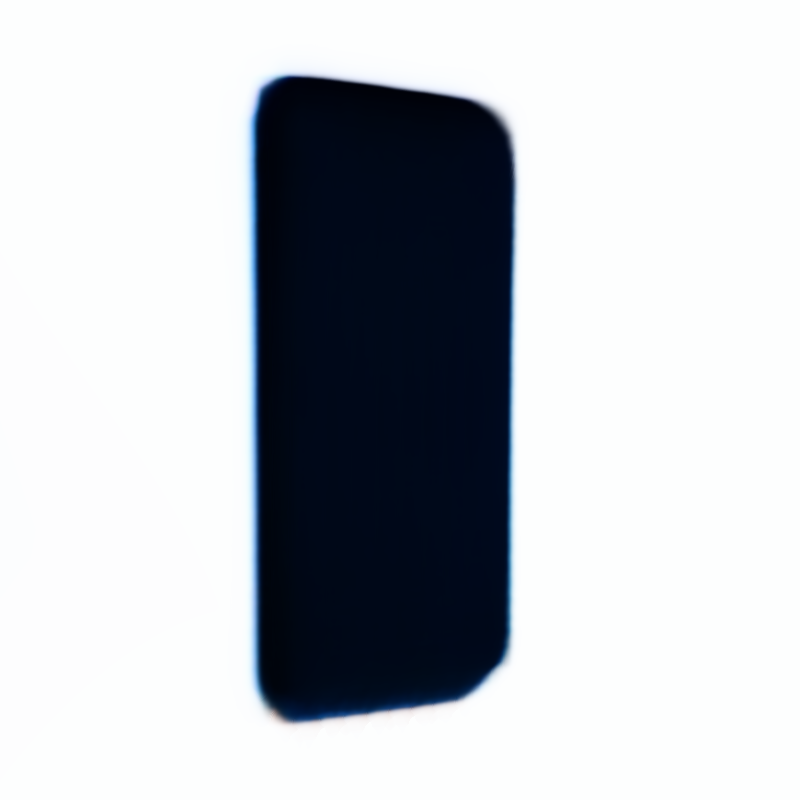}
    \includegraphics[width=0.12\textwidth]{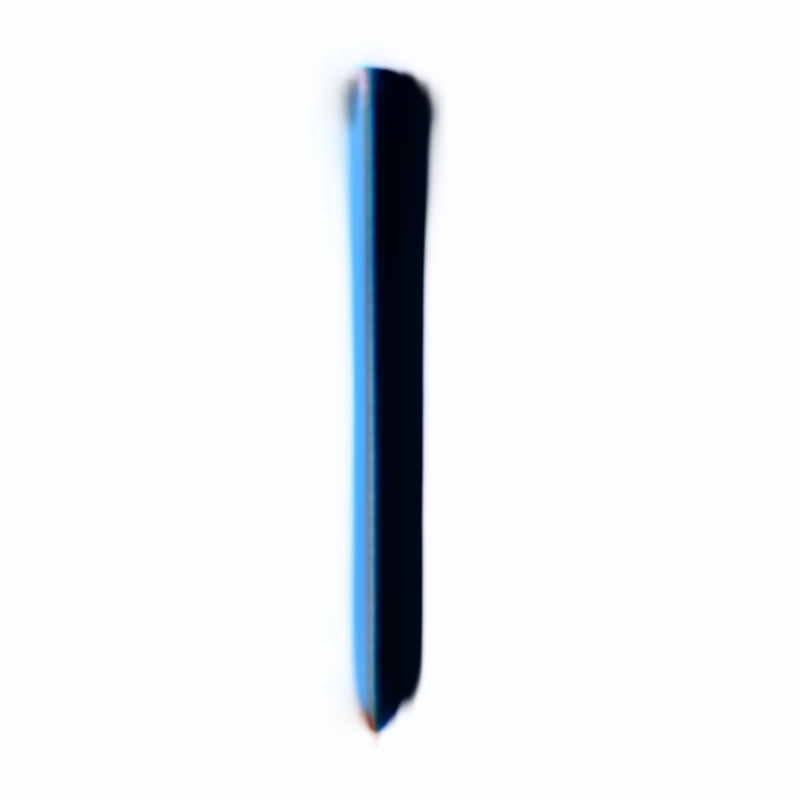}
    \includegraphics[width=0.12\textwidth]{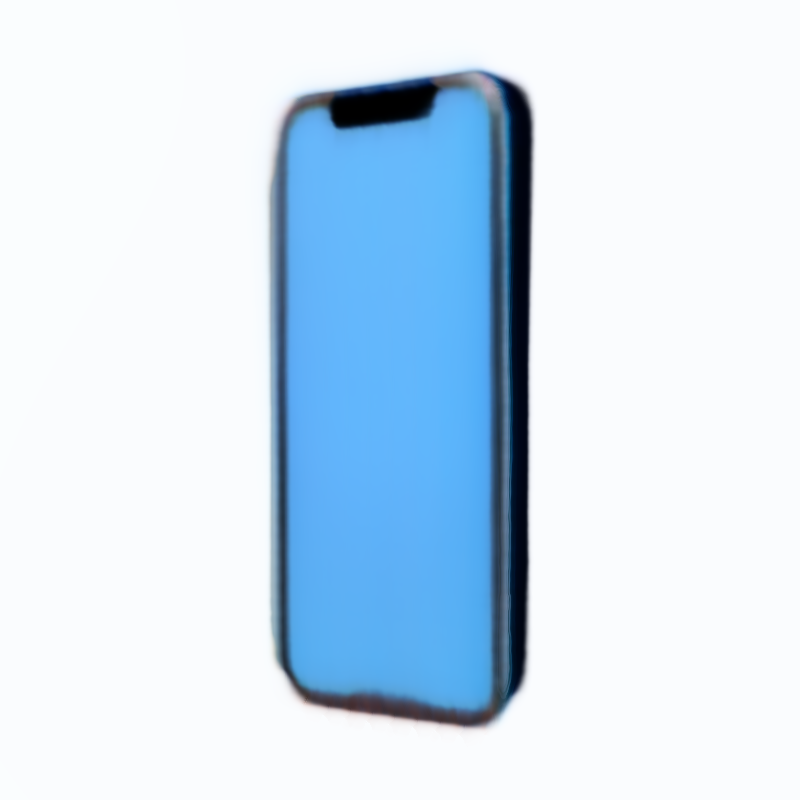}\\
    \includegraphics[width=0.12\textwidth]{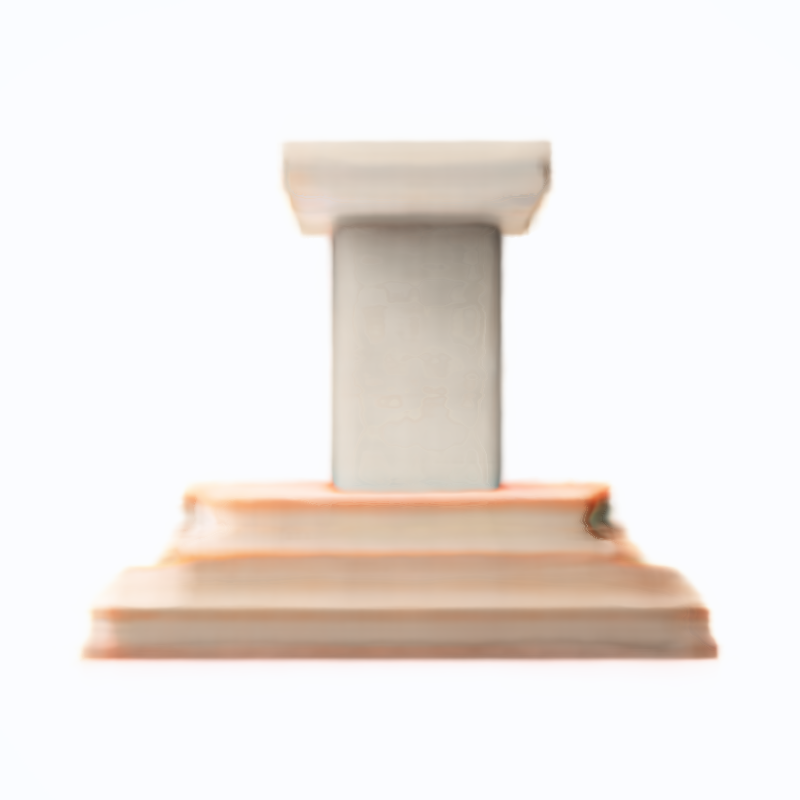}
    \includegraphics[width=0.12\textwidth]{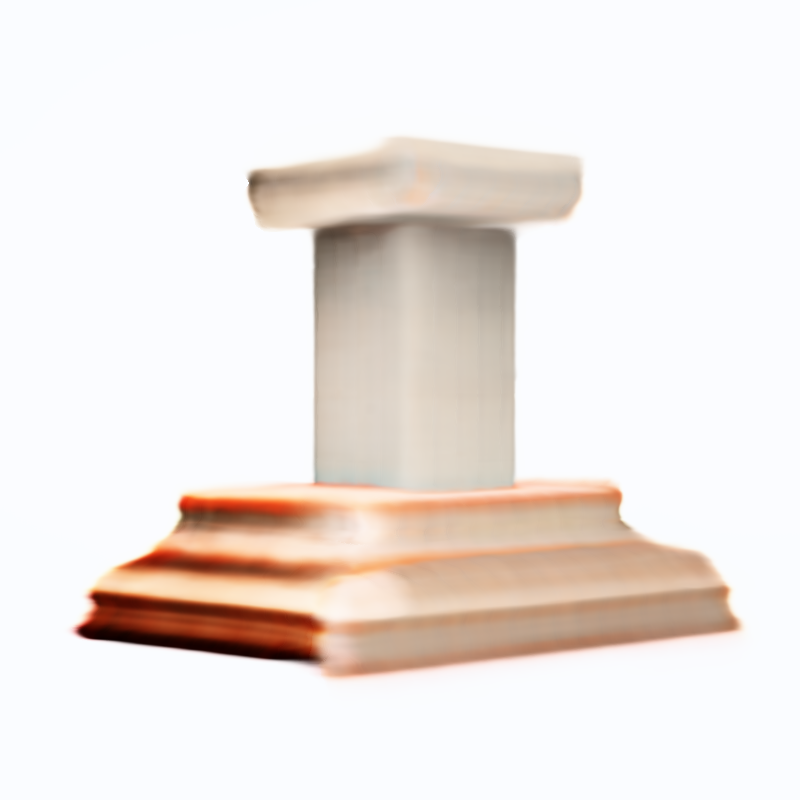}
    \includegraphics[width=0.12\textwidth]{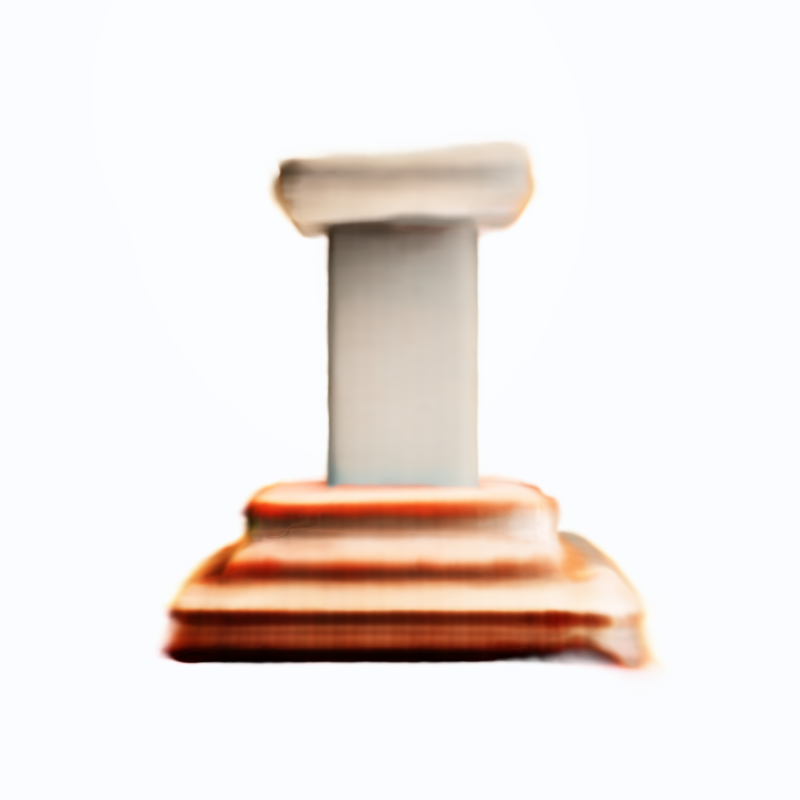}
    \includegraphics[width=0.12\textwidth]{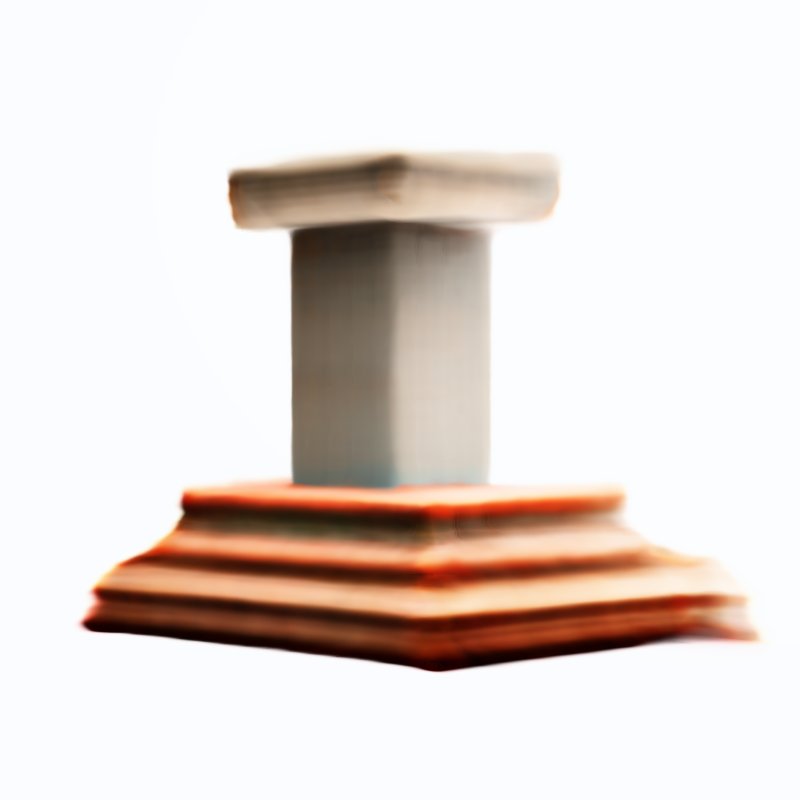}
    \includegraphics[width=0.12\textwidth]{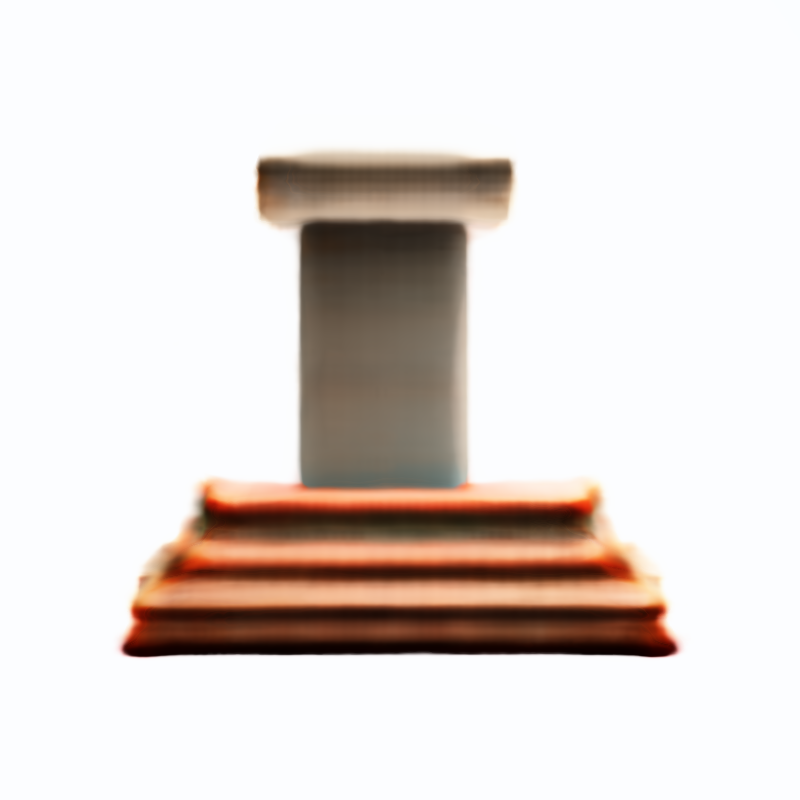}
    \includegraphics[width=0.12\textwidth]{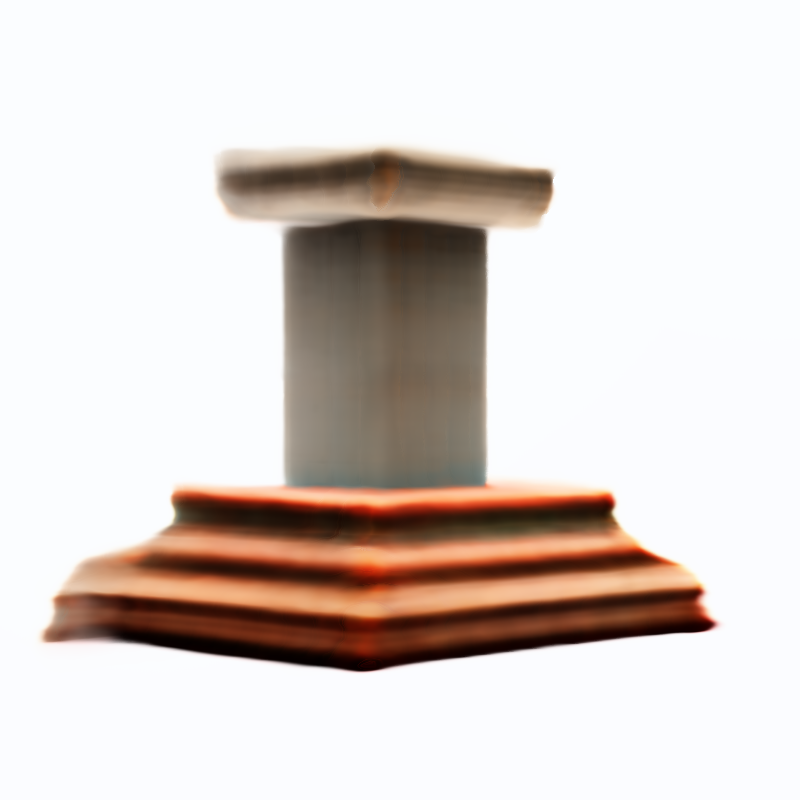}
    \includegraphics[width=0.12\textwidth]{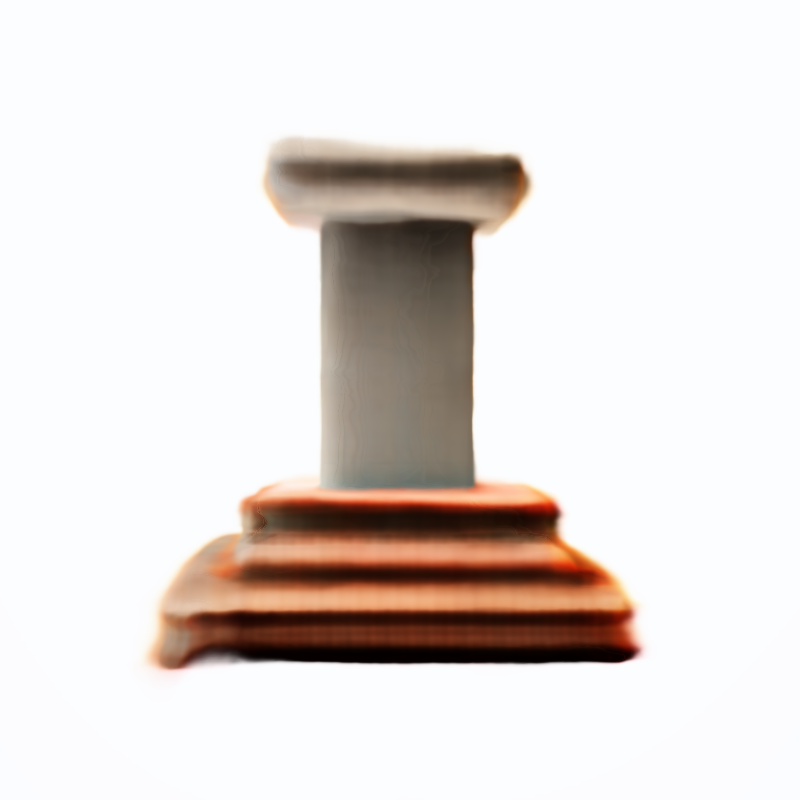}
    \includegraphics[width=0.12\textwidth]{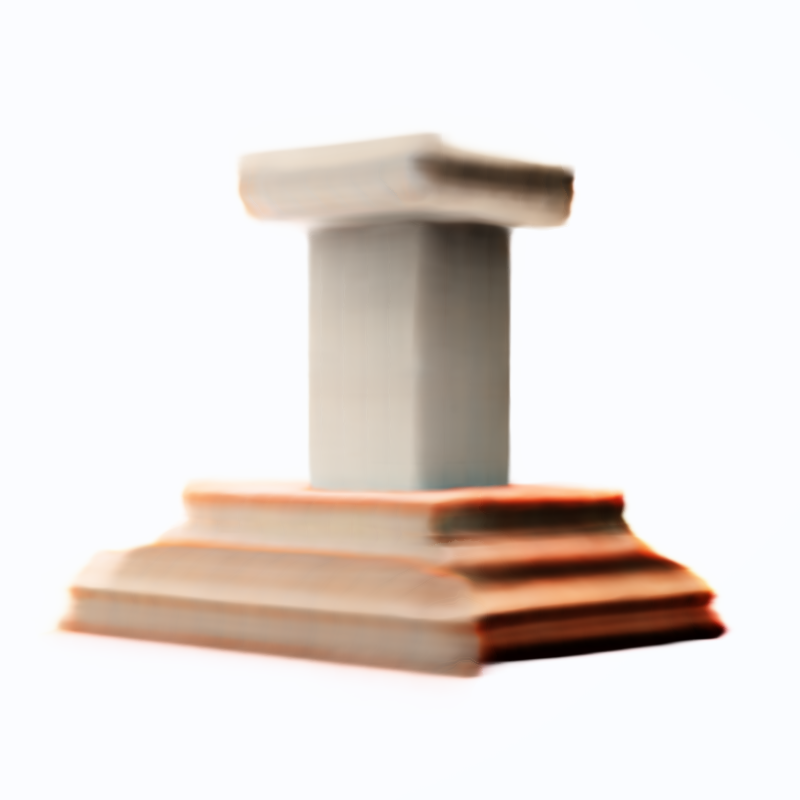}
    \caption{{\bf Multi-view results of out-of-distribution inference from image prompts}.
    We reuse the samples shown in the experiment results section in out-of-distribution inference. We display the 8 camera views generated by our NeRF obtained for each scene. Prompts used are the same in previous experiment results.}
    \label{fig:supp-multi-view-out-dist-img}
\end{figure*}

\begin{figure*}[ht]
    \centering
    \includegraphics[width=0.12\textwidth]{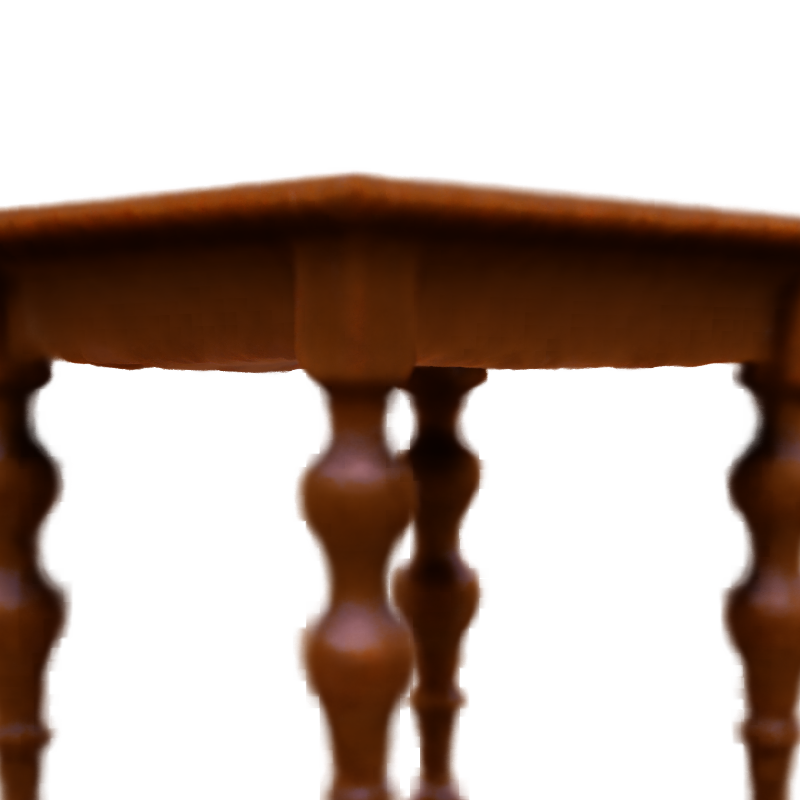}
    \includegraphics[width=0.12\textwidth]{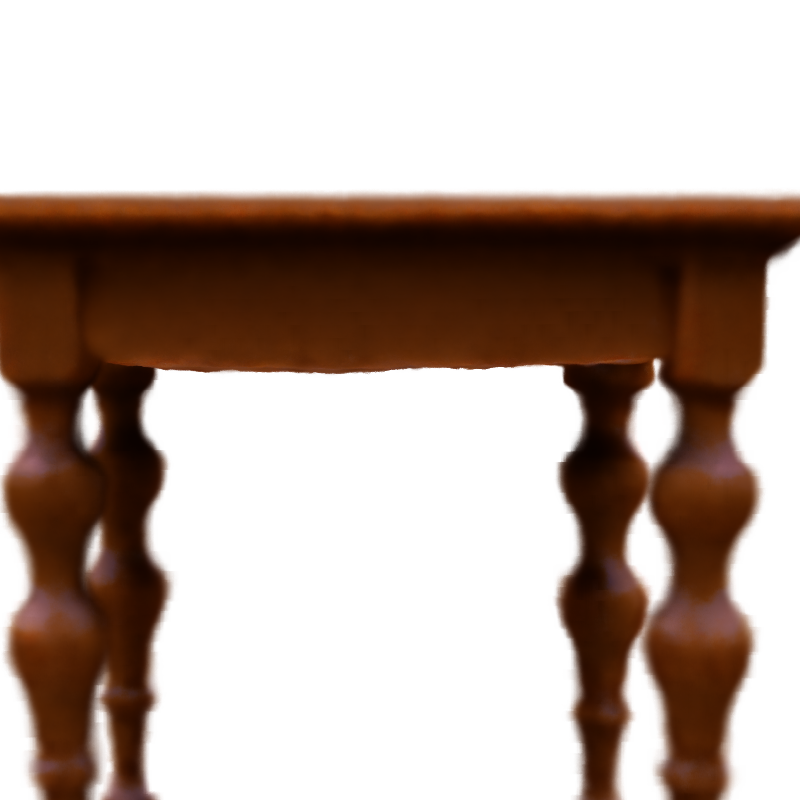}
    \includegraphics[width=0.12\textwidth]{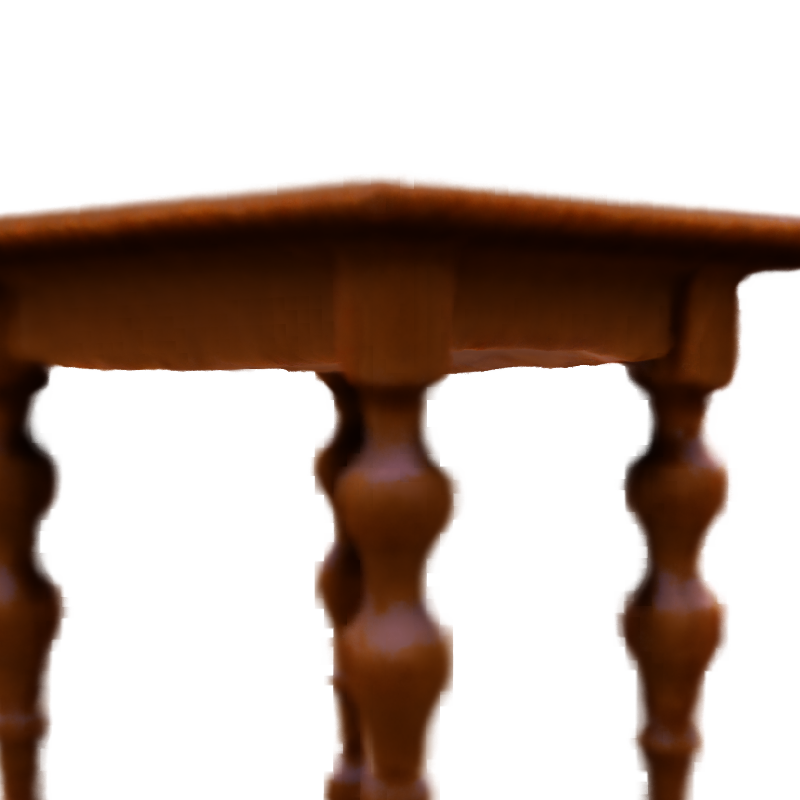}
    \includegraphics[width=0.12\textwidth]{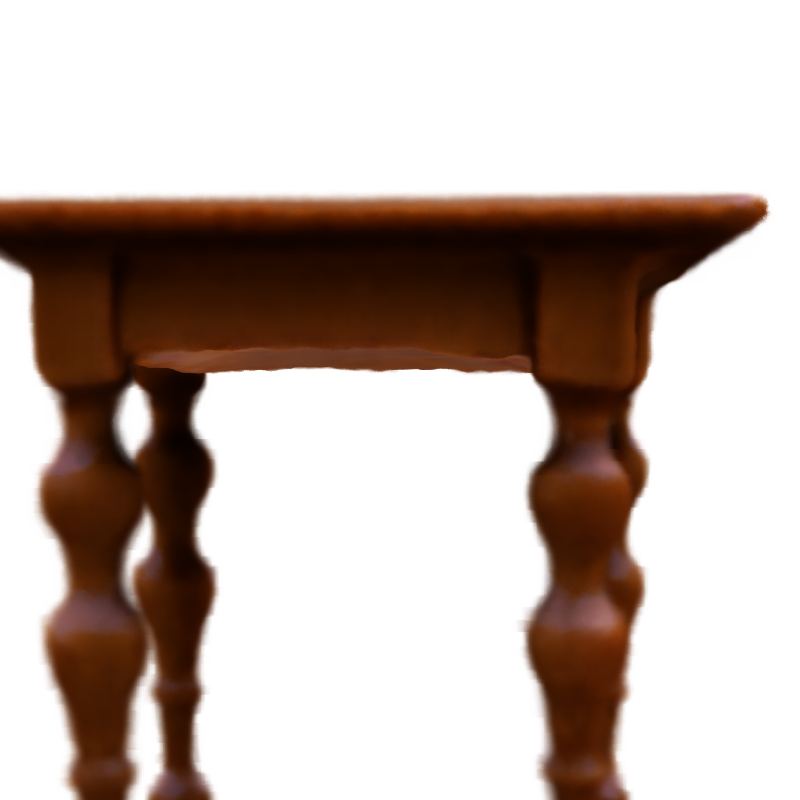}
    \includegraphics[width=0.12\textwidth]{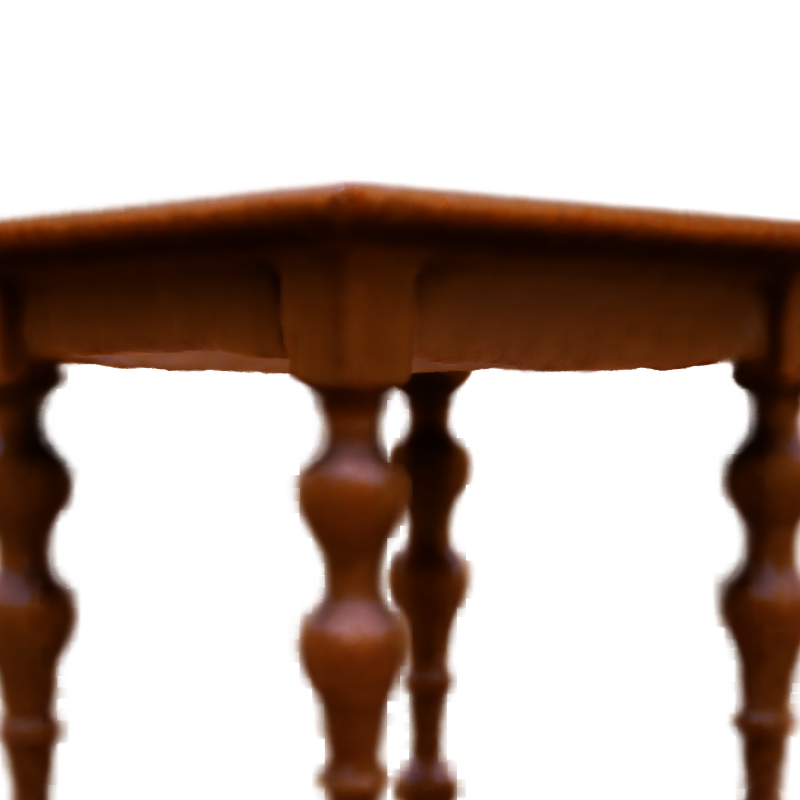}
    \includegraphics[width=0.12\textwidth]{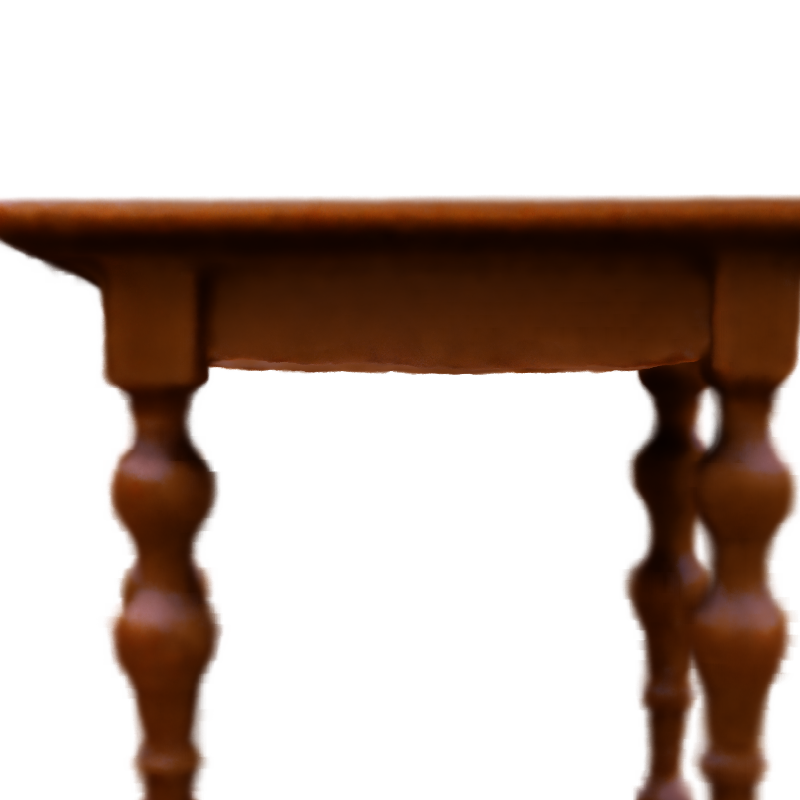}
    \includegraphics[width=0.12\textwidth]{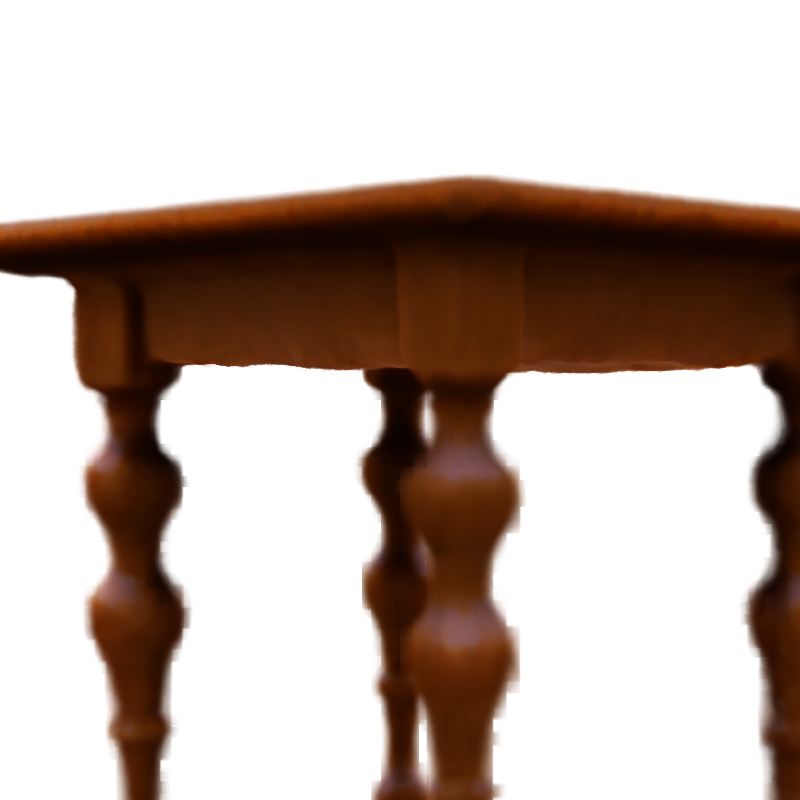}
    \includegraphics[width=0.12\textwidth]{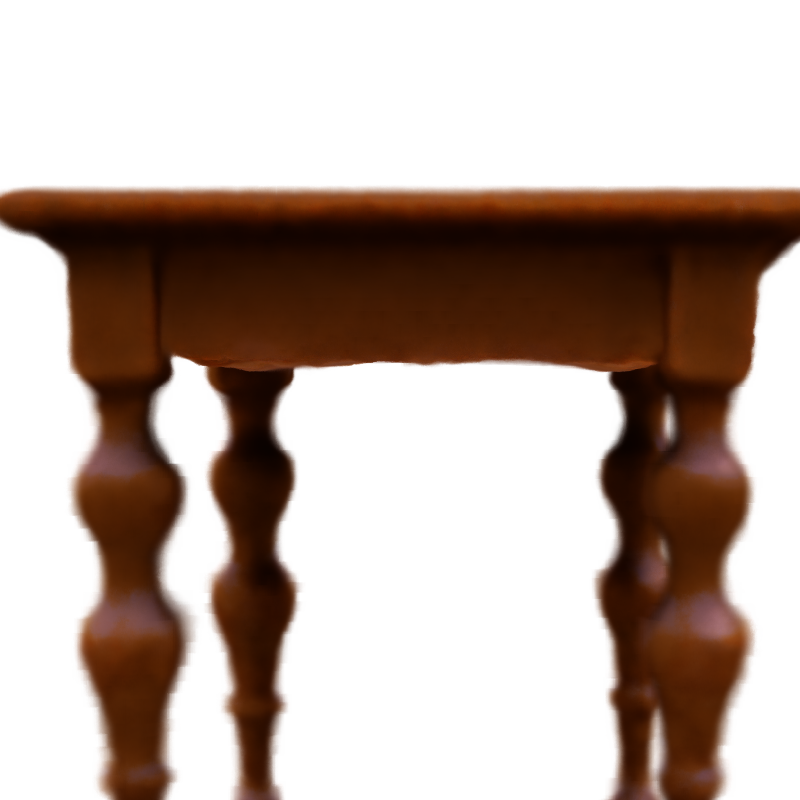}\\
    \includegraphics[width=0.12\textwidth]{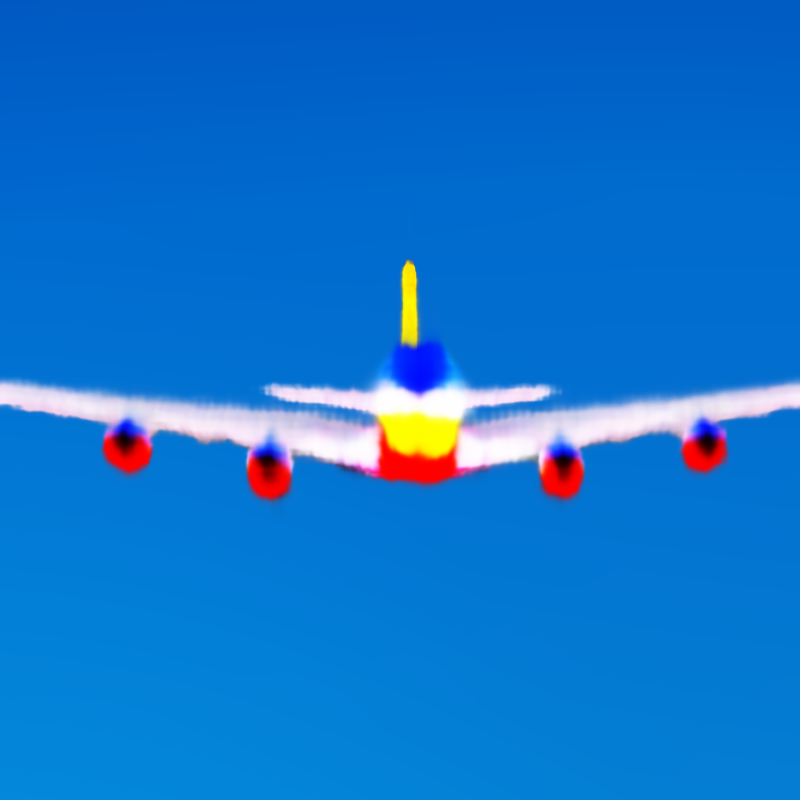}
    \includegraphics[width=0.12\textwidth]{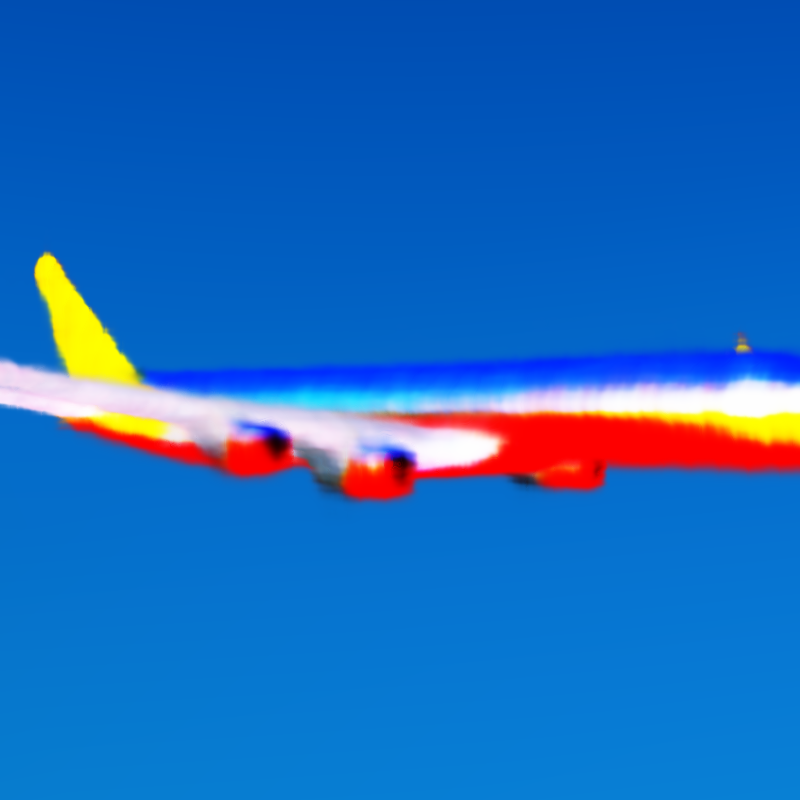}
    \includegraphics[width=0.12\textwidth]{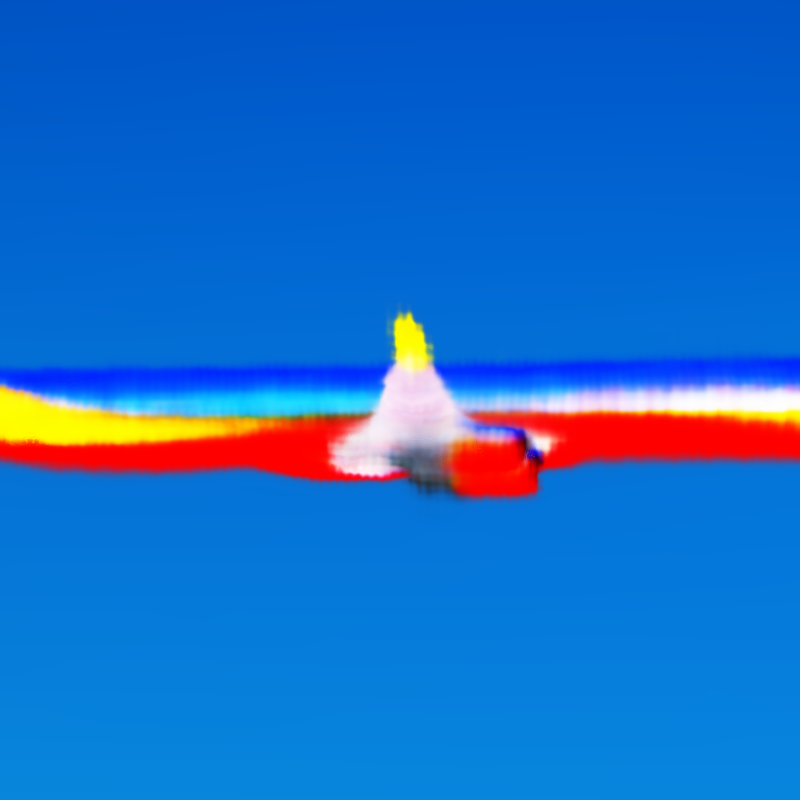}
    \includegraphics[width=0.12\textwidth]{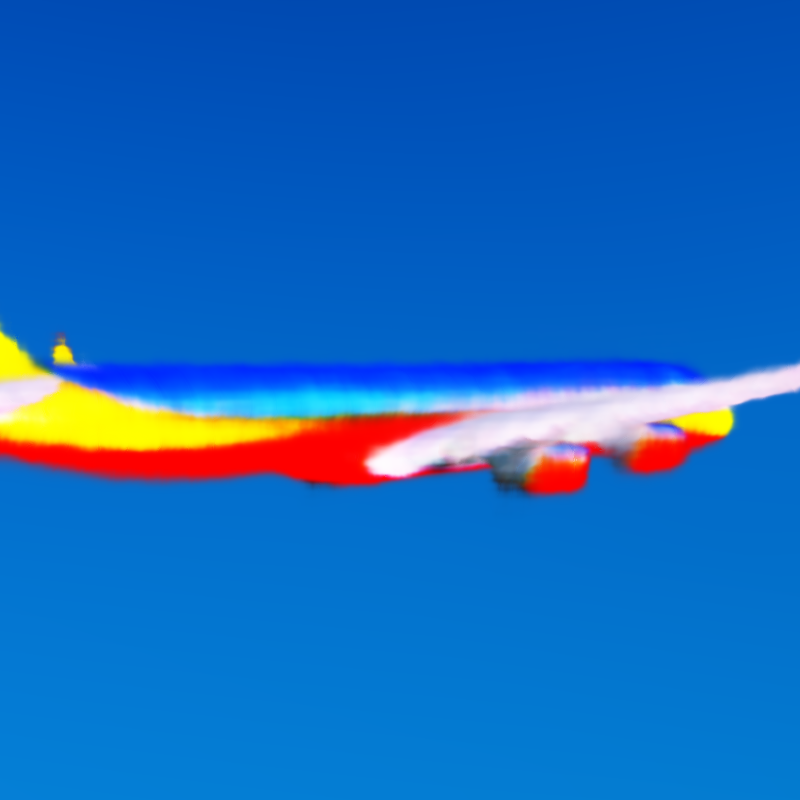}
    \includegraphics[width=0.12\textwidth]{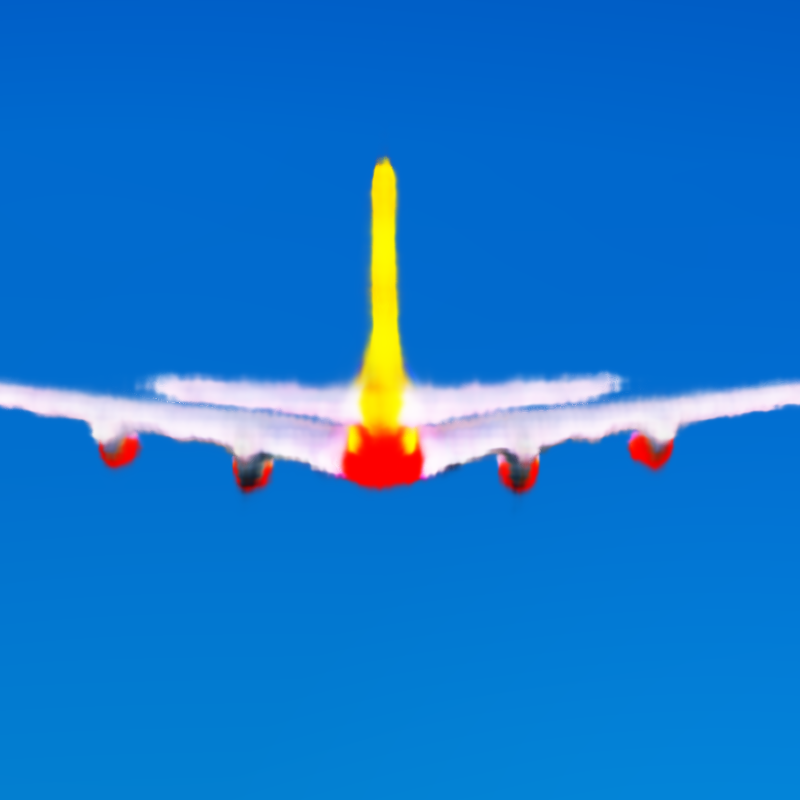}
    \includegraphics[width=0.12\textwidth]{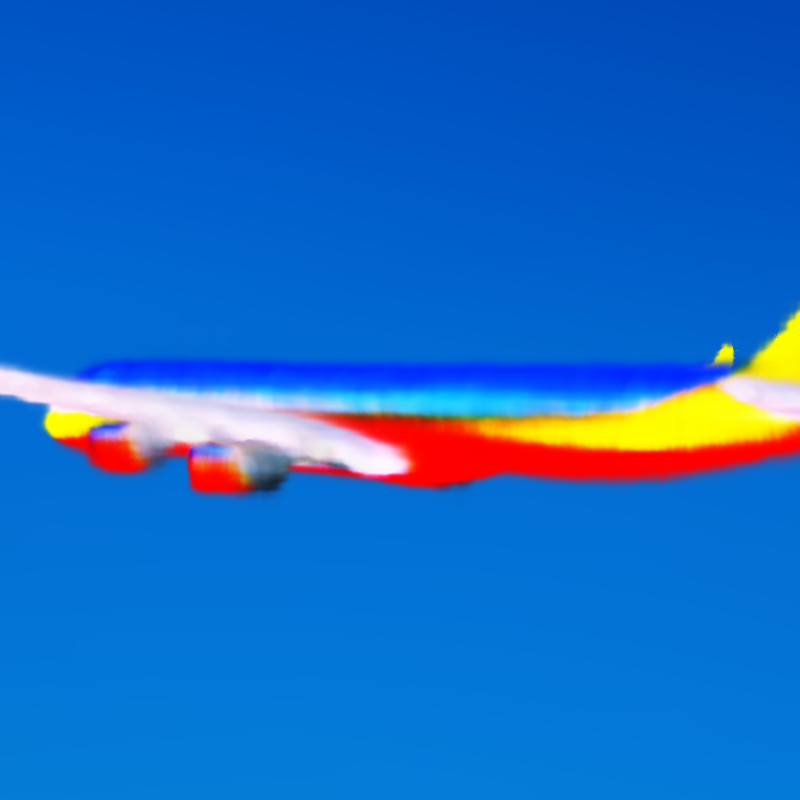}
    \includegraphics[width=0.12\textwidth]{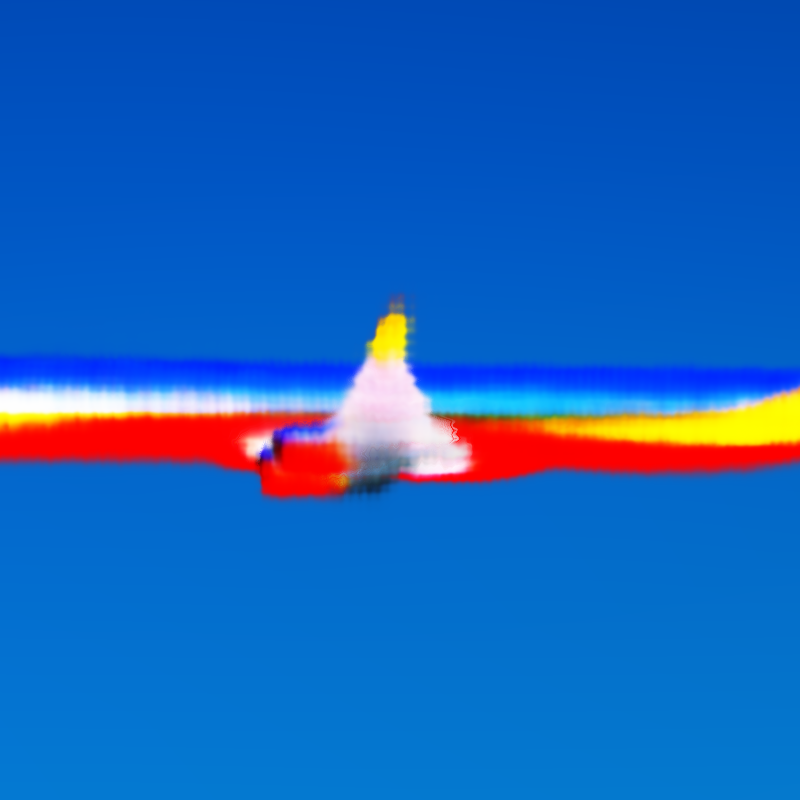}
    \includegraphics[width=0.12\textwidth]{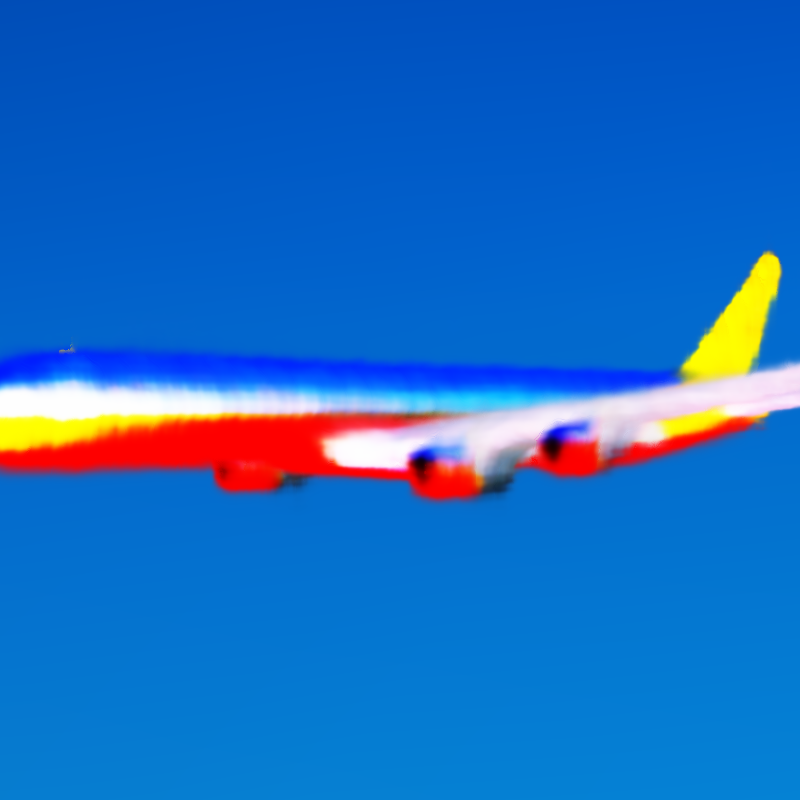}\\
    \includegraphics[width=0.12\textwidth]{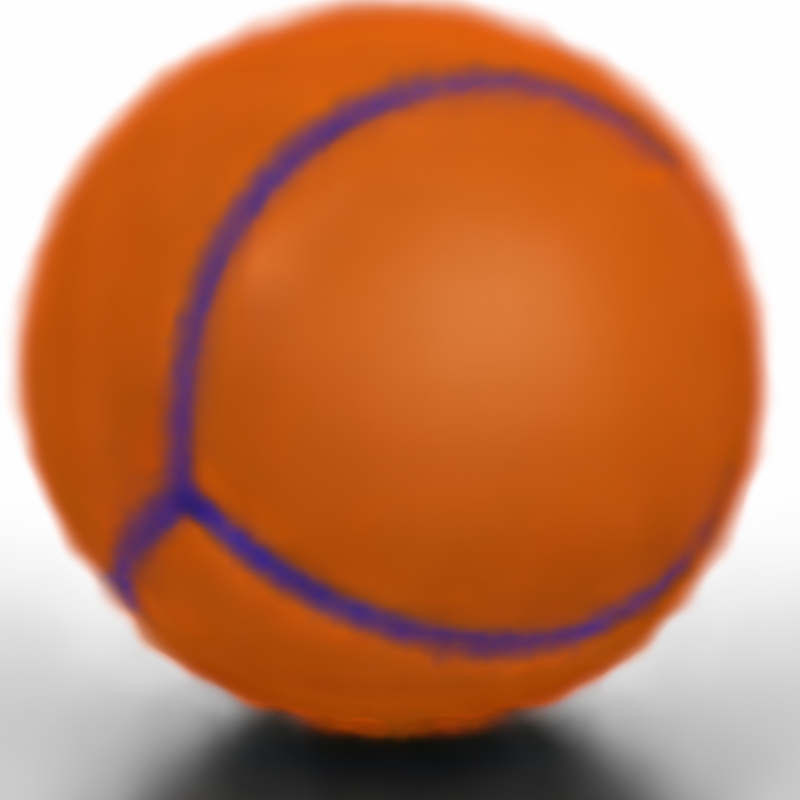}
    \includegraphics[width=0.12\textwidth]{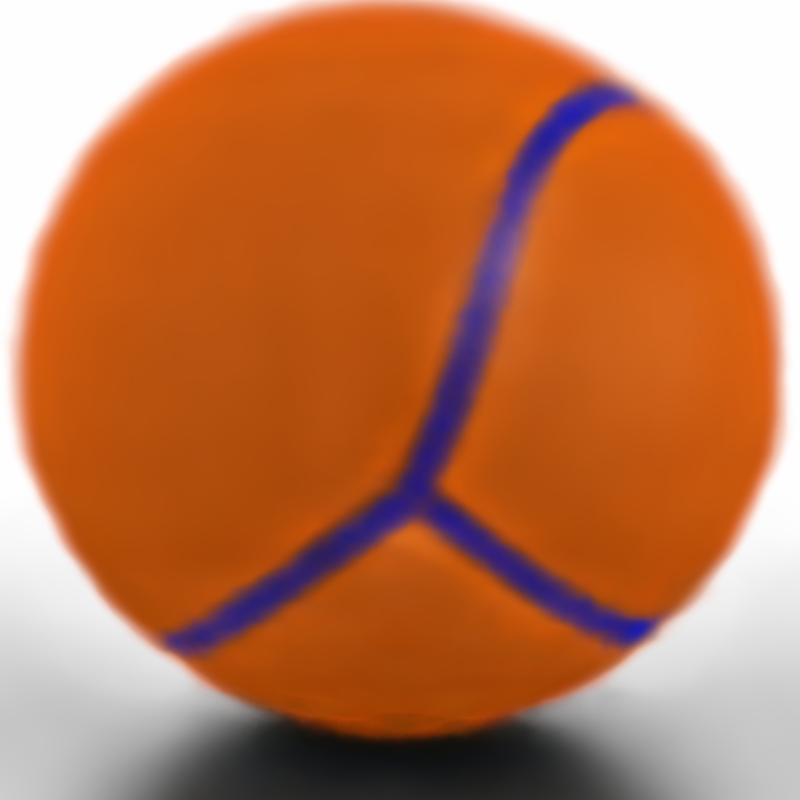}
    \includegraphics[width=0.12\textwidth]{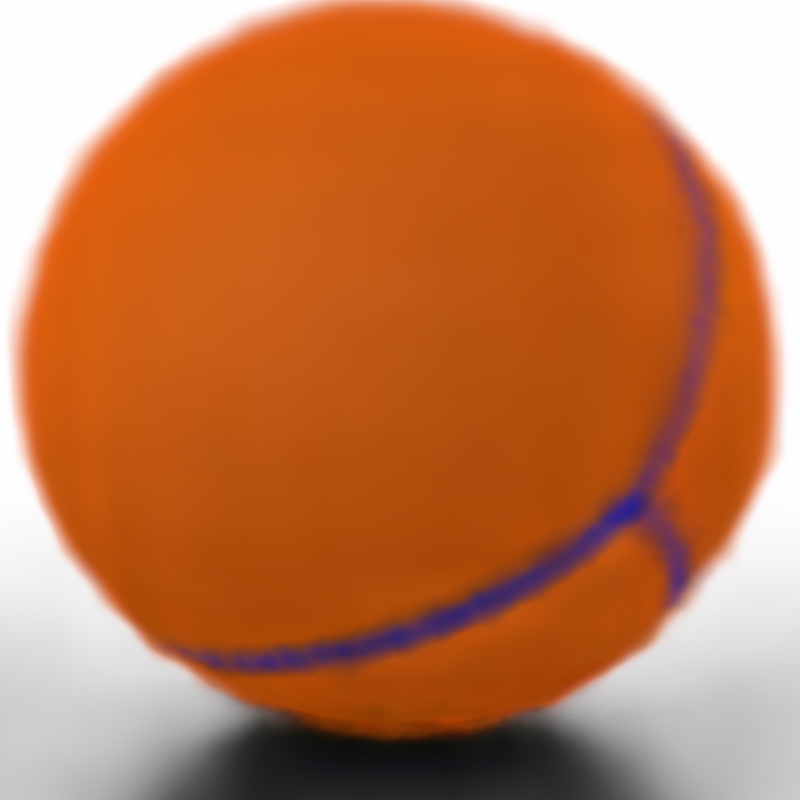}
    \includegraphics[width=0.12\textwidth]{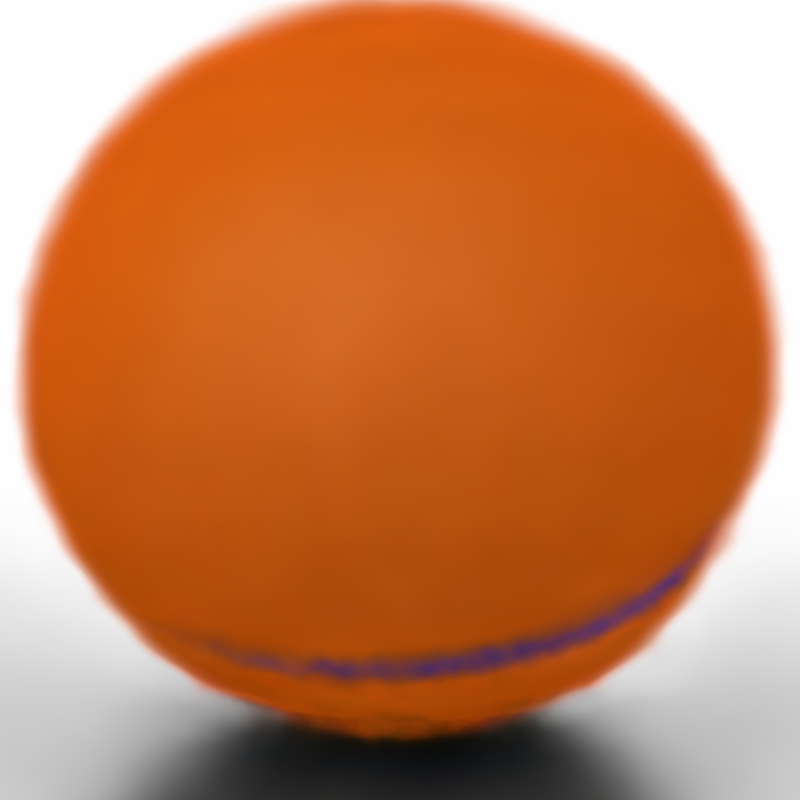}
    \includegraphics[width=0.12\textwidth]{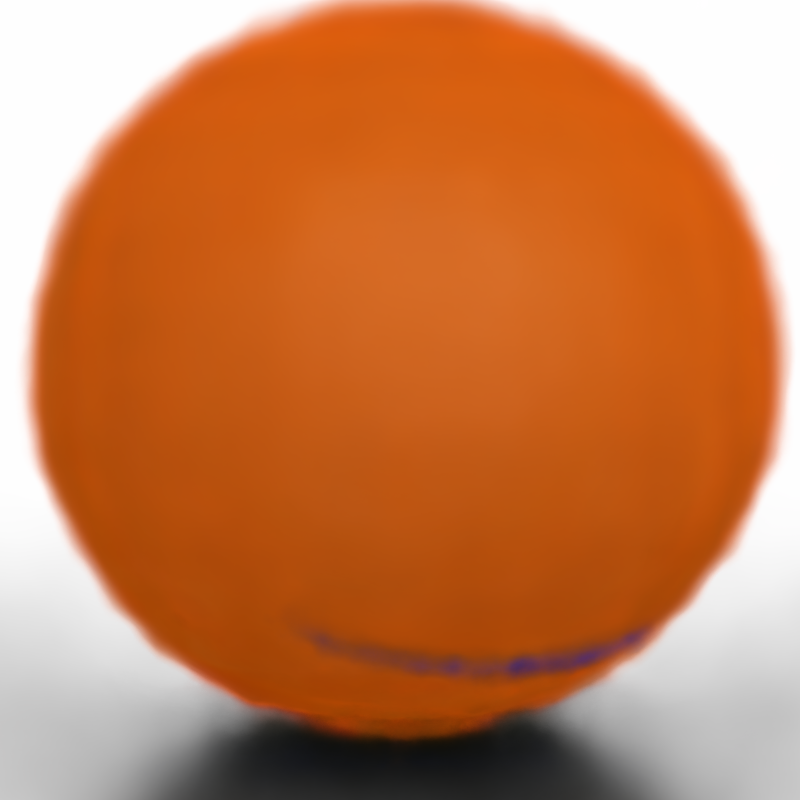}
    \includegraphics[width=0.12\textwidth]{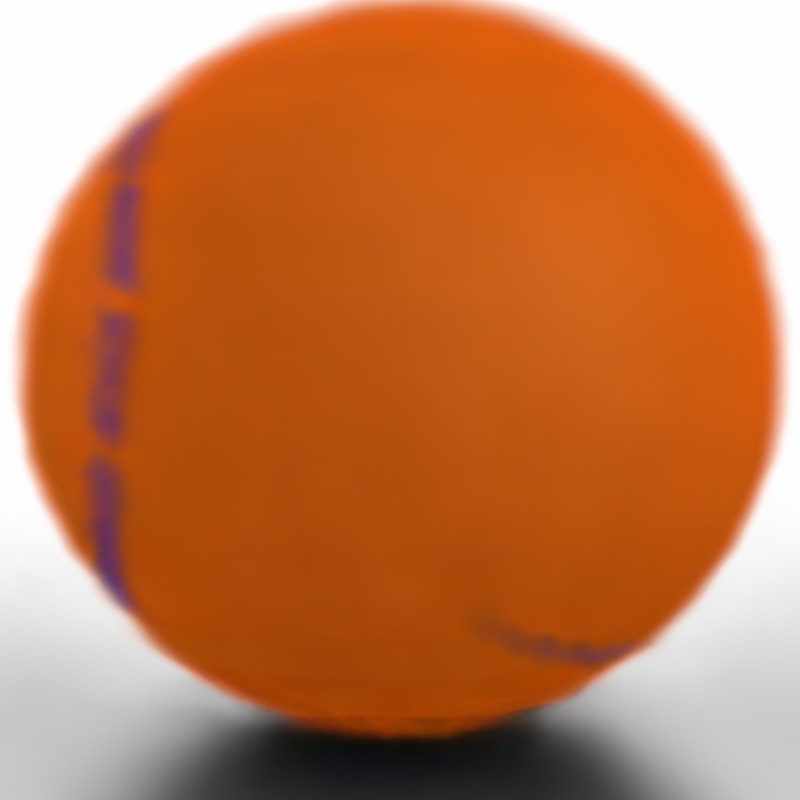}
    \includegraphics[width=0.12\textwidth]{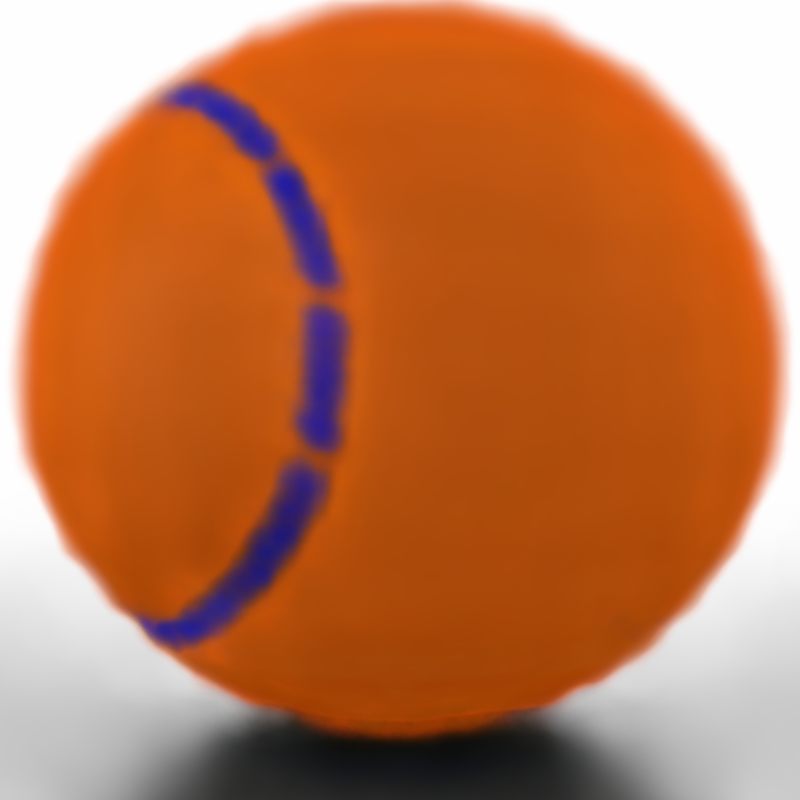}
    \includegraphics[width=0.12\textwidth]{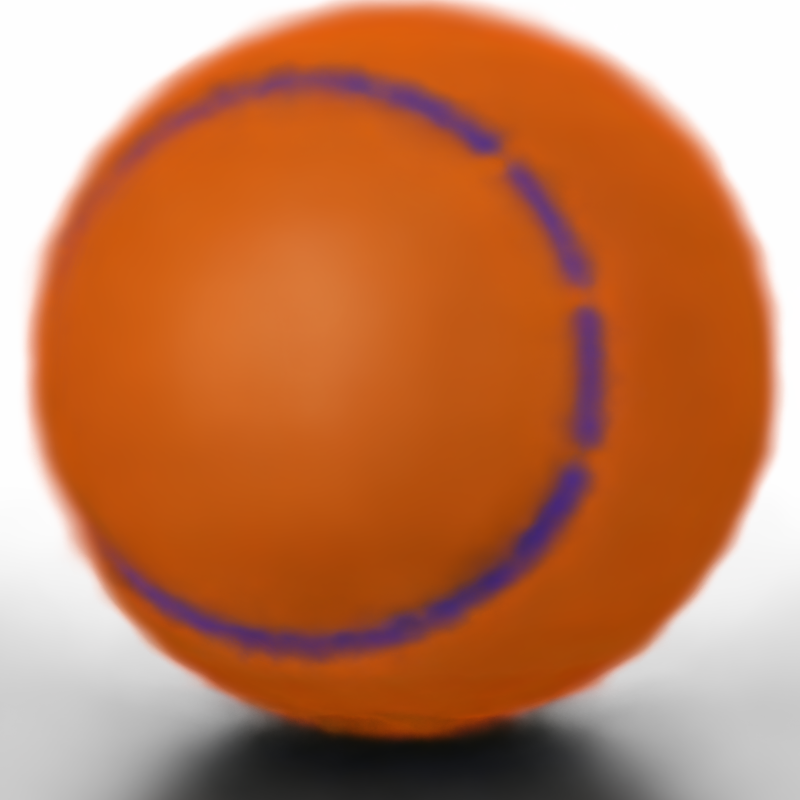}\\
    \includegraphics[width=0.12\textwidth]{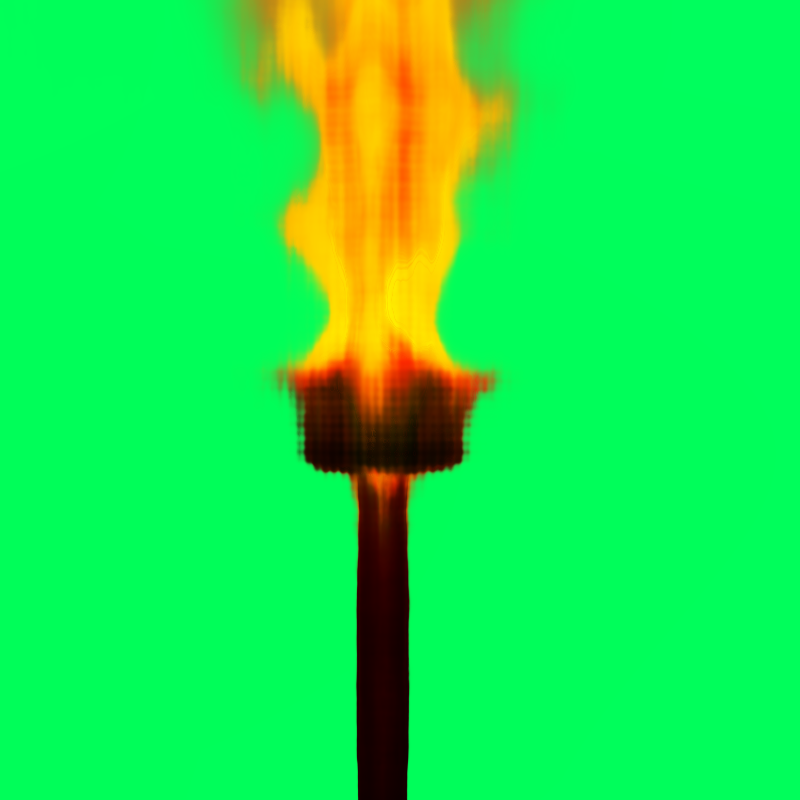}
    \includegraphics[width=0.12\textwidth]{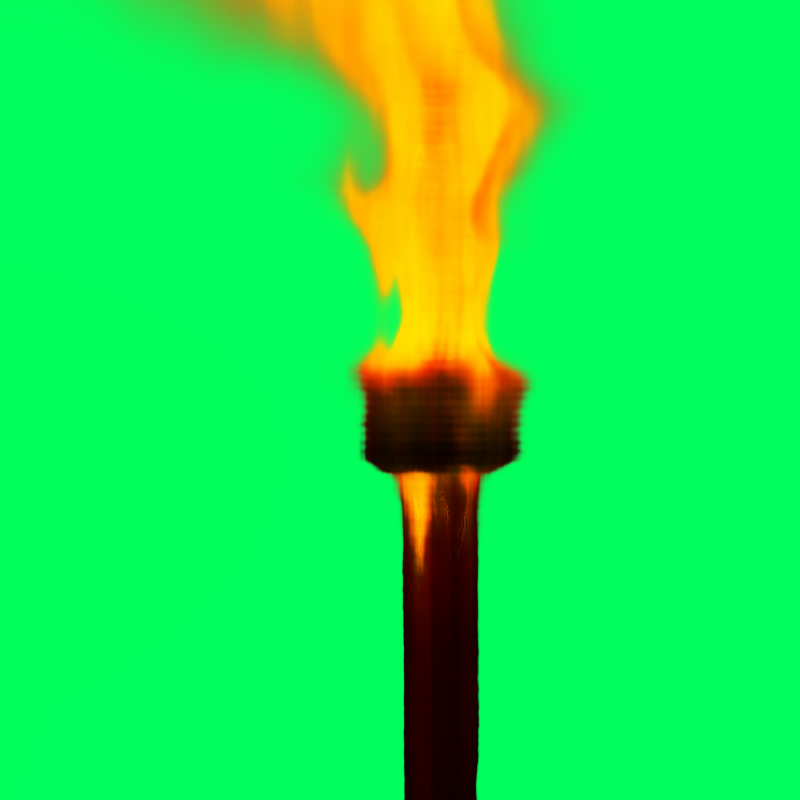}
    \includegraphics[width=0.12\textwidth]{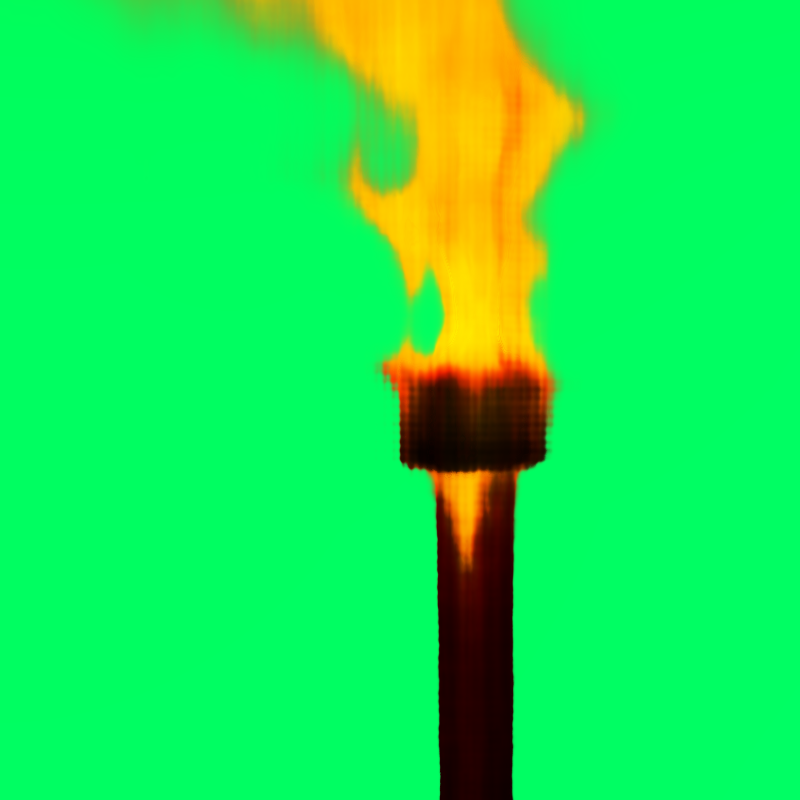}
    \includegraphics[width=0.12\textwidth]{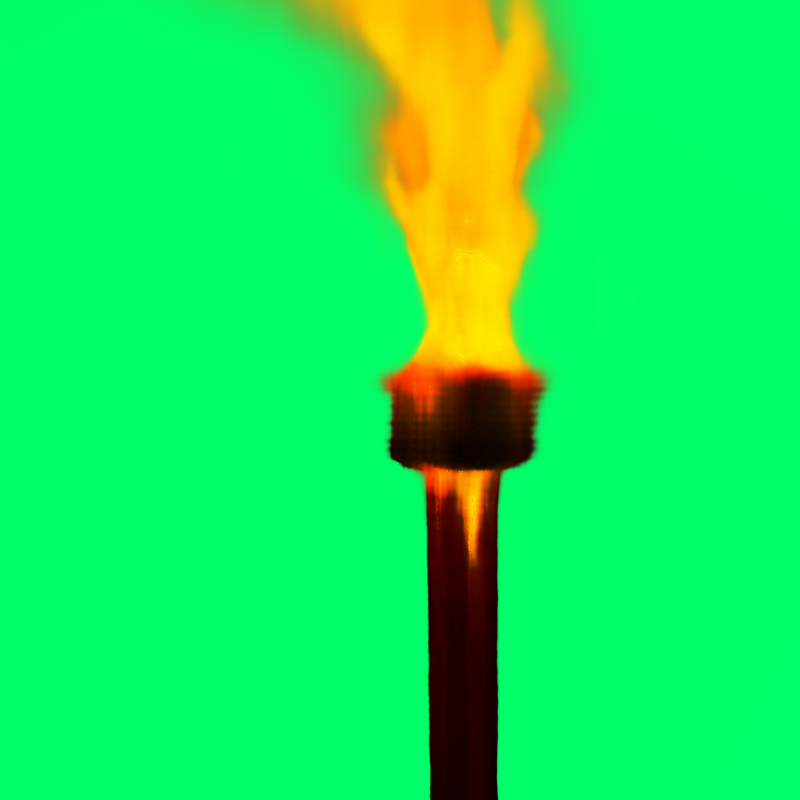}
    \includegraphics[width=0.12\textwidth]{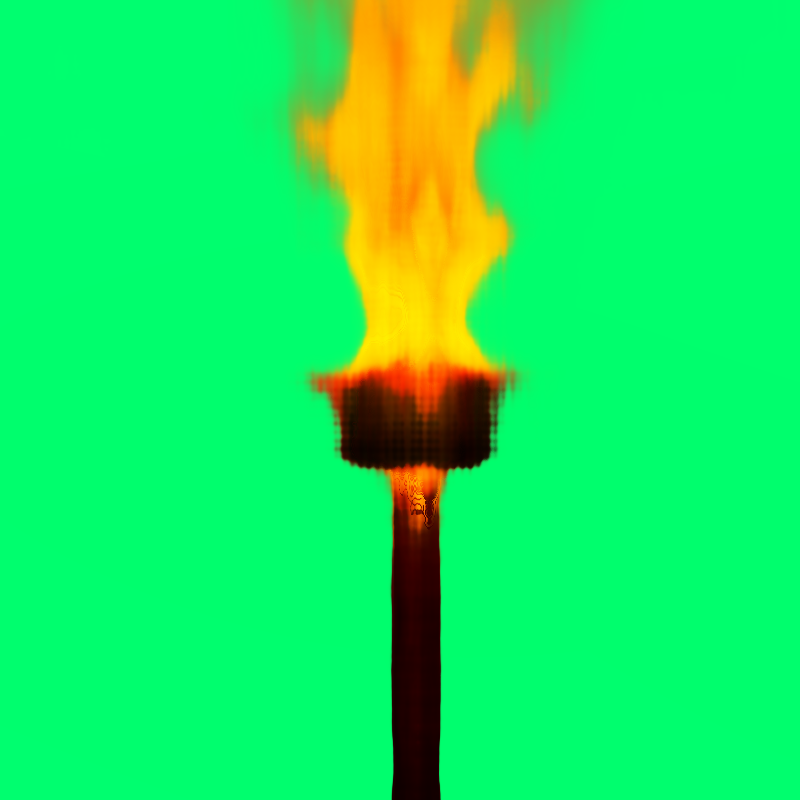}
    \includegraphics[width=0.12\textwidth]{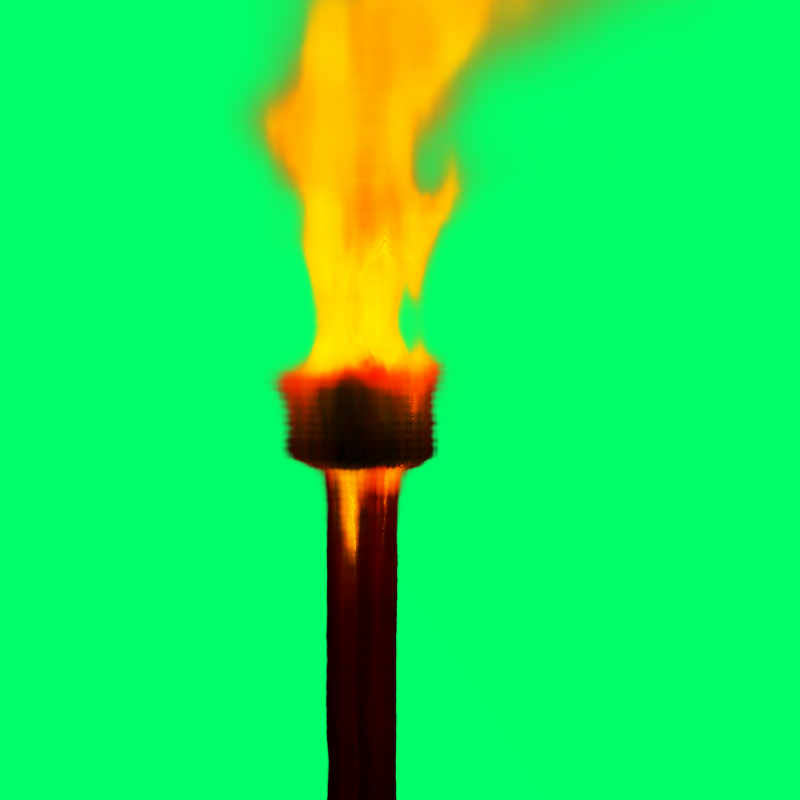}
    \includegraphics[width=0.12\textwidth]{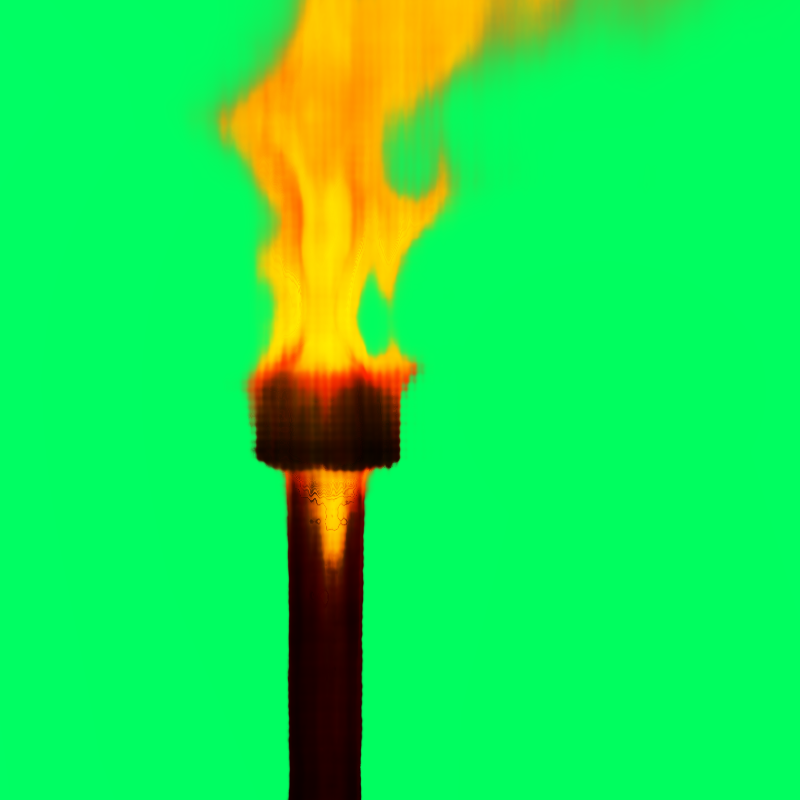}
    \includegraphics[width=0.12\textwidth]{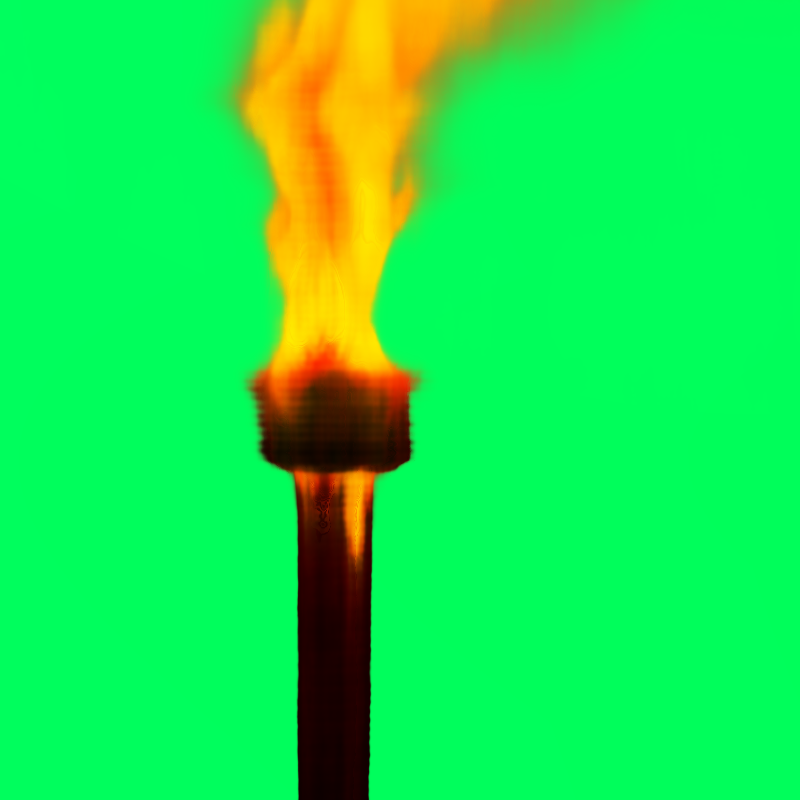}
    \caption{{\bf Multi-view results of out-of-distribution inference from text prompts}.
    We reuse the samples shown in the experiment results section in out-of-distribution inference. We display the 8 camera views generated by our NeRF obtained for each scene. Prompts used (from top to bottom): ``A brown table.'', ``A colorful Boeing passenger plane.'', ``A basketball.'', ``A burning torch.''}
    \label{fig:supp-multi-view-out-dist-text}
\end{figure*}

We randomly select some samples from our test split and display them in~\cref{fig:supp-multi-view-in-dist-1,fig:supp-multi-view-in-dist-2}. For out-of-distribution image or text inference results, we display them in~\cref{fig:supp-multi-view-out-dist-img} and~\cref{fig:supp-multi-view-out-dist-text}, respectively.

\begin{figure*}[ht]
    \centering
    \includegraphics[width=0.49\linewidth,trim={2cm 2cm 0 0},clip]{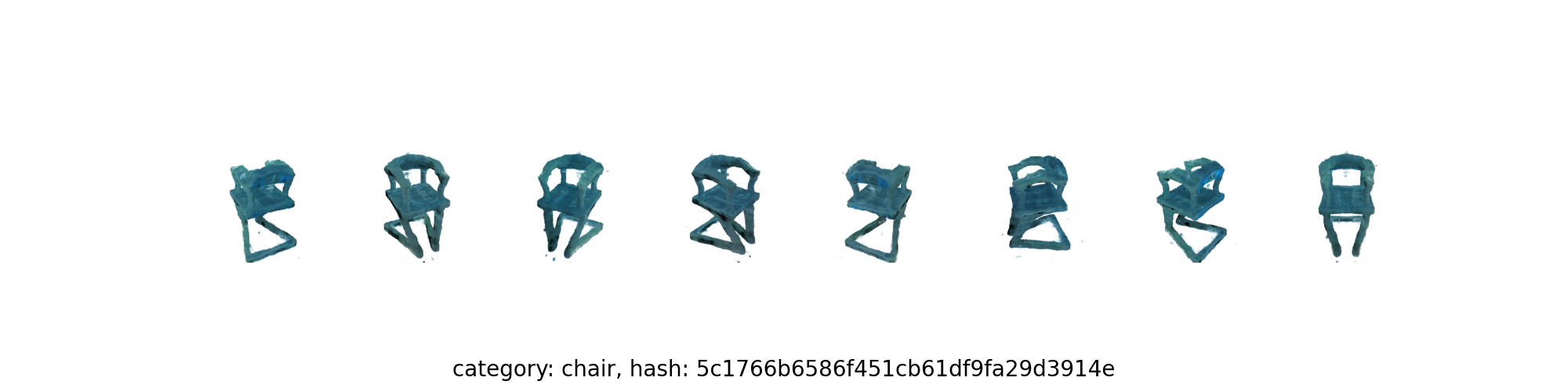}
    \includegraphics[width=0.49\linewidth,trim={2cm 2cm 0 0},clip]{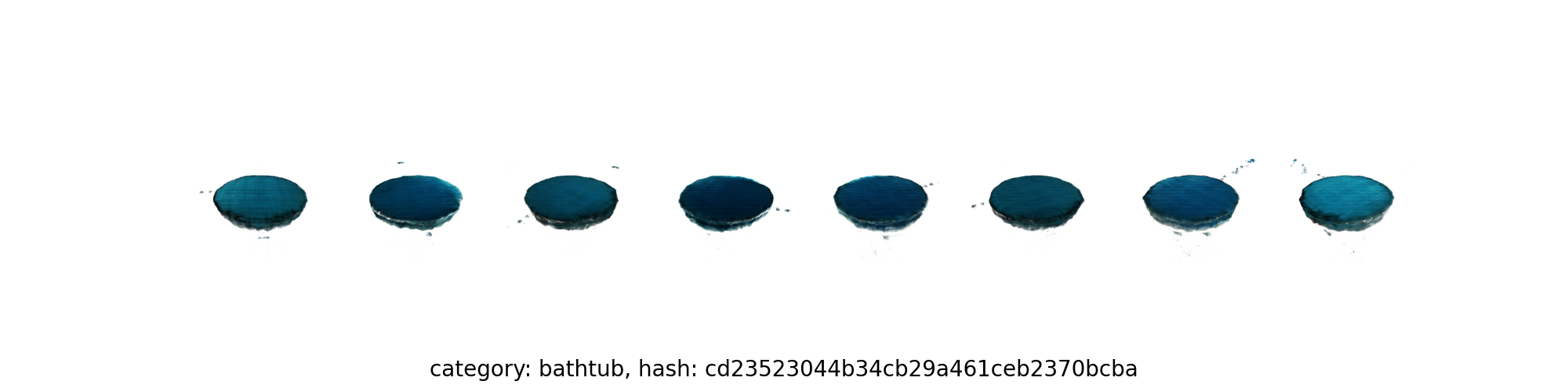}
    \includegraphics[width=0.49\linewidth,trim={2cm 2cm 0 0},clip]{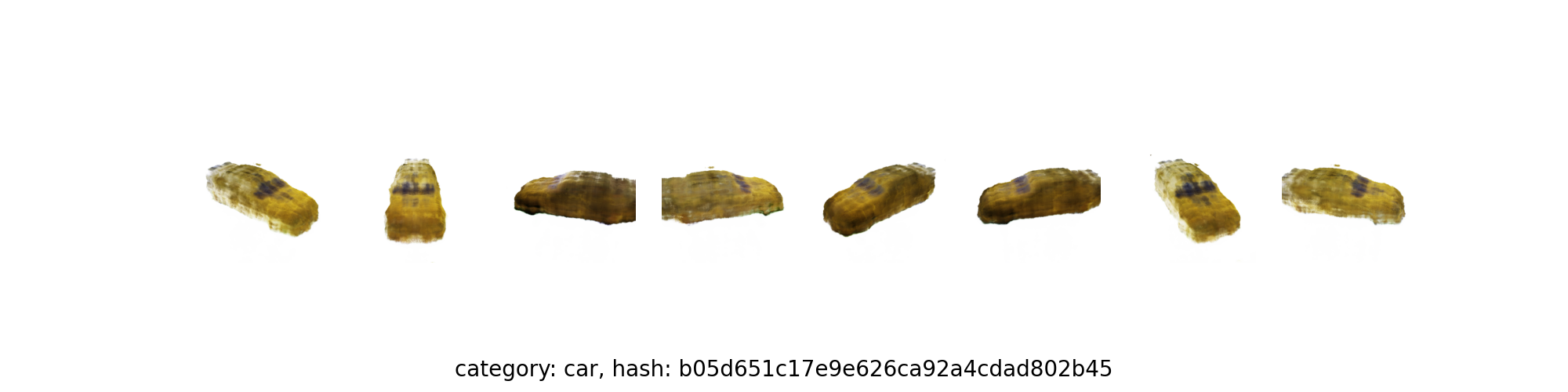}
    \includegraphics[width=0.49\linewidth,trim={2cm 2cm 0 0},clip]{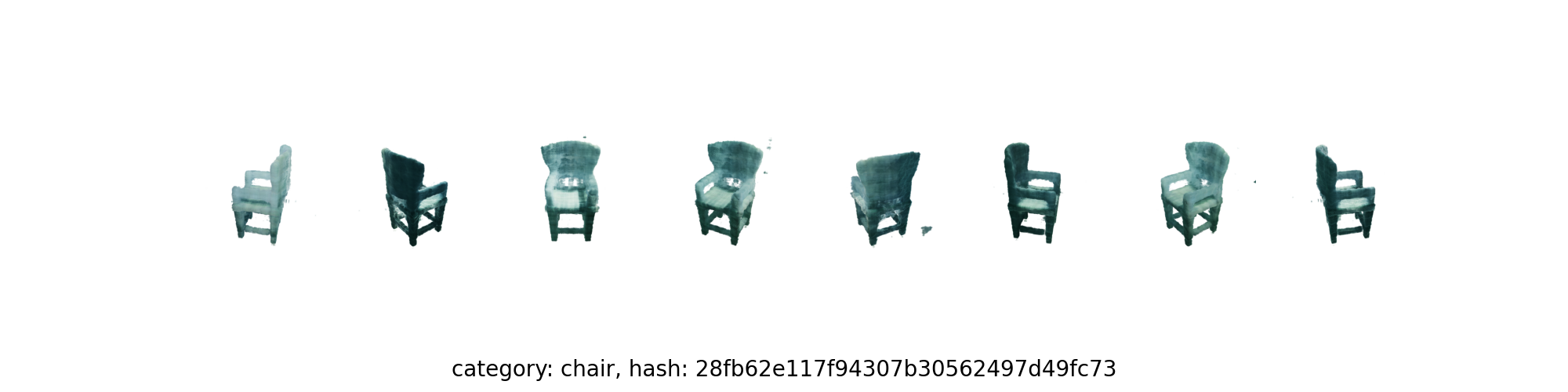}
    \includegraphics[width=0.49\linewidth,trim={2cm 2cm 0 0},clip]{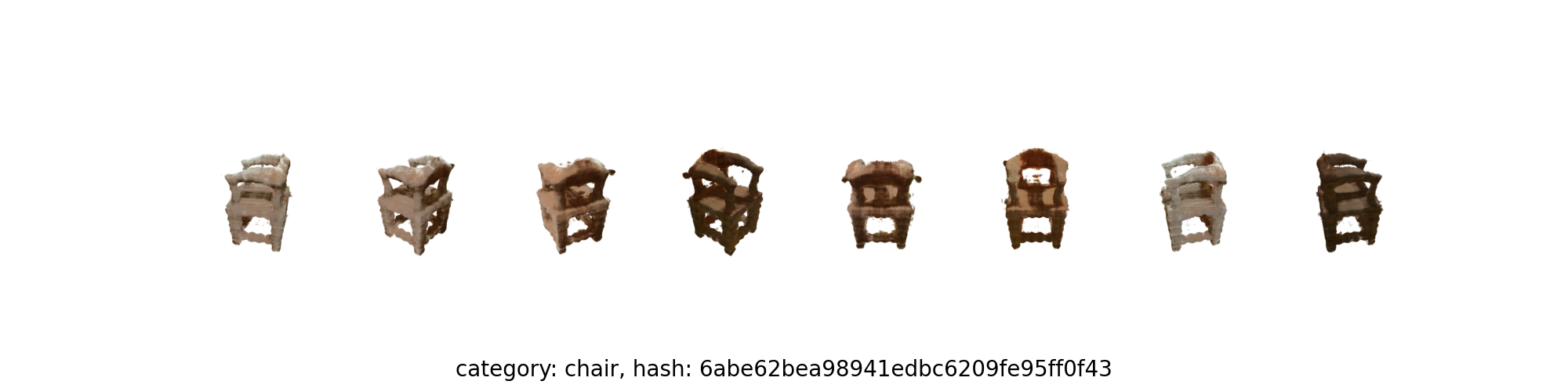}
    \includegraphics[width=0.49\linewidth,trim={2cm 2cm 0 0},clip]{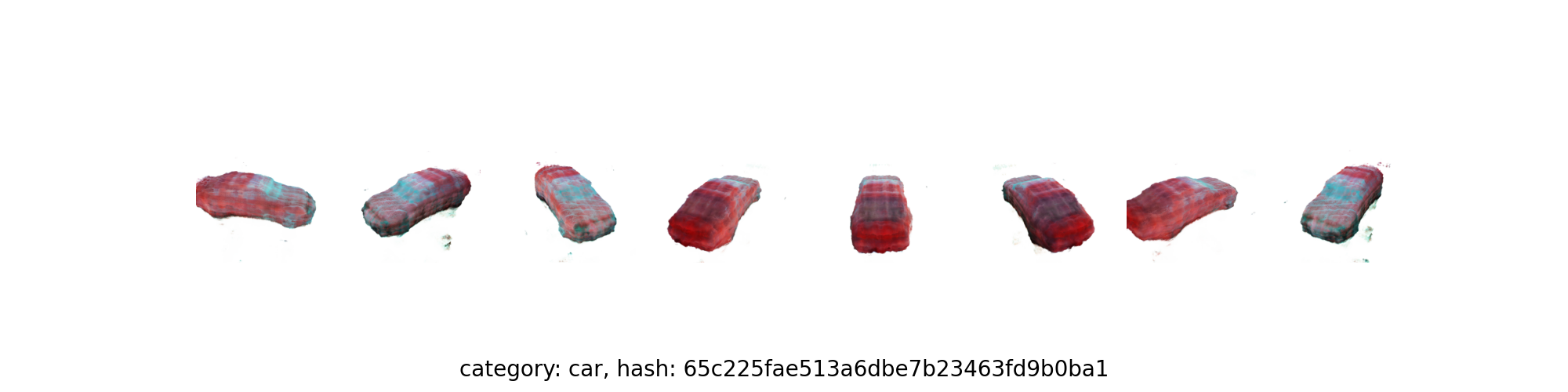}
    \includegraphics[width=0.49\linewidth,trim={2cm 2cm 0 0},clip]{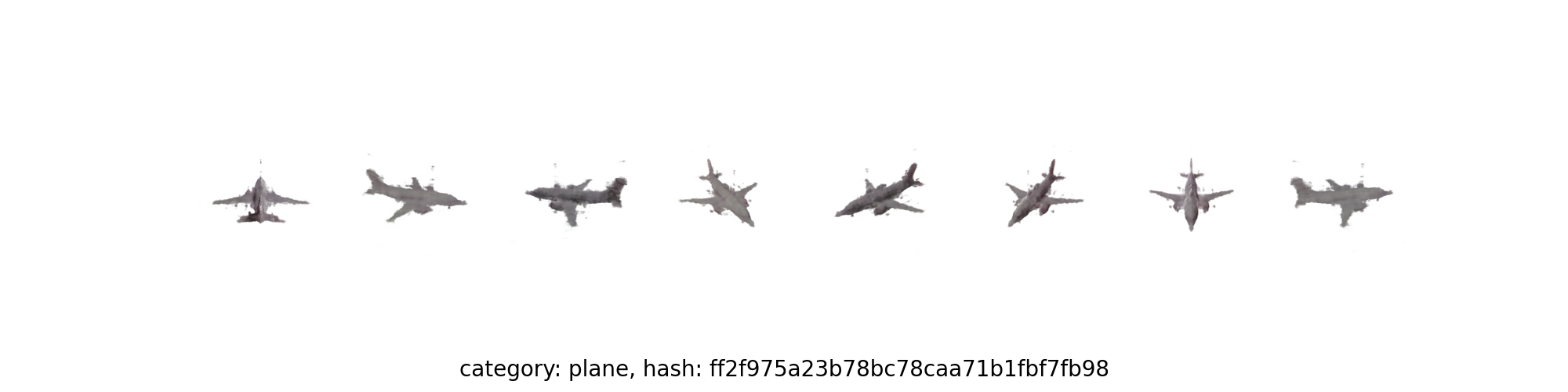}
    \includegraphics[width=0.49\linewidth,trim={2cm 2cm 0 0},clip]{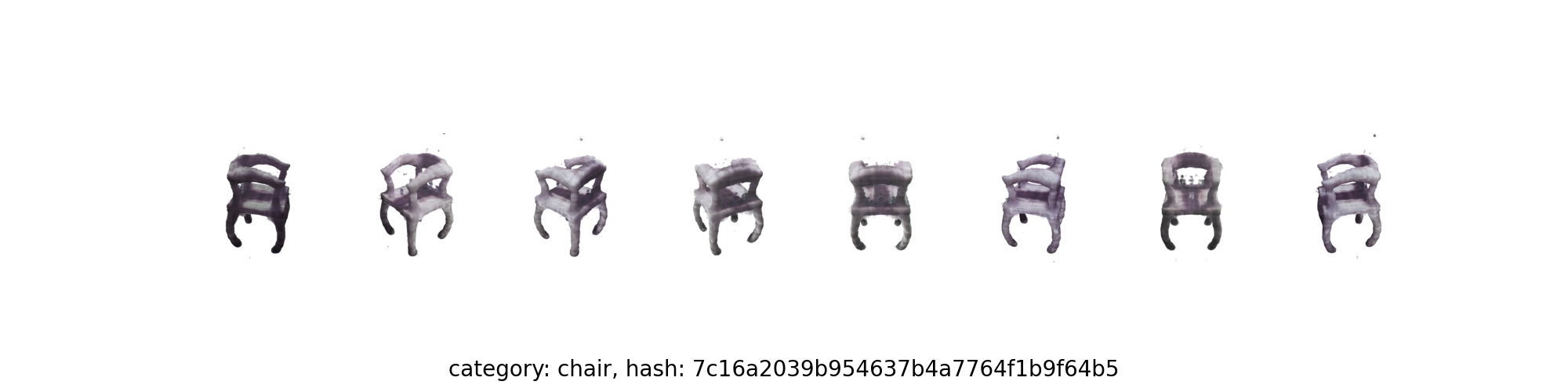}
    \includegraphics[width=0.49\linewidth,trim={2cm 2cm 0 0},clip]{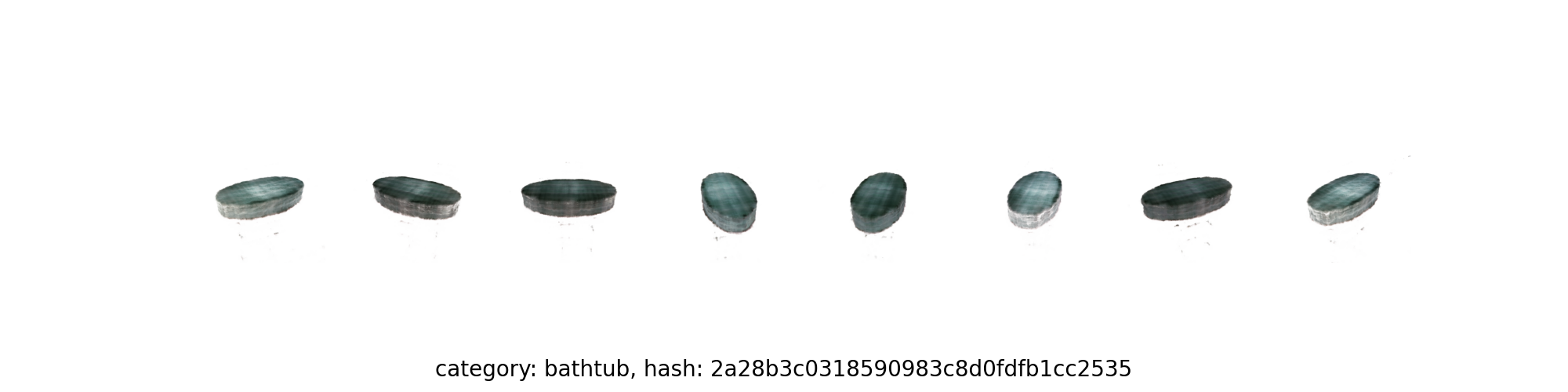}
    \includegraphics[width=0.49\linewidth,trim={2cm 2cm 0 0},clip]{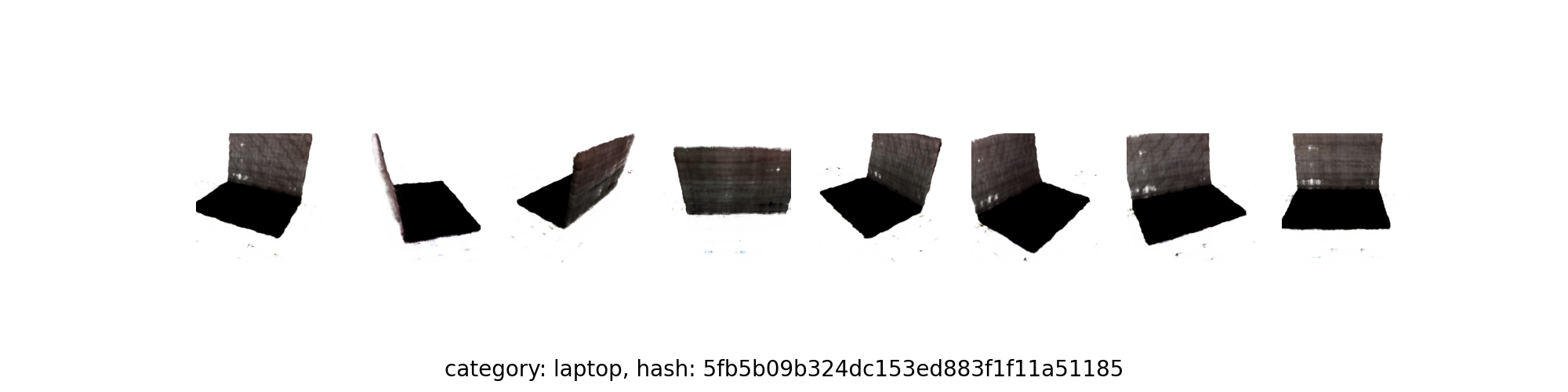}
    \includegraphics[width=0.49\linewidth,trim={2cm 2cm 0 0},clip]{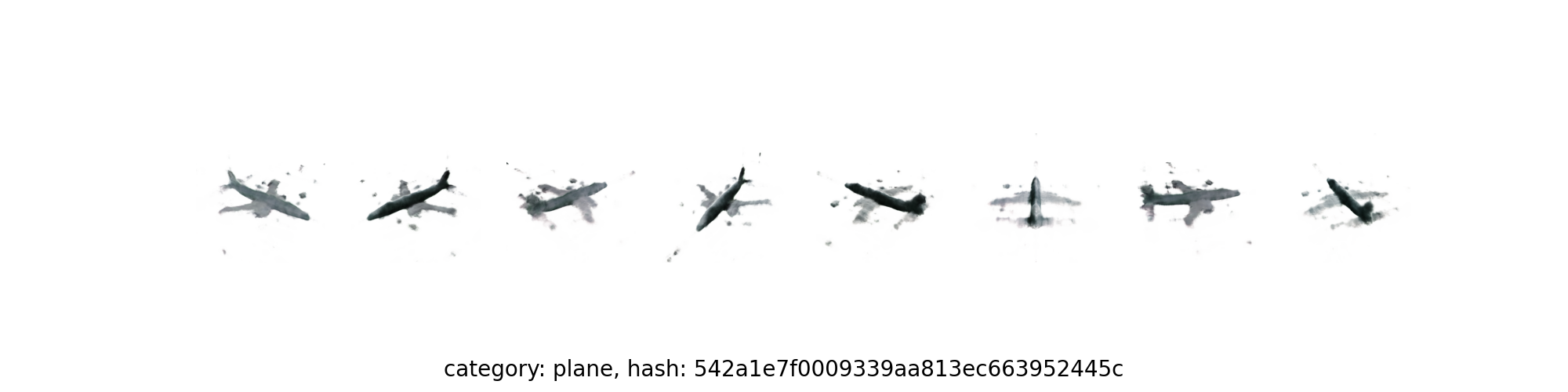}
    \includegraphics[width=0.49\linewidth,trim={2cm 2cm 0 0},clip]{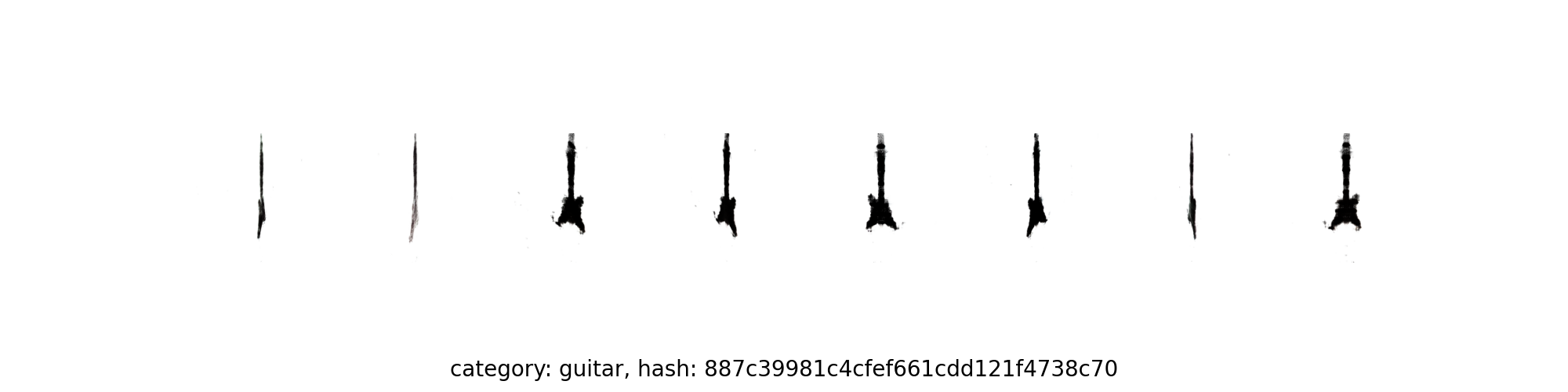}
    \includegraphics[width=0.49\linewidth,trim={2cm 2cm 0 0},clip]{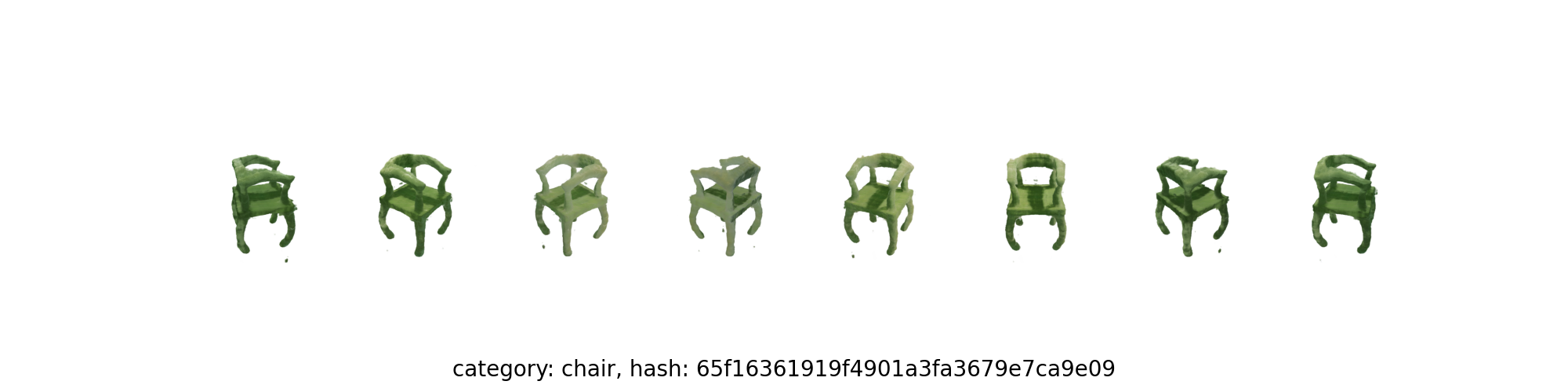}
    \includegraphics[width=0.49\linewidth,trim={2cm 2cm 0 0},clip]{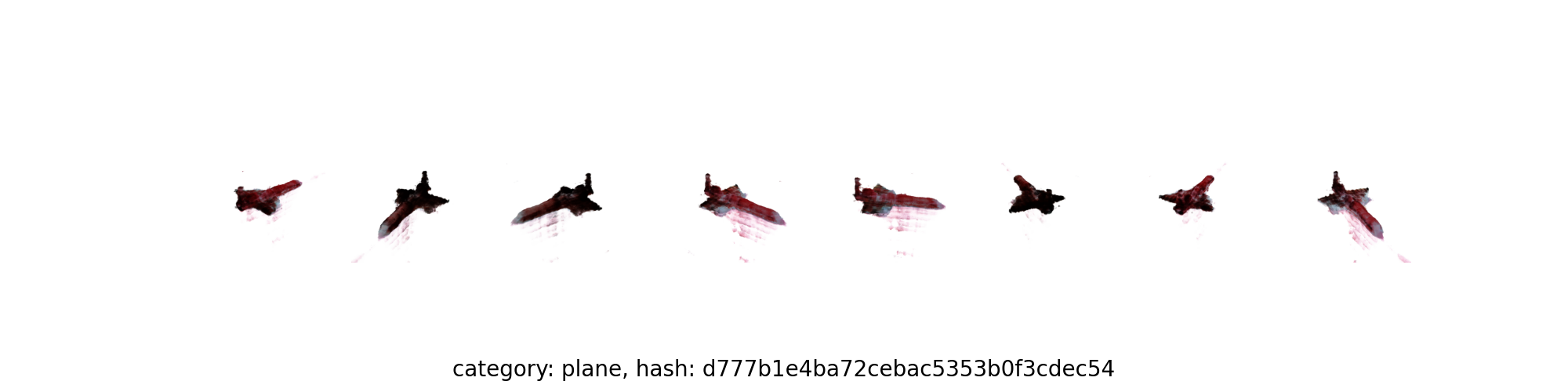}
    \includegraphics[width=0.49\linewidth,trim={2cm 2cm 0 0},clip]{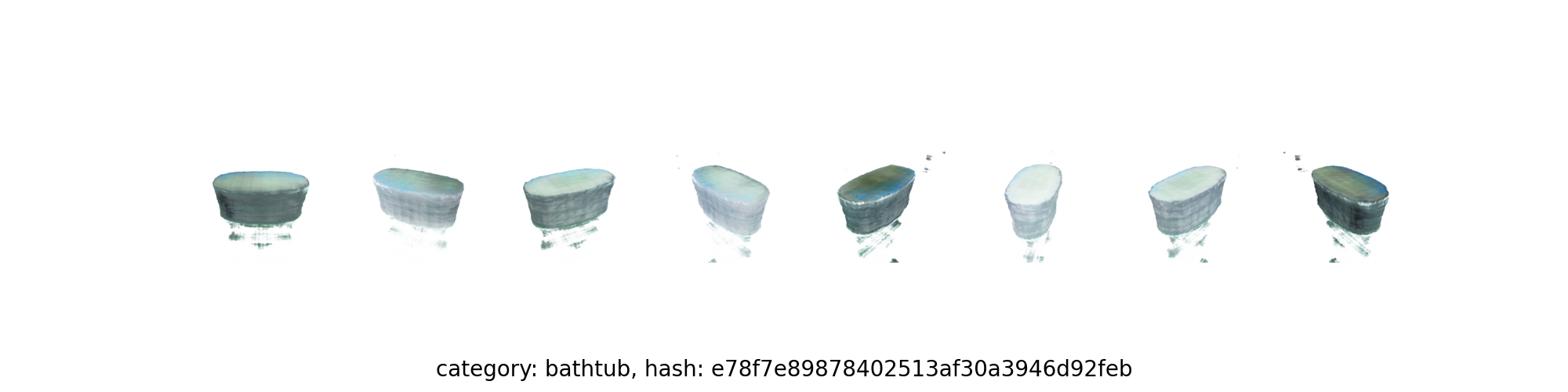}
    \includegraphics[width=0.49\linewidth,trim={2cm 2cm 0 0},clip]{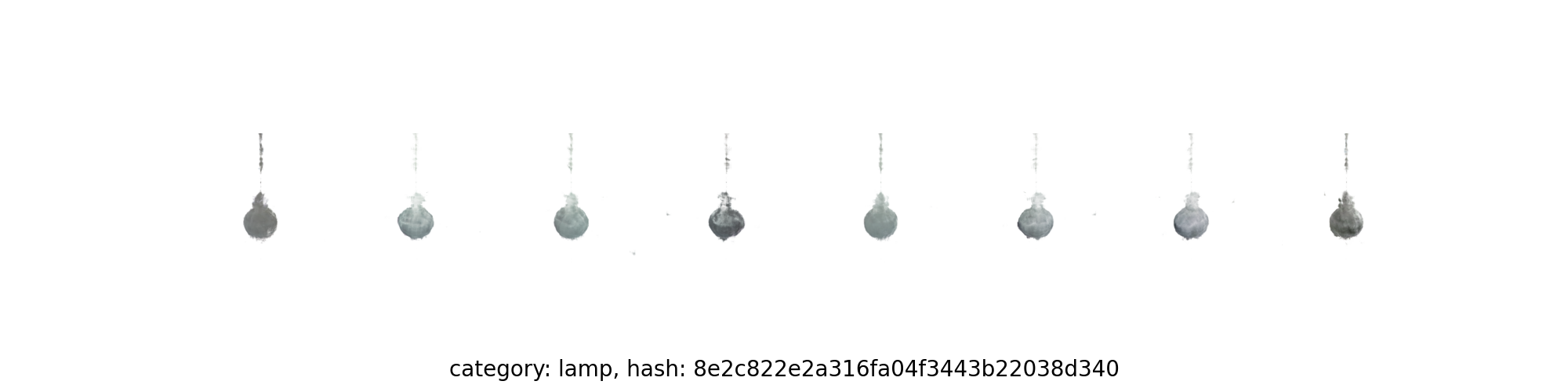}
    \includegraphics[width=0.49\linewidth,trim={2cm 2cm 0 0},clip]{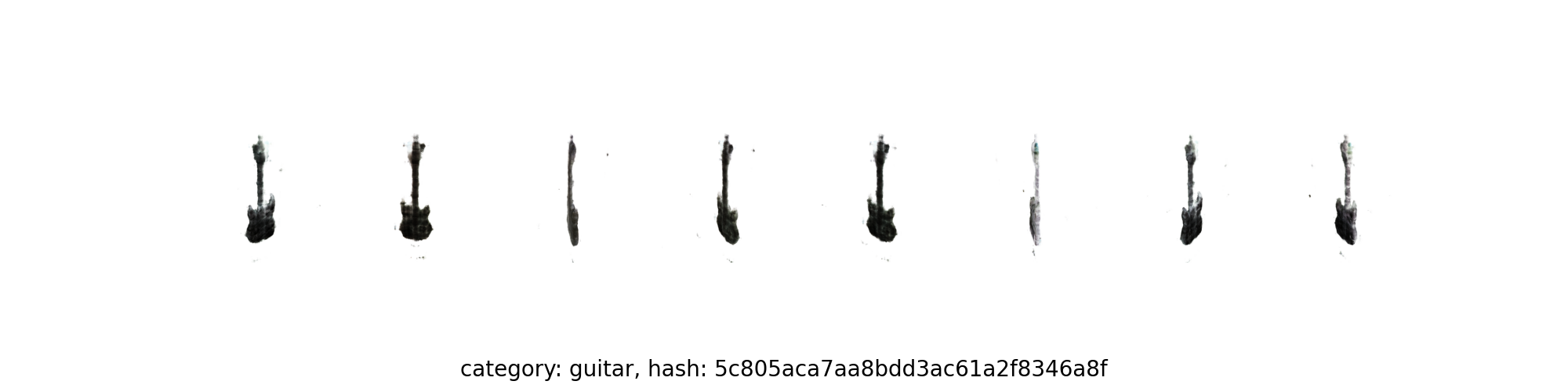}
    \includegraphics[width=0.49\linewidth,trim={2cm 2cm 0 0},clip]{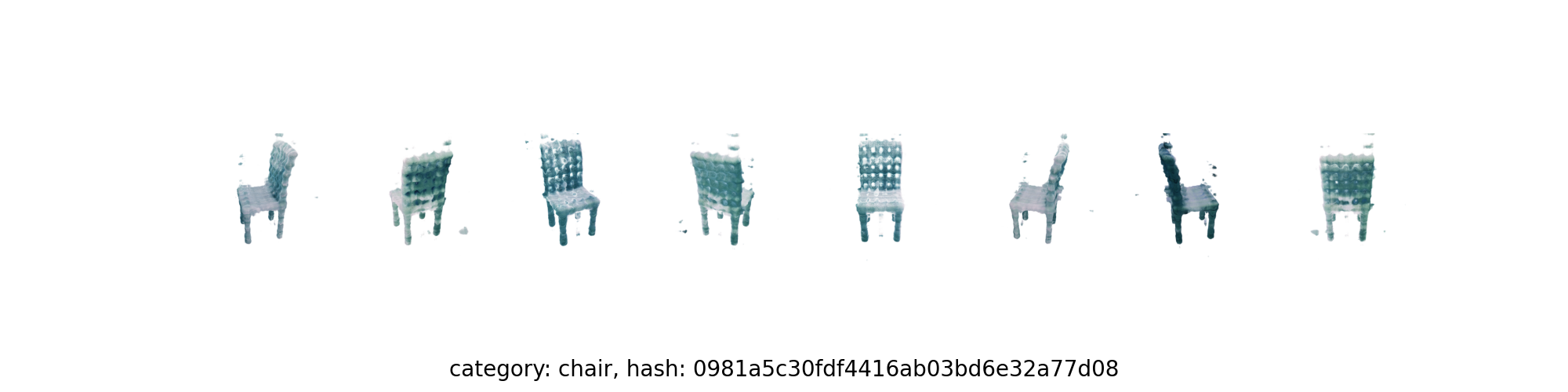}
    \includegraphics[width=0.49\linewidth,trim={2cm 2cm 0 0},clip]{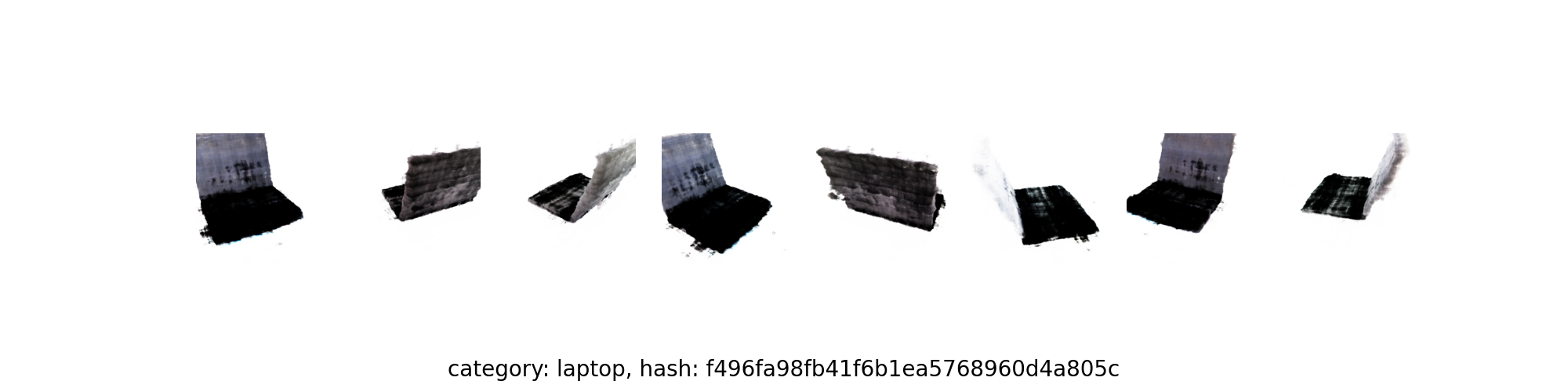}
    \includegraphics[width=0.49\linewidth,trim={2cm 2cm 0 0},clip]{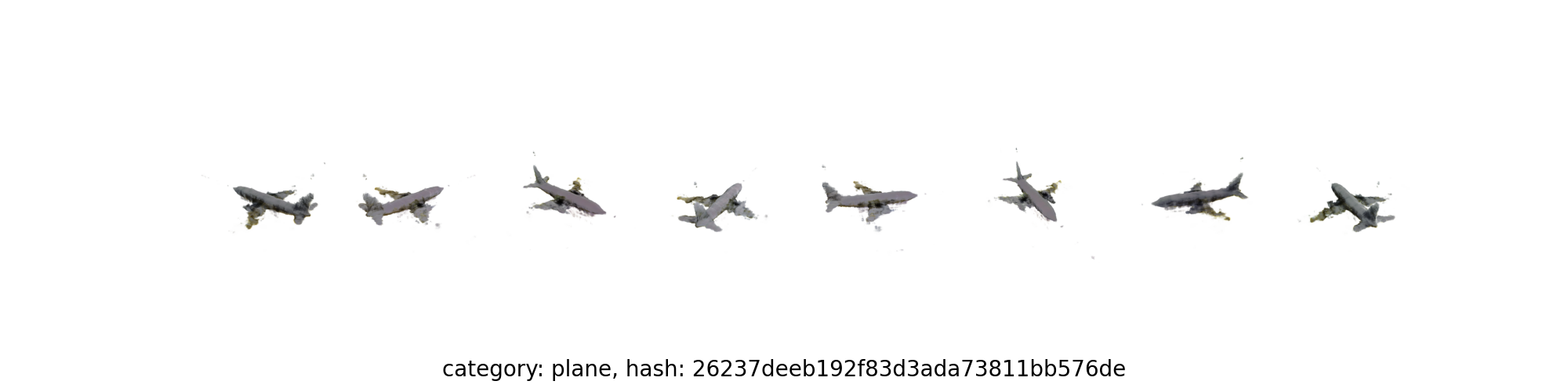}
    \includegraphics[width=0.49\linewidth,trim={2cm 2cm 0 0},clip]{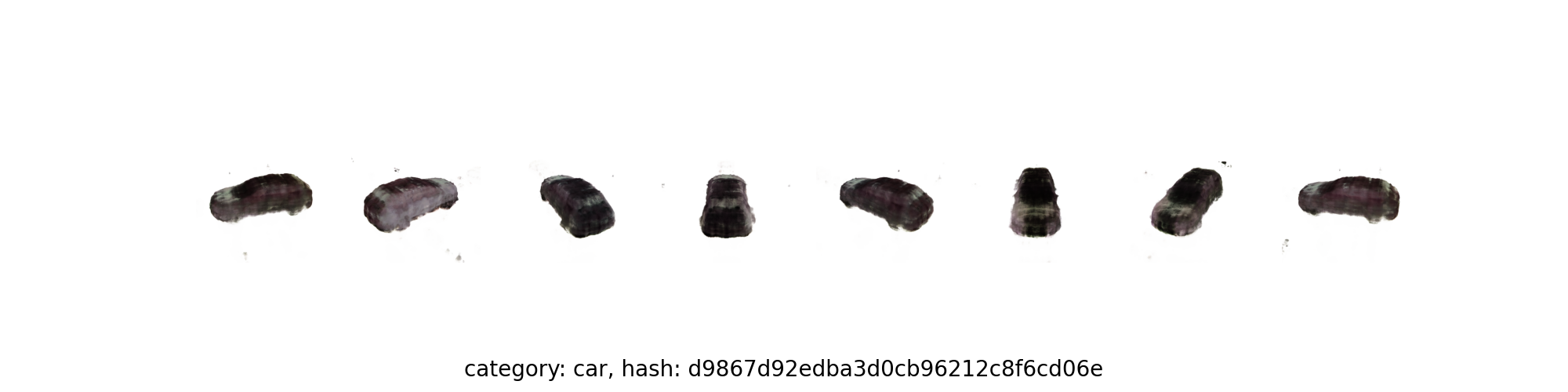}
    \includegraphics[width=0.49\linewidth,trim={2cm 2cm 0 0},clip]{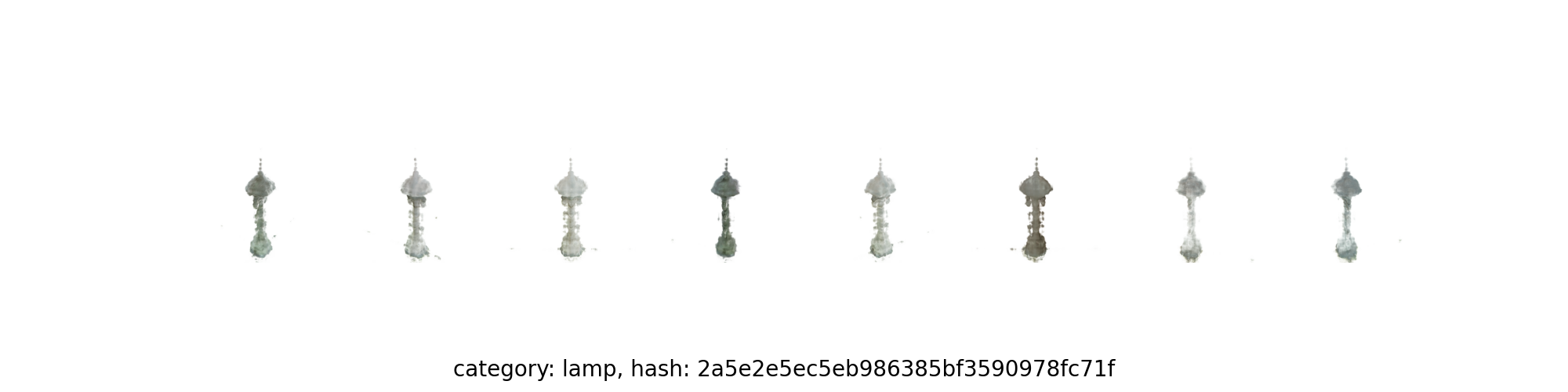}
    \caption{{\bf Multi-view results of in-distribution inference from image prompts}.
    Samples are randomly selected from our test split. For each scene, we display the 8 camera views generated by our NeRF obtained.}
    \label{fig:supp-multi-view-in-dist-1}
\end{figure*}

\begin{figure*}[ht]
    \centering
    \includegraphics[width=0.49\linewidth,trim={2cm 2cm 0 0},clip]{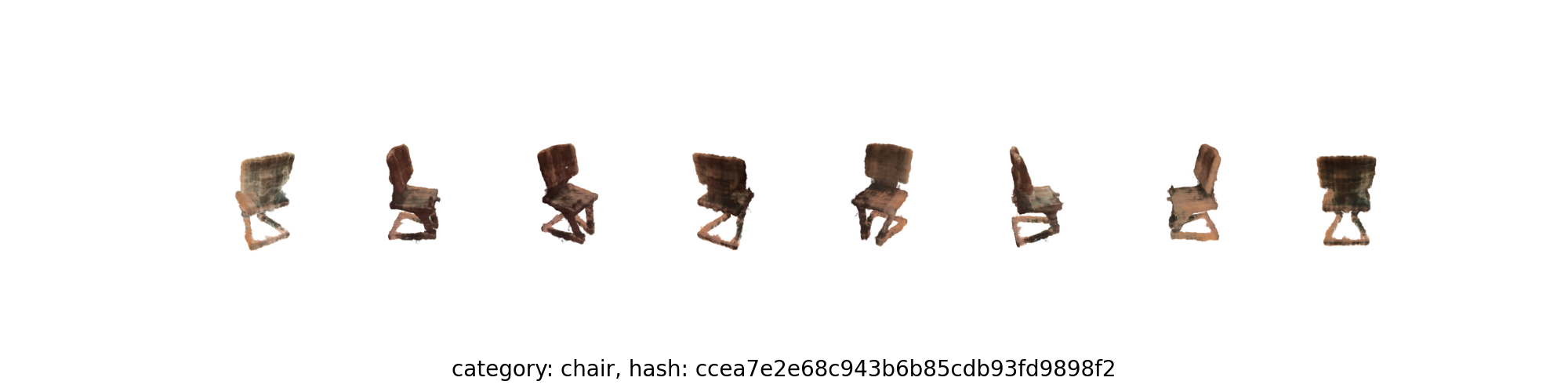}
    \includegraphics[width=0.49\linewidth,trim={2cm 2cm 0 0},clip]{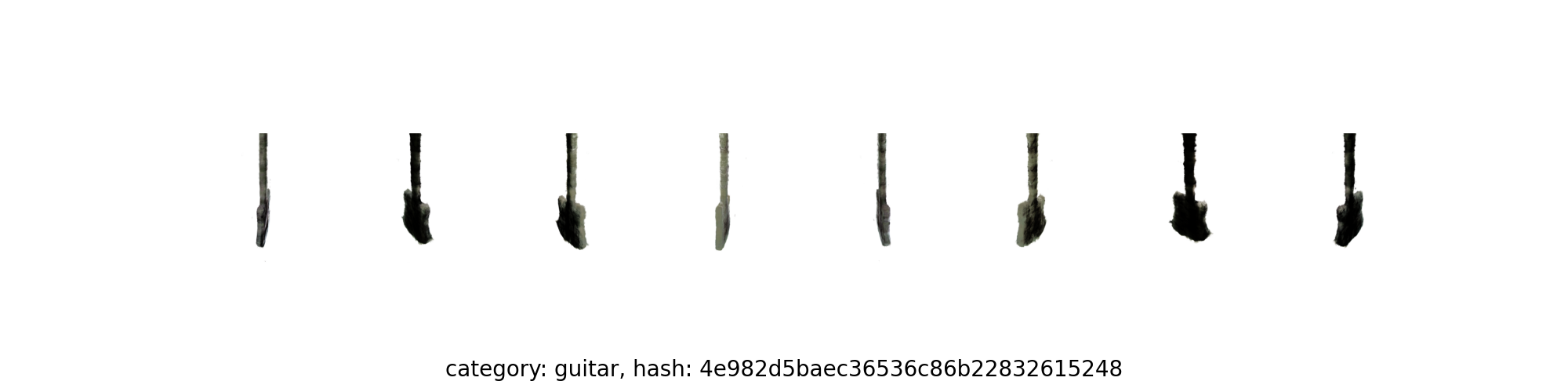}
    \includegraphics[width=0.49\linewidth,trim={2cm 2cm 0 0},clip]{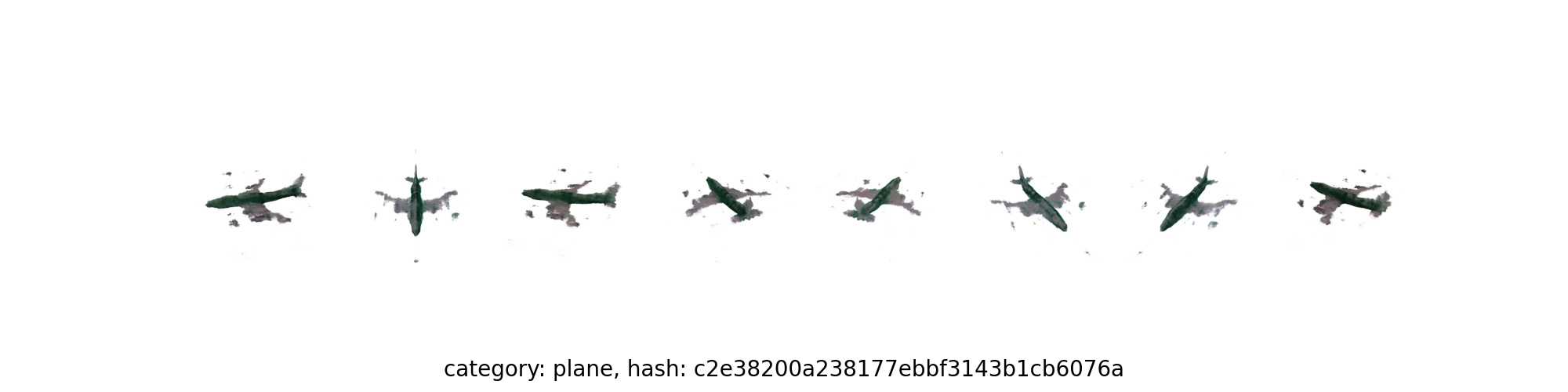}
    \includegraphics[width=0.49\linewidth,trim={2cm 2cm 0 0},clip]{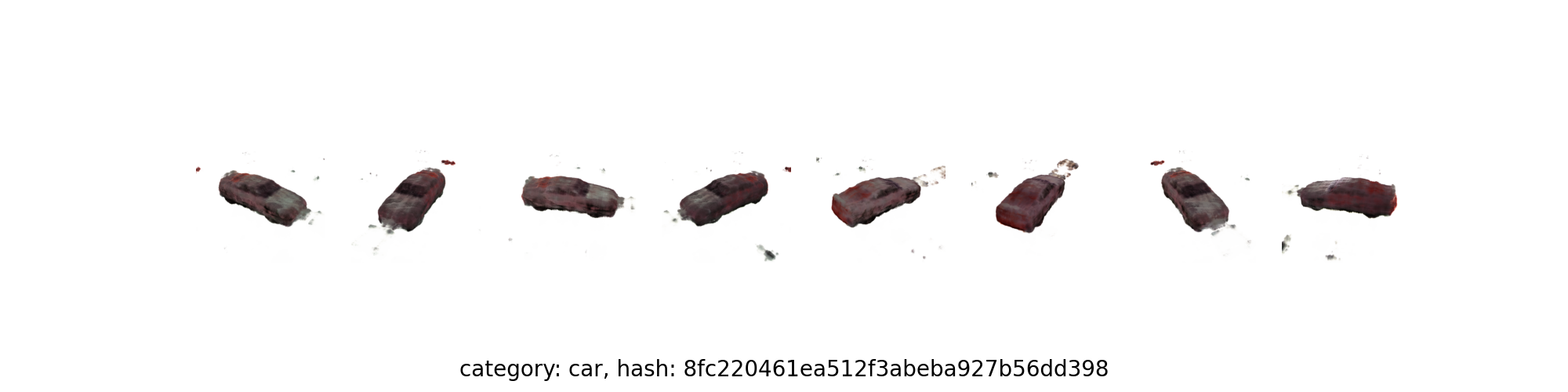}
    \includegraphics[width=0.49\linewidth,trim={2cm 2cm 0 0},clip]{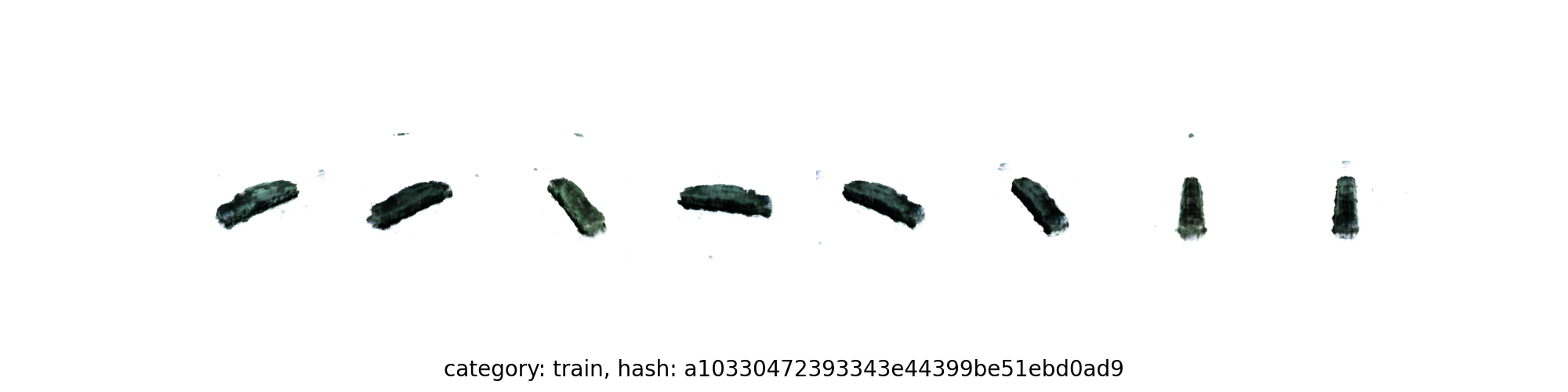}
    \includegraphics[width=0.49\linewidth,trim={2cm 2cm 0 0},clip]{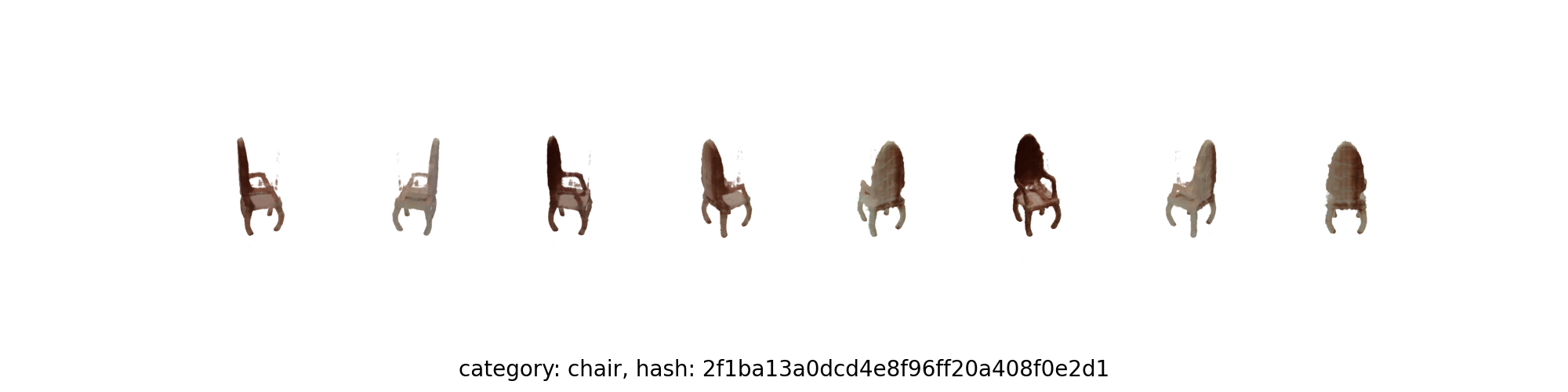}
    \includegraphics[width=0.49\linewidth,trim={2cm 2cm 0 0},clip]{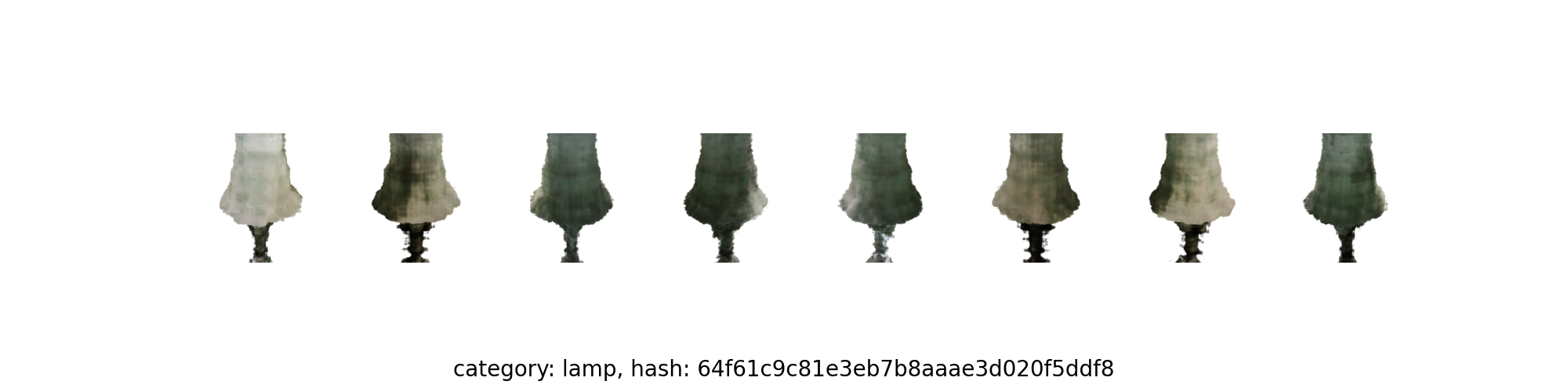}
    \includegraphics[width=0.49\linewidth,trim={2cm 2cm 0 0},clip]{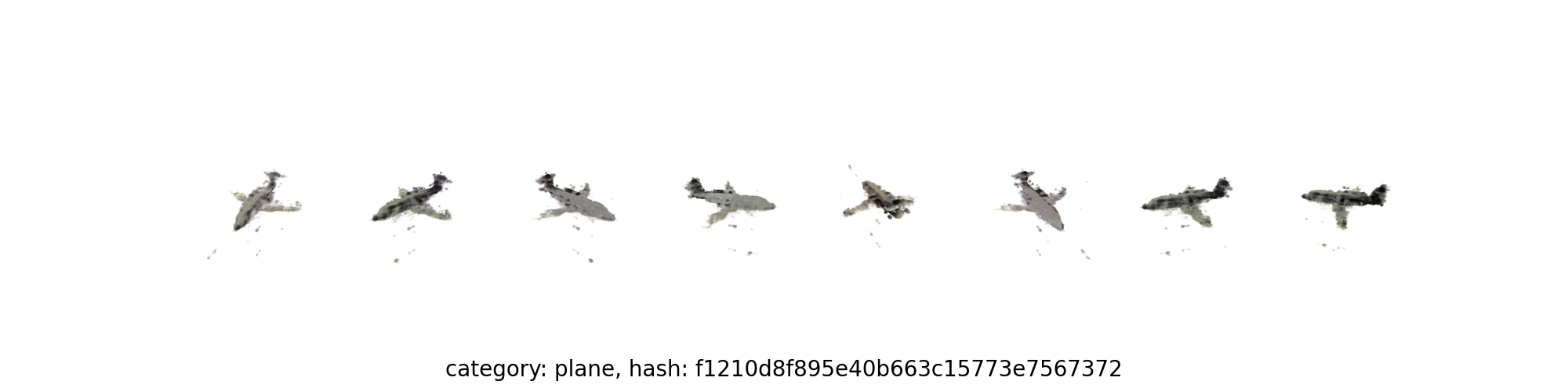}
    \includegraphics[width=0.49\linewidth,trim={2cm 2cm 0 0},clip]{images/supp-in-dist-images/plane_ff2f975a23b78bc78caa71b1fbf7fb98.png}
    \includegraphics[width=0.49\linewidth,trim={2cm 2cm 0 0},clip]{images/supp-in-dist-images/chair_5c1766b6586f451cb61df9fa29d3914e.png}
    \includegraphics[width=0.49\linewidth,trim={2cm 2cm 0 0},clip]{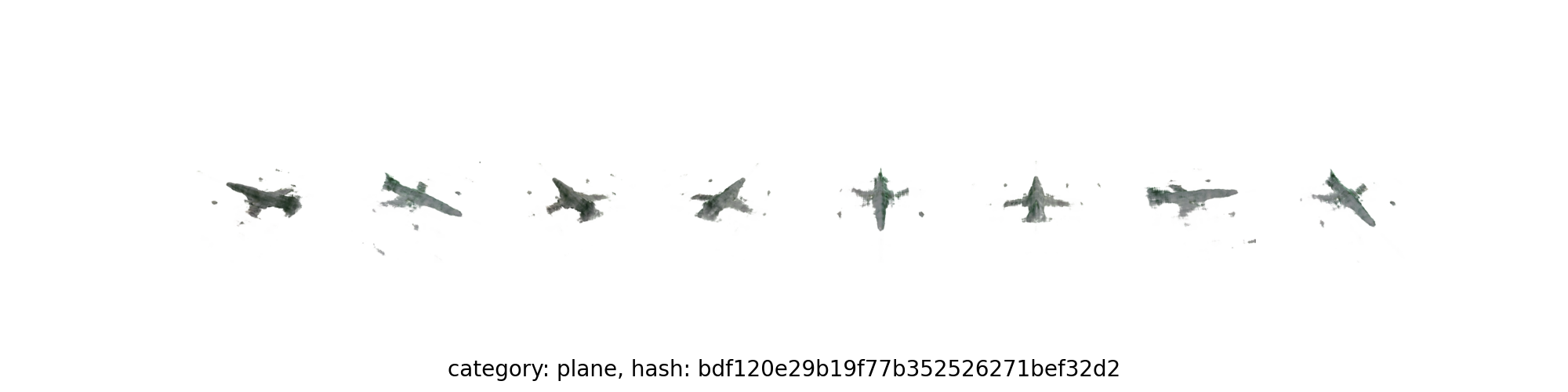}
    \includegraphics[width=0.49\linewidth,trim={2cm 2cm 0 0},clip]{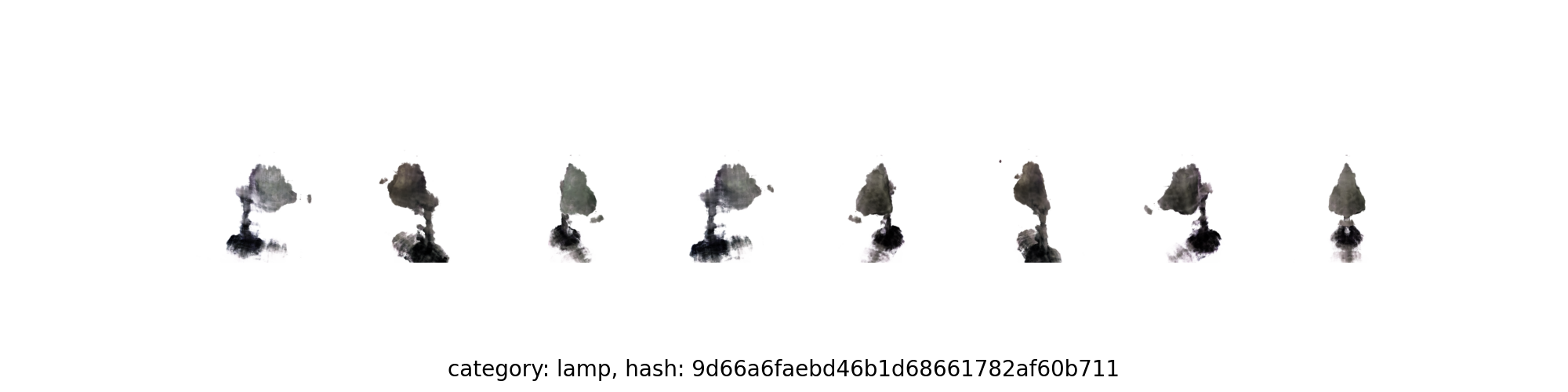}
    \includegraphics[width=0.49\linewidth,trim={2cm 2cm 0 0},clip]{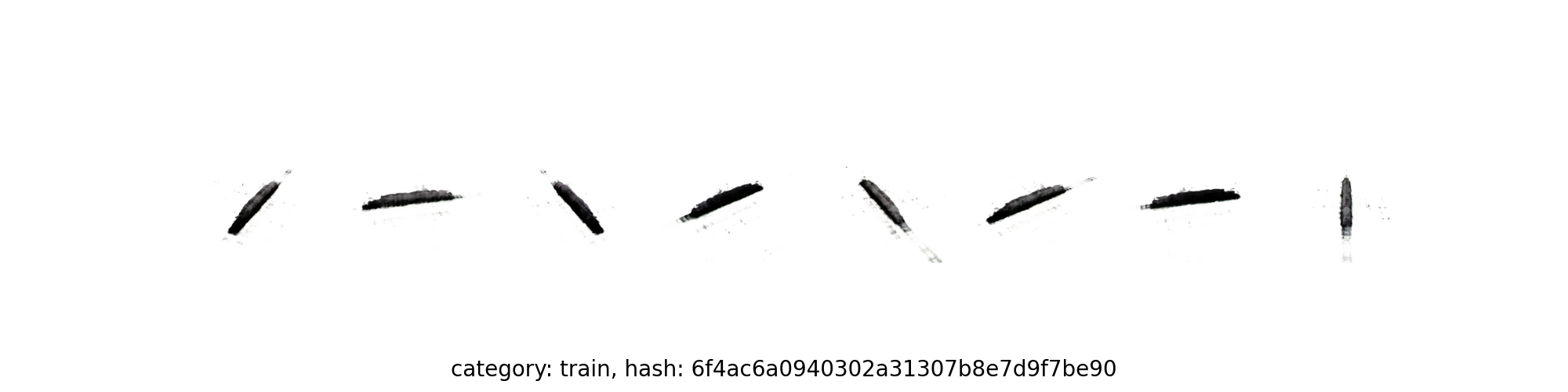}
    \includegraphics[width=0.49\linewidth,trim={2cm 2cm 0 0},clip]{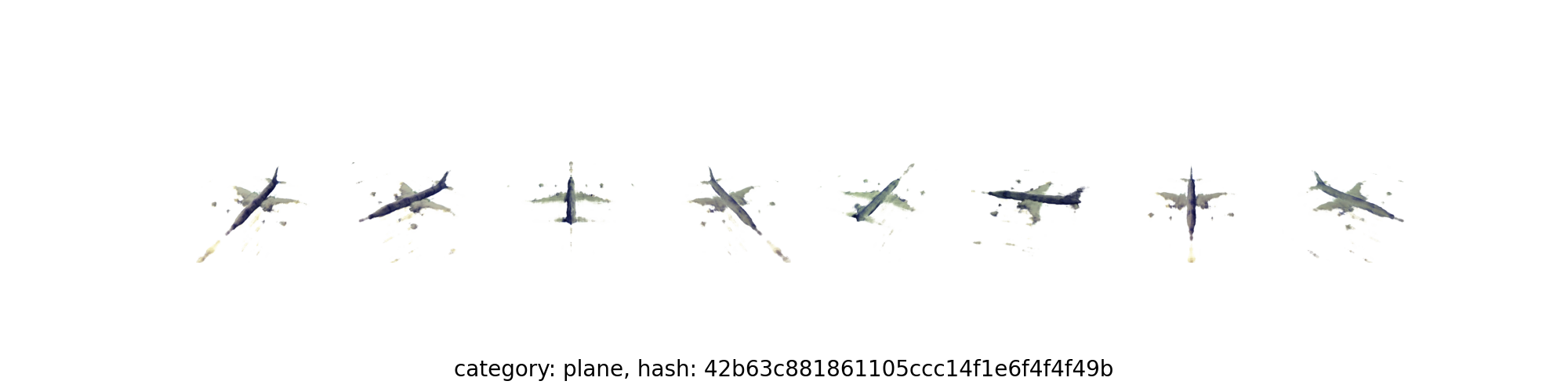}
    \includegraphics[width=0.49\linewidth,trim={2cm 2cm 0 0},clip]{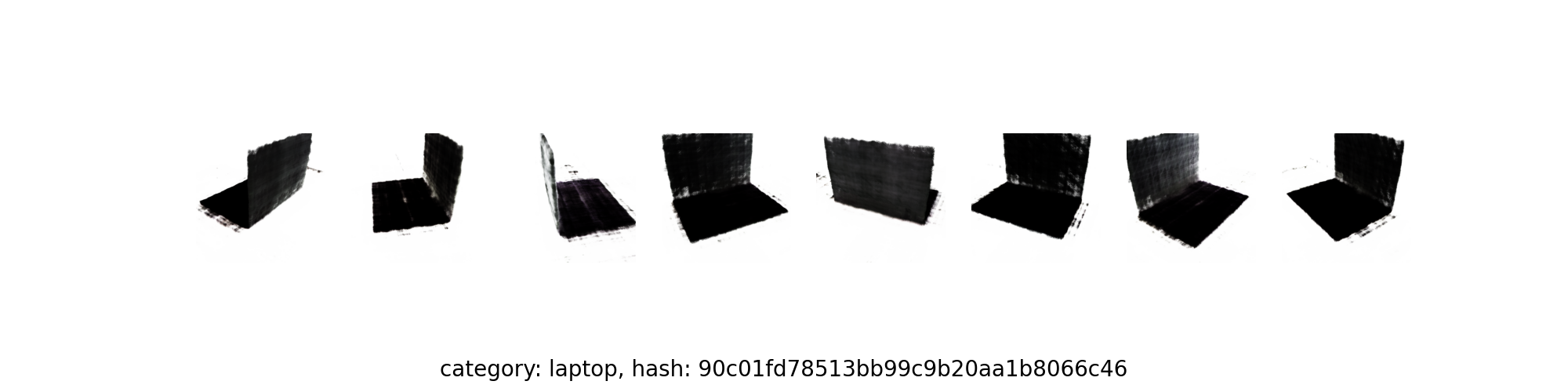}
    \includegraphics[width=0.49\linewidth,trim={2cm 2cm 0 0},clip]{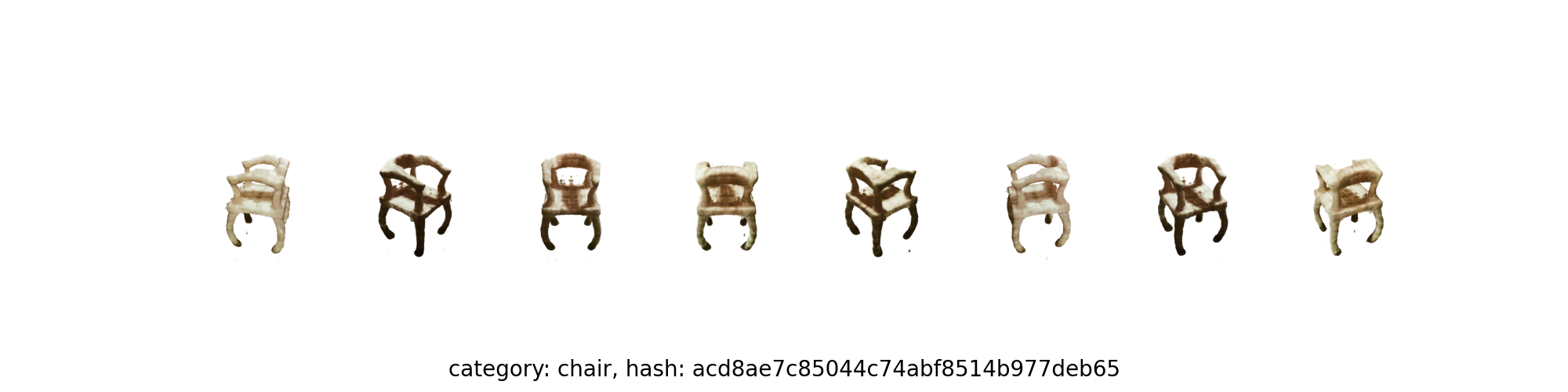}
    \includegraphics[width=0.49\linewidth,trim={2cm 2cm 0 0},clip]{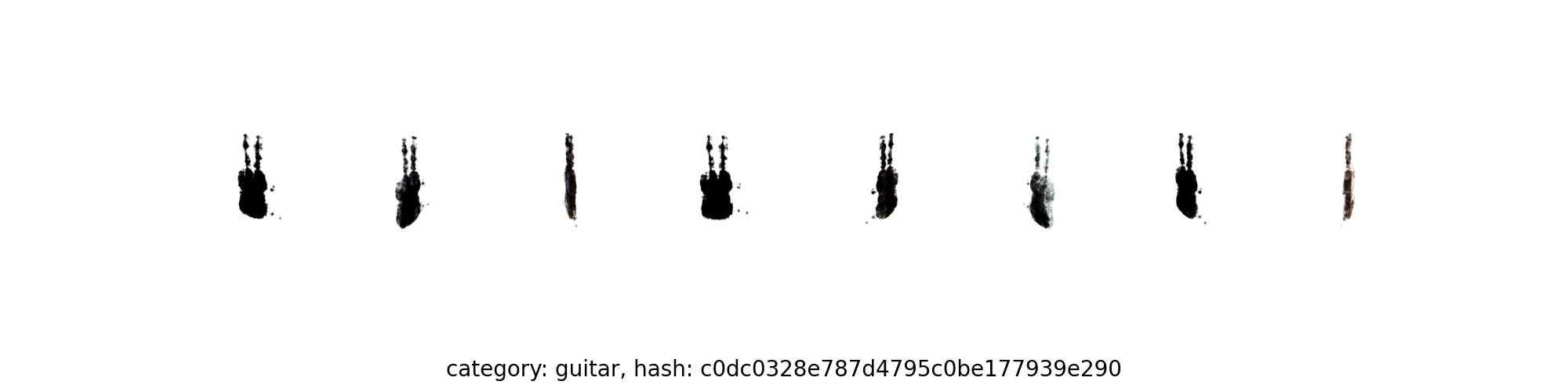}
    \includegraphics[width=0.49\linewidth,trim={2cm 2cm 0 0},clip]{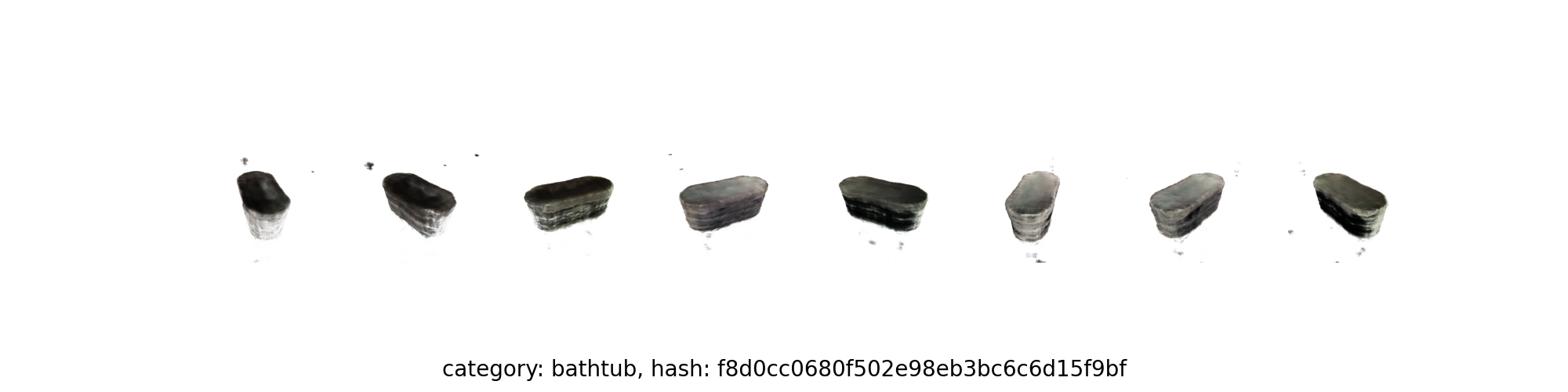}
    \includegraphics[width=0.49\linewidth,trim={2cm 2cm 0 0},clip]{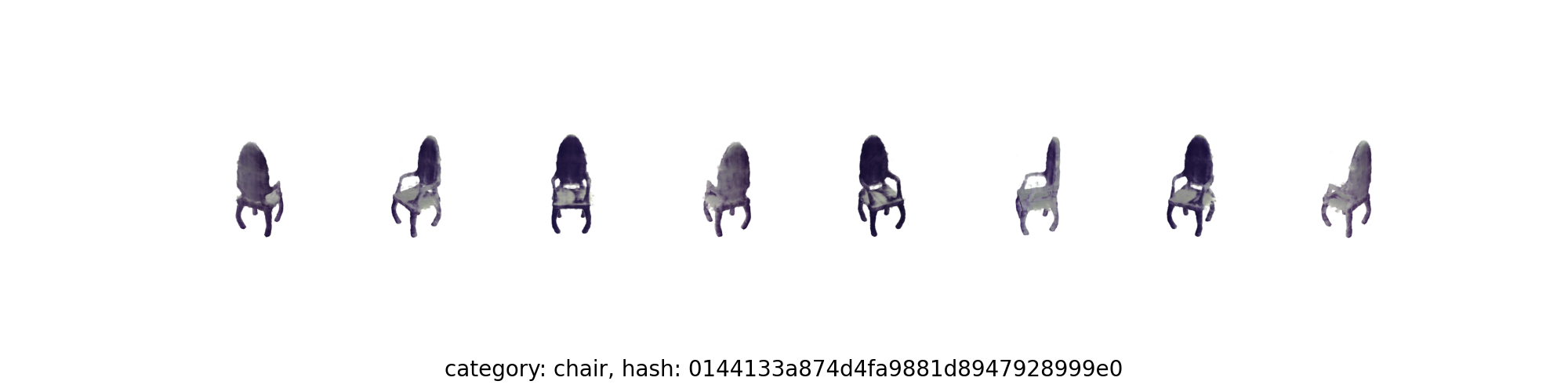}
    \includegraphics[width=0.49\linewidth,trim={2cm 2cm 0 0},clip]{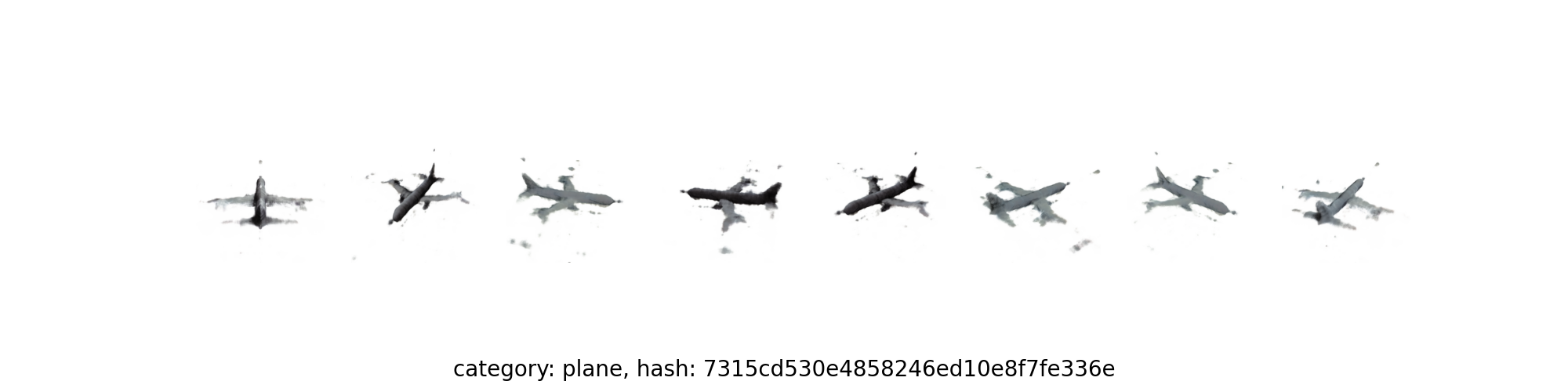}
    \includegraphics[width=0.49\linewidth,trim={2cm 2cm 0 0},clip]{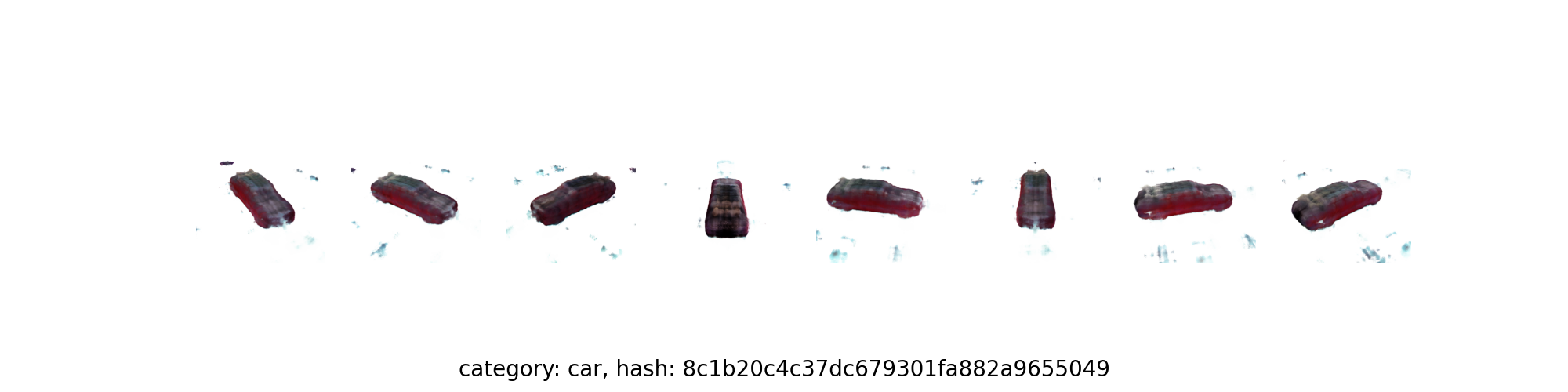}
    \includegraphics[width=0.49\linewidth,trim={2cm 2cm 0 0},clip]{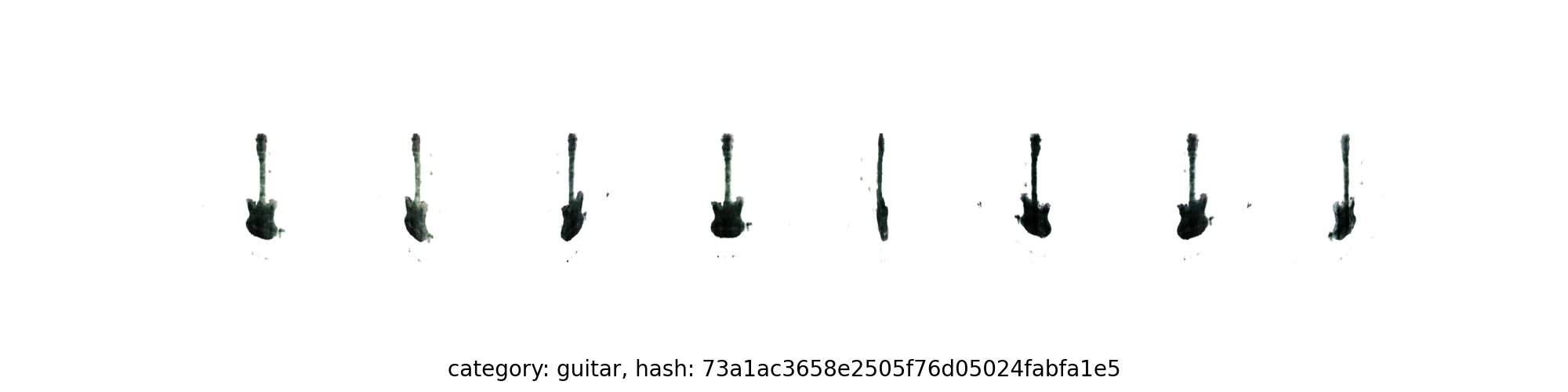}
    \caption{({\it Cont.}) {\bf Multi-view results of in-distribution inference from image prompts}.}
    \label{fig:supp-multi-view-in-dist-2}
\end{figure*}

    

\end{document}